\documentclass{article} 
\usepackage[final]{colm2026_conference}

\usepackage{microtype}
\usepackage{hyperref}
\usepackage{url}
\usepackage{booktabs}
\usepackage{float}
\usepackage{lineno}
\usepackage{makecell}
\usepackage{pifont}
\usepackage{xcolor}
\usepackage{colortbl}
\usepackage{lscape}
\usepackage{pdflscape}
\usepackage{rotating}
\usepackage{array}
\usepackage{graphicx}
\usepackage{caption}
\usepackage{amsmath}
\usepackage[most]{tcolorbox}
\usepackage{tikz}
\usepackage{enumitem}
\usepackage{wrapfig}
\usepackage{subcaption}
\usetikzlibrary{positioning, fit, backgrounds, calc}

\definecolor{darkblue}{rgb}{0, 0, 0.5}
\hypersetup{colorlinks=true, citecolor=darkblue, linkcolor=darkblue, urlcolor=darkblue}

\definecolor{darkgray}{HTML}{404040}
\definecolor{midgray}{HTML}{969696}

\newcommand{\figorbox}[1]{%
  \IfFileExists{#1}{\includegraphics[width=\textwidth]{#1}}%
  {\setlength{\fboxsep}{0pt}\fcolorbox{midgray}{gray!8}{%
     \parbox[c][3.0cm][c]{0.995\textwidth}{\centering
       {\ttfamily\small\detokenize{#1}}\\[4pt]
       {\itshape\small figure not yet regenerated}}}}}

\definecolor{c1}{HTML}{1a9850}
\definecolor{c2}{HTML}{66bd63}
\definecolor{c3}{HTML}{a6d96a}
\definecolor{c4}{HTML}{d9ef8b}
\definecolor{c5}{HTML}{fee08b}
\definecolor{c6}{HTML}{fdae61}
\definecolor{c7}{HTML}{f46d43}
\definecolor{c8}{HTML}{d73027}
\definecolor{c9}{HTML}{a50026}

\definecolor{colI}{HTML}{4575B4}   
\definecolor{colS}{HTML}{1A9850}   
\definecolor{colF}{HTML}{E66101}   
\definecolor{colC}{HTML}{CA0020}   
\definecolor{colM}{HTML}{7570B3}   

\definecolor{cpos}{HTML}{2ca02c}   
\definecolor{cneg}{HTML}{d62728}   

\newcommand{\yes}{{\small$\bullet$}}
\newcommand{\no}{\textcolor{midgray}{---}}

\newcommand{\hc}[1]{%
  \ifdim #1 pt < 0.20pt \cellcolor{c1!45}%
  \else\ifdim #1 pt < 0.40pt \cellcolor{c2!40}%
  \else\ifdim #1 pt < 0.60pt \cellcolor{c3!40}%
  \else\ifdim #1 pt < 0.80pt \cellcolor{c4!45}%
  \else\ifdim #1 pt < 1.00pt \cellcolor{c5!55}%
  \else\ifdim #1 pt < 1.20pt \cellcolor{c6!50}%
  \else\ifdim #1 pt < 1.50pt \cellcolor{c7!45}%
  \else\ifdim #1 pt < 2.00pt \cellcolor{c8!40}%
  \else \cellcolor{c9!35}%
  \fi\fi\fi\fi\fi\fi\fi\fi
  #1%
}

\newcommand{\hcu}[1]{%
  \ifdim #1 pt < 0.20pt \cellcolor{c1!45}%
  \else\ifdim #1 pt < 0.40pt \cellcolor{c2!40}%
  \else\ifdim #1 pt < 0.60pt \cellcolor{c3!40}%
  \else\ifdim #1 pt < 0.80pt \cellcolor{c4!45}%
  \else\ifdim #1 pt < 1.00pt \cellcolor{c5!55}%
  \else\ifdim #1 pt < 1.20pt \cellcolor{c6!50}%
  \else\ifdim #1 pt < 1.50pt \cellcolor{c7!45}%
  \else\ifdim #1 pt < 2.00pt \cellcolor{c8!40}%
  \else \cellcolor{c9!35}%
  \fi\fi\fi\fi\fi\fi\fi\fi
  \underline{#1}%
}

\newcommand{\hcbu}[1]{%
  \ifdim #1 pt < 0.20pt \cellcolor{c1!45}%
  \else\ifdim #1 pt < 0.40pt \cellcolor{c2!40}%
  \else\ifdim #1 pt < 0.60pt \cellcolor{c3!40}%
  \else\ifdim #1 pt < 0.80pt \cellcolor{c4!45}%
  \else\ifdim #1 pt < 1.00pt \cellcolor{c5!55}%
  \else\ifdim #1 pt < 1.20pt \cellcolor{c6!50}%
  \else\ifdim #1 pt < 1.50pt \cellcolor{c7!45}%
  \else\ifdim #1 pt < 2.00pt \cellcolor{c8!40}%
  \else \cellcolor{c9!35}%
  \fi\fi\fi\fi\fi\fi\fi\fi
  \underline{\textbf{#1}}%
}

\newcommand{\hcd}[1]{\hc{#1}$^{\dagger}$}

\newcommand{\hcbud}[1]{\hcbu{#1}$^{\dagger}$}

\newcommand{\hcn}{\cellcolor{white}{---}}

\newsavebox{\ctrtblbox}
\newlength{\ctrtblwd}

\newcommand{\hcR}[1]{%
  \ifdim #1 pt < 0.05pt \cellcolor{c9!35}%
  \else\ifdim #1 pt < 0.12pt \cellcolor{c8!40}%
  \else\ifdim #1 pt < 0.18pt \cellcolor{c7!45}%
  \else\ifdim #1 pt < 0.25pt \cellcolor{c6!50}%
  \else\ifdim #1 pt < 0.32pt \cellcolor{c5!55}%
  \else\ifdim #1 pt < 0.40pt \cellcolor{c4!45}%
  \else\ifdim #1 pt < 0.48pt \cellcolor{c3!40}%
  \else\ifdim #1 pt < 0.55pt \cellcolor{c2!40}%
  \else \cellcolor{c1!45}%
  \fi\fi\fi\fi\fi\fi\fi\fi
  #1%
}
\newcommand{\hcRu}[1]{%
  \ifdim #1 pt < 0.05pt \cellcolor{c9!35}%
  \else\ifdim #1 pt < 0.12pt \cellcolor{c8!40}%
  \else\ifdim #1 pt < 0.18pt \cellcolor{c7!45}%
  \else\ifdim #1 pt < 0.25pt \cellcolor{c6!50}%
  \else\ifdim #1 pt < 0.32pt \cellcolor{c5!55}%
  \else\ifdim #1 pt < 0.40pt \cellcolor{c4!45}%
  \else\ifdim #1 pt < 0.48pt \cellcolor{c3!40}%
  \else\ifdim #1 pt < 0.55pt \cellcolor{c2!40}%
  \else \cellcolor{c1!45}%
  \fi\fi\fi\fi\fi\fi\fi\fi
  \underline{#1}%
}
\newcommand{\hcRbu}[1]{%
  \ifdim #1 pt < 0.05pt \cellcolor{c9!35}%
  \else\ifdim #1 pt < 0.12pt \cellcolor{c8!40}%
  \else\ifdim #1 pt < 0.18pt \cellcolor{c7!45}%
  \else\ifdim #1 pt < 0.25pt \cellcolor{c6!50}%
  \else\ifdim #1 pt < 0.32pt \cellcolor{c5!55}%
  \else\ifdim #1 pt < 0.40pt \cellcolor{c4!45}%
  \else\ifdim #1 pt < 0.48pt \cellcolor{c3!40}%
  \else\ifdim #1 pt < 0.55pt \cellcolor{c2!40}%
  \else \cellcolor{c1!45}%
  \fi\fi\fi\fi\fi\fi\fi\fi
  \underline{\textbf{#1}}%
}

\newcommand{\zc}[2]{\cellcolor{#1}#2}

\newtcolorbox{promptbox}[1][]{%
  enhanced,
  boxrule=0.5pt,
  colframe=midgray,
  colback=white,
  fontupper=\ttfamily\fontsize{6.5}{8.5}\selectfont\color{darkgray},
  left=4pt, right=4pt, top=3pt, bottom=3pt,
  arc=0pt, outer arc=0pt,
  boxsep=2pt,
  title={\sffamily\fontsize{7}{8}\selectfont\bfseries\color{darkgray} #1},
  attach boxed title to top left={yshift=-1mm, xshift=3mm},
  boxed title style={%
    colback=white, colframe=white, boxrule=0pt,
    size=small, left=2pt, right=2pt, top=0.5pt, bottom=0.5pt},
}

\title{Small Foundation Models of Human Cognition\\and Behaviour}

\author{Nick Oh \\
\raisebox{-0.15em}{\includegraphics[height=1em]{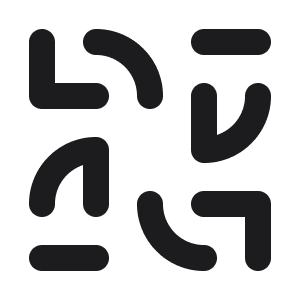}} socius labs\\ 
London, UK\\
\texttt{nick.sh.oh@socius.org}
\And
Fernand Gobet \\
\raisebox{-0.15em}{\includegraphics[height=1em]{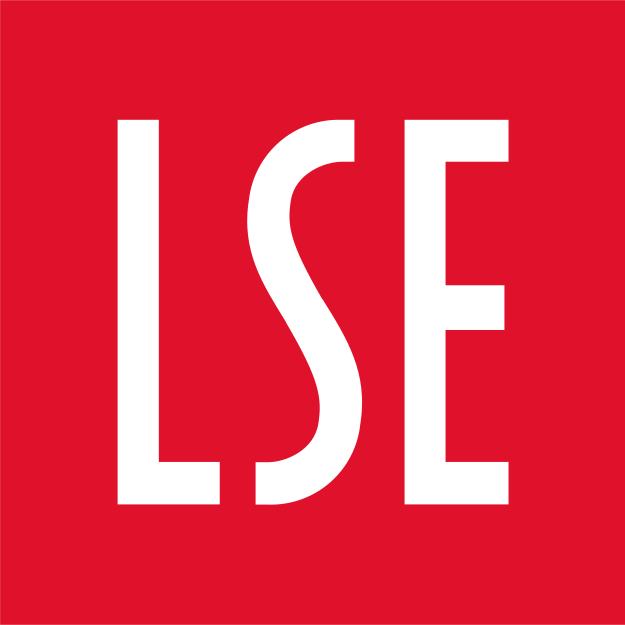}}
London School of Economics and Political Science \\
London, UK\\
\texttt{f.gobet@lse.ac.uk}
}

\begin{document}

\ifcolmsubmission
\linenumbers
\fi

\maketitle

\begin{abstract}
Large language models fine-tuned on human behavioural data have emerged as general-purpose cognitive proxies, but the scale this requires, and whether these models process task structure or exploit statistical shortcuts, remain open questions. We train fourteen models from 135M to 14B parameters across four architecture families on Psych-101, a dataset of 10.7 million trial-level choices from 160 experiments. For in-distribution simulations, scale barely matters. The models fall within a narrow band, as though against a ceiling, and 0.6B to 1B parameters suffice to match a 70B baseline on held-out participants. Out-of-distribution, that band opens into a markedly steeper scaling gradient, with larger models clearly advantaged in generalisation to novel task structure. To determine what information these models use, we run two diagnostics. We progressively strip four prompt channels -- task instructions, experimental stimuli, outcome feedback, and choice history -- across 27 experiments, and permute trial order. Masking the content of stimuli and feedback destroys 75.7\% of learned information and pushes models below chance, demonstrating that choice history alone does not account for performance. Permutation reveals invariance on tasks with independent trials but sensitivity where trial order is determined by prior responses. Small cognitively fine-tuned models therefore show promise as noise ceiling estimators for psychological experiments, though their scope remains bounded by the paradigms seen in training.
\end{abstract}


\begin{center}
\begin{tabular}{@{}c@{\hspace{2em}}c@{}}
  \raisebox{-0.15em}{\includegraphics[height=1em]{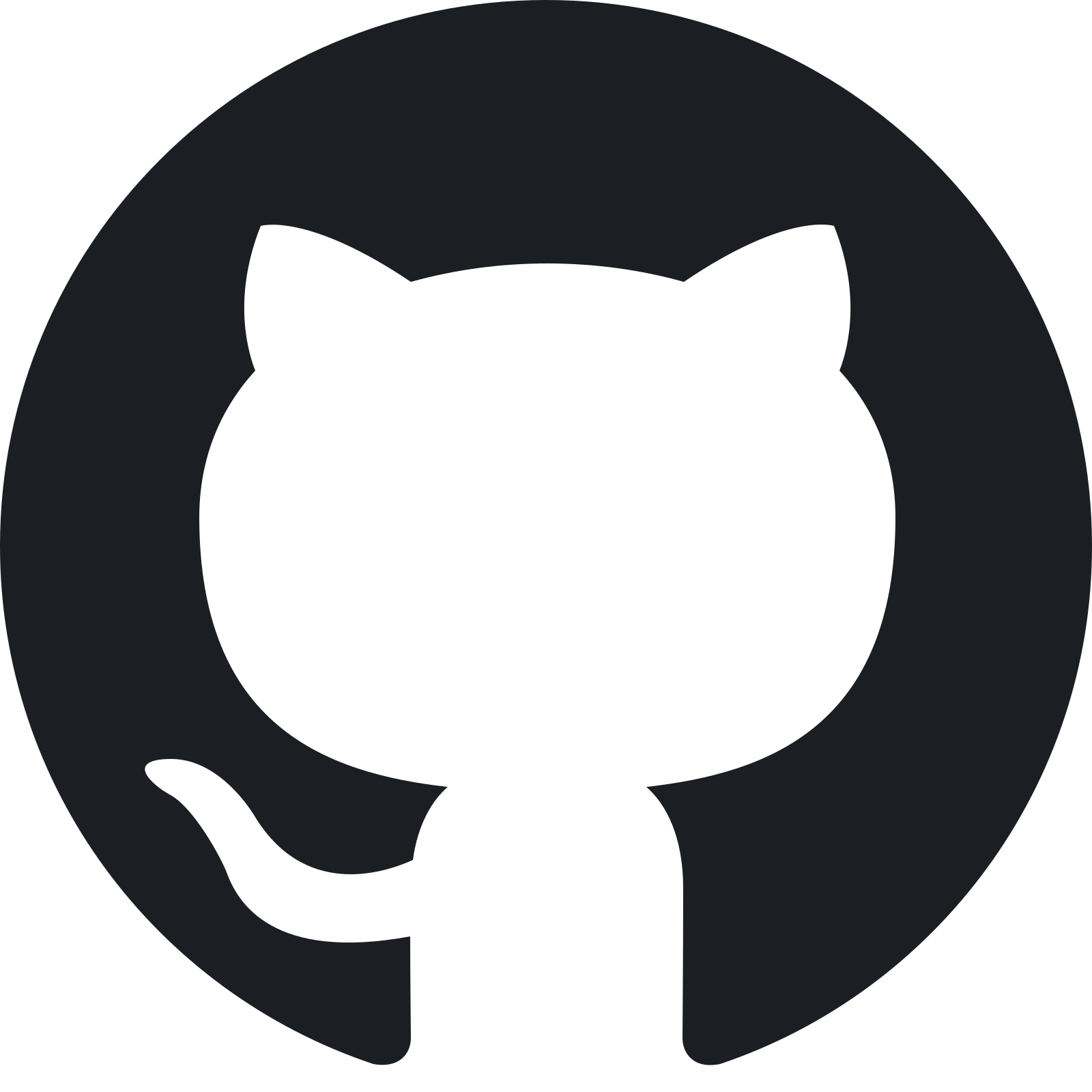}} &
  \raisebox{-0.15em}{\includegraphics[height=1em]{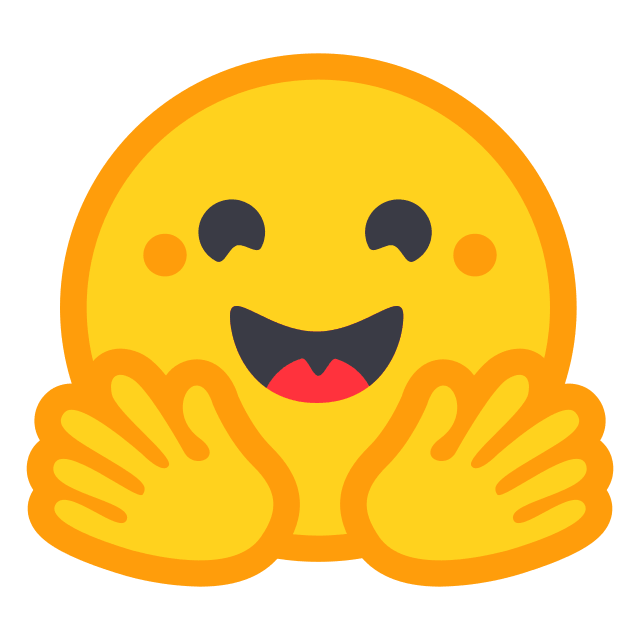}}\\[2pt]
  {\small\url{https://github.com/socius-org/Centauri}} &
  {\small\url{https://hf.co/collections/socius/centauri}}\\[1pt]
\end{tabular}
\end{center}

\section{Introduction}
\label{sec:introduction}

Unified theories of cognition \citep{newell1990unified} aspire to account for the breadth of human mental function within a single framework, and decades of work on symbolic and hybrid architectures such as ACT-R \citep{anderson1998atomic}, CHREST \citep{gobet2001chunking}, and Soar \citep{laird2012soar} have laid a robust foundation. Yet a comprehensive account remains elusive, in part because each new experimental paradigm typically demands new model components, and cross-domain comparison requires reconciling incommensurable formalisms \citep{busemeyer2010cognitive}.

Deep neural networks (DNNs) trained on human behavioural data offer a complementary path, learning regularities directly from choice data and serving as benchmarks against which interpretable models can be evaluated \citep{agrawal2020scaling, peterson2021using, kuperwajs2023using}. \texttt{Centaur} \citep{binz2025foundation} scaled this idea to a general-purpose cognitive proxy by fine-tuning \texttt{Llama-3.1-70B} on Psych-101, a dataset of 10.7 million trial-level choices from over 60,000 participants across 160 experiments. Subsequent work has extended this paradigm to new domains including social science surveys \citep{kolluri2025finetuning}, behavioural economics \citep{xie2025fm}, and social media-derived behavioural prediction \citep{lei2026humanllm}, as well as new post-training methods \citep{zhu2025using} and in-context model generation \citep{rmus2025generating}.

Two questions about this paradigm remain open, and we address them in turn. The first concerns \textit{scale}. Asking whether 70B parameters are necessary is not by itself a well-posed question, because predictive accuracy could be limited by the capacity of the base model, the capacity of the adapter that fine-tuning is permitted to move, the quantity of behavioural training data, or the irreducible stochasticity of human responses. We therefore vary these factors separately, training fourteen models from 135M to 14B parameters across four families -- \texttt{Llama-3.1/3.2} \citep{grattafiori2024llama3}, \texttt{Qwen3} \citep{yang2025qwen3}, \texttt{SmolLM2/3} \citep{allal2025smollm2smolgoesbig, bakouch2025smollm3}, and \texttt{OLMo2/3} \citep{olmo20252olmo2, olmo2026olmo3} -- sweeping LoRA rank from 4 to 64, and fine-tuning on nested subsets of Psych-101. In-distribution, model scale is not what limits prediction. The size needed to match a reproduced \texttt{Centaur-70B} on held-out participants falls from 8B to 0.6B as adapter rank rises from 4 to 32. Out-of-distribution the advantage of scale reappears, so what additional parameters buy is not a better fit to Psych-101 but better transfer beyond it. Model size is therefore a choice about transfer rather than about in-distribution fit (Section~\ref{sec:cogbench}).

The second question concerns \textit{what these models learn}. Recent work has raised the concern that they exploit statistical shortcuts in choice sequences rather than process task structure \citep{xie2025centaur, liu2025can, namazova2025not, schlegelmilch2026snake}, yet no previous study has systematically removed subcomponents of the prompts. We decompose the prompts into four information channels (instruction $I$, stimuli $S$, feedback $F$, and choice history $C$) and conduct structural ablations across 32 experiments, alongside order permutation tests on 2 experiments with contrasting exchangeability structure. The results show a pattern of task-adaptive information use that is difficult to reconcile with pure shortcut learning (Section~\ref{sec:prompt_decomposition}).

Taken together, these results support a modest reading of what \texttt{Centaur}-style supervised fine-tuning produces. These models are not theories of cognition in the sense that ACT-R, CHREST, and Soar are, since they posit no architecture, commit to no mechanism, and explain nothing about how a choice is generated. What they offer instead is an estimate of the noise ceiling\footnote{A noise ceiling estimates the best achievable prediction performance on a dataset after accounting for irreducible stochasticity in human responses.} \citep{agrawal2020scaling, kuperwajs2023using}, an empirical bound on how much of the behaviour in a given paradigm is predictable at all, and thus on how much room a process-level theory of that paradigm still has to improve. Such a bound is informative only if the model reaches it by using the information the experiment actually provides -- which our study establishes -- and it extends only as far as the paradigms the training data covers. We return to this position in Section~\ref{sec:discussion}.

\section{Related work}
\label{sec:related_work}

\subsection{Neural network as a cognitive and behavioural proxy}
\label{sec:cognitive_proxy}

Cognitive models have traditionally been handcrafted and refined through cycles of intuition and model comparison \citep{kuperwajs2023using}. Neural networks trained on human behavioural data are reshaping this process by providing a data-driven reference point against which interpretable theories can be benchmarked. \citet{frank2026cognitive} argue that the key advance of modern AI models is their stimulus computability, the ability to operate over the same stimuli that humans experience, which enables their use as scientific models of mind. \citet{agrawal2020scaling} formalised one instance of this idea as \emph{scientific regret minimisation}, in which the per-datapoint gap between what a DNN can predict and what an interpretable theory predicts identifies precisely where theoretical improvement is possible. 

This approach has proven informative across domains. In \textit{planning}, \citet{kuperwajs2023using} trained feedforward DNNs on 82.8 million moves made by 1.2 million players of 4-in-a-row, establishing a noise ceiling that exposed systematic failures in a heuristic-search cognitive model undetectable from raw data alone. \citet{jensen2024recurrent} took a different but complementary approach, training a meta-reinforcement learning agent that learns when to plan, allocating more deliberation to novel situations and producing rollout patterns resembling rodent hippocampal replays. In \textit{decision-making}, \citet{peterson2021using} found that a DNN trained on approximately 13,000 risky choice problems captured regularities that all 21 existing theories missed, independently recovering Prospect Theory's S-shaped utility function while revealing unformalised context-sensitivity in human value assignment. \citet{zhu2025capturing} extended this to strategic interaction across 90,000 decisions in 2,400 matrix games, showing context-dependent variation in strategic sophistication beyond what fixed-parameter theories can accommodate.

\subsection{Cognitive and behavioural foundation models}
\label{sec:cogfm}

Where the preceding work uses neural networks as tools for improving domain-specific cognitive models, a recent line of work goes further, fine-tuning large language models directly on human behavioural data to create general-purpose foundation models of human cognition and behaviour. The resulting systems vary widely in their training data, adaptation strategy, and target granularity of prediction. A structured comparison of these models, along with full training, infrastructure, and evaluation details, is provided in Appendix~\ref{app:cogfm_comparison}.

\texttt{Centaur}~\citep{binz2025foundation} established the paradigm by fine-tuning \texttt{Llama-3.1-70B} on Psych-101, outperforming domain-specific cognitive models on held-out participants while generalising to unseen tasks and domains. Subsequent work has extended this approach in two directions. The first broadens the data foundation: \texttt{Be.FM}~\citep{xie2025fm}, \texttt{Socrates}~\citep{kolluri2025finetuning}, and \texttt{HumanLLM}~\citep{lei2026humanllm} move beyond controlled experiments to behavioural science literature, large-scale social science surveys with demographic metadata, and social media corpora respectively, targeting capabilities from population-level distributional alignment to individualised behaviour prediction. The second explores alternatives to supervised fine-tuning: \citet{zhu2025using} applied reinforcement learning (GRPO) to risky choice data, showing that chain-of-thought traces are causally linked to prediction accuracy, while \texttt{GeCCo}~\citep{rmus2025generating} avoided fine-tuning entirely, using in-context learning to generate interpretable parametric cognitive models as executable code that matched or exceeded both \texttt{Centaur} and the best handcrafted models.

\section{Methods}
\label{sec:methods}

\textbf{Data.} We train on Psych-101~\citep{binz2025foundation}, a corpus of 10.7 million trial-level choices from over 60,000 participants across 160 psychological experiments, transcribed into natural language. Each prompt encodes a single participant's complete trial-by-trial session, and we use the same train--test splits and prompt templates as \texttt{Centaur} without modification. Experiments fall into six task types, namely decision-making, bandit, Markov decision processes (MDP), memory, supervised learning, and miscellaneous.

\textbf{Models.} We fine-tune fourteen pre-trained base\footnote{We fine-tune base rather than post-trained models. \citet{binz2026posttrainingmakeslargelanguage} find that post-training reduces alignment with human behaviour across model families, sizes, and objectives; \citet{gao2025increasing} find that alignment with human brain responses improves monotonically with scale in base models, while instruction-tuned variants show no advantage over base models of the same size.} (non-instruct) models drawn from four families: five \texttt{Qwen3-Base} models (0.6B, 1.7B, 4B, 8B, 14B)~\citep{yang2025qwen3}, three \texttt{Llama-3} models (1B, 3B, 8B)~\citep{grattafiori2024llama3}, four \texttt{SmolLM} models (\texttt{SmolLM2} at 135M, 360M, and 1.7B~\citep{allal2025smollm2smolgoesbig}, \texttt{SmolLM3} at 3B~\citep{bakouch2025smollm3}), and two \texttt{OLMo} models (\texttt{OLMo-2} at 1B~\citep{olmo20252olmo2}, \texttt{OLMo-3} at 7B~\citep{olmo2026olmo3}). We refer to the resulting fine-tuned models as \textbf{\texttt{Qwentaur}}, \textbf{\texttt{Llama-Centaur}}, \textbf{\texttt{Smoltaur}}, and \textbf{\texttt{Olmotaur}} respectively.

\textbf{Training.} Following \citet{binz2025foundation}, we apply supervised fine-tuning with cross-entropy loss masked to human response tokens only, so that the model is not optimised on task instructions or context. We use rank-stabilised LoRA applied to all linear layers, trained for one epoch with AdamW (learning rate $5 \times 10^{-5}$, weight decay 0.01) and a linear warmup of 100 steps. All models are trained in bf16 precision on a single NVIDIA A100 80GB GPU, with training times ranging from 3 to 49 hours.

\textbf{Varying capacity and data.} Model scale, adapter capacity, and training-set size are varied independently, each while the other two are held. Adapter capacity is swept over $r \in \{4, 8, 16, 32, 64\}$ at full data. Training-set size is varied over nested, experiment-stratified subsets of Psych-101 (1/16, 1/8, 1/4, 1/2, and the full set) at fixed rank ($r$=16). Subsets are nested so that each larger fraction strictly contains the smaller ones, and stratified so that all 160 experiments remain represented at every size, ensuring that reducing the quantity of training data does not also reduce paradigm coverage.

\textbf{Comparability.} Psych-101 sessions are long, with a median of $4{,}183$ tokens and a maximum of $44{,}599$, so a model's context window determines how much of a session it can condition on. \texttt{Qwen3}, \texttt{Llama-3}, \texttt{SmolLM3} and \texttt{OLMo-3} are fine-tuned/evaluated at $32{,}768$ tokens and lose $2.5\%$ of trials to truncation, whereas \texttt{SmolLM2} is capped at $8{,}192$ ($25.0\%$) and \texttt{OLMo-2} at $4{,}096$ ($39.6\%$). In both families that span generations, \texttt{SmolLM2/3} and \texttt{OLMo-2/3}, the larger model is also the newer generation with the larger window, so scaling within those families confounds parameter count with generation and context. We therefore mainly focus on the eight \texttt{Qwentaur} and \texttt{Llama-Centaur} models in Section~\ref{sec:experiment}, comparing \texttt{Smoltaur} and \texttt{Olmotaur} against them at matched size where it is informative.

\section{Experiments}
\label{sec:experiment}

With fourteen models trained on Psych-101, we ask two questions. The first is how much of \texttt{Centaur}'s predictive accuracy survives at small scale (Section~\ref{sec:cogbench}), and the second is what information in the prompt the models actually use (Section~\ref{sec:prompt_decomposition}). Adapter rank and training-set size vary only in Section~\ref{sec:cogbench_psych101}, where they are themselves the object of study. Sections~\ref{sec:cogbench_psych201} and~\ref{sec:prompt_decomposition} use one configuration throughout, LoRA rank 16 on the full dataset.

\subsection{Cognitive benchmark performance}
\label{sec:cogbench}

We evaluate on two benchmarks. Psych-101 tests prediction for held out participants in-distribution; Psych-201-RT, a subset of Psych-201 \citep{binz2026posttrainingmakeslargelanguage}, tests transfer to out-of-distribution experiments. Full per-experiment results for both are reported in Appendices~\ref{app:full_results_psych101} and~\ref{app:full_results_psych201}. Figures report mean negative log-likelihood, in nats, over the 38 of 46 Psych-101 tasks for which \citet{binz2025foundation} publish a domain-specific cognitive model. Lower is better throughout.

\subsubsection{Psych-101 (in-distribution performance)}
\label{sec:cogbench_psych101}

Comparing each fine-tuned model with its own base checkpoint of the same size separates what fine-tuning contributes from what the base model already provides. \texttt{Qwen3-0.6B} scores $0.731$ before fine-tuning and $0.528$ after, a gain of $0.203$ nats; across the eight models the average gain is $0.188$ (range $0.138$ to $0.295$). Scale buys far less. Going from \texttt{Qwentaur-0.6B} to \texttt{Qwentaur-14B} gains $0.019$ nats, and from \texttt{Llama-Centaur-1B} to \texttt{Llama-Centaur-8B} $0.024$ nats. On Psych-101, whether a model has been fine-tuned matters roughly ten times more than how large it is. The fine-tuning gain shrinks steadily as the base model grows, from $0.203$ at \texttt{Qwentaur-0.6B} to $0.138$ at \texttt{Qwentaur-14B}, and from $0.295$ at \texttt{Llama-Centaur-1B} to $0.158$ at \texttt{Llama-Centaur-8B}. A larger model already predicts Psych-101 better before any fine-tuning, so less is left for fine-tuning to supply.

At rank 8, the adapter capacity \texttt{Centaur} itself used (QLoRA, 4-bit base weights), four models match or beat our reproduction of \texttt{Centaur-70B} ($0.524$; see Appendix~\ref{app:70b_discrepancy}): \texttt{Qwentaur-8B} and \texttt{Llama-Centaur-8B} (both $0.517$), \texttt{Qwentaur-4B} ($0.520$) and \texttt{Smoltaur-3B} ($0.523$), with \texttt{Qwentaur-1.7B} close behind ($0.528$). Three to eight billion parameters therefore suffice to match a seventy-billion-parameter model trained on the same data with the same adapter budget, though at higher base-weight precision.

\begin{figure}[t]
\centering
\includegraphics[width=\textwidth]{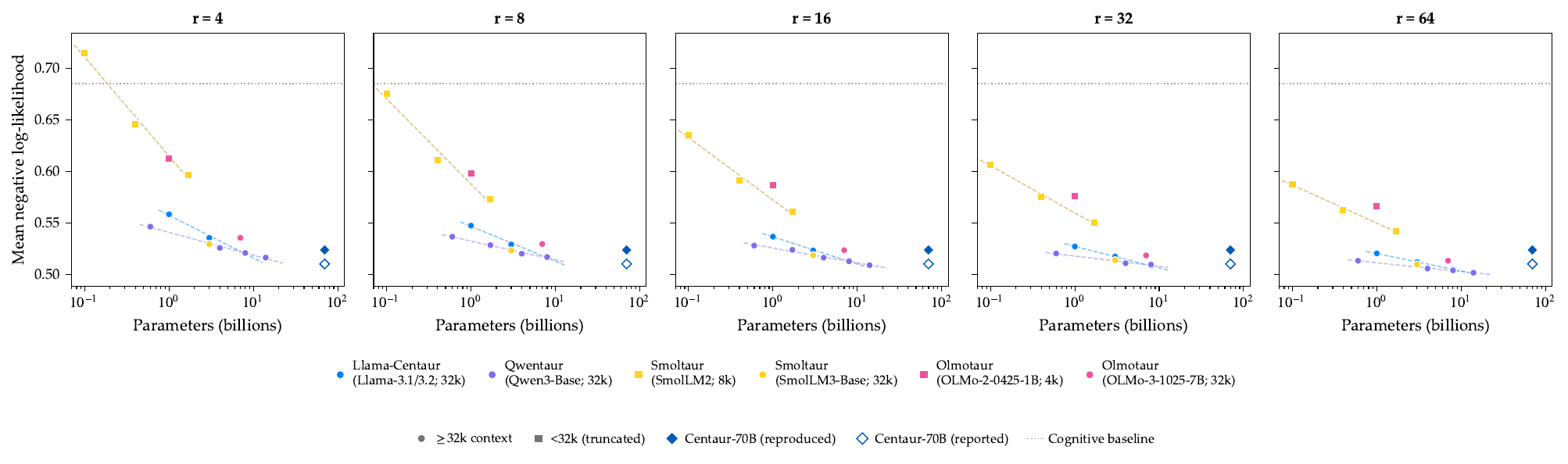}
\caption{\textbf{Adapter rank against model size on Psych-101.} Mean negative log-likelihood over the 38 of 46 Psych-101 tasks for which \citet{binz2025foundation} publish a domain-specific cognitive model, plotted against parameter count, one panel per adapter rank ($r=4$--$64$). Trend lines are fitted only within groups matched on generation and context window, so \texttt{Olmotaur} has none and the \texttt{Smoltaur} fit covers the \texttt{SmolLM2} models only.}
\label{fig:rank_sweep}
\end{figure}

That threshold is not a fixed property of the task. Adapter capacity substitutes for parameters, so a small model at high rank reaches a much larger model at low rank in both matched families. \texttt{Qwentaur-0.6B} at $r{=}64$ ($0.513$) matches \texttt{Qwentaur-14B} at $r{=}4$ ($0.516$) with twenty-three times fewer parameters, and \texttt{Llama-Centaur-1B} at $r{=}64$ ($0.520$) matches \texttt{Llama-Centaur-8B} at $r{=}4$ ($0.521$) with eight times fewer. The size needed to match \texttt{Centaur-70B} moves accordingly, from 8B at $r{=}4$ to 3B at $r{=}8$, 1.7B at $r{=}16$, and 0.6B at $r{=}32$ and above (Figure~\ref{fig:rank_sweep}).

Training data is a third lever on the same quantity, and its returns are visibly diminishing. Across nested subsets of Psych-101 the gain from each doubling declines steadily, with \texttt{Qwentaur-14B} gaining $0.016$, $0.013$, $0.009$ and $0.008$ nats over successive doublings, so further participants of the same kind would add little (Appendix~\ref{app:rank_data}).

\subsubsection{Psych-201 (out-of-distribution performance)}
\label{sec:cogbench_psych201}

\begin{figure}[t]
\centering
\includegraphics[width=\textwidth]{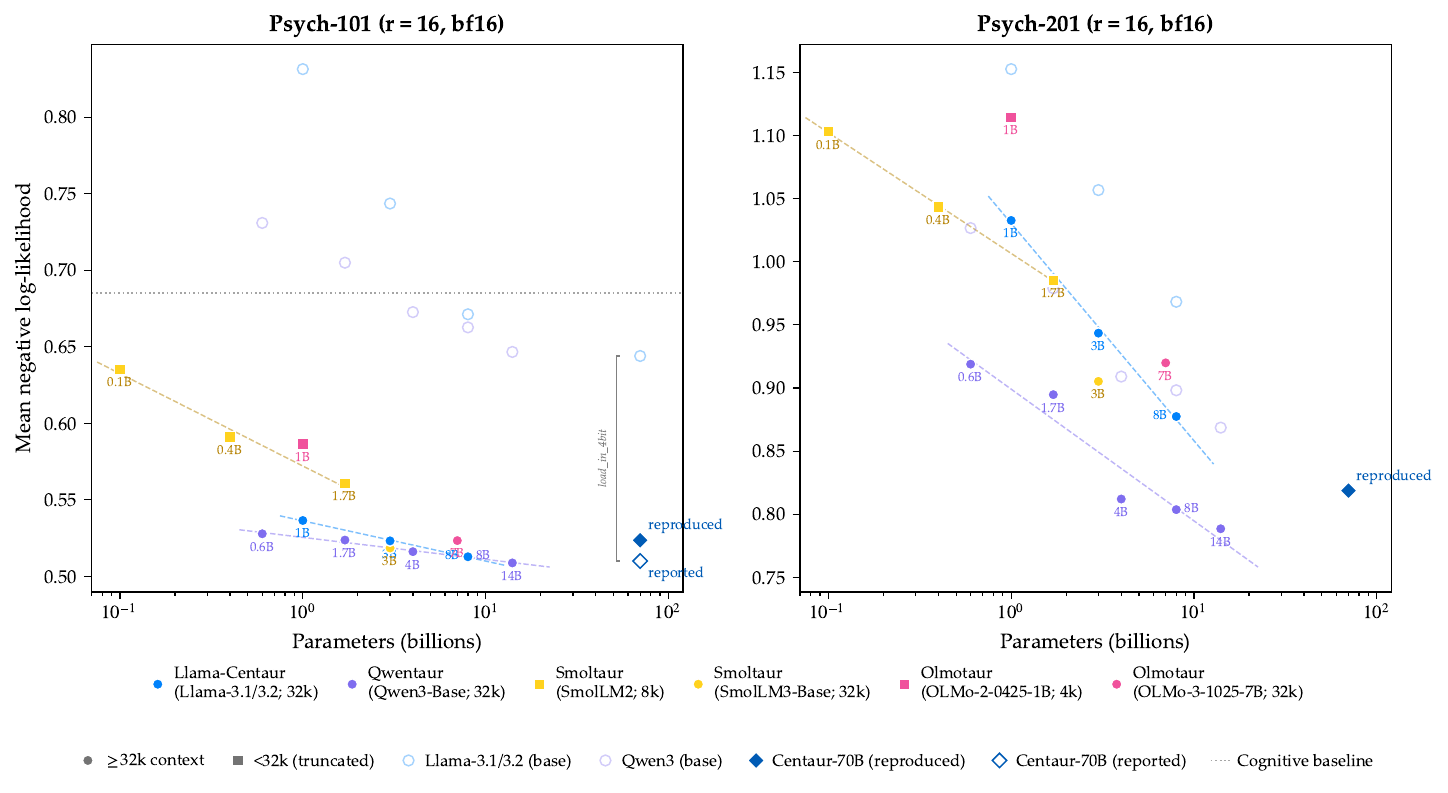}
\caption{\textbf{In-distribution versus out-of-distribution scaling at rank 16.} \textbf{(a)}~Psych-101, mean NLL over the 38 tasks with a reported cognitive model. \textbf{(b)}~Psych-201-RT, all 18 held-out experiments; no cognitive baseline is published for Psych-201. Trend lines are fitted only within groups matched on generation and context window.}
\label{fig:scaling_r16_ood}
\end{figure}

The in-distribution results leave the models looking almost interchangeable. On Psych-101 the eight matched models (at rank 16 on the full dataset) span just $0.028$ nats, from $0.509$ to $0.537$, a band that also contains our \texttt{Centaur-70B} reproduction. On the eighteen held-out experiments of Psych-201-RT the same eight models run from $0.788$ to $1.033$, a spread of $0.244$ (Figure~\ref{fig:scaling_r16_ood}).

Before fine-tuning, a larger model is better on both benchmarks. A tenfold larger \texttt{Qwen3} predicts Psych-101 $0.062$ nats better and Psych-201 $0.118$ better; for \texttt{Llama-3} the figures are $0.177$ and $0.204$. After fine-tuning, the in-distribution advantage nearly disappears ($0.014$ and $0.026$) while the out-of-distribution advantage largely persists ($0.104$ and $0.172$). In other words, once a model has been fine-tuned on Psych-101, making it larger buys very little further accuracy on Psych-101 itself, but as much accuracy on unseen experiments as it did before fine-tuning. Model size should therefore be chosen for transfer rather than for in-distribution fit.

Set against each other, the two benchmarks show that what bounds these models as instruments is the range of paradigms they were trained on rather than their capacity. In-distribution they sit in a narrow band, as though against a ceiling; out-of-distribution they remain far apart, and the largest model in each family is still clearly the best, with no sign of a ceiling. All fourteen fine-tuned models beat the best base model in the study in-distribution, an un-fine-tuned \texttt{Qwen3-14B} at $0.647$, whereas only three do out of distribution, against the same model at $0.869$. A noise-ceiling estimate taken from one of these models is therefore trustworthy for paradigms in Psych-101, and unvalidated beyond them. We cannot, however, separate the benchmark from the training objective. Psych-101 may contain structure that supervised fine-tuning on next-choice prediction is unable to extract, in which case the limit belongs to the method rather than the data (Section~\ref{sec:limitation}).

%


\subsection{Prompt decomposition and diagnostic testing}
\label{sec:prompt_decomposition}

A model is useful as an instrument only if its predictions rest on the information the participant used, and aggregate fit cannot separate that from exploiting whichever regularity in the prompt is most predictive \citep{bowers2025misuse, orr2025not}. For \texttt{Centaur}-style supervised fine-tuning the concern is concrete, since recent work reports that most of its accuracy survives reducing the prompt to choice history or removing task instructions \citep{xie2025centaur, liu2025can}. We therefore decompose the Psych-101 prompt into information channels and ask which of them the model relies on (Section~\ref{sec:structural_ablation}) and whether it uses them as each task's structure requires (Section~\ref{sec:order_permutation}).

\subsubsection{Decomposing the training objective}
\label{sec:training_objective}

\citet{xie2025centaur} formalise \texttt{Centaur}'s training objective as maximising $P(C_{t+1} \mid C_{1:t}, T)$ at the token level, where each $C$ represents a tokenised participant choice and $T$ denotes task-related information in the prompt.
This formulation bundles all non-choice information into a single variable~$T$, which conflates conceptually distinct sources of information that can be independently manipulated.
We propose a finer-grained decomposition of the prompt into four information channels, $P\!\bigl(C_{t+1} \mid I\vphantom{S_{1:t+1}}, S_{1:t+1}, F_{1:t}, C_{1:t})$, where $I$ is the \textbf{task instruction}, $S_{1:t+1}$ the \textbf{experimental stimuli} including those of the upcoming trial, $F_{1:t}$ the \textbf{outcome feedback} after each past choice, and $C_{1:t}$ the participant's \textbf{choice history}.


The decomposition lets us state the shortcut hypothesis precisely. In its strongest form it asserts \textit{history sufficiency}, that choice history already carries all the information available about the next choice, $C_{t+1} \perp\!\!\!\perp \{I, S, F\} \mid C_{1:t}$, so a model could reach its full-prompt accuracy while ignoring instruction, stimuli and feedback outright. Weaker forms allow the model to read the remaining channels but deny that it uses their content, and they differ in which channel they exempt. The one our design is built to test is \textit{format sufficiency}, that the model locates the choice from the recurring arrangement of stimulus, choice and feedback statements without using what any of those statements say.

\begin{figure*}[t]
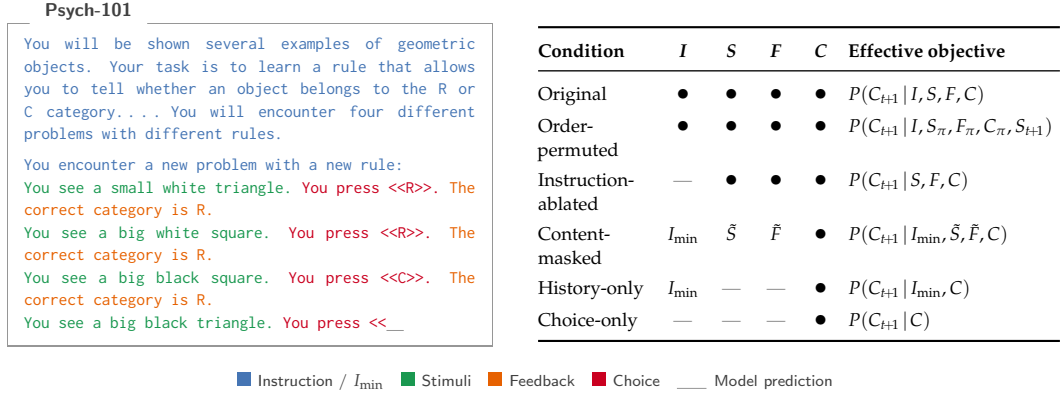

\centering
\begin{minipage}[t]{0.46\textwidth}
\vspace{2pt}%
\begin{promptbox}[Psych-101]
\textcolor{colI}{You will be shown several examples of geometric objects. Your task is to learn a rule that allows you to tell whether an object belongs to the R or C category.\,\dots\;You will encounter four different problems with different rules.}\par\smallskip
\textcolor{colI}{You encounter a new problem with a new rule:}\par
\textcolor{colS}{You see a small white triangle.} \textcolor{colC}{You press <<R>>.} \textcolor{colF}{ The correct category is R.}\par
\textcolor{colS}{You see a big white square.} \textcolor{colC}{You press <<R>>.} \textcolor{colF}{ The correct category is R.}\par
\textcolor{colS}{You see a big black square.} \textcolor{colC}{You press <<C>>.} \textcolor{colF}{ The correct category is R.}\par
\textcolor{colS}{You see a big black triangle.} \textcolor{colC}{You press <<}\textcolor{midgray}{\underline{\phantom{MM}}}
\end{promptbox}
\end{minipage}%
\hfill%
\begin{minipage}[t]{0.50\textwidth}
\vspace{15pt}%
\centering
\setlength{\tabcolsep}{2pt}
\fontsize{7}{9}\selectfont
\renewcommand{\arraystretch}{1.25}
\begin{tabular}{
    @{}
    >{\raggedright\arraybackslash}p{1.5cm}
    >{\centering\arraybackslash}p{0.55cm}
    >{\centering\arraybackslash}p{0.45cm}
    >{\centering\arraybackslash}p{0.45cm}
    >{\centering\arraybackslash}p{0.45cm}
    >{\raggedright\arraybackslash}p{2.8cm}
    @{}
}
\toprule
\textbf{Condition} & $\boldsymbol{I}$ & $\boldsymbol{S}$ & $\boldsymbol{F}$ & $\boldsymbol{C}$ & \textbf{Effective objective} \\
\midrule
Original
    & \yes & \yes & \yes & \yes
    & $P(C_{t\!+\!1} \!\mid\! I, S, F, C)$ \\[1pt]
Order-permuted
    & \yes & \yes & \yes & \yes
    & $P(C_{t\!+\!1} \!\mid\! I, S_{\pi}, F_{\pi}, C_{\pi}, S_{t\!+\!1})$ \\[1pt]
Instruction-ablated
    & \no & \yes & \yes & \yes
    & $P(C_{t\!+\!1} \!\mid\! S, F, C)$ \\[1pt]
Content-masked
    & $I_{\min}$ & $\tilde{S}$ & $\tilde{F}$ & \yes
    & $P(C_{t\!+\!1} \!\mid\! I_{\min}, \tilde{S}, \tilde{F}, C)$ \\[1pt]
History-only
    & $I_{\min}$ & \no & \no & \yes
    & $P(C_{t\!+\!1} \!\mid\! I_{\min}, C)$ \\[1pt]
Choice-only
    & \no & \no & \no & \yes
    & $P(C_{t\!+\!1} \!\mid\! C)$ \\
\bottomrule
\end{tabular}
\end{minipage}
\par\vspace{4pt}
\begin{center}
\sffamily\fontsize{6.5}{8}\selectfont\color{darkgray}
\textcolor{colI}{\rule{5pt}{5pt}}\;\,Instruction / $I_\mathrm{min}$\quad
\textcolor{colS}{\rule{5pt}{5pt}}\;\,Stimuli\quad
\textcolor{colF}{\rule{5pt}{5pt}}\;\,Feedback\quad
\textcolor{colC}{\rule{5pt}{5pt}}\;\,Choice\quad
\textcolor{midgray}{\underline{\phantom{MM}}}\;\,Model prediction
\end{center}
\caption{%
\textbf{Prompt decomposition into four information channels.}
\textbf{(Left)}~Example prompt from Psych-101 \citep{binz2025foundation} with channels colour-coded.
\textbf{(Right)}~Ablation conditions and their effective prediction objectives.
Per-experiment prompts under all four ablation conditions are shown in Figures~\ref{fig:ablation_badham2017deficits}--\ref{fig:ablation_tomov2021multitask} (Appendix~\ref{app:structural_ablation}).
}
\label{fig:prompt_decomposition}
\end{figure*}

Existing studies test the strong form, and only in part. \citet{xie2025centaur} evaluate a history-only condition and \citet{liu2025can} test instruction-ablated, choice-only and adversarial-instruction conditions, but each covers only four experiments, and neither varies the granularity of what is removed, so neither separates content from template (Appendix~\ref{app:prior_ablations}). We extend both scope and granularity (Figure~\ref{fig:prompt_decomposition}). First, we evaluate four structural ablation conditions, original, instruction-ablated, content-masked and history-only, across 32 sequential experiments. The content-masked condition is new to this work and is the one that discriminates the two hypotheses: it replaces specific stimulus values and feedback outcomes with generic placeholders (e.g.\ ``some points'', ``a shape'') and reduces the instruction to a minimal action-space definition ($I_{\min}$), while preserving each trial's formatting, so whatever degradation it causes is attributable to informational content rather than to positional template. Second, we test order permutation on two experiments, one exchangeable and one adaptive as a negative control. Only one exchangeable task is available, as most Psych-101 experiments carry sequential dependencies.

\subsubsection{Structural ablation (sequential tasks)}
\label{sec:structural_ablation}

\begin{figure}[t]
\centering
\includegraphics[width=\textwidth]{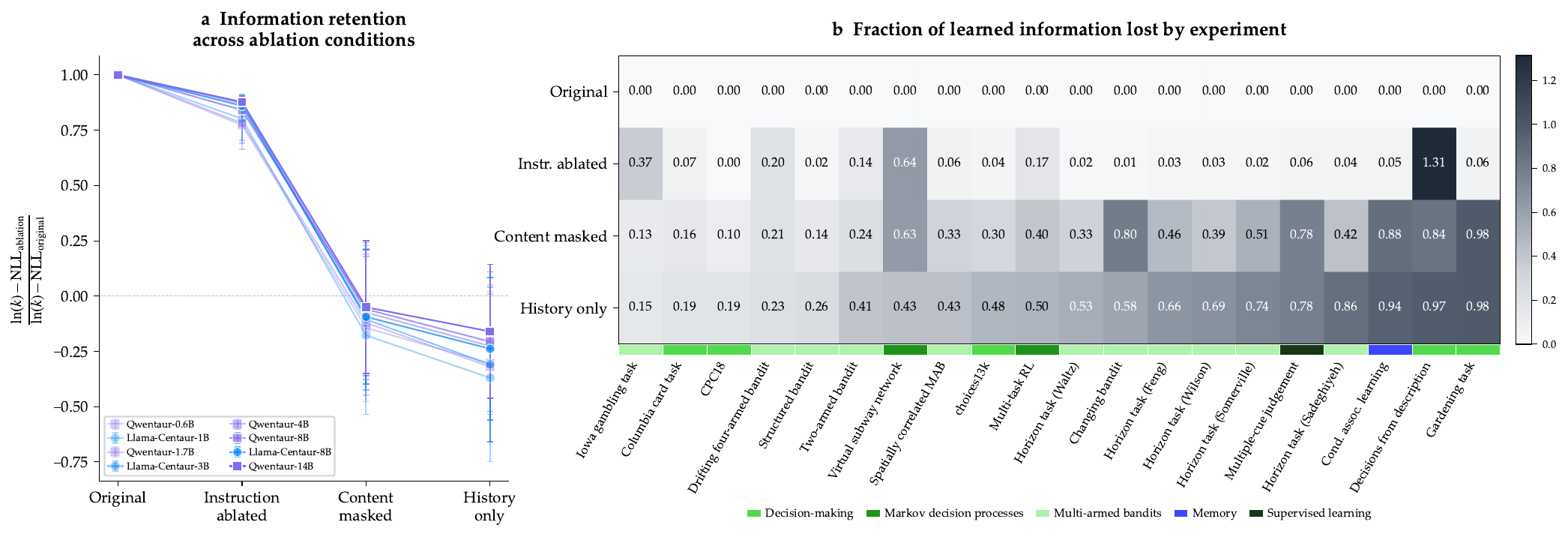}
\caption{%
  \textbf{Structural ablation results across eight models and 27 experiments with well-defined response options.} Five experiments with mixed or continuous formats are excluded. \textbf{(a)}~Mean information retention $R = (\ln k - \mathcal{L}c)/(\ln k - \mathcal{L}{\mathrm{orig}})$ by condition, averaged over experiments; $R=1$: no loss, $R=0$: chance. Circles: \texttt{Llama-Centaur}; squares: \texttt{Qwentaur}; colour intensity scales with model size; error bars: SEM. \textbf{(b)}~Per-experiment fraction of learned information lost, $\delta = 1 - R$, averaged across models. Columns sorted by history-only $\delta$; colour strip indicates task type. Experiments with $\delta > 1$ under the history-only condition are omitted; see Figure~\ref{fig:structural_ablation_app} (Appendix~\ref{app:structural_ablation}) for the complete set.
}
\label{fig:structural_ablation}
\end{figure}

We measure each condition by information retention $R = (\ln k - \mathrm{NLL}_{\mathrm{ablation}})/(\ln k - \mathrm{NLL}_{\mathrm{original}})$, where $\ln k$ is the chance-level cross-entropy for $k$ response options, so $R=1$ when nothing is lost and $R=0$ at chance. Retention falls at every step (Figure~\ref{fig:structural_ablation}a), from original ($1.00$) to instruction-ablated ($0.84$), content-masked ($-0.12$) and history-only ($-0.26$), and each drop is significant (Friedman $\chi^2(3)=63.76$, $p=9.3\times10^{-14}$; pairwise Wilcoxon $p<2\times10^{-3}$ throughout) and holds for all eight models. Accordingly, removing the instruction costs little, whereas masking the content of stimuli and feedback costs almost everything and pushes models below chance. Neither form of the shortcut hypothesis survives; \textit{choice history alone does not suffice, and neither does the trial template}.

Partitioning the total loss across the three steps, the instruction accounts for $12.5\%$, the content of stimuli and feedback for $75.7\%$, and the template for the remaining $11.7\%$. What the models use is what the participant saw and was told. This follows two earlier findings. The small cost of removing instructions reported by \citet{liu2025can} replicates here, but does not extend to the other channels; and the accuracy \citet{xie2025centaur} recover from choice history alone holds on the two sequential tasks they examined, both of which are in our set, but not on average, since mean retention under history-only is below chance across the 27 experiments.

\begin{figure}[t]
    \centering
    \includegraphics[width=\textwidth]{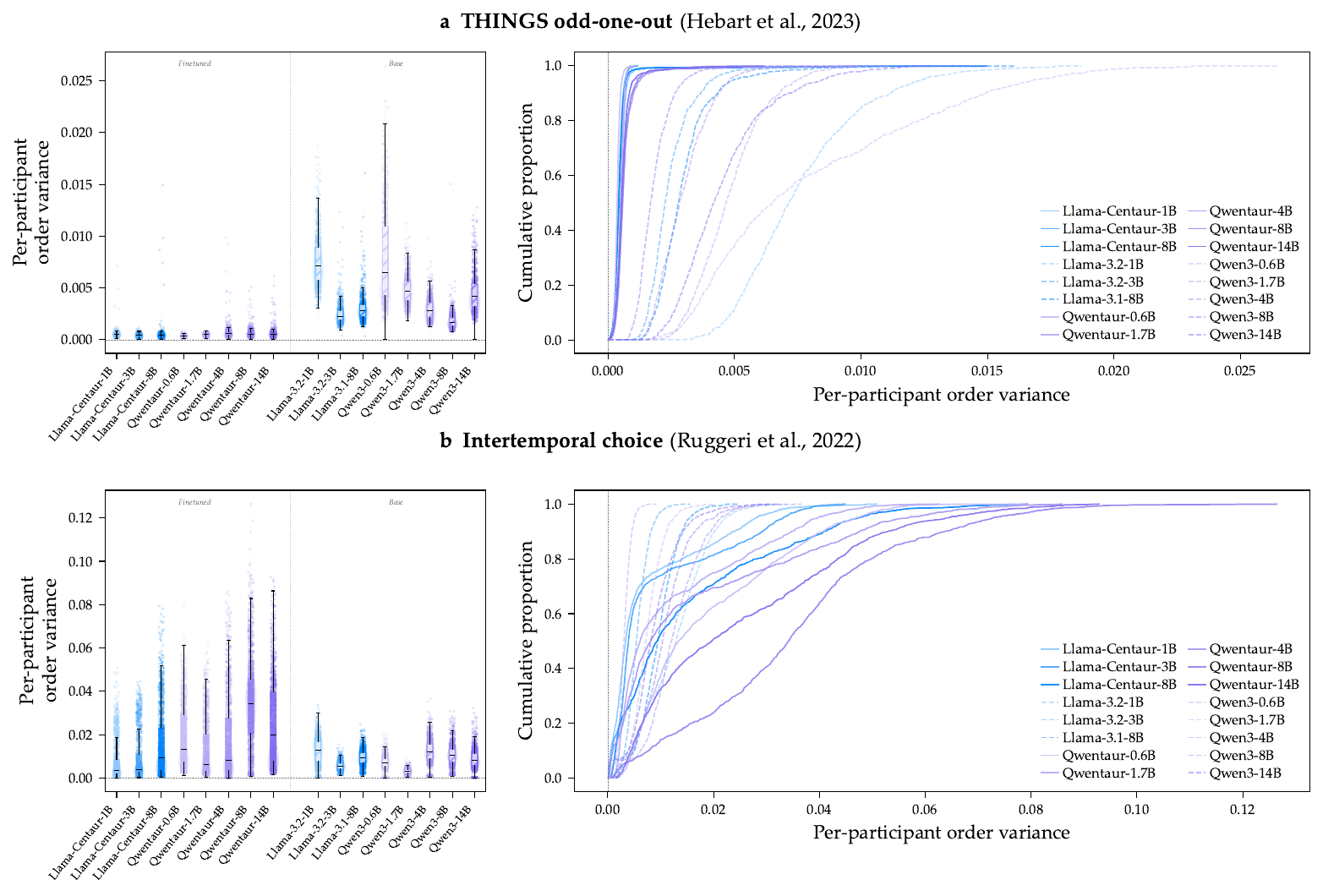}
    \caption{%
    \textbf{Per-participant order variance across models and experiments.} \textbf{(a)}~THINGS odd-one-out (exchangeable; primary test). \textbf{(b)}~Intertemporal choice (adaptive staircase; negative control). \textbf{Left:} violin and box plots; fine-tuned (solid) vs.\ base (hatched); $y$-axis scales differ. \textbf{Right:} ECDFs; solid = fine-tuned, dashed = base; steeper curves near zero indicate greater order invariance.
    \label{fig:order_permutation}
    }
\end{figure}

At the per-task level, 18 of 27 tasks (67\%) degrade strictly monotonically across the four conditions (Figure~\ref{fig:structural_ablation}b). Of the nine exceptions, five invert only between two conditions that are already at or below chance and four are genuine reversals (further analysis in Appendix~\ref{app:nonmonotonic}).

\subsubsection{Order permutation (exchangeability)}
\label{sec:order_permutation}


The ablations show which channels the model reads, not whether it reads them in a way the task requires. A model can depend heavily on stimuli and still treat the trials as an ordered sequence where their order carries no information \citep{namazova2025not, schlegelmilch2026snake}. Exchangeability makes this testable. Where trials are independent, permuting the context must leave the prediction unchanged; where each trial follows from the previous response, it must not. For each (participant, target trial) pair we draw 50 random permutations of the context trials and measure the variance of each response-token probability across them, low variance indicating order invariance.

Two experiments are chosen to anchor opposite ends of the exchangeability spectrum. \textbf{THINGS odd-one-out}~\citep{hebart2023things} (primary test; 768 participants) presents triplets of objects for similarity judgement. Triplets are sampled independently, with no feedback, no adaptation, and no sequential dependency. Because the order in which previous triplets appeared carries no information about the current judgement, context trials are exchangeable by construction. A cognitive proxy that has genuinely learned this task structure, rather than defaulting to sequential heuristics, should produce order-invariant predictions. \textbf{Intertemporal choice}~\citep{ruggeri2022globalizability} (negative control; 1,295 participants) uses an adaptive, response-contingent procedure to locate each participant's indifference point, where the value offered on each trial depends on the previous response. Permuting trial order breaks the staircase logic entirely, so a faithful behavioural proxy should be order-\textit{sensitive}.

Figure~\ref{fig:order_permutation} matches both expectations. On THINGS odd-one-out, fine-tuned models cluster near zero variance, so they have learned the exchangeability of independent similarity judgements. On intertemporal choice, fine-tuned models are order-sensitive, as the design requires, since each offer is generated from the answer before it and permutation then yields a sequence the procedure itself could never have produced.

Together with the structural ablation results, the picture that emerges is one of \textit{task-adaptive information use}. Cognitive fine-tuning produces proxies that exploit sequential regularities where they are genuinely predictive, process stimuli where stimuli are informative, and respect exchangeability where the experimental design requires it. The critique that these models are merely exploiting statistical regularities is, in one sense, correct, but the regularities they exploit are precisely the ones that reflect the computational structure of the underlying cognitive tasks.

\section{Discussion}
\label{sec:discussion}

Our results bound what these models are for. Scale is not what limits prediction within the paradigms a model was trained on, but it is what limits transfer beyond them, and the ablations show that the predictions rest on the content of what participants were shown and told, which licenses using the models as instruments. That licence also rests on the gain being specific to cognitive data rather than to fine-tuning as such, which the non-cognitive and behavioural controls confirm (Appendix~\ref{app:specificity}). 

Yet predicting held-out choices, however accurately, is not explaining the processes that generate them. In the tradition of \citet{newell1990unified} and \citet{simon1969sciences}, a cognitive model posits explicit mechanisms within a decomposable architecture, and reproducing behavioural regularities alone is what Simon called \textit{explanation from the outside} \citep{simon1969sciences, SimonNewell1971, newell1990unified, simon2000expertise}. Unlike \texttt{ACT-R} \citep{anderson1998atomic} or \texttt{Soar} \citep{laird2012soar}, these models commit to no architecture and no mechanism, and that absence is what makes them general, since they cover many paradigms by relying on scale and learned statistics rather than designed-in assumptions.

What we claim instead is the role of a \textit{noise ceiling estimator} for in-distribution paradigms \citep{agrawal2020scaling, kuperwajs2023using}, which is more substantive than the phrase suggests. Estimated per datapoint, a noise ceiling maps where systematic structure in behaviour exists and where it does not, so the gap between what the model captures and what an interpretable theory captures marks where behaviour is more structured than current theory can account for, and therefore where theory-building has most to gain. A model used only for prediction returns a number; used as a ceiling it returns a map of where an existing theory falls short, which is the logic \citet{agrawal2020scaling} call scientific regret minimisation. It is in this sense, and not as a claim to mechanism, that we read our title's reference to cognition \textit{and} behaviour, since to call the work behaviour alone would understate the stakes and to call it a theory of cognition would overstate them.

Whether fine-tuning moves these models any distance inward is a question for mechanistic and representational analysis, and one worth asking, since networks trained on behaviour have recovered known architectures of memory \citep{salvatore2025sequence, ji2024linking}, prefrontal-like specialisation \citep{yang2019task}, and planning mechanisms with neural counterparts \citep{jensen2024recurrent}.

\section{Limitations and future work}
\label{sec:limitation}

\textbf{Data structure.} Two features of Psych-101 bound what we can conclude from it. The first concerns which axis of the data we varied. Our subsets are drawn at the level of whole participant sessions and stratified so that all 160 experiments contribute to every fraction, which isolates data quantity but holds paradigm coverage fixed by construction. The diminishing returns we report are therefore returns to depth, more participants per experiment, and the return to breadth is unmeasured. Given that the models saturate within trained paradigms and remain far from saturated outside them, breadth is likely the more informative axis, and subsetting by experiment rather than by participant would estimate how many distinct paradigms a cognitive foundation model needs. Psych-201 \citep{binz2026posttrainingmakeslargelanguage} is a promising step in that direction, $3.5$ times the size of Psych-101, but it does not escape the second limitation, which is structural. Participants are nested within experiments in Psych-201 as in Psych-101, so no individual is observed across tasks, which precludes the cross-experiment person-task decomposition that \citet{gobet2000individual} and \citet{orr2025not} identify as necessary for bounding predictability. Datasets in which the same participants complete multiple paradigms would address both limitations at once.

\textbf{Training and architecture.} Adaptation here is low-rank throughout. The rank sweep shows that adapter capacity substitutes for parameters over the range we test, but it does not tell us where either axis ends. Full fine-tuning at the smallest scales would separate a genuine ceiling on the learnable signal from one imposed by low-rank adaptation, and the data ablation shows returns already diminishing, so the three may exhaust at different points. Architecture coverage is narrow in a further sense, since all four families are dense decoder-only transformers, leaving open whether the plateau shifts under mixture-of-experts \citep{jiang2024mixtral}, state-space \citep{gu2024mamba}, or hybrid designs \citep{nvidia2025nemotronh} that allocate capacity differently. Finally, SFT optimises for imitation of the training distribution, at a measurable cost to capabilities distant from it (Appendix~\ref{app:retention}). \citet{kolluri2025finetuning} show that contrastive DPO on demographic preference pairs improves individual response accuracy over SFT, while \citet{zhu2025using} find that GRPO elicits chain-of-thought reasoning that in turn improves choice prediction. RL is unlikely to surface genuinely novel cognitive mechanisms, since it redistributes probability mass over behaviours already represented in the pre-trained weights, but it generalises where SFT memorises \citep{chu2025sft}, which is the regime our Psych-201 results identify as the harder one. Pairing RL with cross-task datasets would allow fine-tuning toward person-level behavioural profiles rather than task-level distributions.

\section{Conclusion}
\label{sec:conclusion}

Seventy billion parameters are not what \texttt{Centaur}'s accuracy on Psych-101 rests on. Across fourteen models from $135$M to $14$B, whether a model has been fine-tuned matters far more than how large it is. Only beyond the training paradigms does scale begin to earn its cost, and there larger models transfer better with no sign of saturation. Accuracy alone, however, would settle nothing; an instrument must also be right for the right reasons. Prediction collapses when the content of stimuli and feedback is masked, and responds to trial order only where the design makes order informative. Small cognitively fine-tuned models are therefore adequate for estimating noise ceilings within the paradigms they have seen, and what binds them is the reach of that coverage rather than their capacity.


\section*{Acknowledgements}

We thank Marcel Binz for answering our questions about Psych-201 and the \texttt{Centaur} evaluation setup, and Seungbin Joo for insightful discussions on the different forms of the shortcut hypothesis.

\section*{Ethics Statement}

All models are trained on Psych-101 \citep{binz2025foundation}, a publicly released dataset of anonymised trial-level responses aggregated from published psychological experiments. We collected no new human data and made no attempts to re-identify participants. Our models predict laboratory task responses and are intended as scientific instruments. We acknowledge that models capable of predicting human choices could in principle be misused; however, the training domain (abstract laboratory paradigms) limit this risk. We release all code and model weights to support reproducibility and scientific progress, and encourage users to employ these models within the scientific context for which they were developed.

However, their utility as scientific instruments is currently bounded by the paradigm coverage of the training data, and whether they extrapolate reliably to conditions producing more extreme outcomes than any in Psych-101 remains untested \citep{schlegelmilch2026snake}. More speculatively, \citet{dalessandro2025ai} caution that widespread reliance on cheap AI-generated behavioural data could induce a ``street light effect'' in the cognitive sciences, with researchers gravitating toward questions that fall within these models' capabilities and neglecting phenomena that resist efficient simulation.

\section*{Reproducibility Statement}

All code for training, evaluation, and visualisation is publicly available at \url{https://github.com/socius-org/Centauri}. This includes the training scripts for all four model families, the evaluation pipelines, and all figure-generation code. Fine-tuned LoRA adapters are released on Hugging Face at \url{https://hf.co/collections/socius/Centauri}. The training data, Psych-101, is publicly available through \citet{binz2025foundation}, and we use the same train–test splits and prompt templates without modification. All models were trained on a single NVIDIA A100 80GB GPU using identical hyperparameters (Table~\ref{tab:training_hyperparams}): rank-stabilised LoRA applied to all linear layers at $r = \alpha \in \{4, 8, 16, 32, 64\}$, with $r{=}16$ as the reference configuration, AdamW with learning rate $5 \times 10^{-5}$, weight decay 0.01, linear warmup of 100 steps, and one epoch of training in bf16 precision. The structural ablation and order permutation experiments use deterministic prompt manipulation scripts included in the GitHub repository; permutation tests use 50 random permutations per (participant, target trial) pair. Appendix~\ref{app:cogfm_comparison} provides full infrastructure details including training times and batch configurations.



\newpage
\bibliography{colm2026_conference}
\bibliographystyle{colm2026_conference}

\clearpage
\appendix
\raggedbottom
\section{Cognitive and behavioural foundation models}
\label{app:cogfm_comparison}

Several recent works have fine-tuned or prompted large language models to predict human behaviour in cognitive tasks. Tables~\ref{tab:cogfm_comparison}--\ref{tab:cogfm_infrastructure} situate our models within this emerging landscape. Table~\ref{tab:cogfm_comparison} compares each model's base LLM, post-training method, training data, and target domain. Table~\ref{tab:training_hyperparams} reports training hyperparameters -- adaptation strategy, optimiser, learning rate, batch configuration, and scheduler -- to facilitate direct methodological comparison and reproducibility. Table~\ref{tab:cogfm_data_pipeline} details the data pipeline and loss configuration, including numerical precision, loss masking strategy, input format, and any use of LLM-assisted data synthesis. Table~\ref{tab:cogfm_infrastructure} records the computational infrastructure, training time, and inference settings for each model.

\begin{table}[H]
\centering
\setlength{\tabcolsep}{2.5pt}
\scriptsize
\renewcommand{\arraystretch}{1.25}
\begin{tabular}{
    @{}
    p{2cm}
    >{\raggedright\arraybackslash}p{3.5cm}
    >{\raggedright\arraybackslash}p{3cm}
    >{\raggedright\arraybackslash}p{2.5cm}
    >{\raggedright\arraybackslash}p{2cm}
    @{}
}
\toprule
\textbf{Model} & \textbf{Base LLM} & \textbf{Post-training} & \textbf{Training data} & \textbf{Domain} \\
\midrule

\texttt{Centaur} \newline {\tiny\citep{binz2025foundation}}
    & {\tiny\texttt{Llama-3.1-70B}}
    & SFT (masked CE on response tokens); rank-stabilised QLoRA ($r{=}\alpha{=}8$), 4-bit quantised
    & \texttt{Psych-101}: 160 expts, 60{,}092 partic., 10.7M choices
    & Decision, memory, learning, planning \\

{\tiny\citet{zhu2025using}}\textsuperscript{1}
    & {\tiny\texttt{Qwen2.5-7B-Instruct}}
    & SFT / Centaur-style SFT / GRPO compared; LoRA ($r{=}\alpha{=}32$)
    & \texttt{choices13k}: 13{,}102 train / 1{,}462 test risky choice problems
    & Decision \\

\texttt{Be.FM} \newline {\tiny\citep{xie2025fm}}
    & {\tiny\texttt{Llama-3.1-\{8,70\}B-Instruct}}
    & SFT with LoRA (all layers); 8-bit quantised\textsuperscript{2}
    & \texttt{AER}: 2{,}703 papers.; \texttt{MobLab}: 68{,}779 subj., 82{,}057 obs.; \texttt{Big Five}: 17{,}667 subj.
    & Behavioural science, economic game, personality \\

\texttt{Socrates} \newline {\tiny\citep{kolluri2025finetuning}}
    & {\tiny\texttt{Llama-3-8B-Instruct}} \newline {\tiny\texttt{Qwen2.5-14B-Instruct}}
    & SFT / SFT + oracle reasoning traces / contrastive DPO compared; full fine-tuning
    & \texttt{SocSci210}: 210 TESS\textsuperscript{3} expts, 400{,}491 partic., 2.9M individual responses
    & Social sciences (economics, psychology, political science) \\

\texttt{HumanLLM} \newline {\tiny\citep{lei2026humanllm}}
    & {\tiny\texttt{Qwen2.5-\{3,7\}B-Instruct}} \newline {\tiny\texttt{Qwen3-8B}}\textsuperscript{4} \newline {\tiny\texttt{Llama-3.1-8B-Instruct}} \newline {\tiny\texttt{Phi-3-mini-128k-instruct}}
    & SFT (masked non-response tokens); full fine-tuning; 1:1 weight merge with base (LM-Cocktail)
    & \texttt{Cognitive Genome}: Reddit 2.8M, Twitter 673K, Blogger 368K, Amazon 1.7M; 1.2M train samples
    & Social intelligence (personalised behaviour) \\

\texttt{GeCCo} \newline {\tiny\citep{rmus2025generating}}
    & {\tiny\texttt{Llama-3.1-70B-Instruct}} \newline {\tiny\texttt{DeepSeek-R1-Distill-Llama-3.1-70B}} \newline {\tiny\texttt{Qwen2.5-72B-Instruct}}
    & No fine-tuning; in-context learning with iterative BIC-based refinement ($10 \times 5$ runs)
    & Behavioural data from 4 cognitive domains (in-context, not for training)
    & Decision, learning, planning, working memory \\

\midrule

\texttt{\textbf{Ours}\newline\tiny\texttt{Llama-Centaur}\newline\tiny\texttt{Qwentaur}\newline\tiny\texttt{Smoltaur}\newline\tiny\texttt{Olmotaur}}
    & {\tiny\texttt{Llama-3.2-\{1,3\}B}} \newline {\tiny\texttt{Llama-3.1-8B}} \newline {\tiny\texttt{Qwen3-\{0.6,1.7,4,8,14\}B-Base}} \newline {\tiny\texttt{SmolLM2-\{135,360\}M}} \newline {\tiny\texttt{SmolLM2-1.7B}} \newline {\tiny\texttt{SmolLM3-3B-Base}} \newline {\tiny\texttt{OLMo-2-0425-1B}} \newline {\tiny\texttt{OLMo-3-1025-7B}}
    & SFT (masked CE on response tokens, Centaur-style); rank-stabilised LoRA ($r{=}\alpha \in \{4,8,16,32,64\}$)
    & \texttt{Psych-101}: 160 expts, 60{,}092 partic., 10.7M choices
    & Decision, memory, learning, planning \\

\midrule
\multicolumn{5}{@{}p{13.5cm}@{}}{\tiny
    \textsuperscript{1}No named model; methods (SFT v. RL) comparison only.\;
    \textsuperscript{2}8-bit quantisation applies to 70B variant only; 8B is unquantised.\;
    \textsuperscript{3}TESS: NSF's Time-sharing Experiments for the Social Sciences, a repository of peer-reviewed social science experiments conducted on nationally representative samples.\;
    \textsuperscript{4}\citet{lei2026humanllm} states ``Qwen3-8B'' without Instruct suffix; base/instruct status unspecified.
} \\
\bottomrule
\end{tabular}
\caption{%
    \textbf{Core comparison of LLM-based cognitive and behavioural foundation models.}
}
\label{tab:cogfm_comparison}
\end{table}


\begin{table}[H]
\centering
\setlength{\tabcolsep}{4pt}
\scriptsize
\renewcommand{\arraystretch}{1.25}
\begin{tabular}{
    @{}
    p{2cm}
    >{\raggedright\arraybackslash}p{2.0cm}
    >{\raggedright\arraybackslash}p{1.5cm}
    >{\raggedright\arraybackslash}p{1cm}
    >{\raggedright\arraybackslash}p{1.5cm}
    >{\raggedright\arraybackslash}p{2cm}
    >{\raggedright\arraybackslash}p{1.5cm}
    @{}
}
\toprule
\textbf{Model} & \textbf{Adaptation} & \textbf{Optimiser} & \textbf{Epochs} & \textbf{Learning rate} & \textbf{Eff.\ batch size} \newline {\tiny(PD${\times}$GA${\times}d$)} & \textbf{Scheduler} \\
\midrule

\texttt{Centaur}\newline\tiny\citep{binz2025foundation}
    & QLoRA ($r{=}\alpha{=}8$), all linear layers
    & 8-bit AdamW
    & 1\textsuperscript{1}
    & $5{\times}10^{-5}$
    & $1{\times}32{\times}1$
    & Cosine (WU 100 steps) \\

{\tiny\citet{zhu2025using}}
    & LoRA $r{=}\alpha{=}32$, all linear layers, dropout 0.05
    & AdamW
    & SFT: 6; RL: 3
    & SFT: $10^{-5}$;\newline RL: $3{\times}10^{-6}$
    & SFT: PD${\times}8{\times}1$;\newline RL: PD${\times}8{\times}4$
    & SFT: fixed; RL: cosine \\

\texttt{Be.FM}\newline\tiny\citep{xie2025fm}
    & LoRA, all layers
    & Not specified
    & 3
    & $10^{-4}$
    & $1{\times}8{\times}d$
    & Cosine (WU 0.1) \\

\texttt{Socrates}\newline\tiny\citep{kolluri2025finetuning}
    & Full fine-tuning (no LoRA)
    & Not specified
    & 1
    & SFT: $10^{-5}$; DPO: $10^{-6}$
    & PD${\times}$GA${\times}8{=}256$
    & Cosine (WU 0.05) \\

\texttt{HumanLLM}\newline\tiny\citep{lei2026humanllm}
    & Full fine-tuning (no LoRA)
    & Not specified
    & 3
    & $5{\times}10^{-6}$
    & PD${\times}$GA${\times}8{=}64$
    & Cosine (WU 0.5) \\

\texttt{GeCCo}\newline\tiny\citep{rmus2025generating}
    & N/A (no training)\textsuperscript{2}
    & N/A
    & N/A
    & N/A
    & N/A
    & N/A \\

\midrule

\texttt{Qwentaur-0.6B}
    & LoRA ($r{=}\alpha \in \{4,8,16,32,64\}$), all linear layers
    & AdamW
    & 1
    & $5{\times}10^{-5}$
    & $2{\times}16{\times}1$
    & Linear warmup (100 steps) \\
\texttt{Qwentaur-1.7B}
    & LoRA ($r{=}\alpha \in \{8,16\}$), all linear layers
    & AdamW
    & 1
    & $5{\times}10^{-5}$
    & $1{\times}32{\times}1$
    & Linear warmup (100 steps) \\
\texttt{Qwentaur-4B}
    & LoRA ($r{=}\alpha \in \{4,8,16,32,64\}$), all linear layers
    & AdamW
    & 1
    & $5{\times}10^{-5}$
    & $1{\times}32{\times}1$
    & Linear warmup (100 steps) \\
\texttt{Qwentaur-8B}
    & LoRA ($r{=}\alpha \in \{4,8,16,32,64\}$), all linear layers
    & AdamW
    & 1
    & $5{\times}10^{-5}$
    & $1{\times}32{\times}1$
    & Linear warmup (100 steps) \\
\texttt{Qwentaur-14B}
    & LoRA ($r{=}\alpha \in \{4,16,64\}$), all linear layers
    & AdamW
    & 1
    & $5{\times}10^{-5}$
    & $1{\times}32{\times}1$
    & Linear warmup (100 steps) \\
\texttt{Llama-Centaur-1B}
    & LoRA ($r{=}\alpha \in \{4,8,16,32,64\}$), all linear layers
    & AdamW
    & 1
    & $5{\times}10^{-5}$
    & $2{\times}16{\times}1$
    & Linear warmup (100 steps) \\
\texttt{Llama-Centaur-3B}
    & LoRA ($r{=}\alpha \in \{4,8,16,32,64\}$), all linear layers
    & AdamW
    & 1
    & $5{\times}10^{-5}$
    & $1{\times}32{\times}1$
    & Linear warmup (100 steps) \\
\texttt{Llama-Centaur-8B}
    & LoRA ($r{=}\alpha \in \{4,8,16,32,64\}$), all linear layers
    & AdamW
    & 1
    & $5{\times}10^{-5}$
    & $1{\times}32{\times}1$
    & Linear warmup (100 steps) \\
\texttt{Smoltaur-0.1B}
    & LoRA ($r{=}\alpha \in \{4,8,16,32,64\}$), all linear layers
    & AdamW
    & 1
    & $5{\times}10^{-5}$
    & $8{\times}4{\times}1$
    & Linear warmup (100 steps) \\
\texttt{Smoltaur-0.4B}
    & LoRA ($r{=}\alpha \in \{4,8,16,32,64\}$), all linear layers
    & AdamW
    & 1
    & $5{\times}10^{-5}$
    & $4{\times}8{\times}1$
    & Linear warmup (100 steps) \\
\texttt{Smoltaur-1.7B}
    & LoRA ($r{=}\alpha \in \{4,8,16,32,64\}$), all linear layers
    & AdamW
    & 1
    & $5{\times}10^{-5}$
    & $2{\times}16{\times}1$
    & Linear warmup (100 steps) \\
\texttt{Smoltaur-3B}
    & LoRA ($r{=}\alpha \in \{4,8,16,32,64\}$), all linear layers
    & AdamW
    & 1
    & $5{\times}10^{-5}$
    & $1{\times}32{\times}1$
    & Linear warmup (100 steps) \\
\texttt{Olmotaur-1B}
    & LoRA ($r{=}\alpha \in \{4,8,16,32,64\}$), all linear layers
    & AdamW
    & 1
    & $5{\times}10^{-5}$
    & $2{\times}16{\times}1$
    & Linear warmup (100 steps) \\
\texttt{Olmotaur-7B}
    & LoRA ($r{=}\alpha \in \{4,8,16,32,64\}$), all linear layers
    & AdamW
    & 1
    & $5{\times}10^{-5}$
    & $1{\times}32{\times}1$
    & Linear warmup (100 steps) \\

\midrule
\multicolumn{7}{@{}p{13cm}@{}}{\tiny
    \textit{Abbreviations}: PD = per-device batch size; GA = gradient accumulation steps; $d$ = number of GPU devices; WU = warmup ratio.\;\newline
    \textsuperscript{1}The published training script (\url{https://github.com/marcelbinz/Llama-3.1-Centaur-70B/blob/main/scripts/cluster_train.sh}) specifies 5 epochs. \citet{binz2025foundation} report training for 1 epoch, suggesting early checkpoint selection.\;\textsuperscript{2}\texttt{GeCCo} \citep{rmus2025generating} uses in-context learning with iterative BIC-based feedback over 10 sampling iterations $\times$ 5 independent runs.
    Models are fitted to held-out data with SciPy \texttt{minimize} (20 random restarts) and evaluated by BIC.\newline
    All post-training methods use supervised fine-tuning unless otherwise noted;
    \citet{zhu2025using} additionally compare RL (GRPO with 12 candidate completions per step, max 1024 tokens; reward $= 1 - |o_B - p_B|$ + format bonus up to 0.5, no standard-deviation normalisation);
    \texttt{Socrates} \citep{kolluri2025finetuning} additionally compares contrastive DPO (preference pairs constructed by varying the demographic persona under the same experimental condition and outcome question).
} \\
\bottomrule
\end{tabular}
\caption{%
    \textbf{Training hyperparameters for LLM-based cognitive and behavioural models.}
    Adaptation strategy refers to the parameter-efficient or full fine-tuning method applied to the base LLM.
    All our variants share identical hyperparameters except for per-device batch size, gradient accumulation steps, and the set of adapter ranks trained.
    Entries marked ``Not specified'' indicate that the original paper did not report the corresponding detail.
}
\label{tab:training_hyperparams}
\end{table}


\begin{table}[H]
\centering
\setlength{\tabcolsep}{2.5pt}
\scriptsize
\renewcommand{\arraystretch}{1.25}
\begin{tabular}{
    @{}
    p{2cm}
    >{\raggedright\arraybackslash}p{1.5cm}
    >{\raggedright\arraybackslash}p{1.0cm}
    >{\raggedright\arraybackslash}p{2cm}
    >{\raggedright\arraybackslash}p{2.5cm}
    >{\raggedright\arraybackslash}p{2.5cm}
    @{}
}
\toprule
\textbf{Model} & \textbf{Precision} & \textbf{Weight decay} & \textbf{Loss masking} & \textbf{Data format} & \textbf{Data synthesis} \\
\midrule

\texttt{Centaur}\newline\tiny\citep{binz2025foundation}
    & 4-bit NF4
    & 0.01
    & Human response tokens only
    & NL trial-by-trial prompts (${\sim}$32K tokens)
    & Template-based prompt construction\textsuperscript{1} \\

{\tiny\citet{zhu2025using}}
    & Not specified
    & Not specified
    & SFT: standard; Centaur-style. GRPO
    & JSON aggregated choice proportions per problem (empirical \% rounded to nearest integer, e.g.\ \texttt{\{"A": 29, "B": 71\}})
    & Reformatted from \texttt{choices13k} empirical choice frequencies (problem-level rather than individual-participant-level prediction) \\

\texttt{Be.FM}\newline\tiny\citep{xie2025fm}
    & 8-bit (70B base, bitsandbytes); bf16 (8B)
    & Not specified
    & Standard (Alpaca template)
    & Alpaca template \{instruction, input, output\}
    & \texttt{GPT-4o} for research workflow extraction \\

\texttt{Socrates}\newline\tiny\citep{kolluri2025finetuning}
    & Not specified
    & 0.1
    & SFT: Response token only. DPO
    & \{persona, stimuli, outcome, response\}
    & \texttt{o4-mini-high} (dataset agent\textsuperscript{2}); \newline\texttt{GPT-4o-mini} (reasoning traces\textsuperscript{3}) \\

\texttt{HumanLLM}\newline\tiny\citep{lei2026humanllm}
    & Not specified
    & Not specified
    & Non-response positions masked
    & ShareGPT format
    & \texttt{Llama-3.3-70B} (extraction); \texttt{GPT-4o} (quality validation) \\

\texttt{GeCCo}\newline\tiny\citep{rmus2025generating}
    & N/A
    & N/A
    & N/A
    & NL prompt + Python function template
    & N/A \\

\midrule

\texttt{\textbf{Ours}}
    & bf16
    & 0.01
    & Human response tokens only (Centaur-style)
    & NL trial-by-trial prompts (${\sim}$32K tokens)
    & Psych-101 dataset (unmodified) \\

\midrule
\multicolumn{6}{@{}p{13cm}@{}}{\tiny
    \textsuperscript{1}Each experiment is converted into natural-language trial-by-trial prompts via author-written scripts that map structured experimental data (participant responses, stimuli, feedback) to verbalised narratives; for an example, see \url{https://github.com/marcelbinz/Psych-201/blob/main/binz2022heuristics/generate_prompts.py}.\;
    \textsuperscript{2}An LLM-based agent (\texttt{o4-mini-high}) parses raw TESS datasets into structured \{persona, stimuli, response\} tuples; see Appendix~A of \citet{kolluri2025finetuning} for details.\;
    \textsuperscript{3}Given an experimental prompt and the corresponding human response, \texttt{GPT-4o-mini} generates reasoning traces explaining the human decision from a social scientist's perspective; see Appendix~E of \citet{kolluri2025finetuning}.
} \\
\bottomrule
\end{tabular}
\caption{%
    \textbf{Data pipeline and loss configuration for LLM-based cognitive and behavioural models.}
    Precision refers to the numerical format used during training: quantised formats (4-bit NF4, 8-bit) apply to base model weights while LoRA adapters and forward/backward computation use bf16.
    Loss masking describes which tokens contribute to the training loss; masked approaches restrict gradient updates to human response tokens, avoiding optimisation on task instructions and context.
    Data format describes the input representation seen by the model during training.
    Data synthesis indicates whether and how any computer-assisted tools were used to construct or augment the training corpus.
}
\label{tab:cogfm_data_pipeline}
\end{table}


\begin{table}[H]
\centering
\setlength{\tabcolsep}{2.5pt}
\scriptsize
\renewcommand{\arraystretch}{1.25}
\begin{tabular}{
    @{}
    p{2.5cm}
    >{\raggedright\arraybackslash}p{1.5cm}
    >{\raggedright\arraybackslash}p{2cm}
    >{\raggedright\arraybackslash}p{2cm}
    >{\raggedright\arraybackslash}p{1.5cm}
    >{\raggedright\arraybackslash}p{1.0cm}
    >{\raggedright\arraybackslash}p{1.5cm}
    @{}
}
\toprule
\textbf{Model} & \textbf{Training Framework} & \textbf{Key packages} & \textbf{Hardware} & \textbf{Train time} & \textbf{Max seq.\ len.} & \textbf{Inference} \\
\midrule

\texttt{Centaur}\newline\tiny\citep{binz2025foundation}
    & unsloth
    & unsloth
    & 1$\times$ A100 80GB
    & ${\sim}$5 days
    & ${\sim}$32{,}768
    & Not specified \\

{\tiny\citet{zhu2025using}}
    & Not specified
    & vLLM (inference)
    & RL: 4$\times$ H100;\newline SFT: 1$\times$ A100
    & RL: ${\sim}$80\,h; SFT: ${\sim}$5\,h
    & 1{,}024 (RL); 30 (SFT inf.)
    & $T{=}0.7$, top-$p{=}0.95$, top-$k{=}0.5$ \\

\texttt{Be.FM}\newline\tiny\citep{xie2025fm}
    & LlamaFactory
    & LlamaFactory, bitsandbytes
    & Not specified
    & Not specified
    & Not specified
    & Not specified \\

\texttt{Socrates}\newline\tiny\citep{kolluri2025finetuning}
    & LlamaFactory
    & LlamaFactory
    & 8$\times$ A100 80GB
    & 4--24\,h
    & 4{,}096 (inf.)
    & $T{=}0.6$, top-$p{=}0.9$ \\

\texttt{HumanLLM}\newline\tiny\citep{lei2026humanllm}
    & LlamaFactory
    & LlamaFactory, DeepSpeed Zero, vLLM
    & 8$\times$ A100 40GB
    & ${\sim}$120\,h (8B)
    & 8{,}192
    & $T{=}0.7$ \\

\texttt{GeCCo}\newline\tiny\citep{rmus2025generating}
    & N/A\textsuperscript{1}
    & SciPy (\texttt{minimize}); Python \texttt{exec()}
    & 4$\times$ A100 40GB
    & $\leq$8\,h per domain
    & N/A
    & $T$: 0.1--0.2\textsuperscript{2} \\

\midrule

\texttt{Qwentaur-0.6B}
    & unsloth
    & unsloth
    & 1$\times$ A100 80GB
    & 4\,h
    & ${\sim}$32{,}768
    & --- \\

\texttt{Qwentaur-1.7B}
    & unsloth
    & unsloth
    & 1$\times$ A100 80GB
    & 7\,h
    & ${\sim}$32{,}768
    & --- \\

\texttt{Qwentaur-4B}
    & unsloth
    & unsloth
    & 1$\times$ A100 80GB
    & 18\,h
    & ${\sim}$32{,}768
    & --- \\

\texttt{Qwentaur-8B}
    & unsloth
    & unsloth
    & 1$\times$ A100 80GB
    & 1\,d
    & ${\sim}$32{,}768
    & --- \\

\texttt{Qwentaur-14B}
    & unsloth
    & unsloth
    & 1$\times$ A100 80GB
    & 1\,d 15\,h
    & ${\sim}$32{,}768
    & --- \\

\texttt{Llama-Centaur-1B}
    & unsloth
    & unsloth
    & 1$\times$ A100 80GB
    & 7\,h
    & ${\sim}$32{,}768
    & --- \\

\texttt{Llama-Centaur-3B}
    & unsloth
    & unsloth
    & 1$\times$ A100 80GB
    & 12\,h
    & ${\sim}$32{,}768
    & --- \\

\texttt{Llama-Centaur-8B}
    & unsloth
    & unsloth
    & 1$\times$ A100 80GB
    & 21\,h
    & ${\sim}$32{,}768
    & --- \\

\texttt{Smoltaur-0.1B}
    & unsloth
    & unsloth
    & 1$\times$ A100 80GB
    & 3\,h
    & ${\sim}$8{,}192
    & --- \\

\texttt{Smoltaur-0.4B}
    & unsloth
    & unsloth
    & 1$\times$ A100 80GB
    & 4\,h
    & ${\sim}$8{,}192
    & --- \\

\texttt{Smoltaur-1.7B}
    & unsloth
    & unsloth
    & 1$\times$ A100 80GB
    & 7\,h
    & ${\sim}$8{,}192
    & --- \\

\texttt{Smoltaur-3B}
    & unsloth
    & unsloth
    & 1$\times$ A100 80GB
    & 1\,d 4\,h
    & ${\sim}$32{,}768
    & --- \\

\texttt{Olmotaur-1B}
    & unsloth
    & unsloth
    & 1$\times$ A100 80GB
    & 5\,h
    & ${\sim}$4{,}096
    & --- \\

\texttt{Olmotaur-7B}
    & unsloth
    & unsloth
    & 1$\times$ A100 80GB
    & 2\,d 1\,h
    & ${\sim}$32{,}768
    & --- \\

\midrule
\multicolumn{7}{@{}p{13.0cm}@{}}{\tiny
    \textit{Abbreviations}: $T$ = sampling temperature; inf.\ = inference; d = days; h = hours.\;
    \newline \textsuperscript{1}No training framework required; GeCCo generates cognitive models via in-context prompting.
    SciPy is used for parameter fitting of the generated models; Python \texttt{exec()} executes the LLM-generated code.\;
    \textsuperscript{2}Temperature varies by LLM: Llama 0.2, Qwen 0.15, DeepSeek-R1 0.1.\;\newline
    All our models were trained on a single NVIDIA A100 80GB GPU.
    Training times scale approximately linearly with parameter count within each model family.
} \\
\bottomrule
\end{tabular}
\caption{%
    \textbf{Infrastructure, packages, and computational cost for LLM-based cognitive and behavioural models.}
    Hardware refers to GPU resources used during training (or in-context generation for \texttt{GeCCo}).
    Training time reports wall-clock duration as stated in each paper; missing entries indicate the information was not reported.
    Our models were each trained for 1 epoch on Psych-101 on a single A100 80GB, with rank-stabilised LoRA at $r{=}\alpha \in \{4,8,16,32,64\}$.
}
\label{tab:cogfm_infrastructure}
\end{table}

\newpage

\section{Adapter rank and training-set size}
\label{app:rank_data}

\begin{figure}[H]
\centering
\includegraphics[width=\textwidth]{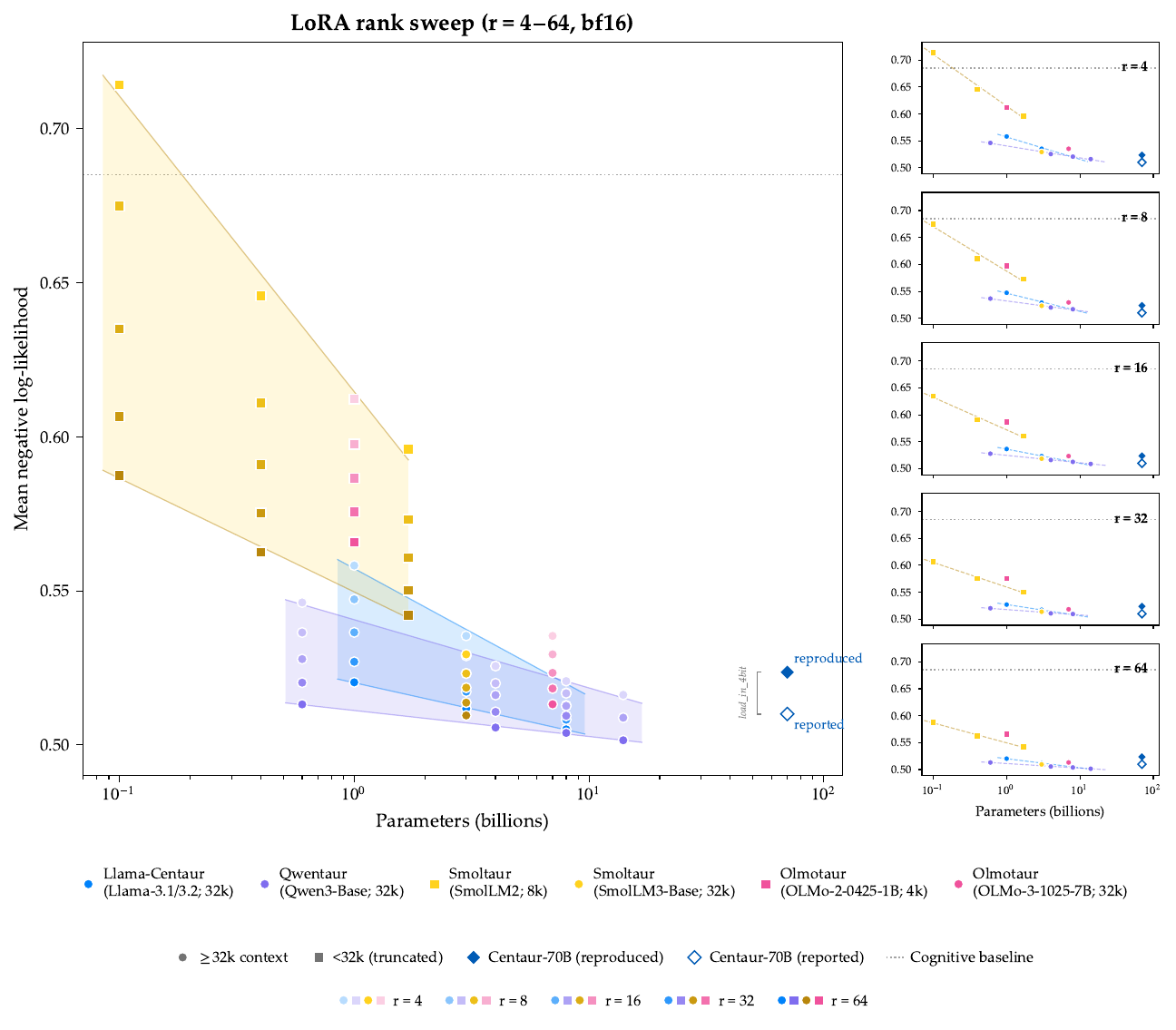}
\caption{%
\textbf{Adapter rank against model size on Psych-101.}
Mean negative log-likelihood over the 38 of 46 Psych-101 tasks for which \citet{binz2025foundation} publish a domain-specific cognitive model baseline, against parameter count, with colour intensity encoding LoRA rank and right-hand panels showing one rank at a time.
Trend lines and shaded envelopes are fitted only within groups matched on model generation and context window, so \texttt{Olmotaur} has none and the \texttt{Smoltaur} fit covers the \texttt{SmolLM2} models only.
Marker shape indicates context window (circle $\geq$ 32k, square $<$ 32k); colour indicates family.
}
\label{fig:rank_sweep_app}
\end{figure}

\begin{figure}[H]
\centering
\includegraphics[width=0.82\textwidth]{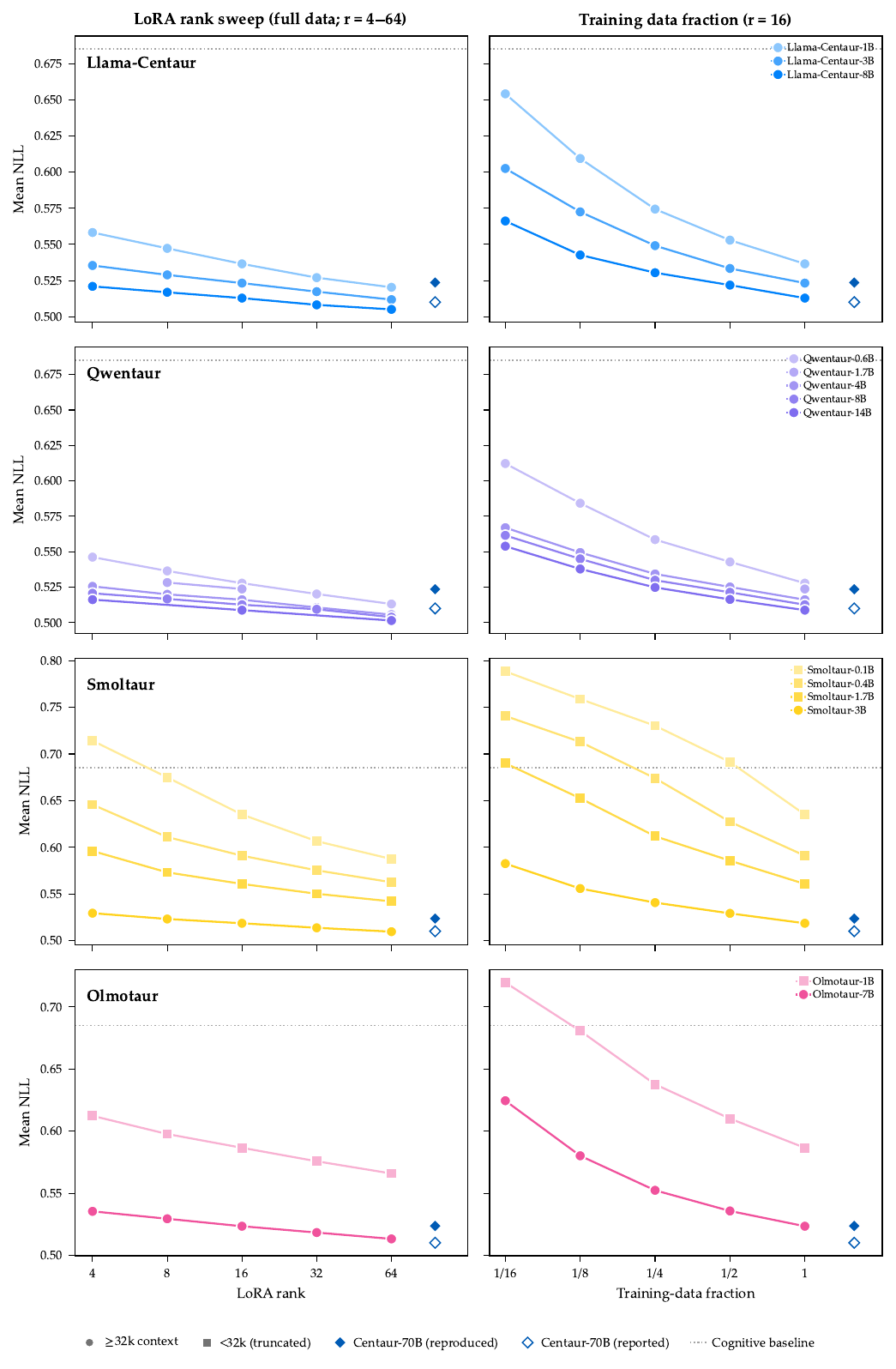}
\caption{%
\textbf{Adapter rank and training-set size by family.}
\textbf{(Left)}~Mean NLL against LoRA rank at full data.
\textbf{(Right)}~Mean NLL against the fraction of Psych-101 used for training, at $r{=}16$.
Rows are the four model families; the dotted line is the cognitive-model baseline and the diamonds are the reproduced and reported \texttt{Centaur-70B} values.
Subsets are nested and experiment-stratified, so every fraction covers all 160 experiments and reducing data quantity does not reduce paradigm coverage.
}
\label{fig:datasize_ablation}
\end{figure}

Adapter rank is swept over $r \in \{4, 8, 16, 32, 64\}$ at full data, and training-set size over nested, experiment-stratified subsets of Psych-101 ($1/16$, $1/8$, $1/4$, $1/2$, and the full set) at fixed rank $r{=}16$. Figure~\ref{fig:rank_sweep_app} shows the rank sweep across all fourteen models on a single axis, which makes the substitution between adapter capacity and parameter count visible directly: a small model at high rank overlaps a much larger model at low rank. Figure~\ref{fig:datasize_ablation} separates the two axes by family, with the rank sweep on the left and the data ablation on the right.

\section{Full per-experiment results}
\label{app:full_results}

\subsection{Psych-101 (in-distribution)}
\label{app:full_results_psych101}

This appendix presents the full per-experiment results on Psych-101. Table~\ref{tab:psych101_normalised} summarises performance by task type using the normalised metric $(\ln k - \mathrm{NLL})/\ln k$, restricted to the 34 experiments admitting both a cognitive model baseline and a well-defined chance level. Table~\ref{tab:psych101_pe_bf16_base_v_ft} then reports base versus fine-tuned NLL per experiment for the \texttt{Qwen3} and \texttt{Llama} families, isolating the effect of cognitive fine-tuning at each scale, and Table~\ref{tab:psych101_pe_ft_all} gives per-experiment NLL for all four fine-tuned families side by side. Table~\ref{tab:psych101_pe_control} compares our models against other behavioural foundation models (\texttt{Be.FM} \citep{xie2025fm} and \texttt{Socrates} \citep{kolluri2025finetuning}) and non-cognitive controls (\texttt{Nemotron} \citep{bercovich2025nemotron} and \texttt{Hermes} \citep{teknium2024hermes3technicalreport, teknium2025hermes4technicalreport}), grouped by base-model family. All tables include a rightmost column reporting the chance-level baseline $\ln(k)$, where $k$ is the number of discrete response options.

\paragraph{Experiment exclusions.} Of the 46 experiments in Psych-101, 12 are excluded from the normalised analysis reported in Table~\ref{tab:psych101_normalised}; their raw NLL values remain in all appendix tables. These exclusions fall into four categories.

\textit{Continuous or free-form responses.} Two experiments elicit numeric responses on a continuous scale, for which no discrete uniform baseline exists: the aversive learning task \citep{wise2019computational} (supervised learning) asks participants to type probability estimates, and the probabilistic reasoning task \citep{zhu2020bayesian} (miscellaneous) similarly requires free-form percentage judgements.

\textit{Mixed response types.} Seven experiments interleave multiple response formats within a single session, making it impossible to assign a single $k$: the episodic long-term memory task \citep{popov2023intent} (memory) combines binary size judgements with free word recall and arithmetic answers; the experiential-symbolic task \citep{garcia2023experiential} (decision-making) mixes binary letter choices with probability estimates; the recall and recognition task \citep{cox2018information} (memory) interleaves binary recognition with open-vocabulary cued recall; the grammar judgement task \citep{jansen2021rational} (miscellaneous) combines free numeric self-assessments with five-alternative grammar questions whose labels vary across items; the Medin categorisation task \citep{levering2020revisiting} (supervised learning) pairs binary categorisation with 9-point typicality ratings; the risky choice task \citep{krueger2024identifying} (decision-making) requires multi-stage responses including gamble selection from six options, colour queries, and a stop action; and the decisions from experience task \citep{wulff2018sampling} (multi-armed bandits) switches between a three-option sampling phase (two lotteries plus stop) and a two-option final choice.

\textit{Degenerate or trial-varying action spaces.} Two experiments have action spaces that prevent a meaningful $\ln(k)$ computation: the go/no-go task \citep{enkavi2019large} (decision-making) only tokenises go-responses, yielding $\ln(1) = 0$; and the tile-revealing task \citep{kumar2023disentangling} (Markov decision processes) uses grid coordinates on a $7 \times 7$ board where the set of valid positions shrinks as cells are revealed, making $k$ variable across trials.

\textit{State-dependent action space.} The virtual subway network task \citep{tomov2020discovery} (Markov decision processes) nominally offers five response tokens (four cardinal directions plus an end-round key), but walls in the navigation grid constrain which directions are valid on any given trial. We include this experiment in the appendix tables using $\ln(5) \approx 1.61$ as a nominal ceiling, but exclude it from the normalised analysis (Table~\ref{tab:psych101_normalised}) since the effective action space varies per state.

\clearpage

\begin{table}[t]
\centering
\resizebox{\columnwidth}{!}{%
\renewcommand{\arraystretch}{1.2}
\begin{tabular}{@{}l   r r r r r   r r r   r r r r   r r   r r r   r r r@{}}
\toprule
& \multicolumn{5}{c}{\textbf{\texttt{Qwentaur}}} & \multicolumn{3}{c}{\textbf{\texttt{Llama-Centaur}}} & \multicolumn{4}{c}{\textbf{\texttt{Smoltaur}}} & \multicolumn{2}{c}{\textbf{\texttt{Olmotaur}}} & \multicolumn{3}{c}{\textbf{Base}} & \multicolumn{3}{c}{\citet{binz2025foundation}} \\
\cmidrule(lr){2-6}\cmidrule(lr){7-9}\cmidrule(lr){10-13}\cmidrule(lr){14-15}\cmidrule(lr){16-18}\cmidrule(lr){19-21}
\textbf{Task type} & \texttt{0.6B} & \texttt{1.7B} & \texttt{4B} & \texttt{8B} & \texttt{14B} & \texttt{1B} & \texttt{3B} & \texttt{8B} & \texttt{0.1B} & \texttt{0.4B} & \texttt{1.7B} & \texttt{3B} & \texttt{1B} & \texttt{7B} & \texttt{Q-8B} & \texttt{Q-14B} & \texttt{L-8B} & \texttt{70B}$_p$ & \texttt{70B}$_r$ & Cog.$_p$ \\
\midrule
Decision (8) & \hcR{0.53} & \hcR{0.53} & \hcR{0.54} & \hcR{0.54} & \hcRbu{0.54} & \hcR{0.53} & \hcR{0.53} & \hcRu{0.54} & \hcR{0.41} & \hcR{0.46} & \hcR{0.48} & \hcR{0.54} & \hcR{0.47} & \hcR{0.53} & \hcR{0.34} & \hcR{0.35} & \hcR{0.33} & \hcR{0.56} & \hcR{0.52} & \hcR{0.27} \\
MDP (5) & \hcR{0.28} & \hcR{0.30} & \hcR{0.31} & \hcR{0.31} & \hcRbu{0.32} & \hcR{0.28} & \hcR{0.30} & \hcR{0.31} & \hcR{0.18} & \hcR{0.20} & \hcR{0.26} & \hcR{0.30} & \hcR{0.20} & \hcR{0.30} & \hcR{0.15} & \hcR{0.17} & \hcR{0.19} & \hcR{0.32} & \hcRu{0.31} & \hcR{0.11} \\
Bandit (12) & \hcR{0.48} & \hcR{0.48} & \hcR{0.49} & \hcRu{0.49} & \hcRbu{0.50} & \hcR{0.46} & \hcR{0.48} & \hcR{0.49} & \hcR{0.36} & \hcR{0.41} & \hcR{0.44} & \hcR{0.49} & \hcR{0.42} & \hcR{0.48} & \hcR{0.35} & \hcR{0.36} & \hcR{0.33} & \hcR{0.46} & \hcR{0.45} & \hcR{0.39} \\
Memory (4) & \hcR{0.57} & \hcR{0.57} & \hcR{0.58} & \hcR{0.58} & \hcRu{0.58} & \hcR{0.56} & \hcR{0.57} & \hcR{0.58} & \hcR{0.46} & \hcR{0.51} & \hcR{0.55} & \hcR{0.57} & \hcR{0.53} & \hcR{0.57} & \hcR{0.42} & \hcR{0.47} & \hcR{0.43} & \hcR{0.58} & \hcRbu{0.58} & \hcR{0.36} \\
Misc. (2) & \hcR{0.57} & \hcR{0.57} & \hcR{0.58} & \hcR{0.58} & \hcRbu{0.58} & \hcR{0.56} & \hcR{0.57} & \hcRu{0.58} & \hcR{0.45} & \hcR{0.53} & \hcR{0.55} & \hcR{0.58} & \hcR{0.55} & \hcR{0.57} & \hcR{0.41} & \hcR{0.42} & \hcR{0.40} & \hcR{0.57} & \hcR{0.57} & \hcR{0.55} \\
Sup.\ learn. (3) & \hcR{0.30} & \hcR{0.29} & \hcR{0.31} & \hcR{0.31} & \hcRu{0.31} & \hcR{0.29} & \hcR{0.31} & \hcR{0.31} & \hcR{0.18} & \hcR{0.23} & \hcR{0.25} & \hcR{0.30} & \hcR{0.20} & \hcRbu{0.31} & \hcR{0.24} & \hcR{0.24} & \hcR{0.23} & \hcR{0.30} & \hcR{0.30} & \hcR{0.11} \\
\midrule
\textbf{Mean (34)} & \hcR{0.46} & \hcR{0.46} & \hcR{0.47} & \hcR{0.48} & \hcRbu{0.48} & \hcR{0.45} & \hcR{0.47} & \hcRu{0.48} & \hcR{0.35} & \hcR{0.39} & \hcR{0.43} & \hcR{0.47} & \hcR{0.40} & \hcR{0.47} & \hcR{0.32} & \hcR{0.34} & \hcR{0.32} & \hcR{0.47} & \hcR{0.46} & \hcR{0.30} \\
\bottomrule
\end{tabular}%
}
\vspace{2pt}
{\tiny\centering
\colorbox{c9!35}{\strut\,} $<$0.05 ~
\colorbox{c8!40}{\strut\,} 0.05--0.12 ~
\colorbox{c7!45}{\strut\,} 0.12--0.18 ~
\colorbox{c6!50}{\strut\,} 0.18--0.25 ~
\colorbox{c5!55}{\strut\,} 0.25--0.32 ~
\colorbox{c4!45}{\strut\,} 0.32--0.40 ~
\colorbox{c3!40}{\strut\,} 0.40--0.48 ~
\colorbox{c2!40}{\strut\,} 0.48--0.55 ~
\colorbox{c1!45}{\strut\,} $\geq$0.55\par}
\caption{\textbf{Fraction of available information captured above chance, $(\ln k - \mathrm{NLL}) \,/\, \ln k$, by task type for finetuned models (bf16).} A value of 0 indicates chance-level performance; 1 indicates perfect prediction. Restricted to 34 experiments (of 46) with both a cognitive model baseline and a well-defined discrete response space ($\ln k > 0$); 12 experiments are excluded. Finetuned families: \texttt{Qwentaur}, \texttt{Llama-Centaur}, \texttt{Smoltaur}, \texttt{Olmotaur}. Base columns: \texttt{Q-8B}/\texttt{Q-14B} (Qwen3), \texttt{L-8B} (Llama-3.1). Subscript $_r$ denotes our reproducing evaluation of the original Centaur model distributed by \citet{binz2025foundation}, evaluated under identical python library and CUDA versions as our small foundation models for fair comparison. Subscript $_p$ denotes values published by \citet{binz2025foundation}. \texttt{70B}$_p$ is shown for reference but excluded from best/second-best marking, as it was evaluated under different software conditions. \underline{\textbf{Bold+underline}} marks the best model, \underline{underline} the second-best.}
\label{tab:psych101_normalised}
\end{table}

\begin{sidewaystable}[p]
\centering
\sbox{\ctrtblbox}{%
{\tiny
\setlength{\tabcolsep}{3pt}
\renewcommand{\arraystretch}{1.15}
\begin{tabular}{@{}l l  r@{\;\;}r  r@{\;\;}r  r@{\;\;}r  r@{\;\;}r  r@{\;\;}r  r@{\;\;}r  r@{\;\;}r  r@{\;\;}r  r  r  r  r  r@{}}
\toprule
& & \multicolumn{10}{c}{\texttt{\textbf{Qwen3}} family} & \multicolumn{6}{c}{\texttt{\textbf{Llama-3.1/3.2}} family} & \multicolumn{4}{c}{\citet{binz2025foundation}} & \textbf{Chance} \\
\cmidrule(lr){3-12}\cmidrule(lr){13-18}\cmidrule(lr){19-22}
& & \multicolumn{2}{c}{0.6B} & \multicolumn{2}{c}{1.7B} & \multicolumn{2}{c}{4B} & \multicolumn{2}{c}{8B} & \multicolumn{2}{c}{14B} & \multicolumn{2}{c}{1B} & \multicolumn{2}{c}{3B} & \multicolumn{2}{c}{8B} & & & & & \\
\cmidrule(lr){3-4} \cmidrule(lr){5-6} \cmidrule(lr){7-8} \cmidrule(lr){9-10} \cmidrule(lr){11-12} \cmidrule(lr){13-14} \cmidrule(lr){15-16} \cmidrule(lr){17-18}
\textbf{Experiment} & \textbf{Type} & {\fontsize{4}{5}\selectfont base} & {\fontsize{4}{5}\selectfont ft} & {\fontsize{4}{5}\selectfont base} & {\fontsize{4}{5}\selectfont ft} & {\fontsize{4}{5}\selectfont base} & {\fontsize{4}{5}\selectfont ft} & {\fontsize{4}{5}\selectfont base} & {\fontsize{4}{5}\selectfont ft} & {\fontsize{4}{5}\selectfont base} & {\fontsize{4}{5}\selectfont ft} & {\fontsize{4}{5}\selectfont base} & {\fontsize{4}{5}\selectfont ft} & {\fontsize{4}{5}\selectfont base} & {\fontsize{4}{5}\selectfont ft} & {\fontsize{4}{5}\selectfont base} & {\fontsize{4}{5}\selectfont ft} & L-70B$_p$ & C$_r$ & C$_p$ & Cog.$_p$ & $\ln(k)$ \\
\midrule
Balloon analog risk task {\fontsize{4}{5}\selectfont\citep{frey2017risk}} & Decision & \hc{0.09} & \hc{0.07} & \hc{0.09} & \hc{0.07} & \hc{0.08} & \hc{0.07} & \hc{0.08} & \hc{0.06} & \hc{0.08} & \hcu{0.06} & \hc{0.10} & \hc{0.07} & \hc{0.08} & \hc{0.07} & \hc{0.08} & \hcbu{0.06} & \hc{0.08} & \hc{0.07} & \hc{0.06} & \hc{0.09} & \hc{0.69} \\
CPC18 {\fontsize{4}{5}\selectfont\citep{plonsky2018when}} & Decision & \hc{0.41} & \hc{0.35} & \hc{0.40} & \hc{0.34} & \hc{0.40} & \hc{0.34} & \hc{0.40} & \hc{0.34} & \hc{0.40} & \hcbu{0.33} & \hc{0.43} & \hc{0.36} & \hc{0.41} & \hc{0.35} & \hc{0.39} & \hcu{0.34} & \hc{0.41} & \hc{0.35} & \hc{0.34} & \hc{0.66} & \hc{0.69} \\
Columbia card task {\fontsize{4}{5}\selectfont\citep{frey2017risk}} & Decision & \hc{0.27} & \hc{0.20} & \hc{0.26} & \hc{0.20} & \hc{0.26} & \hc{0.20} & \hc{0.25} & \hc{0.20} & \hc{0.25} & \hcu{0.20} & \hc{0.30} & \hc{0.21} & \hc{0.27} & \hc{0.20} & \hc{0.25} & \hcbu{0.20} & \hc{0.23} & \hc{0.21} & \hc{0.19} & \hc{0.26} & \hc{0.69} \\
Decisions from description {\fontsize{4}{5}\selectfont\citep{wulff2018sampling}} & Decision & \hc{1.08} & \hc{0.62} & \hc{1.05} & \hc{0.63} & \hc{1.01} & \hc{0.62} & \hc{0.99} & \hc{0.61} & \hc{1.01} & \hc{0.60} & \hc{1.55} & \hc{0.62} & \hc{1.20} & \hc{0.63} & \hc{0.96} & \hcu{0.60} & \hc{0.76} & \hcbu{0.59} & \hc{0.53} & \hc{0.61} & \hc{0.69} \\
Experiential-symbolic task {\fontsize{4}{5}\selectfont\citep{garcia2023experiential}} & Decision & \hc{0.76} & \hc{0.48} & \hc{0.68} & \hc{0.47} & \hc{0.67} & \hc{0.46} & \hc{0.64} & \hc{0.46} & \hc{0.57} & \hcu{0.46} & \hc{0.78} & \hc{0.47} & \hc{0.69} & \hc{0.47} & \hc{0.64} & \hc{0.46} & \hc{0.70} & \hcbu{0.46} & \hc{0.45} & \hcn & \hcn \\
Gardening task {\fontsize{4}{5}\selectfont\citep{flesch2018comparing}} & Decision & \hc{0.52} & \hc{0.39} & \hc{0.52} & \hc{0.39} & \hc{0.49} & \hc{0.38} & \hc{0.47} & \hc{0.38} & \hc{0.45} & \hcu{0.38} & \hc{0.81} & \hc{0.40} & \hc{0.64} & \hc{0.39} & \hc{0.48} & \hcbu{0.38} & \hc{0.50} & \hc{0.49} & \hc{0.38} & \hc{0.91} & \hc{0.69} \\
Go/no-go {\fontsize{4}{5}\selectfont\citep{enkavi2019large}} & Decision & \hc{0.01} & \hcbu{0.00} & \hc{0.01} & \hcbu{0.00} & \hc{0.00} & \hcbu{0.00} & \hc{0.01} & \hcbu{0.00} & \hc{0.00} & \hcbu{0.00} & \hc{0.01} & \hcbu{0.00} & \hc{0.01} & \hcbu{0.00} & \hcu{0.00} & \hcbu{0.00} & \hc{0.01} & \hcbu{0.00} & \hc{0.00} & \hc{0.08} & \hc{0.00} \\
Intertemporal choice {\fontsize{4}{5}\selectfont\citep{ruggeri2022globalizability}} & Decision & \hc{0.77} & \hc{0.44} & \hc{0.80} & \hc{0.44} & \hc{0.71} & \hcu{0.44} & \hc{0.69} & \hc{0.44} & \hc{0.70} & \hc{0.44} & \hc{0.85} & \hc{0.44} & \hc{0.75} & \hcu{0.44} & \hc{0.76} & \hc{0.44} & \hc{0.73} & \hcbu{0.44} & \hc{0.43} & \hc{0.66} & \hc{0.69} \\
Multi-attribute DM {\fontsize{4}{5}\selectfont\citep{hilbig2014generalized}} & Decision & \hc{0.49} & \hc{0.09} & \hc{0.34} & \hc{0.08} & \hc{0.20} & \hc{0.09} & \hc{0.19} & \hc{0.08} & \hc{0.19} & \hc{0.08} & \hc{0.63} & \hcu{0.08} & \hc{0.45} & \hc{0.08} & \hc{0.23} & \hc{0.08} & \hc{0.15} & \hcbu{0.06} & \hc{0.06} & \hc{0.19} & \hc{0.69} \\
Risky choice {\fontsize{4}{5}\selectfont\citep{krueger2024identifying}} & Decision & \hc{0.92} & \hc{0.48} & \hc{0.83} & \hc{0.47} & \hc{0.79} & \hc{0.44} & \hc{0.74} & \hc{0.43} & \hc{0.67} & \hcbu{0.42} & \hc{0.99} & \hc{0.49} & \hc{0.85} & \hc{0.46} & \hc{0.76} & \hcu{0.43} & \hc{0.65} & \hc{0.43} & \hc{0.43} & \hcn & \hcn \\
choices13k {\fontsize{4}{5}\selectfont\citep{peterson2021using}} & Decision & \hc{0.62} & \hc{0.44} & \hc{0.60} & \hc{0.44} & \hc{0.59} & \hc{0.43} & \hc{0.59} & \hc{0.43} & \hc{0.55} & \hcbu{0.43} & \hc{0.67} & \hc{0.44} & \hc{0.59} & \hc{0.44} & \hc{0.55} & \hcu{0.43} & \hc{0.53} & \hc{0.43} & \hc{0.43} & \hc{0.66} & \hc{0.69} \\
\midrule
Multi-task RL {\fontsize{4}{5}\selectfont\citep{tomov2021multitask}} & MDP & \hc{0.70} & \hc{0.60} & \hc{0.72} & \hc{0.59} & \hc{0.68} & \hc{0.58} & \hc{0.70} & \hc{0.58} & \hc{0.68} & \hcbu{0.57} & \hc{0.80} & \hc{0.60} & \hc{0.70} & \hc{0.59} & \hc{0.66} & \hc{0.57} & \hc{0.66} & \hcu{0.57} & \hc{0.57} & \hc{1.04} & \hc{1.10} \\
Tile-revealing task {\fontsize{4}{5}\selectfont\citep{kumar2023disentangling}} & MDP & \hc{2.95} & \hc{0.02} & \hc{2.81} & \hc{0.01} & \hc{2.67} & \hc{0.00} & \hc{2.73} & \hc{0.07} & \hc{2.60} & \hcu{0.00} & \hc{4.86} & \hc{0.02} & \hc{3.98} & \hc{0.00} & \hc{2.92} & \hcbu{0.00} & \hc{2.74} & \hc{2.29} & \hc{1.87} & \hcn & \hcn \\
Two-step task {\fontsize{4}{5}\selectfont\citep{kool2017cost}} & MDP & \hc{0.67} & \hc{0.54} & \hc{0.64} & \hc{0.53} & \hc{0.64} & \hcbu{0.53} & \hc{0.65} & \hcu{0.53} & \hc{0.63} & \hcbu{0.53} & \hc{0.67} & \hc{0.55} & \hc{0.62} & \hc{0.53} & \hc{0.61} & \hcbu{0.53} & \hc{0.61} & \hc{0.53} & \hcd{0.50} & \hcd{0.60} & \hc{0.69} \\
Two-step task {\fontsize{4}{5}\selectfont\citep{kool2016when}} & MDP & \hc{0.69} & \hc{0.51} & \hc{0.64} & \hc{0.50} & \hc{0.64} & \hc{0.49} & \hc{0.65} & \hc{0.49} & \hc{0.63} & \hcu{0.48} & \hc{0.67} & \hc{0.52} & \hc{0.62} & \hc{0.50} & \hc{0.61} & \hc{0.49} & \hc{0.61} & \hcbu{0.48} & \hcd{0.50} & \hcd{0.60} & \hc{0.69} \\
Two-step task {\fontsize{4}{5}\selectfont\citep{zorowitz2023data}} & MDP & \hc{0.62} & \hc{0.52} & \hc{0.59} & \hc{0.51} & \hc{0.60} & \hc{0.51} & \hc{0.61} & \hcbu{0.51} & \hc{0.59} & \hc{0.51} & \hc{0.64} & \hc{0.52} & \hc{0.60} & \hcu{0.51} & \hc{0.58} & \hc{0.51} & \hc{0.61} & \hc{0.51} & \hcd{0.50} & \hcd{0.60} & \hc{0.69} \\
Virtual subway network {\fontsize{4}{5}\selectfont\citep{tomov2020discovery}} & MDP & \hc{1.68} & \hc{1.16} & \hc{1.55} & \hc{1.15} & \hc{1.57} & \hc{1.14} & \hc{1.55} & \hcu{1.13} & \hc{1.57} & \hcbu{1.12} & \hc{1.68} & \hc{1.19} & \hc{1.60} & \hc{1.16} & \hc{1.53} & \hc{1.13} & \hc{1.53} & \hc{1.16} & \hc{1.13} & \hcn & \hc{1.61} \\
Zoopermarket {\fontsize{4}{5}\selectfont\citep{ludwig2023human}} & MDP & \hc{0.61} & \hc{0.54} & \hc{0.61} & \hc{0.52} & \hc{0.60} & \hc{0.51} & \hc{0.60} & \hc{0.50} & \hc{0.59} & \hcbu{0.48} & \hc{0.64} & \hc{0.54} & \hc{0.62} & \hc{0.52} & \hc{0.60} & \hc{0.50} & \hc{0.60} & \hcu{0.50} & \hc{0.48} & \hc{0.60} & \hc{0.69} \\
\midrule
Cond.\ assoc.\ learning {\fontsize{4}{5}\selectfont\citep{collins2014working}} & Memory & \hc{0.76} & \hc{0.54} & \hc{0.71} & \hc{0.53} & \hc{0.71} & \hc{0.52} & \hc{0.69} & \hc{0.52} & \hc{0.61} & \hcbu{0.51} & \hc{0.82} & \hc{0.55} & \hc{0.73} & \hc{0.53} & \hc{0.70} & \hc{0.52} & \hc{0.64} & \hcu{0.52} & \hc{0.54} & \hc{0.86} & \hc{1.10} \\
Digit span {\fontsize{4}{5}\selectfont\citep{enkavi2019large}} & Memory & \hc{0.94} & \hc{0.60} & \hc{0.89} & \hc{0.59} & \hc{0.72} & \hc{0.58} & \hc{0.70} & \hc{0.57} & \hc{0.69} & \hcbu{0.57} & \hc{1.01} & \hc{0.60} & \hc{0.84} & \hc{0.59} & \hc{0.75} & \hc{0.58} & \hc{0.66} & \hcu{0.57} & \hc{0.55} & \hc{0.94} & \hc{2.30} \\
Episodic long-term memory {\fontsize{4}{5}\selectfont\citep{popov2023intent}} & Memory & \hc{1.72} & \hc{1.27} & \hc{1.45} & \hc{1.05} & \hc{1.33} & \hc{0.96} & \hc{1.30} & \hc{0.94} & \hc{1.18} & \hcu{0.90} & \hc{1.97} & \hc{1.28} & \hc{1.65} & \hc{1.02} & \hc{1.43} & \hc{0.92} & \hc{1.13} & \hcbu{0.89} & \hc{0.87} & \hcn & \hc{0.69} \\
N-back {\fontsize{4}{5}\selectfont\citep{enkavi2019large}} & Memory & \hc{0.58} & \hc{0.42} & \hc{0.58} & \hc{0.42} & \hc{0.55} & \hc{0.41} & \hc{0.55} & \hc{0.41} & \hc{0.53} & \hcu{0.40} & \hc{0.59} & \hc{0.42} & \hc{0.57} & \hc{0.42} & \hc{0.57} & \hc{0.41} & \hc{0.52} & \hcbu{0.40} & \hc{0.40} & \hc{0.58} & \hc{0.69} \\
Recall and recognition {\fontsize{4}{5}\selectfont\citep{cox2018information}} & Memory & \hc{1.73} & \hc{1.24} & \hc{1.57} & \hc{1.18} & \hc{1.44} & \hc{1.13} & \hc{1.41} & \hc{1.12} & \hc{1.36} & \hcu{1.09} & \hc{1.87} & \hc{1.22} & \hc{1.56} & \hc{1.14} & \hc{1.37} & \hc{1.10} & \hc{1.38} & \hcbu{1.07} & \hc{1.06} & \hcn & \hcn \\
Recent probes {\fontsize{4}{5}\selectfont\citep{enkavi2019large}} & Memory & \hc{0.70} & \hc{0.26} & \hc{0.61} & \hc{0.26} & \hc{0.39} & \hc{0.26} & \hc{0.41} & \hc{0.26} & \hc{0.35} & \hcu{0.26} & \hc{0.74} & \hc{0.27} & \hc{0.56} & \hc{0.26} & \hc{0.34} & \hcbu{0.26} & \hc{0.34} & \hc{0.26} & \hc{0.26} & \hc{0.39} & \hc{0.69} \\
\bottomrule
\end{tabular}%
}
}%
\setlength{\ctrtblwd}{\wd\ctrtblbox}%
\ifdim\ctrtblwd>\dimexpr\textheight-8pt\relax
  \setlength{\ctrtblwd}{\dimexpr\textheight-8pt\relax}%
\fi
\resizebox{\ctrtblwd}{!}{\usebox{\ctrtblbox}}
\par
\begin{minipage}{\ctrtblwd}
\vspace{4pt}
\begin{center}\footnotesize
\colorbox{c1!45}{\strut\hspace{5pt}} \scriptsize$<$0.20 \quad
\colorbox{c2!40}{\strut\hspace{5pt}} \scriptsize 0.20--0.40 \quad
\colorbox{c3!40}{\strut\hspace{5pt}} \scriptsize 0.40--0.60 \quad
\colorbox{c4!45}{\strut\hspace{5pt}} \scriptsize 0.60--0.80 \quad
\colorbox{c5!55}{\strut\hspace{5pt}} \scriptsize 0.80--1.00 \quad
\colorbox{c6!50}{\strut\hspace{5pt}} \scriptsize 1.00--1.20 \quad
\colorbox{c7!45}{\strut\hspace{5pt}} \scriptsize 1.20--1.50 \quad
\colorbox{c8!40}{\strut\hspace{5pt}} \scriptsize 1.50--2.00 \quad
\colorbox{c9!35}{\strut\hspace{5pt}} \scriptsize$>$2.00
\end{center}
\vspace{-2pt}
{\tiny \underline{\textbf{Bold+underline}}\,=\,best; \underline{underline}\,=\,second-best (C$_p$ excluded; see footnote). NLL under bf16 (half-precision) inference. Reference columns report 4-bit values as originally published. Lower is better. ``---''\,=\,no cognitive baseline reported. Size labels in billions of parameters. Subscript $_r$ denotes our reproducing evaluation of the original Centaur model under identical library and CUDA versions as our models; $_p$ denotes values published by \citet{binz2025foundation}. C$_p$ is included for reference but excluded from best/second-best marking, as it was evaluated under different software conditions. $^\dagger$Horizon task (5 experiments) and Two-step task (3 experiments) are reported as single merged values by \citet{binz2025foundation}. Rounded to 2\,d.p.}
\end{minipage}
\captionsetup{width=\ctrtblwd}
\caption{\textbf{Base vs.\ finetuned NLL on Psych-101 under bf16 inference.} Within each model size, the left column (base) shows the pretrained model and the right column (ft) shows the cognitively finetuned variant. Cell colour encodes performance (expressed as NLL); the consistent colour shift from base to ft demonstrates the effect of cognitive finetuning across all scales.}
\label{tab:psych101_pe_bf16_base_v_ft}
\end{sidewaystable}

\begin{sidewaystable}[p]
\ContinuedFloat
\centering
\sbox{\ctrtblbox}{%
{\tiny
\setlength{\tabcolsep}{3pt}
\renewcommand{\arraystretch}{1.15}
\begin{tabular}{@{}l l  r@{\;\;}r  r@{\;\;}r  r@{\;\;}r  r@{\;\;}r  r@{\;\;}r  r@{\;\;}r  r@{\;\;}r  r@{\;\;}r  r  r  r  r  r@{}}
\toprule
& & \multicolumn{10}{c}{\texttt{\textbf{Qwen3}} family} & \multicolumn{6}{c}{\texttt{\textbf{Llama-3.1/3.2}} family} & \multicolumn{4}{c}{\citet{binz2025foundation}} & \textbf{Chance} \\
\cmidrule(lr){3-12}\cmidrule(lr){13-18}\cmidrule(lr){19-22}
& & \multicolumn{2}{c}{0.6B} & \multicolumn{2}{c}{1.7B} & \multicolumn{2}{c}{4B} & \multicolumn{2}{c}{8B} & \multicolumn{2}{c}{14B} & \multicolumn{2}{c}{1B} & \multicolumn{2}{c}{3B} & \multicolumn{2}{c}{8B} & & & & & \\
\cmidrule(lr){3-4} \cmidrule(lr){5-6} \cmidrule(lr){7-8} \cmidrule(lr){9-10} \cmidrule(lr){11-12} \cmidrule(lr){13-14} \cmidrule(lr){15-16} \cmidrule(lr){17-18}
\textbf{Experiment} & \textbf{Type} & {\fontsize{4}{5}\selectfont base} & {\fontsize{4}{5}\selectfont ft} & {\fontsize{4}{5}\selectfont base} & {\fontsize{4}{5}\selectfont ft} & {\fontsize{4}{5}\selectfont base} & {\fontsize{4}{5}\selectfont ft} & {\fontsize{4}{5}\selectfont base} & {\fontsize{4}{5}\selectfont ft} & {\fontsize{4}{5}\selectfont base} & {\fontsize{4}{5}\selectfont ft} & {\fontsize{4}{5}\selectfont base} & {\fontsize{4}{5}\selectfont ft} & {\fontsize{4}{5}\selectfont base} & {\fontsize{4}{5}\selectfont ft} & {\fontsize{4}{5}\selectfont base} & {\fontsize{4}{5}\selectfont ft} & L-70B$_p$ & C$_r$ & C$_p$ & Cog.$_p$ & $\ln(k)$ \\
\midrule
Grammar judgement {\fontsize{4}{5}\selectfont\citep{jansen2021rational}} & Misc. & \hc{2.12} & \hc{1.47} & \hc{2.02} & \hc{1.45} & \hc{2.00} & \hc{1.45} & \hc{1.89} & \hc{1.44} & \hc{1.91} & \hcu{1.43} & \hc{2.33} & \hc{1.48} & \hc{2.16} & \hc{1.45} & \hc{2.12} & \hc{1.44} & \hc{1.99} & \hc{1.44} & \hc{1.44} & \hcbu{1.41} & \hcn \\
Probabilistic reasoning {\fontsize{4}{5}\selectfont\citep{zhu2020bayesian}} & Misc. & \hc{2.75} & \hc{2.45} & \hc{2.69} & \hc{2.40} & \hc{2.61} & \hc{2.36} & \hc{2.57} & \hcu{2.35} & \hc{2.58} & \hcbu{2.34} & \hc{2.89} & \hc{2.56} & \hc{2.80} & \hc{2.47} & \hc{2.71} & \hc{2.43} & \hc{2.64} & \hc{2.40} & \hc{2.37} & \hcn & \hcn \\
Serial reaction time task {\fontsize{4}{5}\selectfont\citep{wu2023chunking}} & Misc. & \hc{0.21} & \hc{0.17} & \hc{0.19} & \hc{0.17} & \hc{0.19} & \hc{0.17} & \hc{0.19} & \hcu{0.17} & \hc{0.19} & \hcbu{0.17} & \hc{0.31} & \hc{0.17} & \hc{0.23} & \hc{0.17} & \hc{0.20} & \hc{0.17} & \hc{0.19} & \hc{0.18} & \hc{0.17} & \hc{0.20} & \hc{1.39} \\
THINGS odd-one-out {\fontsize{4}{5}\selectfont\citep{hebart2023things}} & Misc. & \hc{1.21} & \hc{0.82} & \hc{1.21} & \hc{0.81} & \hc{1.17} & \hc{0.80} & \hc{1.15} & \hc{0.79} & \hc{1.13} & \hcbu{0.79} & \hc{1.31} & \hc{0.82} & \hc{1.18} & \hc{0.82} & \hc{1.16} & \hcu{0.79} & \hc{1.14} & \hc{0.80} & \hc{0.81} & \hc{0.83} & \hc{1.10} \\
\midrule
Changing bandit {\fontsize{4}{5}\selectfont\citep{xiong2023neural}} & Bandit & \hc{0.42} & \hc{0.30} & \hc{0.38} & \hc{0.30} & \hc{0.36} & \hcu{0.29} & \hc{0.35} & \hc{0.29} & \hc{0.34} & \hcbu{0.29} & \hc{1.34} & \hc{0.32} & \hc{0.90} & \hc{0.31} & \hc{0.39} & \hc{0.29} & \hc{0.38} & \hc{0.45} & \hc{0.30} & \hc{0.44} & \hc{0.69} \\
Decisions from experience {\fontsize{4}{5}\selectfont\citep{wulff2018sampling}} & Bandit & \hc{0.74} & \hc{0.49} & \hc{0.66} & \hc{0.49} & \hc{0.69} & \hc{0.49} & \hc{0.78} & \hc{0.49} & \hc{0.61} & \hc{0.48} & \hc{0.81} & \hc{0.50} & \hc{0.71} & \hc{0.49} & \hc{0.65} & \hc{0.48} & \hcbu{0.43} & \hcu{0.47} & \hc{0.37} & \hc{0.54} & \hcn \\
Drifting four-armed bandit {\fontsize{4}{5}\selectfont\citep{bahrami2020four}} & Bandit & \hc{0.98} & \hc{0.73} & \hc{0.95} & \hc{0.72} & \hc{0.92} & \hc{0.71} & \hc{0.91} & \hc{0.71} & \hc{0.91} & \hcbu{0.71} & \hc{0.95} & \hc{0.73} & \hc{0.94} & \hc{0.72} & \hc{0.90} & \hc{0.71} & \hc{0.88} & \hcu{0.71} & \hc{0.70} & \hc{0.90} & \hc{1.39} \\
Horizon task {\fontsize{4}{5}\selectfont\citep{feng2021dynamics}} & Bandit & \hc{0.47} & \hc{0.23} & \hc{0.43} & \hc{0.23} & \hc{0.39} & \hc{0.22} & \hc{0.36} & \hc{0.22} & \hc{0.34} & \hcu{0.22} & \hc{0.81} & \hc{0.26} & \hc{0.67} & \hc{0.24} & \hc{0.44} & \hcbu{0.22} & \hc{0.52} & \hc{0.40} & \hcd{0.40} & \hcd{0.36} & \hc{0.69} \\
Horizon task {\fontsize{4}{5}\selectfont\citep{sadeghiyeh2020temporal}} & Bandit & \hc{0.67} & \hc{0.59} & \hc{0.72} & \hc{0.58} & \hc{0.66} & \hc{0.58} & \hc{0.65} & \hc{0.58} & \hc{0.65} & \hc{0.58} & \hc{0.67} & \hc{0.59} & \hc{0.68} & \hc{0.58} & \hc{0.66} & \hc{0.58} & \hcu{0.52} & \hc{0.58} & \hcd{0.40} & \hcbud{0.36} & \hc{0.69} \\
Horizon task {\fontsize{4}{5}\selectfont\citep{somerville2017charting}} & Bandit & \hc{0.54} & \hc{0.34} & \hc{0.55} & \hc{0.34} & \hc{0.53} & \hc{0.34} & \hc{0.47} & \hcu{0.34} & \hc{0.49} & \hcbu{0.33} & \hc{0.59} & \hc{0.35} & \hc{0.57} & \hc{0.34} & \hc{0.51} & \hc{0.34} & \hc{0.52} & \hc{0.35} & \hcd{0.40} & \hcd{0.36} & \hc{0.69} \\
Horizon task {\fontsize{4}{5}\selectfont\citep{waltz2020differential}} & Bandit & \hc{0.33} & \hc{0.15} & \hc{0.33} & \hc{0.15} & \hc{0.31} & \hc{0.15} & \hc{0.27} & \hc{0.15} & \hc{0.24} & \hc{0.15} & \hc{0.37} & \hc{0.16} & \hc{0.37} & \hc{0.16} & \hc{0.30} & \hcbu{0.14} & \hc{0.52} & \hcu{0.15} & \hcd{0.40} & \hcd{0.36} & \hc{0.69} \\
Horizon task {\fontsize{4}{5}\selectfont\citep{wilson2014humans}} & Bandit & \hc{0.57} & \hc{0.46} & \hc{0.58} & \hc{0.46} & \hc{0.56} & \hc{0.45} & \hc{0.53} & \hc{0.45} & \hc{0.54} & \hcu{0.45} & \hc{0.64} & \hc{0.47} & \hc{0.62} & \hc{0.46} & \hc{0.55} & \hc{0.45} & \hc{0.52} & \hc{0.48} & \hcd{0.40} & \hcbud{0.36} & \hc{0.69} \\
Iowa gambling task {\fontsize{4}{5}\selectfont\citep{steingroever2015data}} & Bandit & \hc{1.07} & \hc{0.93} & \hc{1.06} & \hc{0.93} & \hc{1.07} & \hc{0.91} & \hc{1.07} & \hc{0.91} & \hc{1.07} & \hc{0.91} & \hc{1.10} & \hc{0.94} & \hc{1.06} & \hc{0.92} & \hc{1.04} & \hcu{0.91} & \hc{0.99} & \hcbu{0.91} & \hc{0.89} & \hc{1.16} & \hc{1.39} \\
Prob.\ instrumental learning {\fontsize{4}{5}\selectfont\citep{lefebvre2017behavioural}} & Bandit & \hc{0.62} & \hc{0.50} & \hc{0.60} & \hc{0.50} & \hc{0.57} & \hcbu{0.49} & \hc{0.59} & \hcu{0.49} & \hc{0.56} & \hc{0.49} & \hc{0.62} & \hc{0.51} & \hc{0.58} & \hc{0.50} & \hc{0.56} & \hc{0.49} & \hc{0.54} & \hc{0.50} & \hc{0.49} & \hc{0.50} & \hc{0.69} \\
Spatially correlated MAB {\fontsize{4}{5}\selectfont\citep{wu2018generalization}} & Bandit & \hc{2.61} & \hc{2.01} & \hc{2.55} & \hc{1.96} & \hc{2.50} & \hc{1.93} & \hc{2.44} & \hc{1.87} & \hc{2.46} & \hcu{1.86} & \hc{2.76} & \hc{2.16} & \hc{2.64} & \hc{2.00} & \hc{2.58} & \hc{1.93} & \hc{2.45} & \hcbu{1.82} & \hc{1.83} & \hc{2.76} & \hc{3.40} \\
Structured bandit {\fontsize{4}{5}\selectfont\citep{schulz2020finding}} & Bandit & \hc{0.92} & \hc{0.67} & \hc{0.92} & \hc{0.66} & \hc{0.88} & \hc{0.65} & \hc{0.87} & \hc{0.65} & \hc{0.83} & \hcbu{0.64} & \hc{0.94} & \hc{0.67} & \hc{0.89} & \hc{0.66} & \hc{0.85} & \hc{0.65} & \hc{0.81} & \hcu{0.64} & \hc{0.64} & \hc{1.05} & \hc{2.08} \\
Two-armed bandit {\fontsize{4}{5}\selectfont\citep{gershman2018deconstructing}} & Bandit & \hc{0.43} & \hc{0.30} & \hc{0.41} & \hc{0.30} & \hc{0.40} & \hc{0.29} & \hc{0.39} & \hcu{0.29} & \hc{0.39} & \hcbu{0.29} & \hc{0.43} & \hc{0.30} & \hc{0.43} & \hc{0.29} & \hc{0.40} & \hc{0.29} & \hc{0.38} & \hc{0.30} & \hc{0.30} & \hc{0.42} & \hc{0.69} \\
\midrule
Aversive learning {\fontsize{4}{5}\selectfont\citep{wise2019computational}} & Sup.\ learn. & \hc{5.20} & \hc{4.71} & \hc{5.06} & \hc{4.55} & \hc{4.95} & \hc{4.27} & \hc{4.89} & \hc{4.24} & \hc{4.75} & \hcu{3.81} & \hc{5.66} & \hc{4.58} & \hc{5.52} & \hc{4.80} & \hc{5.25} & \hcbu{3.66} & \hc{5.11} & \hc{4.13} & \hc{4.07} & \hcn & \hcn \\
Medin categorization {\fontsize{4}{5}\selectfont\citep{levering2020revisiting}} & Sup.\ learn. & \hc{0.60} & \hc{0.51} & \hc{0.60} & \hc{0.51} & \hc{0.60} & \hc{0.50} & \hc{0.59} & \hc{0.50} & \hc{0.55} & \hcbu{0.50} & \hc{0.71} & \hc{0.52} & \hc{0.60} & \hc{0.50} & \hc{0.57} & \hcu{0.50} & \hc{0.58} & \hc{0.50} & \hc{0.50} & \hc{0.53} & \hcn \\
Multiple-cue judgment {\fontsize{4}{5}\selectfont\citep{collsioo2023numerical}} & Sup.\ learn. & \hc{1.46} & \hc{1.17} & \hc{1.38} & \hc{1.16} & \hc{1.28} & \hc{1.14} & \hc{1.30} & \hc{1.14} & \hc{1.26} & \hcu{1.13} & \hc{1.69} & \hc{1.17} & \hc{1.50} & \hc{1.15} & \hc{1.32} & \hcbu{1.12} & \hc{1.28} & \hc{1.14} & \hc{1.12} & \hc{1.92} & \hc{2.20} \\
Shepard categorization {\fontsize{4}{5}\selectfont\citep{badham2017deficits}} & Sup.\ learn. & \hc{0.63} & \hc{0.55} & \hc{0.61} & \hc{0.55} & \hc{0.59} & \hc{0.54} & \hc{0.58} & \hc{0.53} & \hc{0.59} & \hcbu{0.53} & \hc{0.67} & \hc{0.56} & \hc{0.62} & \hc{0.54} & \hc{0.60} & \hcu{0.53} & \hc{0.58} & \hc{0.54} & \hc{0.54} & \hc{0.61} & \hc{0.69} \\
Weather prediction task {\fontsize{4}{5}\selectfont\citep{speekenbrink2008learning}} & Sup.\ learn. & \hc{0.62} & \hc{0.55} & \hc{0.59} & \hc{0.56} & \hc{0.61} & \hcu{0.54} & \hc{0.58} & \hc{0.55} & \hc{0.58} & \hc{0.55} & \hc{0.62} & \hc{0.55} & \hc{0.59} & \hcbu{0.54} & \hc{0.58} & \hc{0.56} & \hc{0.57} & \hc{0.56} & \hc{0.55} & \hc{0.63} & \hc{0.69} \\
\midrule
\textbf{Mean (all 48)} &  & \hc{0.99} & \hc{0.69} & \hc{0.94} & \hc{0.68} & \hc{0.90} & \hc{0.66} & \hc{0.89} & \hc{0.66} & \hc{0.87} & \hcbu{0.64} & \hc{1.14} & \hc{0.70} & \hc{1.02} & \hc{0.68} & \hc{0.92} & \hcu{0.64} & \hc{0.88} & \hc{0.71} & \hc{0.69} & \hc{0.69} & \hc{0.98} \\
\bottomrule
\end{tabular}%
}
}%
\setlength{\ctrtblwd}{\wd\ctrtblbox}%
\ifdim\ctrtblwd>\dimexpr\textheight-8pt\relax
  \setlength{\ctrtblwd}{\dimexpr\textheight-8pt\relax}%
\fi
\resizebox{\ctrtblwd}{!}{\usebox{\ctrtblbox}}
\par
\begin{minipage}{\ctrtblwd}
\vspace{4pt}
\begin{center}\footnotesize
\colorbox{c1!45}{\strut\hspace{5pt}} \scriptsize$<$0.20 \quad
\colorbox{c2!40}{\strut\hspace{5pt}} \scriptsize 0.20--0.40 \quad
\colorbox{c3!40}{\strut\hspace{5pt}} \scriptsize 0.40--0.60 \quad
\colorbox{c4!45}{\strut\hspace{5pt}} \scriptsize 0.60--0.80 \quad
\colorbox{c5!55}{\strut\hspace{5pt}} \scriptsize 0.80--1.00 \quad
\colorbox{c6!50}{\strut\hspace{5pt}} \scriptsize 1.00--1.20 \quad
\colorbox{c7!45}{\strut\hspace{5pt}} \scriptsize 1.20--1.50 \quad
\colorbox{c8!40}{\strut\hspace{5pt}} \scriptsize 1.50--2.00 \quad
\colorbox{c9!35}{\strut\hspace{5pt}} \scriptsize$>$2.00
\end{center}
\vspace{-2pt}
{\tiny \underline{\textbf{Bold+underline}}\,=\,best; \underline{underline}\,=\,second-best (C$_p$ excluded; see footnote). NLL under bf16 (half-precision) inference. Reference columns report 4-bit values as originally published. Lower is better. ``---''\,=\,no cognitive baseline reported. Size labels in billions of parameters. Subscript $_r$ denotes our reproducing evaluation of the original Centaur model under identical library and CUDA versions as our models; $_p$ denotes values published by \citet{binz2025foundation}. C$_p$ is included for reference but excluded from best/second-best marking, as it was evaluated under different software conditions. $^\dagger$Horizon task (5 experiments) and Two-step task (3 experiments) are reported as single merged values by \citet{binz2025foundation}. Rounded to 2\,d.p.}
\end{minipage}
\captionsetup{width=\ctrtblwd}
\caption{\textbf{Base vs.\ finetuned NLL on Psych-101 under bf16 inference (continued).} Within each model size, the left column (base) shows the pretrained model and the right column (ft) the cognitively finetuned variant.}
\label{tab:psych101_pe_bf16_base_v_ft_cont}
\end{sidewaystable}

\begin{sidewaystable}[p]
\centering
\sbox{\ctrtblbox}{%
{\tiny
\setlength{\tabcolsep}{3pt}
\renewcommand{\arraystretch}{1.15}
\begin{tabular}{@{}l l  r  r  r  r  r  r  r  r  r  r  r  r  r  r  rrrr  r@{}}
\toprule
& & \multicolumn{5}{c}{\texttt{\textbf{Qwentaur}}} & \multicolumn{3}{c}{\texttt{\textbf{Llama-Centaur}}} & \multicolumn{4}{c}{\texttt{\textbf{Smoltaur}}} & \multicolumn{2}{c}{\texttt{\textbf{Olmotaur}}} & \multicolumn{4}{c}{\citet{binz2025foundation}} & \textbf{Chance} \\
\cmidrule(lr){3-7}\cmidrule(lr){8-10}\cmidrule(lr){11-14}\cmidrule(lr){15-16}
\textbf{Experiment} & \textbf{Type} & 0.6B & 1.7B & 4B & 8B & 14B & 1B & 3B & 8B & 0.1B & 0.4B & 1.7B & 3B & 1B & 7B & L-70B$_p$ & C$_r$ & C$_p$ & Cog.$_p$ & $\ln(k)$ \\
\midrule
Balloon analog risk task {\fontsize{4}{5}\selectfont\citep{frey2017risk}} & Decision & \hc{0.07} & \hc{0.07} & \hc{0.07} & \hc{0.06} & \hcu{0.06} & \hc{0.07} & \hc{0.07} & \hcbu{0.06} & \hc{0.08} & \hc{0.07} & \hc{0.07} & \hc{0.07} & \hc{0.07} & \hc{0.07} & \hc{0.08} & \hc{0.07} & \hc{0.06} & \hc{0.09} & \hc{0.69} \\
CPC18 {\fontsize{4}{5}\selectfont\citep{plonsky2018when}} & Decision & \hc{0.35} & \hc{0.34} & \hc{0.34} & \hc{0.34} & \hcbu{0.33} & \hc{0.36} & \hc{0.35} & \hcu{0.34} & \hc{0.42} & \hc{0.41} & \hc{0.40} & \hc{0.34} & \hc{0.44} & \hc{0.34} & \hc{0.41} & \hc{0.35} & \hc{0.34} & \hc{0.66} & \hc{0.69} \\
Columbia card task {\fontsize{4}{5}\selectfont\citep{frey2017risk}} & Decision & \hc{0.20} & \hc{0.20} & \hc{0.20} & \hc{0.20} & \hcu{0.20} & \hc{0.21} & \hc{0.20} & \hcbu{0.20} & \hc{0.24} & \hc{0.22} & \hc{0.21} & \hc{0.20} & \hc{0.22} & \hc{0.20} & \hc{0.23} & \hc{0.21} & \hc{0.19} & \hc{0.26} & \hc{0.69} \\
Decisions from description {\fontsize{4}{5}\selectfont\citep{wulff2018sampling}} & Decision & \hc{0.62} & \hc{0.63} & \hc{0.62} & \hc{0.61} & \hc{0.60} & \hc{0.62} & \hc{0.63} & \hcu{0.60} & \hc{0.68} & \hc{0.66} & \hc{0.64} & \hc{0.61} & \hc{0.63} & \hc{0.62} & \hc{0.76} & \hcbu{0.59} & \hc{0.53} & \hc{0.61} & \hc{0.69} \\
Experiential-symbolic task {\fontsize{4}{5}\selectfont\citep{garcia2023experiential}} & Decision & \hc{0.48} & \hc{0.47} & \hc{0.46} & \hc{0.46} & \hc{0.46} & \hc{0.47} & \hc{0.47} & \hc{0.46} & \hc{0.47} & \hcu{0.45} & \hcbu{0.43} & \hc{0.47} & \hc{0.47} & \hc{0.47} & \hc{0.70} & \hc{0.46} & \hc{0.45} & \hcn & \hcn \\
Gardening task {\fontsize{4}{5}\selectfont\citep{flesch2018comparing}} & Decision & \hc{0.39} & \hc{0.39} & \hc{0.38} & \hc{0.38} & \hcu{0.38} & \hc{0.40} & \hc{0.39} & \hcbu{0.38} & \hc{0.57} & \hc{0.54} & \hc{0.53} & \hc{0.38} & \hc{0.58} & \hc{0.38} & \hc{0.50} & \hc{0.49} & \hc{0.38} & \hc{0.91} & \hc{0.69} \\
Go/no-go {\fontsize{4}{5}\selectfont\citep{enkavi2019large}} & Decision & \hcbu{0.00} & \hcbu{0.00} & \hcbu{0.00} & \hcbu{0.00} & \hcbu{0.00} & \hcbu{0.00} & \hcbu{0.00} & \hcbu{0.00} & \hc{0.00} & \hc{0.00} & \hc{0.00} & \hcu{0.00} & \hc{0.00} & \hc{0.00} & \hc{0.01} & \hcbu{0.00} & \hc{0.00} & \hc{0.08} & \hc{0.00} \\
Intertemporal choice {\fontsize{4}{5}\selectfont\citep{ruggeri2022globalizability}} & Decision & \hc{0.44} & \hc{0.44} & \hcu{0.44} & \hc{0.44} & \hc{0.44} & \hc{0.44} & \hcu{0.44} & \hc{0.44} & \hc{0.47} & \hc{0.45} & \hc{0.44} & \hc{0.44} & \hc{0.44} & \hc{0.44} & \hc{0.73} & \hcbu{0.44} & \hc{0.43} & \hc{0.66} & \hc{0.69} \\
Multi-attribute DM {\fontsize{4}{5}\selectfont\citep{hilbig2014generalized}} & Decision & \hc{0.09} & \hc{0.08} & \hc{0.09} & \hc{0.08} & \hc{0.08} & \hc{0.08} & \hc{0.08} & \hc{0.08} & \hc{0.31} & \hc{0.19} & \hc{0.16} & \hcu{0.08} & \hc{0.10} & \hc{0.09} & \hc{0.15} & \hcbu{0.06} & \hc{0.06} & \hc{0.19} & \hc{0.69} \\
Risky choice {\fontsize{4}{5}\selectfont\citep{krueger2024identifying}} & Decision & \hc{0.48} & \hc{0.47} & \hc{0.44} & \hc{0.43} & \hcbu{0.42} & \hc{0.49} & \hc{0.46} & \hcu{0.43} & \hc{0.72} & \hc{0.61} & \hc{0.53} & \hc{0.45} & \hc{0.58} & \hc{0.46} & \hc{0.65} & \hc{0.43} & \hc{0.43} & \hcn & \hcn \\
choices13k {\fontsize{4}{5}\selectfont\citep{peterson2021using}} & Decision & \hc{0.44} & \hc{0.44} & \hc{0.43} & \hc{0.43} & \hcbu{0.43} & \hc{0.44} & \hc{0.44} & \hcu{0.43} & \hc{0.49} & \hc{0.46} & \hc{0.44} & \hc{0.44} & \hc{0.45} & \hc{0.44} & \hc{0.53} & \hc{0.43} & \hc{0.43} & \hc{0.66} & \hc{0.69} \\
\midrule
Multi-task RL {\fontsize{4}{5}\selectfont\citep{tomov2021multitask}} & MDP & \hc{0.60} & \hc{0.59} & \hc{0.58} & \hc{0.58} & \hcbu{0.57} & \hc{0.60} & \hc{0.59} & \hc{0.57} & \hc{0.68} & \hc{0.66} & \hc{0.63} & \hc{0.59} & \hc{0.78} & \hc{0.59} & \hc{0.66} & \hcu{0.57} & \hc{0.57} & \hc{1.04} & \hc{1.10} \\
Tile-revealing task {\fontsize{4}{5}\selectfont\citep{kumar2023disentangling}} & MDP & \hc{0.02} & \hc{0.01} & \hc{0.00} & \hc{0.07} & \hcu{0.00} & \hc{0.02} & \hc{0.00} & \hcbu{0.00} & \hc{2.69} & \hc{1.99} & \hc{0.04} & \hc{0.00} & \hc{0.06} & \hc{0.09} & \hc{2.74} & \hc{2.29} & \hc{1.87} & \hcn & \hcn \\
Two-step task {\fontsize{4}{5}\selectfont\citep{kool2017cost}} & MDP & \hc{0.54} & \hc{0.53} & \hcbu{0.53} & \hcu{0.53} & \hcbu{0.53} & \hc{0.55} & \hc{0.53} & \hcbu{0.53} & \hc{0.60} & \hc{0.60} & \hc{0.56} & \hc{0.53} & \hc{0.57} & \hc{0.53} & \hc{0.61} & \hc{0.53} & \hcd{0.50} & \hcd{0.60} & \hc{0.69} \\
Two-step task {\fontsize{4}{5}\selectfont\citep{kool2016when}} & MDP & \hc{0.51} & \hc{0.50} & \hc{0.49} & \hc{0.49} & \hcu{0.48} & \hc{0.52} & \hc{0.50} & \hc{0.49} & \hc{0.59} & \hc{0.58} & \hc{0.52} & \hc{0.50} & \hc{0.55} & \hc{0.50} & \hc{0.61} & \hcbu{0.48} & \hcd{0.50} & \hcd{0.60} & \hc{0.69} \\
Two-step task {\fontsize{4}{5}\selectfont\citep{zorowitz2023data}} & MDP & \hc{0.52} & \hc{0.51} & \hc{0.51} & \hcbu{0.51} & \hc{0.51} & \hc{0.52} & \hcu{0.51} & \hc{0.51} & \hc{0.60} & \hc{0.58} & \hc{0.52} & \hc{0.52} & \hc{0.54} & \hc{0.51} & \hc{0.61} & \hc{0.51} & \hcd{0.50} & \hcd{0.60} & \hc{0.69} \\
Virtual subway network {\fontsize{4}{5}\selectfont\citep{tomov2020discovery}} & MDP & \hc{1.16} & \hc{1.15} & \hc{1.14} & \hcu{1.13} & \hcbu{1.12} & \hc{1.19} & \hc{1.16} & \hc{1.13} & \hc{1.43} & \hc{1.33} & \hc{1.30} & \hc{1.14} & \hc{1.32} & \hc{1.15} & \hc{1.53} & \hc{1.16} & \hc{1.13} & \hcn & \hc{1.61} \\
Zoopermarket {\fontsize{4}{5}\selectfont\citep{ludwig2023human}} & MDP & \hc{0.54} & \hc{0.52} & \hc{0.51} & \hc{0.50} & \hcbu{0.48} & \hc{0.54} & \hc{0.52} & \hc{0.50} & \hc{0.63} & \hc{0.61} & \hc{0.56} & \hc{0.51} & \hc{0.61} & \hc{0.52} & \hc{0.60} & \hcu{0.50} & \hc{0.48} & \hc{0.60} & \hc{0.69} \\
\midrule
Cond.\ assoc.\ learning {\fontsize{4}{5}\selectfont\citep{collins2014working}} & Memory & \hc{0.54} & \hc{0.53} & \hc{0.52} & \hc{0.52} & \hcbu{0.51} & \hc{0.55} & \hc{0.53} & \hc{0.52} & \hc{0.74} & \hc{0.64} & \hc{0.58} & \hc{0.53} & \hc{0.65} & \hc{0.53} & \hc{0.64} & \hcu{0.52} & \hc{0.54} & \hc{0.86} & \hc{1.10} \\
Digit span {\fontsize{4}{5}\selectfont\citep{enkavi2019large}} & Memory & \hc{0.60} & \hc{0.59} & \hc{0.58} & \hc{0.57} & \hcbu{0.57} & \hc{0.60} & \hc{0.59} & \hc{0.58} & \hc{0.71} & \hc{0.65} & \hc{0.62} & \hc{0.59} & \hc{0.62} & \hc{0.59} & \hc{0.66} & \hcu{0.57} & \hc{0.55} & \hc{0.94} & \hc{2.30} \\
Episodic long-term memory {\fontsize{4}{5}\selectfont\citep{popov2023intent}} & Memory & \hc{1.27} & \hc{1.05} & \hc{0.96} & \hc{0.94} & \hcu{0.90} & \hc{1.28} & \hc{1.02} & \hc{0.92} & \hc{1.88} & \hc{1.56} & \hc{1.37} & \hc{1.01} & \hc{1.23} & \hc{1.05} & \hc{1.13} & \hcbu{0.89} & \hc{0.87} & \hcn & \hc{0.69} \\
N-back {\fontsize{4}{5}\selectfont\citep{enkavi2019large}} & Memory & \hc{0.42} & \hc{0.42} & \hc{0.41} & \hc{0.41} & \hcu{0.40} & \hc{0.42} & \hc{0.42} & \hc{0.41} & \hc{0.49} & \hc{0.47} & \hc{0.44} & \hc{0.41} & \hc{0.44} & \hc{0.42} & \hc{0.52} & \hcbu{0.40} & \hc{0.40} & \hc{0.58} & \hc{0.69} \\
Recall and recognition {\fontsize{4}{5}\selectfont\citep{cox2018information}} & Memory & \hc{1.24} & \hc{1.18} & \hc{1.13} & \hc{1.12} & \hcu{1.09} & \hc{1.22} & \hc{1.14} & \hc{1.10} & \hc{1.69} & \hc{1.50} & \hc{1.25} & \hc{1.15} & \hc{1.41} & \hc{1.17} & \hc{1.38} & \hcbu{1.07} & \hc{1.06} & \hcn & \hcn \\
Recent probes {\fontsize{4}{5}\selectfont\citep{enkavi2019large}} & Memory & \hc{0.26} & \hc{0.26} & \hc{0.26} & \hc{0.26} & \hcu{0.26} & \hc{0.27} & \hc{0.26} & \hcbu{0.26} & \hc{0.33} & \hc{0.28} & \hc{0.27} & \hc{0.26} & \hc{0.27} & \hc{0.26} & \hc{0.34} & \hc{0.26} & \hc{0.26} & \hc{0.39} & \hc{0.69} \\
\bottomrule
\end{tabular}%
}
}%
\setlength{\ctrtblwd}{\wd\ctrtblbox}%
\ifdim\ctrtblwd>\dimexpr\textheight-8pt\relax
  \setlength{\ctrtblwd}{\dimexpr\textheight-8pt\relax}%
\fi
\resizebox{\ctrtblwd}{!}{\usebox{\ctrtblbox}}
\par
\begin{minipage}{\ctrtblwd}
\vspace{4pt}
\begin{center}\footnotesize
\colorbox{c1!45}{\strut\hspace{5pt}} \scriptsize$<$0.20 \quad
\colorbox{c2!40}{\strut\hspace{5pt}} \scriptsize 0.20--0.40 \quad
\colorbox{c3!40}{\strut\hspace{5pt}} \scriptsize 0.40--0.60 \quad
\colorbox{c4!45}{\strut\hspace{5pt}} \scriptsize 0.60--0.80 \quad
\colorbox{c5!55}{\strut\hspace{5pt}} \scriptsize 0.80--1.00 \quad
\colorbox{c6!50}{\strut\hspace{5pt}} \scriptsize 1.00--1.20 \quad
\colorbox{c7!45}{\strut\hspace{5pt}} \scriptsize 1.20--1.50 \quad
\colorbox{c8!40}{\strut\hspace{5pt}} \scriptsize 1.50--2.00 \quad
\colorbox{c9!35}{\strut\hspace{5pt}} \scriptsize$>$2.00
\end{center}
\vspace{-2pt}
{\tiny \underline{\textbf{Bold+underline}}\,=\,best; \underline{underline}\,=\,second-best (C$_p$ excluded; see below). Lower is better. ``---''\,=\,not available. All models finetuned at LoRA rank 16, bf16 inference. Reference columns (\citet{binz2025foundation}): L-70B$_p$\,=\,Llama-3.1-70B base, C$_r$\,=\,our reproduced Centaur-70B, C$_p$\,=\,published Centaur-70B, Cog.$_p$\,=\,domain-specific cognitive baseline. C$_p$ is included for reference but excluded from best/second-best marking, as it was evaluated under different software conditions. $^\dagger$Horizon and Two-step tasks are reported as single merged values by \citet{binz2025foundation}. Rounded to 2\,d.p.}
\end{minipage}
\captionsetup{width=\ctrtblwd}
\caption{\textbf{Per-experiment NLL on Psych-101 for all four finetuned families (bf16, LoRA rank 16).} Cell colour encodes performance (green\,=\,better).}
\label{tab:psych101_pe_ft_all}
\end{sidewaystable}

\begin{sidewaystable}[p]
\ContinuedFloat
\centering
\sbox{\ctrtblbox}{%
{\tiny
\setlength{\tabcolsep}{3pt}
\renewcommand{\arraystretch}{1.15}
\begin{tabular}{@{}l l  r  r  r  r  r  r  r  r  r  r  r  r  r  r  rrrr  r@{}}
\toprule
& & \multicolumn{5}{c}{\texttt{\textbf{Qwentaur}}} & \multicolumn{3}{c}{\texttt{\textbf{Llama-Centaur}}} & \multicolumn{4}{c}{\texttt{\textbf{Smoltaur}}} & \multicolumn{2}{c}{\texttt{\textbf{Olmotaur}}} & \multicolumn{4}{c}{\citet{binz2025foundation}} & \textbf{Chance} \\
\cmidrule(lr){3-7}\cmidrule(lr){8-10}\cmidrule(lr){11-14}\cmidrule(lr){15-16}
\textbf{Experiment} & \textbf{Type} & 0.6B & 1.7B & 4B & 8B & 14B & 1B & 3B & 8B & 0.1B & 0.4B & 1.7B & 3B & 1B & 7B & L-70B$_p$ & C$_r$ & C$_p$ & Cog.$_p$ & $\ln(k)$ \\
\midrule
Grammar judgement {\fontsize{4}{5}\selectfont\citep{jansen2021rational}} & Misc. & \hc{1.47} & \hc{1.45} & \hc{1.45} & \hc{1.44} & \hcu{1.43} & \hc{1.48} & \hc{1.45} & \hc{1.44} & \hc{1.63} & \hc{1.54} & \hc{1.51} & \hc{1.46} & \hc{1.51} & \hc{1.46} & \hc{1.99} & \hc{1.44} & \hc{1.44} & \hcbu{1.41} & \hcn \\
Probabilistic reasoning {\fontsize{4}{5}\selectfont\citep{zhu2020bayesian}} & Misc. & \hc{2.45} & \hc{2.40} & \hc{2.36} & \hcu{2.35} & \hcbu{2.34} & \hc{2.56} & \hc{2.47} & \hc{2.43} & \hc{2.67} & \hc{2.60} & \hc{2.50} & \hc{2.45} & \hc{2.61} & \hc{2.47} & \hc{2.64} & \hc{2.40} & \hc{2.37} & \hcn & \hcn \\
Serial reaction time task {\fontsize{4}{5}\selectfont\citep{wu2023chunking}} & Misc. & \hc{0.17} & \hc{0.17} & \hc{0.17} & \hc{0.17} & \hc{0.17} & \hc{0.17} & \hc{0.17} & \hc{0.17} & \hc{0.17} & \hc{0.16} & \hcbu{0.16} & \hc{0.17} & \hcu{0.16} & \hc{0.17} & \hc{0.19} & \hc{0.18} & \hc{0.17} & \hc{0.20} & \hc{1.39} \\
THINGS odd-one-out {\fontsize{4}{5}\selectfont\citep{hebart2023things}} & Misc. & \hc{0.82} & \hc{0.81} & \hc{0.80} & \hc{0.79} & \hcbu{0.79} & \hc{0.82} & \hc{0.82} & \hcu{0.79} & \hc{1.08} & \hc{0.90} & \hc{0.86} & \hc{0.80} & \hc{0.86} & \hc{0.81} & \hc{1.14} & \hc{0.80} & \hc{0.81} & \hc{0.83} & \hc{1.10} \\
\midrule
Changing bandit {\fontsize{4}{5}\selectfont\citep{xiong2023neural}} & Bandit & \hc{0.30} & \hc{0.30} & \hcu{0.29} & \hc{0.29} & \hcbu{0.29} & \hc{0.32} & \hc{0.31} & \hc{0.29} & \hc{0.40} & \hc{0.38} & \hc{0.35} & \hc{0.30} & \hc{0.39} & \hc{0.30} & \hc{0.38} & \hc{0.45} & \hc{0.30} & \hc{0.44} & \hc{0.69} \\
Decisions from experience {\fontsize{4}{5}\selectfont\citep{wulff2018sampling}} & Bandit & \hc{0.49} & \hc{0.49} & \hc{0.49} & \hc{0.49} & \hc{0.48} & \hc{0.50} & \hc{0.49} & \hc{0.48} & \hc{0.54} & \hc{0.52} & \hc{0.50} & \hc{0.49} & \hc{0.50} & \hc{0.49} & \hcbu{0.43} & \hcu{0.47} & \hc{0.37} & \hc{0.54} & \hcn \\
Drifting four-armed bandit {\fontsize{4}{5}\selectfont\citep{bahrami2020four}} & Bandit & \hc{0.73} & \hc{0.72} & \hc{0.71} & \hc{0.71} & \hcbu{0.71} & \hc{0.73} & \hc{0.72} & \hc{0.71} & \hc{0.83} & \hc{0.78} & \hc{0.74} & \hc{0.72} & \hc{0.74} & \hc{0.72} & \hc{0.88} & \hcu{0.71} & \hc{0.70} & \hc{0.90} & \hc{1.39} \\
Horizon task {\fontsize{4}{5}\selectfont\citep{feng2021dynamics}} & Bandit & \hc{0.23} & \hc{0.23} & \hc{0.22} & \hc{0.22} & \hcu{0.22} & \hc{0.26} & \hc{0.24} & \hcbu{0.22} & \hc{0.40} & \hc{0.34} & \hc{0.30} & \hc{0.22} & \hc{0.35} & \hc{0.22} & \hc{0.52} & \hc{0.40} & \hcd{0.40} & \hcd{0.36} & \hc{0.69} \\
Horizon task {\fontsize{4}{5}\selectfont\citep{sadeghiyeh2020temporal}} & Bandit & \hc{0.59} & \hc{0.58} & \hc{0.58} & \hc{0.58} & \hc{0.58} & \hc{0.59} & \hc{0.58} & \hc{0.58} & \hc{0.62} & \hc{0.62} & \hc{0.60} & \hc{0.58} & \hc{0.61} & \hc{0.58} & \hcu{0.52} & \hc{0.58} & \hcd{0.40} & \hcbud{0.36} & \hc{0.69} \\
Horizon task {\fontsize{4}{5}\selectfont\citep{somerville2017charting}} & Bandit & \hc{0.34} & \hc{0.34} & \hc{0.34} & \hcu{0.34} & \hcbu{0.33} & \hc{0.35} & \hc{0.34} & \hc{0.34} & \hc{0.48} & \hc{0.40} & \hc{0.38} & \hc{0.34} & \hc{0.40} & \hc{0.34} & \hc{0.52} & \hc{0.35} & \hcd{0.40} & \hcd{0.36} & \hc{0.69} \\
Horizon task {\fontsize{4}{5}\selectfont\citep{waltz2020differential}} & Bandit & \hc{0.15} & \hc{0.15} & \hc{0.15} & \hc{0.15} & \hc{0.15} & \hc{0.16} & \hc{0.16} & \hcbu{0.14} & \hc{0.26} & \hc{0.21} & \hc{0.18} & \hc{0.15} & \hc{0.19} & \hc{0.15} & \hc{0.52} & \hcu{0.15} & \hcd{0.40} & \hcd{0.36} & \hc{0.69} \\
Horizon task {\fontsize{4}{5}\selectfont\citep{wilson2014humans}} & Bandit & \hc{0.46} & \hc{0.46} & \hc{0.45} & \hc{0.45} & \hcu{0.45} & \hc{0.47} & \hc{0.46} & \hc{0.45} & \hc{0.54} & \hc{0.51} & \hc{0.49} & \hc{0.45} & \hc{0.52} & \hc{0.45} & \hc{0.52} & \hc{0.48} & \hcd{0.40} & \hcbud{0.36} & \hc{0.69} \\
Iowa gambling task {\fontsize{4}{5}\selectfont\citep{steingroever2015data}} & Bandit & \hc{0.93} & \hc{0.93} & \hc{0.91} & \hc{0.91} & \hc{0.91} & \hc{0.94} & \hc{0.92} & \hcu{0.91} & \hc{1.01} & \hc{0.98} & \hc{0.94} & \hc{0.92} & \hc{0.95} & \hc{0.92} & \hc{0.99} & \hcbu{0.91} & \hc{0.89} & \hc{1.16} & \hc{1.39} \\
Prob.\ instrumental learning {\fontsize{4}{5}\selectfont\citep{lefebvre2017behavioural}} & Bandit & \hc{0.50} & \hc{0.50} & \hcbu{0.49} & \hcu{0.49} & \hc{0.49} & \hc{0.51} & \hc{0.50} & \hc{0.49} & \hc{0.54} & \hc{0.53} & \hc{0.51} & \hc{0.50} & \hc{0.51} & \hc{0.50} & \hc{0.54} & \hc{0.50} & \hc{0.49} & \hc{0.50} & \hc{0.69} \\
Spatially correlated MAB {\fontsize{4}{5}\selectfont\citep{wu2018generalization}} & Bandit & \hc{2.01} & \hc{1.96} & \hc{1.93} & \hc{1.87} & \hcu{1.86} & \hc{2.16} & \hc{2.00} & \hc{1.93} & \hc{2.50} & \hc{2.30} & \hc{2.16} & \hc{1.95} & \hc{2.26} & \hc{2.06} & \hc{2.45} & \hcbu{1.82} & \hc{1.83} & \hc{2.76} & \hc{3.40} \\
Structured bandit {\fontsize{4}{5}\selectfont\citep{schulz2020finding}} & Bandit & \hc{0.67} & \hc{0.66} & \hc{0.65} & \hc{0.65} & \hcbu{0.64} & \hc{0.67} & \hc{0.66} & \hc{0.65} & \hc{0.77} & \hc{0.71} & \hc{0.68} & \hc{0.65} & \hc{0.73} & \hc{0.66} & \hc{0.81} & \hcu{0.64} & \hc{0.64} & \hc{1.05} & \hc{2.08} \\
Two-armed bandit {\fontsize{4}{5}\selectfont\citep{gershman2018deconstructing}} & Bandit & \hc{0.30} & \hc{0.30} & \hc{0.29} & \hcu{0.29} & \hcbu{0.29} & \hc{0.30} & \hc{0.29} & \hc{0.29} & \hc{0.37} & \hc{0.33} & \hc{0.31} & \hc{0.29} & \hc{0.31} & \hc{0.30} & \hc{0.38} & \hc{0.30} & \hc{0.30} & \hc{0.42} & \hc{0.69} \\
\midrule
Aversive learning {\fontsize{4}{5}\selectfont\citep{wise2019computational}} & Sup.\ learn. & \hc{4.71} & \hc{4.55} & \hc{4.27} & \hc{4.24} & \hcu{3.81} & \hc{4.58} & \hc{4.80} & \hcbu{3.66} & \hc{5.35} & \hc{5.12} & \hc{4.92} & \hc{3.93} & \hc{5.44} & \hc{4.67} & \hc{5.11} & \hc{4.13} & \hc{4.07} & \hcn & \hcn \\
Medin categorization {\fontsize{4}{5}\selectfont\citep{levering2020revisiting}} & Sup.\ learn. & \hc{0.51} & \hc{0.51} & \hc{0.50} & \hc{0.50} & \hcbu{0.50} & \hc{0.52} & \hc{0.50} & \hcu{0.50} & \hc{0.56} & \hc{0.54} & \hc{0.52} & \hc{0.50} & \hc{0.52} & \hc{0.51} & \hc{0.58} & \hc{0.50} & \hc{0.50} & \hc{0.53} & \hcn \\
Multiple-cue judgment {\fontsize{4}{5}\selectfont\citep{collsioo2023numerical}} & Sup.\ learn. & \hc{1.17} & \hc{1.16} & \hc{1.14} & \hc{1.14} & \hcu{1.13} & \hc{1.17} & \hc{1.15} & \hcbu{1.12} & \hc{1.59} & \hc{1.49} & \hc{1.40} & \hc{1.14} & \hc{1.67} & \hc{1.15} & \hc{1.28} & \hc{1.14} & \hc{1.12} & \hc{1.92} & \hc{2.20} \\
Shepard categorization {\fontsize{4}{5}\selectfont\citep{badham2017deficits}} & Sup.\ learn. & \hc{0.55} & \hc{0.55} & \hc{0.54} & \hc{0.53} & \hcbu{0.53} & \hc{0.56} & \hc{0.54} & \hcu{0.53} & \hc{0.62} & \hc{0.58} & \hc{0.56} & \hc{0.54} & \hc{0.56} & \hc{0.54} & \hc{0.58} & \hc{0.54} & \hc{0.54} & \hc{0.61} & \hc{0.69} \\
Weather prediction task {\fontsize{4}{5}\selectfont\citep{speekenbrink2008learning}} & Sup.\ learn. & \hc{0.55} & \hc{0.56} & \hc{0.54} & \hc{0.55} & \hc{0.55} & \hc{0.55} & \hcu{0.54} & \hc{0.56} & \hc{0.58} & \hc{0.56} & \hc{0.56} & \hc{0.55} & \hc{0.57} & \hcbu{0.53} & \hc{0.57} & \hc{0.56} & \hc{0.55} & \hc{0.63} & \hc{0.69} \\
\midrule
\textbf{Mean} &  & \hc{0.69} & \hc{0.68} & \hc{0.66} & \hc{0.66} & \hcbu{0.64} & \hc{0.70} & \hc{0.68} & \hcu{0.64} & \hc{0.89} & \hc{0.82} & \hc{0.73} & \hc{0.66} & \hc{0.77} & \hc{0.68} & \hc{0.88} & \hc{0.71} & \hc{0.69} & \hc{0.69} & \hc{0.98} \\
\bottomrule
\end{tabular}%
}
}%
\setlength{\ctrtblwd}{\wd\ctrtblbox}%
\ifdim\ctrtblwd>\dimexpr\textheight-8pt\relax
  \setlength{\ctrtblwd}{\dimexpr\textheight-8pt\relax}%
\fi
\resizebox{\ctrtblwd}{!}{\usebox{\ctrtblbox}}
\par
\begin{minipage}{\ctrtblwd}
\vspace{4pt}
\begin{center}\footnotesize
\colorbox{c1!45}{\strut\hspace{5pt}} \scriptsize$<$0.20 \quad
\colorbox{c2!40}{\strut\hspace{5pt}} \scriptsize 0.20--0.40 \quad
\colorbox{c3!40}{\strut\hspace{5pt}} \scriptsize 0.40--0.60 \quad
\colorbox{c4!45}{\strut\hspace{5pt}} \scriptsize 0.60--0.80 \quad
\colorbox{c5!55}{\strut\hspace{5pt}} \scriptsize 0.80--1.00 \quad
\colorbox{c6!50}{\strut\hspace{5pt}} \scriptsize 1.00--1.20 \quad
\colorbox{c7!45}{\strut\hspace{5pt}} \scriptsize 1.20--1.50 \quad
\colorbox{c8!40}{\strut\hspace{5pt}} \scriptsize 1.50--2.00 \quad
\colorbox{c9!35}{\strut\hspace{5pt}} \scriptsize$>$2.00
\end{center}
\vspace{-2pt}
{\tiny \underline{\textbf{Bold+underline}}\,=\,best; \underline{underline}\,=\,second-best (C$_p$ excluded; see below). Lower is better. ``---''\,=\,not available. All models finetuned at LoRA rank 16, bf16 inference. Reference columns (\citet{binz2025foundation}): L-70B$_p$\,=\,Llama-3.1-70B base, C$_r$\,=\,our reproduced Centaur-70B, C$_p$\,=\,published Centaur-70B, Cog.$_p$\,=\,domain-specific cognitive baseline. C$_p$ is included for reference but excluded from best/second-best marking, as it was evaluated under different software conditions. $^\dagger$Horizon and Two-step tasks are reported as single merged values by \citet{binz2025foundation}. Rounded to 2\,d.p.}
\end{minipage}
\captionsetup{width=\ctrtblwd}
\caption{\textbf{Per-experiment NLL on Psych-101 for all four finetuned families (bf16, LoRA rank 16), continued.} Cell colour encodes performance (green\,=\,better).}
\label{tab:psych101_pe_ft_all_cont}
\end{sidewaystable}

\begin{sidewaystable}
\centering
\sbox{\ctrtblbox}{%
{\tiny
\setlength{\tabcolsep}{2pt}
\renewcommand{\arraystretch}{1.15}
\begin{tabular}{@{}l l  r r r r r r   r r r r r r r r r r   r r  r@{}}
\toprule
& & \multicolumn{6}{c}{\texttt{\textbf{Qwen 2.5/3}} family} & \multicolumn{10}{c}{\texttt{\textbf{Llama 3/3.1/3.2}} family} & \multicolumn{2}{c}{\citet{binz2025foundation}} & \textbf{Chance} \\
\cmidrule(lr){3-8}\cmidrule(lr){9-18}
& & \multicolumn{2}{c}{Base} & \multicolumn{2}{c}{\texttt{Qwentaur}} & \multicolumn{1}{c}{Other cog.} & \multicolumn{1}{c}{Non-cog.} & \multicolumn{2}{c}{Base} & \multicolumn{2}{c}{\texttt{Llama-Centaur}} & \multicolumn{2}{c}{Other cog.} & \multicolumn{4}{c}{Non-cog.} & & \\
\cmidrule(lr){3-4}\cmidrule(lr){5-6}\cmidrule(lr){7-7}\cmidrule(lr){8-8}\cmidrule(lr){9-10}\cmidrule(lr){11-12}\cmidrule(lr){13-14}\cmidrule(lr){15-18}
\textbf{Experiment} & \textbf{Type} & 8B & 14B & 8B & 14B & \texttt{Socr.-14B} & \texttt{Herm.-14B} & 3B & 8B & 3B & 8B & \texttt{Be.FM-8B} & \texttt{Socr.-8B} & \texttt{Herm.-3B} & \texttt{Nemo.-4B} & \texttt{Herm.-8B} & \texttt{Nemo.-8B} & \texttt{C-70B}$_r$ & Cog.$_p$ & $\ln(k)$ \\
\midrule
Balloon analog risk task {\fontsize{4}{5}\selectfont\citep{frey2017risk}} & Decision & \hc{0.08} & \hc{0.08} & \hc{0.06} & \hcu{0.06} & \hc{5.87} & \hc{0.08} & \hc{0.08} & \hc{0.08} & \hc{0.07} & \hcbu{0.06} & \hc{0.10} & \hc{13.52} & \hc{0.09} & \hc{0.17} & \hc{0.09} & \hc{0.18} & \hc{0.07} & \hc{0.09} & \hc{0.69} \\
CPC18 {\fontsize{4}{5}\selectfont\citep{plonsky2018when}} & Decision & \hc{0.40} & \hc{0.40} & \hc{0.34} & \hcbu{0.33} & \hc{8.54} & \hc{0.40} & \hc{0.41} & \hc{0.39} & \hc{0.35} & \hcu{0.34} & \hc{0.47} & \hc{14.34} & \hc{0.44} & \hc{0.95} & \hc{0.42} & \hc{0.64} & \hc{0.35} & \hc{0.66} & \hc{0.69} \\
Columbia card task {\fontsize{4}{5}\selectfont\citep{frey2017risk}} & Decision & \hc{0.25} & \hc{0.25} & \hc{0.20} & \hcu{0.20} & \hc{6.72} & \hc{0.26} & \hc{0.27} & \hc{0.25} & \hc{0.20} & \hcbu{0.20} & \hc{0.26} & \hc{14.07} & \hc{0.27} & \hc{0.48} & \hc{0.26} & \hc{0.49} & \hc{0.21} & \hc{0.26} & \hc{0.69} \\
Decisions from description {\fontsize{4}{5}\selectfont\citep{wulff2018sampling}} & Decision & \hc{0.99} & \hc{1.01} & \hc{0.61} & \hc{0.60} & \hc{10.13} & \hc{1.02} & \hc{1.20} & \hc{0.96} & \hc{0.63} & \hcu{0.60} & \hc{1.28} & \hc{13.59} & \hc{1.34} & \hc{2.99} & \hc{1.17} & \hc{4.44} & \hcbu{0.59} & \hc{0.61} & \hc{0.69} \\
Experiential-symbolic task {\fontsize{4}{5}\selectfont\citep{garcia2023experiential}} & Decision & \hc{0.64} & \hc{0.57} & \hc{0.46} & \hcu{0.46} & \hc{7.86} & \hc{0.58} & \hc{0.69} & \hc{0.64} & \hc{0.47} & \hc{0.46} & \hc{0.73} & \hc{12.95} & \hc{0.79} & \hc{1.37} & \hc{0.67} & \hc{0.95} & \hcbu{0.46} & \hcn & \hcn \\
Gardening task {\fontsize{4}{5}\selectfont\citep{flesch2018comparing}} & Decision & \hc{0.47} & \hc{0.45} & \hc{0.38} & \hcu{0.38} & \hc{10.28} & \hc{0.45} & \hc{0.64} & \hc{0.48} & \hc{0.39} & \hcbu{0.38} & \hc{0.48} & \hc{16.26} & \hc{0.53} & \hc{0.61} & \hc{0.46} & \hc{0.64} & \hc{0.49} & \hc{0.91} & \hc{0.69} \\
Go/no-go {\fontsize{4}{5}\selectfont\citep{enkavi2019large}} & Decision & \hc{0.01} & \hc{0.00} & \hcbu{0.00} & \hcbu{0.00} & \hc{4.24} & \hc{0.00} & \hc{0.01} & \hc{0.00} & \hcbu{0.00} & \hcbu{0.00} & \hc{0.01} & \hc{12.55} & \hc{0.01} & \hc{0.04} & \hcu{0.00} & \hc{0.01} & \hcbu{0.00} & \hc{0.08} & \hc{0.00} \\
Intertemporal choice {\fontsize{4}{5}\selectfont\citep{ruggeri2022globalizability}} & Decision & \hc{0.69} & \hc{0.70} & \hc{0.44} & \hc{0.44} & \hc{12.23} & \hc{0.74} & \hc{0.75} & \hc{0.76} & \hcu{0.44} & \hc{0.44} & \hc{0.76} & \hc{13.82} & \hc{0.72} & \hc{2.75} & \hc{1.04} & \hc{2.72} & \hcbu{0.44} & \hc{0.66} & \hc{0.69} \\
Multi-attribute DM {\fontsize{4}{5}\selectfont\citep{hilbig2014generalized}} & Decision & \hc{0.19} & \hc{0.19} & \hc{0.08} & \hcu{0.08} & \hc{10.45} & \hc{0.20} & \hc{0.45} & \hc{0.23} & \hc{0.08} & \hc{0.08} & \hc{0.30} & \hc{13.86} & \hc{0.45} & \hc{0.61} & \hc{0.23} & \hc{0.54} & \hcbu{0.06} & \hc{0.19} & \hc{0.69} \\
Risky choice {\fontsize{4}{5}\selectfont\citep{krueger2024identifying}} & Decision & \hc{0.74} & \hc{0.67} & \hc{0.43} & \hcbu{0.42} & \hc{7.17} & \hc{0.67} & \hc{0.85} & \hc{0.76} & \hc{0.46} & \hcu{0.43} & \hc{0.85} & \hc{16.10} & \hc{0.94} & \hc{1.79} & \hc{0.82} & \hc{1.35} & \hc{0.43} & \hcn & \hcn \\
choices13k {\fontsize{4}{5}\selectfont\citep{peterson2021using}} & Decision & \hc{0.59} & \hc{0.55} & \hc{0.43} & \hcbu{0.43} & \hc{9.40} & \hc{0.53} & \hc{0.59} & \hc{0.55} & \hc{0.44} & \hcu{0.43} & \hc{0.68} & \hc{13.44} & \hc{0.65} & \hc{1.65} & \hc{0.60} & \hc{1.13} & \hc{0.43} & \hc{0.66} & \hc{0.69} \\
\midrule
Multi-task RL {\fontsize{4}{5}\selectfont\citep{tomov2021multitask}} & MDP & \hc{0.70} & \hc{0.68} & \hc{0.58} & \hcbu{0.57} & \hc{7.95} & \hc{0.66} & \hc{0.70} & \hc{0.66} & \hc{0.59} & \hc{0.57} & \hc{0.82} & \hc{14.70} & \hc{0.81} & \hc{1.20} & \hc{0.70} & \hc{1.06} & \hcu{0.57} & \hc{1.04} & \hc{1.10} \\
Tile-revealing task {\fontsize{4}{5}\selectfont\citep{kumar2023disentangling}} & MDP & \hc{2.73} & \hc{2.60} & \hc{0.07} & \hcu{0.00} & \hc{35.15} & \hc{2.62} & \hc{3.98} & \hc{2.92} & \hc{0.00} & \hcbu{0.00} & \hc{3.03} & \hc{65.75} & \hc{2.93} & \hc{3.48} & \hc{2.85} & \hc{3.37} & \hc{2.29} & \hcn & \hcn \\
Two-step task {\fontsize{4}{5}\selectfont\citep{kool2017cost}} & MDP & \hc{0.65} & \hc{0.63} & \hcu{0.53} & \hcbu{0.53} & \hc{7.90} & \hc{0.63} & \hc{0.62} & \hc{0.61} & \hc{0.53} & \hcbu{0.53} & \hc{0.71} & \hc{14.06} & \hc{0.67} & \hc{1.17} & \hc{0.67} & \hc{0.81} & \hc{0.53} & \hc{0.60} & \hc{0.69} \\
Two-step task {\fontsize{4}{5}\selectfont\citep{kool2016when}} & MDP & \hc{0.65} & \hc{0.63} & \hc{0.49} & \hcu{0.48} & \hc{8.82} & \hc{0.63} & \hc{0.62} & \hc{0.61} & \hc{0.50} & \hc{0.49} & \hc{0.72} & \hc{14.63} & \hc{0.68} & \hc{1.15} & \hc{0.67} & \hc{0.82} & \hcbu{0.48} & \hc{0.60} & \hc{0.69} \\
Two-step task {\fontsize{4}{5}\selectfont\citep{zorowitz2023data}} & MDP & \hc{0.61} & \hc{0.59} & \hcbu{0.51} & \hc{0.51} & \hc{7.68} & \hc{0.58} & \hc{0.60} & \hc{0.58} & \hcu{0.51} & \hc{0.51} & \hc{0.65} & \hc{13.18} & \hc{0.63} & \hc{1.03} & \hc{0.62} & \hc{0.82} & \hc{0.51} & \hc{0.60} & \hc{0.69} \\
Virtual subway network {\fontsize{4}{5}\selectfont\citep{tomov2020discovery}} & MDP & \hc{1.55} & \hc{1.57} & \hcu{1.13} & \hcbu{1.12} & \hc{11.39} & \hc{1.60} & \hc{1.60} & \hc{1.53} & \hc{1.16} & \hc{1.13} & \hc{1.61} & \hc{18.21} & \hc{1.69} & \hc{2.41} & \hc{1.64} & \hc{2.10} & \hc{1.16} & \hcn & \hc{1.61} \\
Zoopermarket {\fontsize{4}{5}\selectfont\citep{ludwig2023human}} & MDP & \hc{0.60} & \hc{0.59} & \hc{0.50} & \hcbu{0.48} & \hc{6.62} & \hc{0.59} & \hc{0.62} & \hc{0.60} & \hc{0.52} & \hc{0.50} & \hc{0.62} & \hc{13.75} & \hc{0.65} & \hc{1.49} & \hc{0.63} & \hc{1.09} & \hcu{0.50} & \hc{0.60} & \hc{0.69} \\
\midrule
Cond.\ assoc.\ learning {\fontsize{4}{5}\selectfont\citep{collins2014working}} & Memory & \hc{0.69} & \hc{0.61} & \hc{0.52} & \hcbu{0.51} & \hc{10.04} & \hc{0.59} & \hc{0.73} & \hc{0.70} & \hc{0.53} & \hc{0.52} & \hc{0.72} & \hc{16.16} & \hc{0.80} & \hc{0.92} & \hc{0.68} & \hc{0.95} & \hcu{0.52} & \hc{0.86} & \hc{1.10} \\
Digit span {\fontsize{4}{5}\selectfont\citep{enkavi2019large}} & Memory & \hc{0.70} & \hc{0.69} & \hc{0.57} & \hcbu{0.57} & \hc{2.96} & \hc{0.70} & \hc{0.84} & \hc{0.75} & \hc{0.59} & \hc{0.58} & \hc{0.83} & \hc{4.65} & \hc{0.87} & \hc{2.16} & \hc{0.83} & \hc{1.02} & \hcu{0.57} & \hc{0.94} & \hc{2.30} \\
Episodic long-term memory {\fontsize{4}{5}\selectfont\citep{popov2023intent}} & Memory & \hc{1.30} & \hc{1.18} & \hc{0.94} & \hcu{0.90} & \hc{9.08} & \hc{1.21} & \hc{1.65} & \hc{1.43} & \hc{1.02} & \hc{0.92} & \hc{1.52} & \hc{17.12} & \hc{1.82} & \hc{3.07} & \hc{1.72} & \hc{2.22} & \hcbu{0.89} & \hcn & \hc{0.69} \\
N-back {\fontsize{4}{5}\selectfont\citep{enkavi2019large}} & Memory & \hc{0.55} & \hc{0.53} & \hc{0.41} & \hcu{0.40} & \hc{9.24} & \hc{0.53} & \hc{0.57} & \hc{0.57} & \hc{0.42} & \hc{0.41} & \hc{0.57} & \hc{14.42} & \hc{0.59} & \hc{0.79} & \hc{0.57} & \hc{0.75} & \hcbu{0.40} & \hc{0.58} & \hc{0.69} \\
Recall and recognition {\fontsize{4}{5}\selectfont\citep{cox2018information}} & Memory & \hc{1.41} & \hc{1.36} & \hc{1.12} & \hcu{1.09} & \hc{9.33} & \hc{1.38} & \hc{1.56} & \hc{1.37} & \hc{1.14} & \hc{1.10} & \hc{1.53} & \hc{24.31} & \hc{1.65} & \hc{4.16} & \hc{1.61} & \hc{2.60} & \hcbu{1.07} & \hcn & \hcn \\
Recent probes {\fontsize{4}{5}\selectfont\citep{enkavi2019large}} & Memory & \hc{0.41} & \hc{0.35} & \hc{0.26} & \hcu{0.26} & \hc{9.92} & \hc{0.34} & \hc{0.56} & \hc{0.34} & \hc{0.26} & \hcbu{0.26} & \hc{0.40} & \hc{14.16} & \hc{0.61} & \hc{1.00} & \hc{0.39} & \hc{1.06} & \hc{0.26} & \hc{0.39} & \hc{0.69} \\
\bottomrule
\end{tabular}%
}
}%
\setlength{\ctrtblwd}{\wd\ctrtblbox}%
\ifdim\ctrtblwd>\dimexpr\textheight-8pt\relax
  \setlength{\ctrtblwd}{\dimexpr\textheight-8pt\relax}%
\fi
\resizebox{\ctrtblwd}{!}{\usebox{\ctrtblbox}}
\par
\begin{minipage}{\ctrtblwd}
\vspace{4pt}
\begin{center}\footnotesize
\colorbox{c1!45}{\strut\hspace{5pt}} \scriptsize$<$0.20 \quad
\colorbox{c2!40}{\strut\hspace{5pt}} \scriptsize 0.20--0.40 \quad
\colorbox{c3!40}{\strut\hspace{5pt}} \scriptsize 0.40--0.60 \quad
\colorbox{c4!45}{\strut\hspace{5pt}} \scriptsize 0.60--0.80 \quad
\colorbox{c5!55}{\strut\hspace{5pt}} \scriptsize 0.80--1.00 \quad
\colorbox{c6!50}{\strut\hspace{5pt}} \scriptsize 1.00--1.20 \quad
\colorbox{c7!45}{\strut\hspace{5pt}} \scriptsize 1.20--1.50 \quad
\colorbox{c8!40}{\strut\hspace{5pt}} \scriptsize 1.50--2.00 \quad
\colorbox{c9!35}{\strut\hspace{5pt}} \scriptsize$>$2.00
\end{center}
\vspace{-2pt}
{\tiny \underline{\textbf{Bold+underline}}\,=\,best; \underline{underline}\,=\,second-best. Lower is better. ``---''\,=\,not available. Models are grouped by base-model family. ``Base'' = pretrained model without finetuning (bf16); ``Other cog.'' = other cognitive/behavioural foundation models finetuned on non-Psych-101 behavioural data; ``Non-cog.'' = models finetuned for general instruction-following or reasoning. \texttt{Herm.} = Hermes \citep{teknium2024hermes3technicalreport}; \texttt{Nemo.} = Nemotron \citep{bercovich2025nemotron}; \texttt{Socr.} = Socrates \citep{kolluri2025finetuning}; \texttt{Be.FM} = \citet{xie2025fm}. Base model versions differ within families: \texttt{Qwentaur} uses Qwen3; \texttt{Socr.-14B} uses Qwen2.5; \texttt{Herm.-14B} uses Qwen2.5. \texttt{Llama-Centaur} uses Llama-3.1/3.2; \texttt{Socr.-8B} uses Llama-3; \texttt{Be.FM} uses Llama-3.1.}
\end{minipage}
\captionsetup{width=\ctrtblwd}
\caption{\textbf{Comparison of cognitively finetuned models with other behavioural foundation models and non-cognitive controls on Psych-101, grouped by base-model family.} Within each family, \texttt{Qwentaur}/\texttt{Llama-Centaur} (cognitive SFT on Psych-101) consistently outperform both other behavioural finetuned models and non-cognitive controls, demonstrating that alignment with human behavioural patterns requires cognitive finetuning on structured experimental data, not merely finetuning per se.}
\label{tab:psych101_pe_control}
\end{sidewaystable}

\begin{sidewaystable}
\ContinuedFloat
\centering
\sbox{\ctrtblbox}{%
{\tiny
\setlength{\tabcolsep}{2pt}
\renewcommand{\arraystretch}{1.15}
\begin{tabular}{@{}l l  r r r r r r   r r r r r r r r r r   r r  r@{}}
\toprule
& & \multicolumn{6}{c}{\texttt{\textbf{Qwen 2.5/3}} family} & \multicolumn{10}{c}{\texttt{\textbf{Llama 3/3.1/3.2}} family} & \multicolumn{2}{c}{\citet{binz2025foundation}} & \textbf{Chance} \\
\cmidrule(lr){3-8}\cmidrule(lr){9-18}
& & \multicolumn{2}{c}{Base} & \multicolumn{2}{c}{\texttt{Qwentaur}} & \multicolumn{1}{c}{Other cog.} & \multicolumn{1}{c}{Non-cog.} & \multicolumn{2}{c}{Base} & \multicolumn{2}{c}{\texttt{Llama-Centaur}} & \multicolumn{2}{c}{Other cog.} & \multicolumn{4}{c}{Non-cog.} & & \\
\cmidrule(lr){3-4}\cmidrule(lr){5-6}\cmidrule(lr){7-7}\cmidrule(lr){8-8}\cmidrule(lr){9-10}\cmidrule(lr){11-12}\cmidrule(lr){13-14}\cmidrule(lr){15-18}
\textbf{Experiment} & \textbf{Type} & 8B & 14B & 8B & 14B & \texttt{Socr.-14B} & \texttt{Herm.-14B} & 3B & 8B & 3B & 8B & \texttt{Be.FM-8B} & \texttt{Socr.-8B} & \texttt{Herm.-3B} & \texttt{Nemo.-4B} & \texttt{Herm.-8B} & \texttt{Nemo.-8B} & \texttt{C-70B}$_r$ & Cog.$_p$ & $\ln(k)$ \\
\midrule
Grammar judgement {\fontsize{4}{5}\selectfont\citep{jansen2021rational}} & Misc. & \hc{1.89} & \hc{1.91} & \hc{1.44} & \hcu{1.43} & \hc{10.37} & \hc{1.94} & \hc{2.16} & \hc{2.12} & \hc{1.45} & \hc{1.44} & \hc{2.11} & \hc{12.88} & \hc{2.27} & \hc{3.90} & \hc{2.53} & \hc{3.12} & \hc{1.44} & \hcbu{1.41} & \hcn \\
Probabilistic reasoning {\fontsize{4}{5}\selectfont\citep{zhu2020bayesian}} & Misc. & \hc{2.57} & \hc{2.58} & \hcu{2.35} & \hcbu{2.34} & \hc{2.92} & \hc{2.59} & \hc{2.80} & \hc{2.71} & \hc{2.47} & \hc{2.43} & \hc{2.84} & \hc{3.40} & \hc{2.98} & \hc{4.05} & \hc{2.97} & \hc{3.23} & \hc{2.40} & \hcn & \hcn \\
Serial reaction time task {\fontsize{4}{5}\selectfont\citep{wu2023chunking}} & Misc. & \hc{0.19} & \hc{0.19} & \hcu{0.17} & \hcbu{0.17} & \hc{5.09} & \hc{0.19} & \hc{0.23} & \hc{0.20} & \hc{0.17} & \hc{0.17} & \hc{0.25} & \hc{14.43} & \hc{0.21} & \hc{0.25} & \hc{0.20} & \hc{0.26} & \hc{0.18} & \hc{0.20} & \hc{1.39} \\
THINGS odd-one-out {\fontsize{4}{5}\selectfont\citep{hebart2023things}} & Misc. & \hc{1.15} & \hc{1.13} & \hc{0.79} & \hcbu{0.79} & \hc{12.15} & \hc{1.15} & \hc{1.18} & \hc{1.16} & \hc{0.82} & \hcu{0.79} & \hc{1.19} & \hc{14.31} & \hc{1.24} & \hc{2.13} & \hc{1.28} & \hc{1.75} & \hc{0.80} & \hc{0.83} & \hc{1.10} \\
\midrule
Changing bandit {\fontsize{4}{5}\selectfont\citep{xiong2023neural}} & Bandit & \hc{0.35} & \hc{0.34} & \hcu{0.29} & \hcbu{0.29} & \hc{7.52} & \hc{0.34} & \hc{0.90} & \hc{0.39} & \hc{0.31} & \hc{0.29} & \hc{0.42} & \hc{14.76} & \hc{0.39} & \hc{0.78} & \hc{0.36} & \hc{0.53} & \hc{0.45} & \hc{0.44} & \hc{0.69} \\
Decisions from experience {\fontsize{4}{5}\selectfont\citep{wulff2018sampling}} & Bandit & \hc{0.78} & \hc{0.61} & \hc{0.49} & \hc{0.48} & \hc{7.18} & \hc{0.64} & \hc{0.71} & \hc{0.65} & \hc{0.49} & \hcu{0.48} & \hc{0.83} & \hc{13.74} & \hc{0.69} & \hc{2.29} & \hc{0.88} & \hc{1.78} & \hcbu{0.47} & \hc{0.54} & \hcn \\
Drifting four-armed bandit {\fontsize{4}{5}\selectfont\citep{bahrami2020four}} & Bandit & \hc{0.91} & \hc{0.91} & \hc{0.71} & \hcbu{0.71} & \hc{8.43} & \hc{0.90} & \hc{0.94} & \hc{0.90} & \hc{0.72} & \hc{0.71} & \hc{0.99} & \hc{13.91} & \hc{0.99} & \hc{2.10} & \hc{0.96} & \hc{1.78} & \hcu{0.71} & \hc{0.90} & \hc{1.39} \\
Horizon task {\fontsize{4}{5}\selectfont\citep{feng2021dynamics}} & Bandit & \hc{0.36} & \hc{0.34} & \hc{0.22} & \hcu{0.22} & \hc{7.55} & \hc{0.35} & \hc{0.67} & \hc{0.44} & \hc{0.24} & \hcbu{0.22} & \hc{0.39} & \hc{15.10} & \hc{0.43} & \hc{0.65} & \hc{0.36} & \hc{0.62} & \hc{0.40} & \hc{0.36} & \hc{0.69} \\
Horizon task {\fontsize{4}{5}\selectfont\citep{sadeghiyeh2020temporal}} & Bandit & \hc{0.65} & \hc{0.65} & \hc{0.58} & \hcu{0.58} & \hc{8.62} & \hc{0.66} & \hc{0.68} & \hc{0.66} & \hc{0.58} & \hc{0.58} & \hc{0.69} & \hc{14.94} & \hc{0.73} & \hc{1.10} & \hc{0.69} & \hc{0.93} & \hc{0.58} & \hcbu{0.36} & \hc{0.69} \\
Horizon task {\fontsize{4}{5}\selectfont\citep{somerville2017charting}} & Bandit & \hc{0.47} & \hc{0.49} & \hcu{0.34} & \hcbu{0.33} & \hc{8.49} & \hc{0.49} & \hc{0.57} & \hc{0.51} & \hc{0.34} & \hc{0.34} & \hc{0.53} & \hc{15.18} & \hc{0.57} & \hc{0.86} & \hc{0.47} & \hc{0.74} & \hc{0.35} & \hc{0.36} & \hc{0.69} \\
Horizon task {\fontsize{4}{5}\selectfont\citep{waltz2020differential}} & Bandit & \hc{0.27} & \hc{0.24} & \hc{0.15} & \hc{0.15} & \hc{7.82} & \hc{0.26} & \hc{0.37} & \hc{0.30} & \hc{0.16} & \hcbu{0.14} & \hc{0.33} & \hc{15.42} & \hc{0.35} & \hc{0.54} & \hc{0.26} & \hc{0.54} & \hcu{0.15} & \hc{0.36} & \hc{0.69} \\
Horizon task {\fontsize{4}{5}\selectfont\citep{wilson2014humans}} & Bandit & \hc{0.53} & \hc{0.54} & \hc{0.45} & \hcu{0.45} & \hc{8.06} & \hc{0.54} & \hc{0.62} & \hc{0.55} & \hc{0.46} & \hc{0.45} & \hc{0.57} & \hc{15.39} & \hc{0.62} & \hc{0.89} & \hc{0.56} & \hc{0.79} & \hc{0.48} & \hcbu{0.36} & \hc{0.69} \\
Iowa gambling task {\fontsize{4}{5}\selectfont\citep{steingroever2015data}} & Bandit & \hc{1.07} & \hc{1.07} & \hc{0.91} & \hc{0.91} & \hc{8.45} & \hc{1.06} & \hc{1.06} & \hc{1.04} & \hc{0.92} & \hcu{0.91} & \hc{1.11} & \hc{13.23} & \hc{1.13} & \hc{2.27} & \hc{1.09} & \hc{1.60} & \hcbu{0.91} & \hc{1.16} & \hc{1.39} \\
Prob.\ instrumental learning {\fontsize{4}{5}\selectfont\citep{lefebvre2017behavioural}} & Bandit & \hc{0.59} & \hc{0.56} & \hcbu{0.49} & \hc{0.49} & \hc{8.30} & \hc{0.53} & \hc{0.58} & \hc{0.56} & \hc{0.50} & \hcu{0.49} & \hc{0.61} & \hc{15.92} & \hc{0.68} & \hc{1.16} & \hc{0.54} & \hc{0.86} & \hc{0.50} & \hc{0.50} & \hc{0.69} \\
Spatially correlated MAB {\fontsize{4}{5}\selectfont\citep{wu2018generalization}} & Bandit & \hc{2.44} & \hc{2.46} & \hc{1.87} & \hcu{1.86} & \hc{3.14} & \hc{2.48} & \hc{2.64} & \hc{2.58} & \hc{2.00} & \hc{1.93} & \hc{2.82} & \hc{5.43} & \hc{2.64} & \hc{3.93} & \hc{2.78} & \hc{3.19} & \hcbu{1.82} & \hc{2.76} & \hc{3.40} \\
Structured bandit {\fontsize{4}{5}\selectfont\citep{schulz2020finding}} & Bandit & \hc{0.87} & \hc{0.83} & \hc{0.65} & \hcbu{0.64} & \hc{1.57} & \hc{0.80} & \hc{0.89} & \hc{0.85} & \hc{0.66} & \hc{0.65} & \hc{0.98} & \hc{2.12} & \hc{0.91} & \hc{1.41} & \hc{0.90} & \hc{1.08} & \hcu{0.64} & \hc{1.05} & \hc{2.08} \\
Two-armed bandit {\fontsize{4}{5}\selectfont\citep{gershman2018deconstructing}} & Bandit & \hc{0.39} & \hc{0.39} & \hcu{0.29} & \hcbu{0.29} & \hc{7.30} & \hc{0.41} & \hc{0.43} & \hc{0.40} & \hc{0.29} & \hc{0.29} & \hc{0.46} & \hc{13.99} & \hc{0.47} & \hc{0.98} & \hc{0.44} & \hc{0.73} & \hc{0.30} & \hc{0.42} & \hc{0.69} \\
\midrule
Aversive learning {\fontsize{4}{5}\selectfont\citep{wise2019computational}} & Sup.\ learn. & \hc{4.89} & \hc{4.75} & \hc{4.24} & \hcu{3.81} & \hc{7.19} & \hc{4.78} & \hc{5.52} & \hc{5.25} & \hc{4.80} & \hcbu{3.66} & \hc{5.49} & \hc{18.42} & \hc{5.52} & \hc{7.03} & \hc{5.33} & \hc{6.14} & \hc{4.13} & \hcn & \hcn \\
Medin categorization {\fontsize{4}{5}\selectfont\citep{levering2020revisiting}} & Sup.\ learn. & \hc{0.59} & \hc{0.55} & \hc{0.50} & \hcbu{0.50} & \hc{11.00} & \hc{0.56} & \hc{0.60} & \hc{0.57} & \hc{0.50} & \hcu{0.50} & \hc{0.63} & \hc{13.53} & \hc{0.63} & \hc{1.15} & \hc{0.62} & \hc{0.90} & \hc{0.50} & \hc{0.53} & \hcn \\
Multiple-cue judgment {\fontsize{4}{5}\selectfont\citep{collsioo2023numerical}} & Sup.\ learn. & \hc{1.30} & \hc{1.26} & \hc{1.14} & \hcu{1.13} & \hc{6.92} & \hc{1.27} & \hc{1.50} & \hc{1.32} & \hc{1.15} & \hcbu{1.12} & \hc{1.38} & \hc{18.06} & \hc{1.51} & \hc{2.56} & \hc{1.37} & \hc{1.85} & \hc{1.14} & \hc{1.92} & \hc{2.20} \\
Shepard categorization {\fontsize{4}{5}\selectfont\citep{badham2017deficits}} & Sup.\ learn. & \hc{0.58} & \hc{0.59} & \hc{0.53} & \hcbu{0.53} & \hc{10.13} & \hc{0.59} & \hc{0.62} & \hc{0.60} & \hc{0.54} & \hcu{0.53} & \hc{0.60} & \hc{12.66} & \hc{0.64} & \hc{0.92} & \hc{0.63} & \hc{0.76} & \hc{0.54} & \hc{0.61} & \hc{0.69} \\
Weather prediction task {\fontsize{4}{5}\selectfont\citep{speekenbrink2008learning}} & Sup.\ learn. & \hc{0.58} & \hc{0.58} & \hcu{0.55} & \hc{0.55} & \hc{10.65} & \hc{0.56} & \hc{0.59} & \hc{0.58} & \hcbu{0.54} & \hc{0.56} & \hc{0.62} & \hc{17.01} & \hc{0.63} & \hc{0.89} & \hc{0.61} & \hc{0.77} & \hc{0.56} & \hc{0.63} & \hc{0.69} \\
\midrule
\textbf{Mean (all 46)} & & \hc{0.89} & \hc{0.87} & \hc{0.66} & \hcbu{0.64} & \hc{8.65} & \hc{0.87} & \hc{1.02} & \hc{0.92} & \hc{0.68} & \hcu{0.64} & \hc{0.99} & \hc{15.07} & \hc{1.02} & \hc{1.72} & \hc{0.98} & \hc{1.43} & \hc{0.71} & \hc{0.69} & \hc{0.98} \\
\bottomrule
\end{tabular}%
}
}%
\setlength{\ctrtblwd}{\wd\ctrtblbox}%
\ifdim\ctrtblwd>\dimexpr\textheight-8pt\relax
  \setlength{\ctrtblwd}{\dimexpr\textheight-8pt\relax}%
\fi
\resizebox{\ctrtblwd}{!}{\usebox{\ctrtblbox}}
\par
\begin{minipage}{\ctrtblwd}
\vspace{4pt}
\begin{center}\footnotesize
\colorbox{c1!45}{\strut\hspace{5pt}} \scriptsize$<$0.20 \quad
\colorbox{c2!40}{\strut\hspace{5pt}} \scriptsize 0.20--0.40 \quad
\colorbox{c3!40}{\strut\hspace{5pt}} \scriptsize 0.40--0.60 \quad
\colorbox{c4!45}{\strut\hspace{5pt}} \scriptsize 0.60--0.80 \quad
\colorbox{c5!55}{\strut\hspace{5pt}} \scriptsize 0.80--1.00 \quad
\colorbox{c6!50}{\strut\hspace{5pt}} \scriptsize 1.00--1.20 \quad
\colorbox{c7!45}{\strut\hspace{5pt}} \scriptsize 1.20--1.50 \quad
\colorbox{c8!40}{\strut\hspace{5pt}} \scriptsize 1.50--2.00 \quad
\colorbox{c9!35}{\strut\hspace{5pt}} \scriptsize$>$2.00
\end{center}
\vspace{-2pt}
{\tiny \underline{\textbf{Bold+underline}}\,=\,best; \underline{underline}\,=\,second-best. Lower is better. ``---''\,=\,not available. Models are grouped by base-model family. ``Base'' = pretrained model without finetuning (bf16); ``Other cog.'' = other cognitive/behavioural foundation models finetuned on non-Psych-101 behavioural data; ``Non-cog.'' = models finetuned for general instruction-following or reasoning. \texttt{Herm.} = Hermes \citep{teknium2024hermes3technicalreport}; \texttt{Nemo.} = Nemotron \citep{bercovich2025nemotron}; \texttt{Socr.} = Socrates \citep{kolluri2025finetuning}; \texttt{Be.FM} = \citet{xie2025fm}. Base model versions differ within families: \texttt{Qwentaur} uses Qwen3; \texttt{Socr.-14B} uses Qwen2.5; \texttt{Herm.-14B} uses Qwen2.5. \texttt{Llama-Centaur} uses Llama-3.1/3.2; \texttt{Socr.-8B} uses Llama-3; \texttt{Be.FM} uses Llama-3.1.}
\end{minipage}
\captionsetup{width=\ctrtblwd}
\caption{\textbf{Comparison of cognitively finetuned models with other behavioural foundation models and non-cognitive controls on Psych-101, grouped by base-model family (continued).}}
\label{tab:psych101_pe_control_cont}
\end{sidewaystable}

\clearpage

\subsection{Psych-201 (out-of-distribution)}
\label{app:full_results_psych201}

While Psych-101 serves as the in-distribution evaluation, Psych-201 \citep{binz2026posttrainingmakeslargelanguage} provides a stronger out-of-distribution test: models must generalise to entirely unseen experiments that were not part of the Psych-101 training set. Our analysis is based on Psych-201-RT, a subset of the Psych-201 dataset (\url{https://hf.co/datasets/socius/Psych-201-RT}). Table~\ref{tab:psych201_normalised} summarises performance by task type using the normalised metric, restricted to the 15 of 18 experiments with a well-defined discrete response space. Table~\ref{tab:psych201_pe_bf16_base_v_ft} then reports base versus fine-tuned NLL per experiment for the \texttt{Qwen3} and \texttt{Llama} families, isolating the effect of cognitive fine-tuning at each scale, and Table~\ref{tab:psych201_pe_ft_all} gives per-experiment NLL for all four fine-tuned families side by side. Both per-experiment tables cover all 18 experiments, including the 3 excluded from normalisation.

The 18 experiments in Psych-201-RT span four task types: decision-making (3), Markov decision processes (2), multi-armed bandits (5), and miscellaneous (8).
Of these, 15 present discrete response options and admit a well-defined chance-level baseline $\ln(k)$.

The majority of included experiments are binary choice tasks ($k = 2$, $\ln(k) \approx 0.693$): the experience-description choice task \citep{anllo2024weird} and the loss aversion task \citep{spektor2024lossaversion} (both decision-making); both two-step tasks \citep{castrorodrigues2022twostep,shahar2019twosteptask} (Markov decision processes); the range adaptation RL task \citep{bavard2021range}, the explore-exploit task \citep{fan2023trait}, the counterfactual learning task \citep{palminteri2017confirmation}, and the contextual reinforcement learning task \citep{vandendriessche2023depression} (all multi-armed bandits); and the Navon task \citep{busch2024navon}, the compound word processing task \citep{gunther2020ts}, the grammaticality judgement task \citep{dentella2023grammaticality}, and the math arithmetic task \citep{xu2023augmenting} (all miscellaneous).

Several of these experiments use more than two unique response labels across the session (e.g., eight symbol pairs in the counterfactual learning task; \citealp{palminteri2017confirmation}), but each individual trial presents exactly two options; we assign the per-trial $k = 2$ in these cases.

Three experiments have $k > 2$: the Stroop task \citep{busch2024stroop} (miscellaneous; $k = 3$, $\ln(k) \approx 1.099$) presents three response options per trial; the pragmatic language task \citep{franke2024bayesian} and the reference game \citep{franke2016reasoning} (both miscellaneous) each present four options per trial drawn from a larger label pool, yielding $k = 4$ ($\ln(k) \approx 1.386$).
Note that the two-step task of \citet{castrorodrigues2022twostep} also presents four directional arrows on every trial ($k = 4$, $\ln(k) \approx 1.386$), while the two-step task of \citet{shahar2019twosteptask} presents two options per stage ($k = 2$).

\paragraph{Experiment exclusions}

Of the 18 experiments in Psych-201-RT, 3 are excluded from the normalised analysis (Table~\ref{tab:psych201_normalised}).
These fall into two categories.

\textit{Mixed response types.}
The risky decision and happiness task \citep{rutledge2023happiness} (decision-making) interleaves binary lottery choices with free numeric happiness ratings on a 0--100 scale, producing a mixed response space within a single session for which no single $k$ can be defined.

\textit{Continuous responses.}
The scalar inference task \citep{tsvilodub2023xorsome} (miscellaneous) asks participants to provide likelihood ratings via an adjustable slider, yielding multiple distinct integer values in a single participant -- effectively a continuous response with no well-defined discrete action space.
Similarly, the aversive reversal learning task \citep{zika2023traitanxiety} (multi-armed bandits) elicits shock probability estimates on a 0--1 scale, producing multiple distinct values in a single participant, confirming that the response space is continuous rather than categorical.

All three experiments retain their raw NLL values in the full per-experiment results (Table~\ref{tab:psych201_pe_bf16_base_v_ft}) but are excluded from any normalised $(\ln k - \mathrm{NLL})/\ln k$ computations.

\clearpage
\newpage

\clearpage

\begin{table}[t]
\centering
\resizebox{\columnwidth}{!}{%
\renewcommand{\arraystretch}{1.2}
\begin{tabular}{@{}l   r r r r r   r r r   r r r r   r r   r r r   r@{}}
\toprule
& \multicolumn{5}{c}{\textbf{\texttt{Qwentaur}}} & \multicolumn{3}{c}{\textbf{\texttt{Llama-Centaur}}} & \multicolumn{4}{c}{\textbf{\texttt{Smoltaur}}} & \multicolumn{2}{c}{\textbf{\texttt{Olmotaur}}} & \multicolumn{3}{c}{\textbf{Base}} & \multicolumn{1}{c}{\citet{binz2025foundation}} \\
\cmidrule(lr){2-6}\cmidrule(lr){7-9}\cmidrule(lr){10-13}\cmidrule(lr){14-15}\cmidrule(lr){16-18}\cmidrule(lr){19-19}
\textbf{Task type} & \texttt{0.6B} & \texttt{1.7B} & \texttt{4B} & \texttt{8B} & \texttt{14B} & \texttt{1B} & \texttt{3B} & \texttt{8B} & \texttt{0.1B} & \texttt{0.4B} & \texttt{1.7B} & \texttt{3B} & \texttt{1B} & \texttt{7B} & \texttt{Q-8B} & \texttt{Q-14B} & \texttt{L-8B} & \texttt{Centaur-70B} \\
\midrule
Decision (2) & \hcR{0.30} & \hcR{0.32} & \hcRu{0.35} & \hcR{0.34} & \hcRbu{0.37} & \hcR{0.27} & \hcR{0.31} & \hcR{0.33} & \hcR{0.20} & \hcR{0.21} & \hcR{0.24} & \hcR{0.34} & \hcR{0.21} & \hcR{0.31} & \hcR{0.28} & \hcR{0.29} & \hcR{0.28} & \hcR{0.33} \\
MDP (2) & \hcR{0.56} & \hcR{0.56} & \hcR{0.58} & \hcRu{0.58} & \hcRbu{0.58} & \hcR{0.48} & \hcR{0.53} & \hcR{0.57} & \hcR{0.39} & \hcR{0.46} & \hcR{0.52} & \hcR{0.57} & \hcR{0.42} & \hcR{0.57} & \hcR{0.51} & \hcR{0.53} & \hcR{0.53} & \hcR{0.50} \\
Bandit (4) & \hcR{0.32} & \hcR{0.35} & \hcR{0.37} & \hcR{0.36} & \hcRu{0.38} & \hcR{0.33} & \hcR{0.33} & \hcR{0.36} & \hcR{0.24} & \hcR{0.23} & \hcR{0.32} & \hcR{0.34} & \hcR{0.28} & \hcR{0.34} & \hcR{0.22} & \hcR{0.25} & \hcR{0.25} & \hcRbu{0.39} \\
Misc. (7) & \hcR{0.28} & \hcR{0.31} & \hcR{0.45} & \hcR{0.46} & \hcRu{0.49} & \hcR{0.14} & \hcR{0.30} & \hcR{0.40} & \hcR{0.03} & \hcR{0.11} & \hcR{0.18} & \hcR{0.36} & \hcR{0.09} & \hcR{0.35} & \hcR{0.37} & \hcR{0.41} & \hcR{0.29} & \hcRbu{0.52} \\
\midrule
\textbf{Mean (15)} & \hcR{0.33} & \hcR{0.35} & \hcR{0.43} & \hcR{0.43} & \hcRu{0.45} & \hcR{0.26} & \hcR{0.34} & \hcR{0.40} & \hcR{0.16} & \hcR{0.20} & \hcR{0.27} & \hcR{0.38} & \hcR{0.20} & \hcR{0.37} & \hcR{0.34} & \hcR{0.37} & \hcR{0.31} & \hcRbu{0.46} \\
\bottomrule
\end{tabular}%
}
\vspace{2pt}
{\tiny\centering
\colorbox{c9!35}{\strut\,} $<$0.05 ~
\colorbox{c8!40}{\strut\,} 0.05--0.12 ~
\colorbox{c7!45}{\strut\,} 0.12--0.18 ~
\colorbox{c6!50}{\strut\,} 0.18--0.25 ~
\colorbox{c5!55}{\strut\,} 0.25--0.32 ~
\colorbox{c4!45}{\strut\,} 0.32--0.40 ~
\colorbox{c3!40}{\strut\,} 0.40--0.48 ~
\colorbox{c2!40}{\strut\,} 0.48--0.55 ~
\colorbox{c1!45}{\strut\,} $\geq$0.55\par}
\caption{Fraction of available information captured above chance, $(\ln k - \mathrm{NLL}) \,/\, \ln k$, by task type on Psych-201 (out-of-distribution). A value of 0 indicates chance-level performance; 1 indicates perfect prediction. Restricted to 15 experiments (of 18) with a well-defined discrete response space ($\ln k > 0$); 3 experiments with continuous or mixed responses are excluded. \underline{\textbf{Bold+underline}} marks the best model, \underline{underline} the second-best.}
\label{tab:psych201_normalised}
\end{table}
\clearpage

\begin{sidewaystable}[p]
\centering
{\tiny
\setlength{\tabcolsep}{3pt}
\renewcommand{\arraystretch}{1.15}
\begin{tabular}{@{}l l  r@{\;\;}r  r@{\;\;}r  r@{\;\;}r  r@{\;\;}r  r@{\;\;}r  r@{\;\;}r  r@{\;\;}r  r@{\;\;}r  r  r@{}}
\toprule
& & \multicolumn{10}{c}{\texttt{\textbf{Qwen3}} family} & \multicolumn{6}{c}{\texttt{\textbf{Llama-3.1/3.2}} family} & \multicolumn{1}{c}{\citet{binz2025foundation}} & \textbf{Chance} \\
\cmidrule(lr){3-12}\cmidrule(lr){13-18}
& & \multicolumn{2}{c}{0.6B} & \multicolumn{2}{c}{1.7B} & \multicolumn{2}{c}{4B} & \multicolumn{2}{c}{8B} & \multicolumn{2}{c}{14B} & \multicolumn{2}{c}{1B} & \multicolumn{2}{c}{3B} & \multicolumn{2}{c}{8B} & & \\
\cmidrule(lr){3-4} \cmidrule(lr){5-6} \cmidrule(lr){7-8} \cmidrule(lr){9-10} \cmidrule(lr){11-12} \cmidrule(lr){13-14} \cmidrule(lr){15-16} \cmidrule(lr){17-18}
\textbf{Experiment} & \textbf{Type} & {\fontsize{4}{5}\selectfont base} & {\fontsize{4}{5}\selectfont ft} & {\fontsize{4}{5}\selectfont base} & {\fontsize{4}{5}\selectfont ft} & {\fontsize{4}{5}\selectfont base} & {\fontsize{4}{5}\selectfont ft} & {\fontsize{4}{5}\selectfont base} & {\fontsize{4}{5}\selectfont ft} & {\fontsize{4}{5}\selectfont base} & {\fontsize{4}{5}\selectfont ft} & {\fontsize{4}{5}\selectfont base} & {\fontsize{4}{5}\selectfont ft} & {\fontsize{4}{5}\selectfont base} & {\fontsize{4}{5}\selectfont ft} & {\fontsize{4}{5}\selectfont base} & {\fontsize{4}{5}\selectfont ft} & 70B & $\ln(k)$ \\
\midrule
Experience-description choice {\fontsize{4}{5}\selectfont\citep{anllo2024weird}} & Decision & \hc{0.50} & \hc{0.42} & \hc{0.49} & \hc{0.40} & \hc{0.46} & \hc{0.39} & \hc{0.46} & \hcu{0.38} & \hc{0.45} & \hcbu{0.38} & \hc{0.49} & \hc{0.44} & \hc{0.46} & \hc{0.41} & \hc{0.46} & \hc{0.41} & \hc{0.42} & \hc{0.69} \\
Risky decision and happiness {\fontsize{4}{5}\selectfont\citep{rutledge2023happiness}} & Decision & \hc{1.46} & \hc{1.38} & \hc{1.45} & \hc{1.40} & \hc{1.43} & \hc{1.34} & \hc{1.39} & \hcbu{1.31} & \hc{1.39} & \hc{1.33} & \hc{1.57} & \hc{1.40} & \hc{1.53} & \hc{1.37} & \hc{1.43} & \hc{1.34} & \hcu{1.32} & \hcn \\
Loss aversion task {\fontsize{4}{5}\selectfont\citep{spektor2024lossaversion}} & Decision & \hc{0.57} & \hc{0.54} & \hc{0.56} & \hc{0.54} & \hc{0.54} & \hc{0.52} & \hc{0.53} & \hc{0.54} & \hc{0.53} & \hcbu{0.50} & \hc{0.66} & \hc{0.57} & \hc{0.61} & \hc{0.55} & \hc{0.54} & \hc{0.52} & \hcu{0.51} & \hc{0.69} \\
\midrule
Two-step task {\fontsize{4}{5}\selectfont\citep{castrorodrigues2022twostep}} & MDP & \hc{0.26} & \hc{0.26} & \hc{0.27} & \hc{0.28} & \hc{0.26} & \hcu{0.24} & \hc{0.26} & \hcbu{0.24} & \hc{0.25} & \hc{0.25} & \hc{1.29} & \hc{0.45} & \hc{0.95} & \hc{0.35} & \hc{0.26} & \hc{0.25} & \hc{0.46} & \hc{1.39} \\
Two-step task {\fontsize{4}{5}\selectfont\citep{shahar2019twosteptask}} & MDP & \hc{0.55} & \hc{0.48} & \hc{0.53} & \hc{0.47} & \hc{0.54} & \hc{0.47} & \hc{0.55} & \hc{0.47} & \hc{0.53} & \hcbu{0.46} & \hc{0.54} & \hc{0.49} & \hc{0.53} & \hc{0.47} & \hc{0.53} & \hc{0.47} & \hcu{0.46} & \hc{0.69} \\
\midrule
Range adaptation RL {\fontsize{4}{5}\selectfont\citep{bavard2021range}} & Bandit & \hc{0.56} & \hc{0.48} & \hc{0.54} & \hc{0.48} & \hc{0.53} & \hc{0.47} & \hc{0.52} & \hc{0.47} & \hc{0.52} & \hc{0.47} & \hc{0.57} & \hc{0.48} & \hc{0.52} & \hc{0.48} & \hc{0.52} & \hcu{0.46} & \hcbu{0.45} & \hc{0.69} \\
Explore-exploit task {\fontsize{4}{5}\selectfont\citep{fan2023trait}} & Bandit & \hc{0.42} & \hc{0.28} & \hc{0.40} & \hc{0.27} & \hc{0.40} & \hc{0.26} & \hc{0.39} & \hc{0.27} & \hc{0.40} & \hcu{0.26} & \hc{0.42} & \hc{0.28} & \hc{0.43} & \hc{0.27} & \hc{0.39} & \hc{0.26} & \hcbu{0.26} & \hc{0.69} \\
Counterfactual learning {\fontsize{4}{5}\selectfont\citep{palminteri2017confirmation}} & Bandit & \hc{0.64} & \hc{0.50} & \hc{0.62} & \hc{0.49} & \hc{0.60} & \hc{0.48} & \hc{0.61} & \hc{0.49} & \hc{0.58} & \hcu{0.47} & \hc{0.62} & \hc{0.53} & \hc{0.58} & \hc{0.49} & \hc{0.58} & \hc{0.48} & \hcbu{0.46} & \hc{0.69} \\
Contextual reinforcement learning {\fontsize{4}{5}\selectfont\citep{vandendriessche2023depression}} & Bandit & \hc{0.69} & \hc{0.61} & \hc{0.69} & \hc{0.57} & \hc{0.61} & \hc{0.54} & \hc{0.63} & \hc{0.55} & \hc{0.57} & \hcu{0.53} & \hc{0.69} & \hc{0.58} & \hc{0.59} & \hc{0.61} & \hc{0.59} & \hc{0.56} & \hcbu{0.51} & \hc{0.69} \\
Aversive reversal learning {\fontsize{4}{5}\selectfont\citep{zika2023traitanxiety}} & Bandit & \hc{3.66} & \hc{3.36} & \hc{3.70} & \hc{3.36} & \hc{3.57} & \hc{3.29} & \hc{3.51} & \hcbu{3.27} & \hc{3.47} & \hcu{3.27} & \hc{4.00} & \hc{3.99} & \hc{3.90} & \hc{3.85} & \hc{3.87} & \hc{3.82} & \hc{3.80} & \hcn \\
\midrule
Navon task {\fontsize{4}{5}\selectfont\citep{busch2024navon}} & Misc. & \hc{0.43} & \hc{0.40} & \hc{0.40} & \hc{0.35} & \hc{0.43} & \hcu{0.32} & \hc{0.41} & \hc{0.32} & \hc{0.37} & \hc{0.32} & \hc{0.47} & \hc{0.59} & \hc{0.44} & \hc{0.45} & \hc{0.39} & \hc{0.34} & \hcbu{0.31} & \hc{0.69} \\
Stroop task {\fontsize{4}{5}\selectfont\citep{busch2024stroop}} & Misc. & \hc{0.53} & \hc{0.38} & \hc{0.28} & \hc{0.63} & \hc{0.21} & \hc{0.18} & \hc{0.20} & \hc{0.17} & \hc{0.21} & \hcu{0.15} & \hc{0.97} & \hc{0.85} & \hc{0.29} & \hc{0.45} & \hc{0.20} & \hc{0.18} & \hcbu{0.15} & \hc{1.10} \\
Pragmatic language task {\fontsize{4}{5}\selectfont\citep{franke2024bayesian}} & Misc. & \hc{1.55} & \hc{1.25} & \hc{1.47} & \hc{1.03} & \hc{1.17} & \hc{0.81} & \hc{1.08} & \hcu{0.76} & \hc{1.07} & \hc{0.77} & \hc{1.64} & \hc{1.49} & \hc{1.75} & \hc{1.25} & \hc{1.85} & \hc{1.07} & \hcbu{0.75} & \hc{1.39} \\
Reference game {\fontsize{4}{5}\selectfont\citep{franke2016reasoning}} & Misc. & \hc{1.50} & \hc{1.21} & \hc{1.33} & \hc{1.12} & \hc{1.04} & \hc{0.83} & \hc{1.14} & \hc{0.85} & \hc{0.97} & \hcu{0.80} & \hc{1.47} & \hc{1.34} & \hc{1.41} & \hc{1.12} & \hc{1.19} & \hc{1.04} & \hcbu{0.80} & \hc{1.39} \\
Compound word processing {\fontsize{4}{5}\selectfont\citep{gunther2020ts}} & Misc. & \hc{0.62} & \hc{0.49} & \hc{0.52} & \hc{0.41} & \hc{0.37} & \hc{0.37} & \hc{0.36} & \hc{0.37} & \hc{0.34} & \hcu{0.33} & \hc{0.49} & \hc{0.47} & \hc{0.42} & \hc{0.38} & \hc{0.37} & \hc{0.34} & \hcbu{0.32} & \hc{0.69} \\
Grammaticality judgement {\fontsize{4}{5}\selectfont\citep{dentella2023grammaticality}} & Misc. & \hc{0.47} & \hc{0.43} & \hc{0.37} & \hc{0.41} & \hc{0.35} & \hc{0.36} & \hc{0.35} & \hc{0.31} & \hc{0.32} & \hcu{0.30} & \hc{0.50} & \hc{0.42} & \hc{0.40} & \hc{0.39} & \hc{0.34} & \hc{0.33} & \hcbu{0.29} & \hc{0.69} \\
Scalar inference {\fontsize{4}{5}\selectfont\citep{tsvilodub2023xorsome}} & Misc. & \hc{3.37} & \hc{3.33} & \hc{3.29} & \hc{3.19} & \hc{3.13} & \hc{3.04} & \hc{3.06} & \hc{3.00} & \hc{2.97} & \hcu{2.95} & \hc{3.62} & \hc{3.50} & \hc{3.48} & \hc{3.38} & \hc{3.23} & \hc{3.20} & \hcbu{2.91} & \hcn \\
Math arithmetic task {\fontsize{4}{5}\selectfont\citep{xu2023augmenting}} & Misc. & \hc{0.72} & \hc{0.72} & \hc{0.71} & \hc{0.71} & \hc{0.71} & \hc{0.70} & \hc{0.69} & \hc{0.70} & \hc{0.68} & \hcu{0.66} & \hc{0.73} & \hc{0.72} & \hc{0.73} & \hc{0.72} & \hc{0.71} & \hc{0.72} & \hcbu{0.55} & \hc{0.69} \\
\midrule
\textbf{Mean (18)} & & \hc{1.03} & \hc{0.92} & \hc{0.98} & \hc{0.89} & \hc{0.91} & \hc{0.81} & \hc{0.90} & \hcu{0.80} & \hc{0.87} & \hcbu{0.79} & \hc{1.15} & \hc{1.03} & \hc{1.06} & \hc{0.94} & \hc{0.97} & \hc{0.88} & \hc{0.82} & \hc{0.86} \\
\bottomrule
\end{tabular}%
}
\vspace{4pt}
\begin{center}\footnotesize
\colorbox{c1!45}{\strut\hspace{5pt}} \scriptsize$<$0.20 \quad
\colorbox{c2!40}{\strut\hspace{5pt}} \scriptsize 0.20--0.40 \quad
\colorbox{c3!40}{\strut\hspace{5pt}} \scriptsize 0.40--0.60 \quad
\colorbox{c4!45}{\strut\hspace{5pt}} \scriptsize 0.60--0.80 \quad
\colorbox{c5!55}{\strut\hspace{5pt}} \scriptsize 0.80--1.00 \quad
\colorbox{c6!50}{\strut\hspace{5pt}} \scriptsize 1.00--1.20 \quad
\colorbox{c7!45}{\strut\hspace{5pt}} \scriptsize 1.20--1.50 \quad
\colorbox{c8!40}{\strut\hspace{5pt}} \scriptsize 1.50--2.00 \quad
\colorbox{c9!35}{\strut\hspace{5pt}} \scriptsize$>$2.00
\end{center}
\vspace{-2pt}
{\tiny \underline{\textbf{Bold+underline}}\,=\,best; \underline{underline}\,=\,second-best. Lower is better. ``---''\,=\,continuous or mixed response space. Size labels in billions of parameters.}
\caption{Per-experiment NLL on Psych-201 (out-of-distribution). Psych-201 was not used during supervised fine-tuning; lower NLL indicates better generalisation. Within each model size, the left column (base) shows the pretrained model and the right column (ft) shows the cognitively finetuned variant. The $\ln(k)$ column shows the random-guessing baseline where $k$ is the number of per-trial response options.}
\label{tab:psych201_pe_bf16_base_v_ft}
\end{sidewaystable}

\begin{sidewaystable}[p]
\centering
{\tiny
\setlength{\tabcolsep}{3pt}
\renewcommand{\arraystretch}{1.15}
\begin{tabular}{@{}l l  r  r  r  r  r  r  r  r  r  r  r  r  r  r  r  r@{}}
\toprule
& & \multicolumn{5}{c}{\texttt{\textbf{Qwentaur}}} & \multicolumn{3}{c}{\texttt{\textbf{Llama-Centaur}}} & \multicolumn{4}{c}{\texttt{\textbf{Smoltaur}}} & \multicolumn{2}{c}{\texttt{\textbf{Olmotaur}}} & \multicolumn{1}{c}{\citet{binz2025foundation}} & \textbf{Chance} \\
\cmidrule(lr){3-7}\cmidrule(lr){8-10}\cmidrule(lr){11-14}\cmidrule(lr){15-16}
\textbf{Experiment} & \textbf{Type} & 0.6B & 1.7B & 4B & 8B & 14B & 1B & 3B & 8B & 0.1B & 0.4B & 1.7B & 3B & 1B & 7B & 70B & $\ln(k)$ \\
\midrule
Experience-description choice {\fontsize{4}{5}\selectfont\citep{anllo2024weird}} & Decision & \hc{0.42} & \hc{0.40} & \hc{0.39} & \hcu{0.38} & \hcbu{0.38} & \hc{0.44} & \hc{0.41} & \hc{0.41} & \hc{0.53} & \hc{0.52} & \hc{0.47} & \hc{0.40} & \hc{0.49} & \hc{0.41} & \hc{0.42} & \hc{0.69} \\
Risky decision and happiness {\fontsize{4}{5}\selectfont\citep{rutledge2023happiness}} & Decision & \hc{1.38} & \hc{1.40} & \hc{1.34} & \hcbu{1.31} & \hc{1.33} & \hc{1.40} & \hc{1.37} & \hc{1.34} & \hc{1.50} & \hc{1.44} & \hc{1.46} & \hc{1.36} & \hc{1.48} & \hc{1.38} & \hcu{1.32} & \hcn \\
Loss aversion task {\fontsize{4}{5}\selectfont\citep{spektor2024lossaversion}} & Decision & \hc{0.54} & \hc{0.54} & \hc{0.52} & \hc{0.54} & \hcbu{0.50} & \hc{0.57} & \hc{0.55} & \hc{0.52} & \hc{0.58} & \hc{0.58} & \hc{0.58} & \hc{0.52} & \hc{0.61} & \hc{0.54} & \hcu{0.51} & \hc{0.69} \\
\midrule
Two-step task {\fontsize{4}{5}\selectfont\citep{castrorodrigues2022twostep}} & MDP & \hc{0.26} & \hc{0.28} & \hcu{0.24} & \hcbu{0.24} & \hc{0.25} & \hc{0.45} & \hc{0.35} & \hc{0.25} & \hc{0.48} & \hc{0.43} & \hc{0.34} & \hc{0.25} & \hc{0.47} & \hc{0.26} & \hc{0.46} & \hc{1.39} \\
Two-step task {\fontsize{4}{5}\selectfont\citep{shahar2019twosteptask}} & MDP & \hc{0.48} & \hc{0.47} & \hc{0.47} & \hc{0.47} & \hcbu{0.46} & \hc{0.49} & \hc{0.47} & \hc{0.47} & \hc{0.61} & \hc{0.54} & \hc{0.50} & \hc{0.47} & \hc{0.58} & \hc{0.47} & \hcu{0.46} & \hc{0.69} \\
\midrule
Range adaptation RL {\fontsize{4}{5}\selectfont\citep{bavard2021range}} & Bandit & \hc{0.48} & \hc{0.48} & \hc{0.47} & \hc{0.47} & \hc{0.47} & \hc{0.48} & \hc{0.48} & \hcu{0.46} & \hc{0.54} & \hc{0.55} & \hc{0.51} & \hc{0.48} & \hc{0.55} & \hc{0.48} & \hcbu{0.45} & \hc{0.69} \\
Explore-exploit task {\fontsize{4}{5}\selectfont\citep{fan2023trait}} & Bandit & \hc{0.28} & \hc{0.27} & \hc{0.26} & \hc{0.27} & \hcu{0.26} & \hc{0.28} & \hc{0.27} & \hc{0.26} & \hc{0.35} & \hc{0.32} & \hc{0.29} & \hc{0.27} & \hc{0.29} & \hc{0.27} & \hcbu{0.26} & \hc{0.69} \\
Counterfactual learning {\fontsize{4}{5}\selectfont\citep{palminteri2017confirmation}} & Bandit & \hc{0.50} & \hc{0.49} & \hc{0.48} & \hc{0.49} & \hcu{0.47} & \hc{0.53} & \hc{0.49} & \hc{0.48} & \hc{0.59} & \hc{0.60} & \hc{0.52} & \hc{0.52} & \hc{0.59} & \hc{0.51} & \hcbu{0.46} & \hc{0.69} \\
Contextual reinforcement learning {\fontsize{4}{5}\selectfont\citep{vandendriessche2023depression}} & Bandit & \hc{0.61} & \hc{0.57} & \hc{0.54} & \hc{0.55} & \hcu{0.53} & \hc{0.58} & \hc{0.61} & \hc{0.56} & \hc{0.64} & \hc{0.68} & \hc{0.58} & \hc{0.57} & \hc{0.57} & \hc{0.58} & \hcbu{0.51} & \hc{0.69} \\
Aversive reversal learning {\fontsize{4}{5}\selectfont\citep{zika2023traitanxiety}} & Bandit & \hc{3.36} & \hc{3.36} & \hc{3.29} & \hcbu{3.27} & \hcu{3.27} & \hc{3.99} & \hc{3.85} & \hc{3.82} & \hc{3.54} & \hc{3.44} & \hc{3.33} & \hc{3.89} & \hc{4.35} & \hc{4.00} & \hc{3.80} & \hcn \\
\midrule
Navon task {\fontsize{4}{5}\selectfont\citep{busch2024navon}} & Misc. & \hc{0.40} & \hc{0.35} & \hcu{0.32} & \hc{0.32} & \hc{0.32} & \hc{0.59} & \hc{0.45} & \hc{0.34} & \hc{0.65} & \hc{0.61} & \hc{0.44} & \hc{0.34} & \hc{0.72} & \hc{0.35} & \hcbu{0.31} & \hc{0.69} \\
Stroop task {\fontsize{4}{5}\selectfont\citep{busch2024stroop}} & Misc. & \hc{0.38} & \hc{0.63} & \hc{0.18} & \hc{0.17} & \hcu{0.15} & \hc{0.85} & \hc{0.45} & \hc{0.18} & \hc{0.67} & \hc{0.76} & \hc{0.68} & \hc{0.29} & \hc{0.88} & \hc{0.39} & \hcbu{0.15} & \hc{1.10} \\
Pragmatic language task {\fontsize{4}{5}\selectfont\citep{franke2024bayesian}} & Misc. & \hc{1.25} & \hc{1.03} & \hc{0.81} & \hcu{0.76} & \hc{0.77} & \hc{1.49} & \hc{1.25} & \hc{1.07} & \hc{1.61} & \hc{1.64} & \hc{1.54} & \hc{1.06} & \hc{1.39} & \hc{1.08} & \hcbu{0.75} & \hc{1.39} \\
Reference game {\fontsize{4}{5}\selectfont\citep{franke2016reasoning}} & Misc. & \hc{1.21} & \hc{1.12} & \hc{0.83} & \hc{0.85} & \hcu{0.80} & \hc{1.34} & \hc{1.12} & \hc{1.04} & \hc{1.92} & \hc{1.35} & \hc{1.45} & \hc{1.14} & \hc{1.38} & \hc{1.04} & \hcbu{0.80} & \hc{1.39} \\
Compound word processing {\fontsize{4}{5}\selectfont\citep{gunther2020ts}} & Misc. & \hc{0.49} & \hc{0.41} & \hc{0.37} & \hc{0.37} & \hcu{0.33} & \hc{0.47} & \hc{0.38} & \hc{0.34} & \hc{0.61} & \hc{0.52} & \hc{0.44} & \hc{0.35} & \hc{0.56} & \hc{0.42} & \hcbu{0.32} & \hc{0.69} \\
Grammaticality judgement {\fontsize{4}{5}\selectfont\citep{dentella2023grammaticality}} & Misc. & \hc{0.43} & \hc{0.41} & \hc{0.36} & \hc{0.31} & \hcu{0.30} & \hc{0.42} & \hc{0.39} & \hc{0.33} & \hc{0.49} & \hc{0.45} & \hc{0.42} & \hc{0.40} & \hc{0.45} & \hc{0.35} & \hcbu{0.29} & \hc{0.69} \\
Scalar inference {\fontsize{4}{5}\selectfont\citep{tsvilodub2023xorsome}} & Misc. & \hc{3.33} & \hc{3.19} & \hc{3.04} & \hc{3.00} & \hcu{2.95} & \hc{3.50} & \hc{3.38} & \hc{3.20} & \hc{3.78} & \hc{3.60} & \hc{3.45} & \hc{3.28} & \hc{3.96} & \hc{3.30} & \hcbu{2.91} & \hcn \\
Math arithmetic task {\fontsize{4}{5}\selectfont\citep{xu2023augmenting}} & Misc. & \hc{0.72} & \hc{0.71} & \hc{0.70} & \hc{0.70} & \hcu{0.66} & \hc{0.72} & \hc{0.72} & \hc{0.72} & \hc{0.78} & \hc{0.75} & \hc{0.74} & \hc{0.72} & \hc{0.75} & \hc{0.72} & \hcbu{0.55} & \hc{0.69} \\
\midrule
\textbf{Mean (18)} &  & \hc{0.92} & \hc{0.89} & \hc{0.81} & \hcu{0.80} & \hcbu{0.79} & \hc{1.03} & \hc{0.94} & \hc{0.88} & \hc{1.10} & \hc{1.04} & \hc{0.99} & \hc{0.91} & \hc{1.11} & \hc{0.92} & \hc{0.82} & \hc{0.86} \\
\bottomrule
\end{tabular}%
}
\vspace{4pt}
\begin{center}\footnotesize
\colorbox{c1!45}{\strut\hspace{5pt}} \scriptsize$<$0.20 \quad
\colorbox{c2!40}{\strut\hspace{5pt}} \scriptsize 0.20--0.40 \quad
\colorbox{c3!40}{\strut\hspace{5pt}} \scriptsize 0.40--0.60 \quad
\colorbox{c4!45}{\strut\hspace{5pt}} \scriptsize 0.60--0.80 \quad
\colorbox{c5!55}{\strut\hspace{5pt}} \scriptsize 0.80--1.00 \quad
\colorbox{c6!50}{\strut\hspace{5pt}} \scriptsize 1.00--1.20 \quad
\colorbox{c7!45}{\strut\hspace{5pt}} \scriptsize 1.20--1.50 \quad
\colorbox{c8!40}{\strut\hspace{5pt}} \scriptsize 1.50--2.00 \quad
\colorbox{c9!35}{\strut\hspace{5pt}} \scriptsize$>$2.00
\end{center}
\vspace{-2pt}
{\tiny \underline{\textbf{Bold+underline}}\,=\,best; \underline{underline}\,=\,second-best (across the four families and Centaur-70B). Lower is better. ``---''\,=\,continuous or mixed response space. Size labels in billions of parameters.}
\caption{Per-experiment NLL on Psych-201 (out-of-distribution) for the four finetuned families. Psych-201 was not used during fine-tuning; lower NLL indicates better generalisation. The $\ln(k)$ column shows the random-guessing baseline where $k$ is the number of per-trial response options.}
\label{tab:psych201_pe_ft_all}
\end{sidewaystable}

\clearpage

\subsection{Centaur-70B (Psych-101) reproduction notes}
\label{app:70b_discrepancy}

\begin{figure}[H]
    \centering
    \includegraphics[width=\columnwidth]{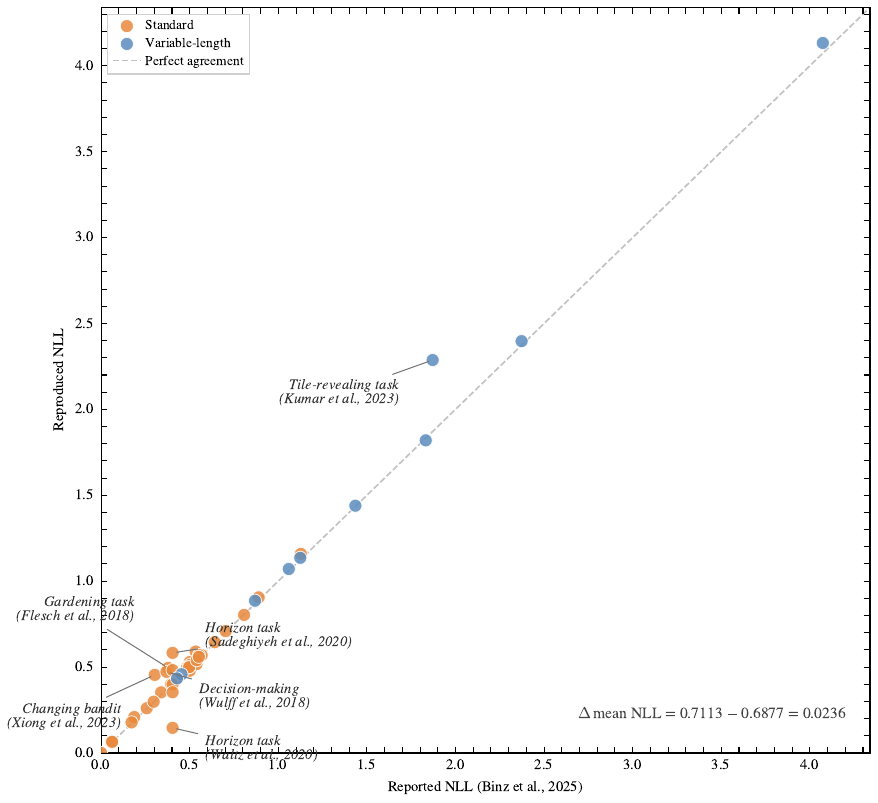}
    \caption{\textbf{Reproducibility of Centaur-70B (4-bit) evaluation across Psych-101.} Reproduced vs.\ reported NLL for Centaur-70B (4-bit) across 46 Psych-101 experiments ($r = 0.994$, mean $\Delta = +0.024$, median $\Delta = +0.006$).}
    \label{fig:70b_discrepancy}
\end{figure}

\texttt{Centaur-70B} \citep{binz2025foundation} uses QLoRA with a frozen 4-bit quantised base model, so base weights are stored in the \texttt{NF4} data type and dequantised to \texttt{bfloat16} on the fly whenever they enter a forward pass. This dequantisation may not be bitwise reproducible across library versions, and our evaluation environment differs from that of \citet{binz2025foundation} in several core packages, most notably \texttt{bitsandbytes} (0.43.1~$\to$~0.49.1), \texttt{PyTorch} (2.3.0~$\to$~2.10.0), and the \texttt{CUDA} toolkit (12.1~$\to$~12.8). Because multiple interacting components differ simultaneously, we cannot attribute the discrepancy to any single library. 

A separate source of apparent deviation concerns shared paradigm aggregates. \citet{binz2025foundation} report results for certain paradigms as a single value across multiple sub-experiments. The ``Horizon task'' entry (reported NLL $= 0.403$) aggregates five sub-experiments \citep{wilson2014humans, sadeghiyeh2020temporal, somerville2017charting, waltz2020differential, feng2021dynamics}, whose individual reproduced NLLs range from 0.146 to 0.582, and the ``Two-step task'' entry (reported NLL $= 0.500$) aggregates three sub-experiments \citep{kool2016when, kool2017cost, zorowitz2023data}, with individual NLLs of 0.478, 0.529, and 0.511 respectively. The apparent large deviations for these sub-experiments (e.g., \citet{waltz2020differential} at 0.146 vs.\ the aggregate 0.403) therefore reflect the difference between individual NLLs and the shared paradigm mean, not environment-induced drift.

\clearpage

\section{Specificity of cognitive fine-tuning}
\label{app:specificity}

\begin{figure}[H]
    \centering
    \includegraphics[width=\columnwidth]{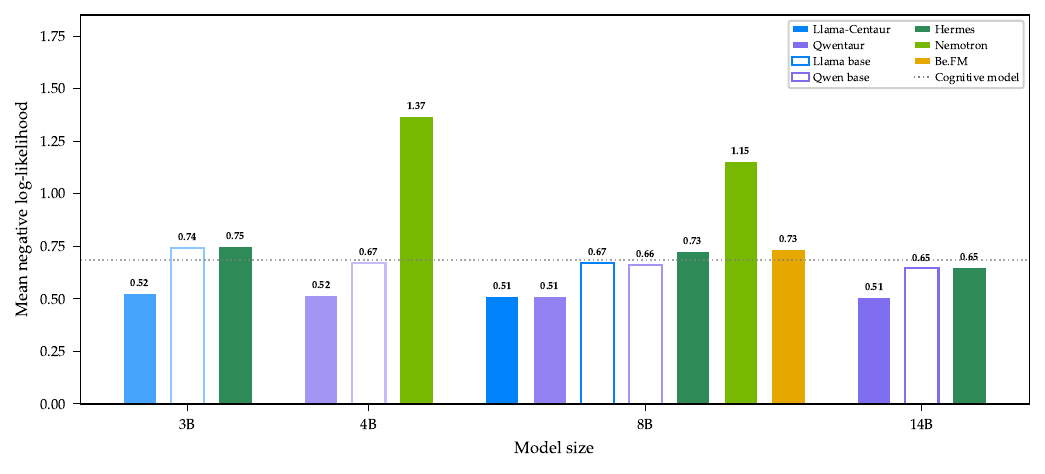}
    \caption{%
    \textbf{Specificity of cognitive fine-tuning: comparison with
    non-cognitive control models.}
    Mean NLL on Psych-101 for cognitively fine-tuned models
    (\texttt{Llama-Centaur}, \texttt{Qwentaur}; LoRA $r{=}16$, full training data) and size-matched non-cognitive and cognitive controls (\texttt{Hermes}, \texttt{Nemotron}, \texttt{Be.FM}).
    Models are grouped by parameter count.
    Cognitively fine-tuned models consistently outperform non-cognitive
    controls at every scale, confirming that the improvement is specific
    to the behavioural signal in Psych-101 and not an artefact of
    fine-tuning per se.
    }
    \label{fig:psych101_control_bar}
\end{figure}

A model's gains on Psych-101 need not come from cognitive data, since extensive post-training on other corpora might confer them incidentally. We test this with two classes of control. The first comprises non-cognitive fine-tuned models, \texttt{Nemotron} \citep{bercovich2025nemotron} and \texttt{Hermes} \citep{teknium2024hermes3technicalreport, teknium2025hermes4technicalreport}; the second comprises behavioural fine-tuned models, \texttt{Socrates} \citep{kolluri2025finetuning} and \texttt{Be.FM} \citep{xie2025fm}. Each is compared against its own base model at matched size (Figure~\ref{fig:psych101_control_bar}), with per-experiment values in Table~\ref{tab:psych101_pe_control}.

None of the non-cognitive controls improve over their base models, replicating \citet{binz2025foundation} at 70B and extending that result downward. Generic supervised fine-tuning does not incidentally produce models that predict human behaviour, at any scale we test. The behavioural controls are more instructive. \texttt{Be.FM} gains modestly on overlapping task domains, whereas \texttt{Socrates} performs \textit{worse} than its base model on several tasks despite a larger and more diverse behavioural corpus. \texttt{Socrates} is optimised for distributional alignment over demographic profiles, learning to predict an aggregate response distribution given a persona and an experimental condition, whereas Psych-101 requires tracking one participant's choices trial by trial, conditioned on the full history of stimuli, feedback and prior responses. Population-level distributional knowledge does not substitute for within-participant sequential prediction, and training on one can interfere with the other. The space of cognitive and behavioural foundation models is therefore not a single continuum from worse to better, but is partitioned by the \textit{prediction target}, which ranges from a population-level distribution to a single participant's trial-by-trial trajectory.

\newpage

\section{Retention of general capabilities}
\label{app:retention}

Table~\ref{tab:retention_app} reports the full per-task impact of cognitive fine-tuning across all benchmarks, with significance assessed via two-sided $z$-tests. All fine-tuned models here are LoRA $r{=}16$ on the full dataset.

The pattern is not one of uniform degradation but of selective reorganisation. Mathematical reasoning is the primary casualty, with GSM8K degrading significantly in six of eight models, from $-0.25$ at \texttt{Qwentaur-0.6B} to $-0.12$ at \texttt{Qwentaur-14B}. The scale dependence suggests a capacity account, in which larger models can accommodate both capabilities while smaller models face a harder trade-off. Formal planning follows the same pattern, with the ACPBench group mean degrading significantly in six of eight models. Cognitive and language tasks, the category most proximate to the training domain, are the most stable, with group means within $\pm 0.02$ at every scale.

\begin{figure}[H]
    \centering
    \includegraphics[width=\columnwidth]{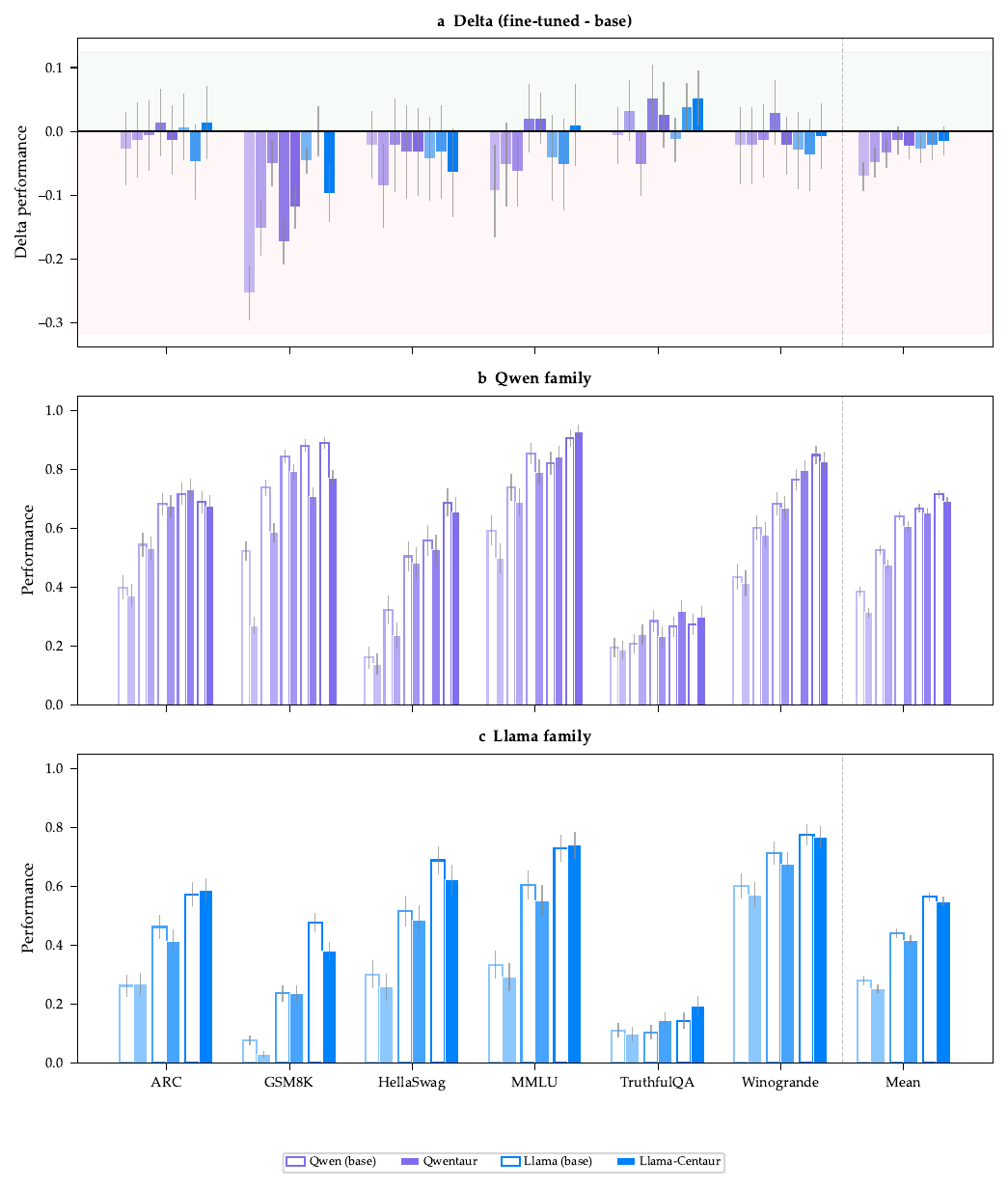}
    \caption{%
    \textbf{Impact of cognitive fine-tuning on MetaBench performance
    ($\Delta$ = fine-tuned $-$ base).}
    Each bar shows the change in accuracy for a matched base--fine-tuned
    pair on six standard LM benchmarks (ARC, GSM8K, HellaSwag, MMLU,
    TruthfulQA, Winogrande) plus their mean.
    Error bars show pooled standard errors.
    Positive values (green region) indicate improvement; negative values
    (red region) indicate degradation.
    Colour intensity scales with model size within each family.
    }
    \label{fig:metabench_combined}
\end{figure}

\begin{figure}[H]
    \centering
    \includegraphics[width=\columnwidth]{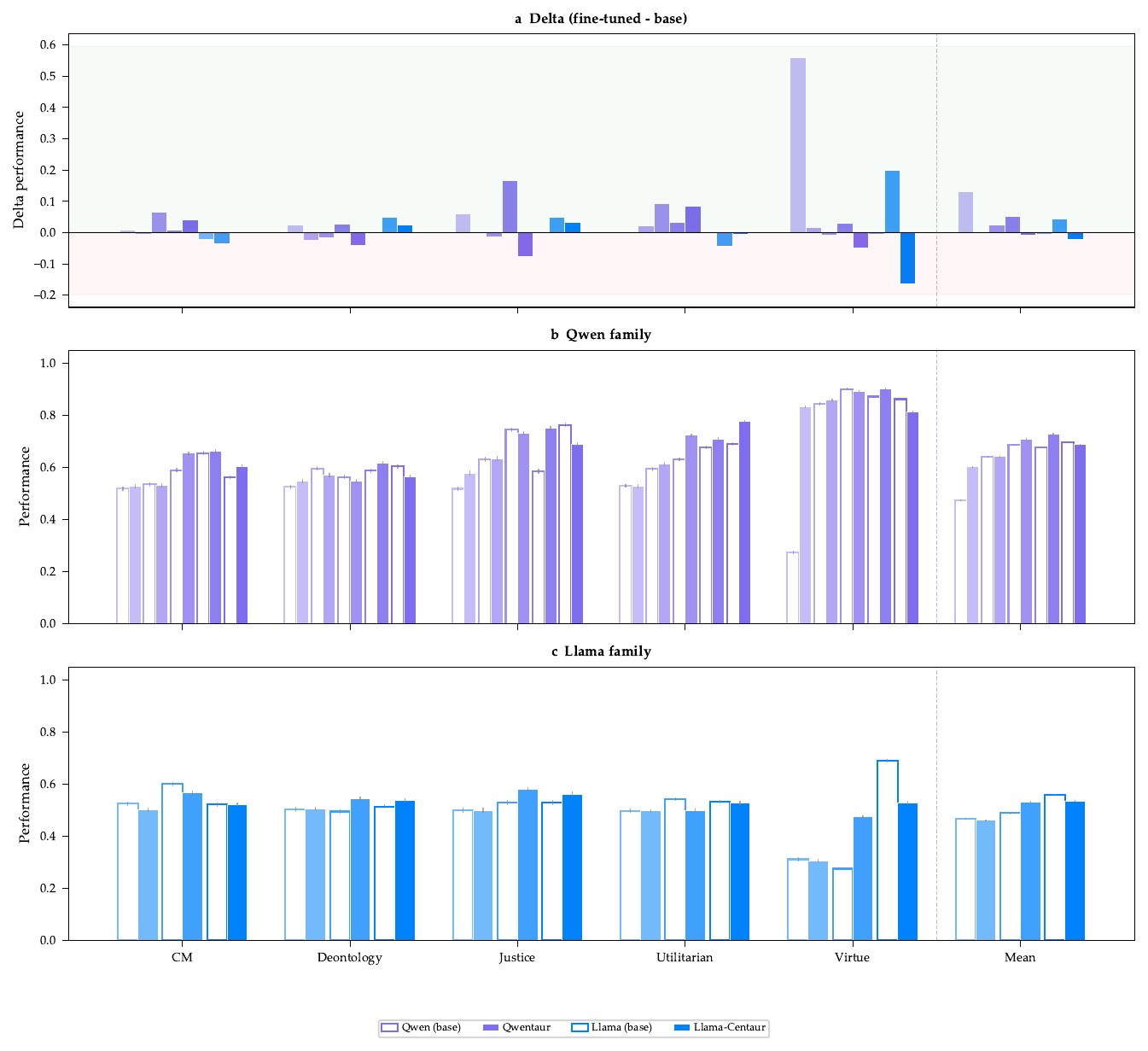}
    \caption{%
    \textbf{Impact of cognitive fine-tuning on Ethics benchmark performance
    ($\Delta$ = fine-tuned $-$ base).}
    Each bar shows the change in accuracy for a matched base--fine-tuned
    pair across five ethical reasoning tasks (commonsense morality,
    deontology, justice, utilitarianism, virtue) plus their mean.
    Error bars show pooled standard errors.
    Positive values (green region) indicate improvement; negative values
    (red region) indicate degradation.
    Colour intensity scales with model size within each family.
    }
    \label{fig:ethics_combined}
\end{figure}

\begin{figure}[H]
    \centering
    \includegraphics[width=\columnwidth]{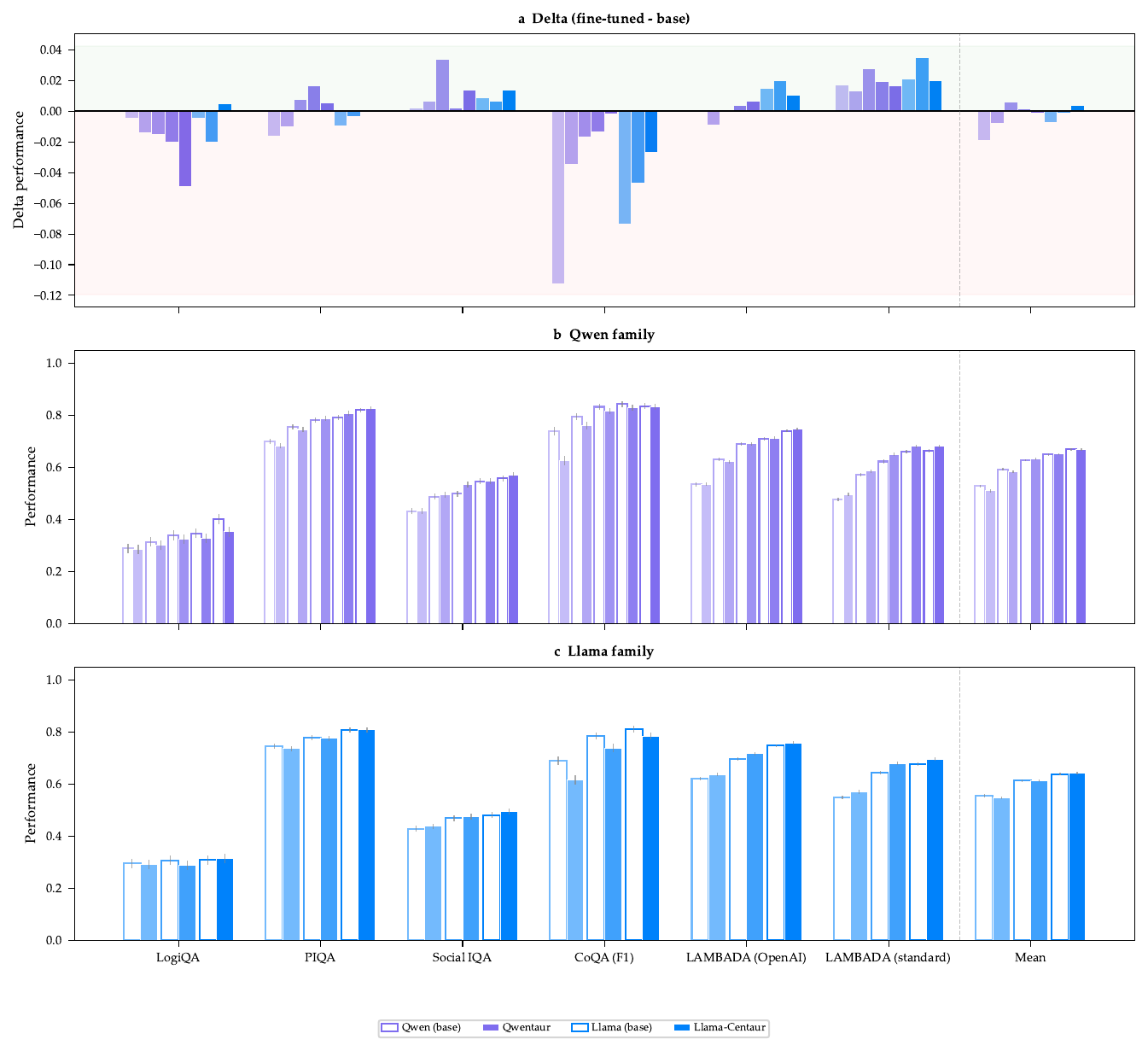}
    \caption{%
    \textbf{Impact of cognitive fine-tuning on cognitive and language
    benchmark performance ($\Delta$ = fine-tuned $-$ base).}
    Each bar shows the change in accuracy for a matched base--fine-tuned
    pair across six tasks (LogiQA, PIQA, Social IQA, CoQA, LAMBADA-OpenAI,
    LAMBADA-Standard) plus their mean.
    Error bars show pooled standard errors.
    Positive values (green region) indicate improvement; negative values
    (red region) indicate degradation.
    Colour intensity scales with model size within each family.
    }
    \label{fig:coglang_combined}
\end{figure}

\begin{sidewaysfigure}
    \centering
    \includegraphics[width=\textheight]{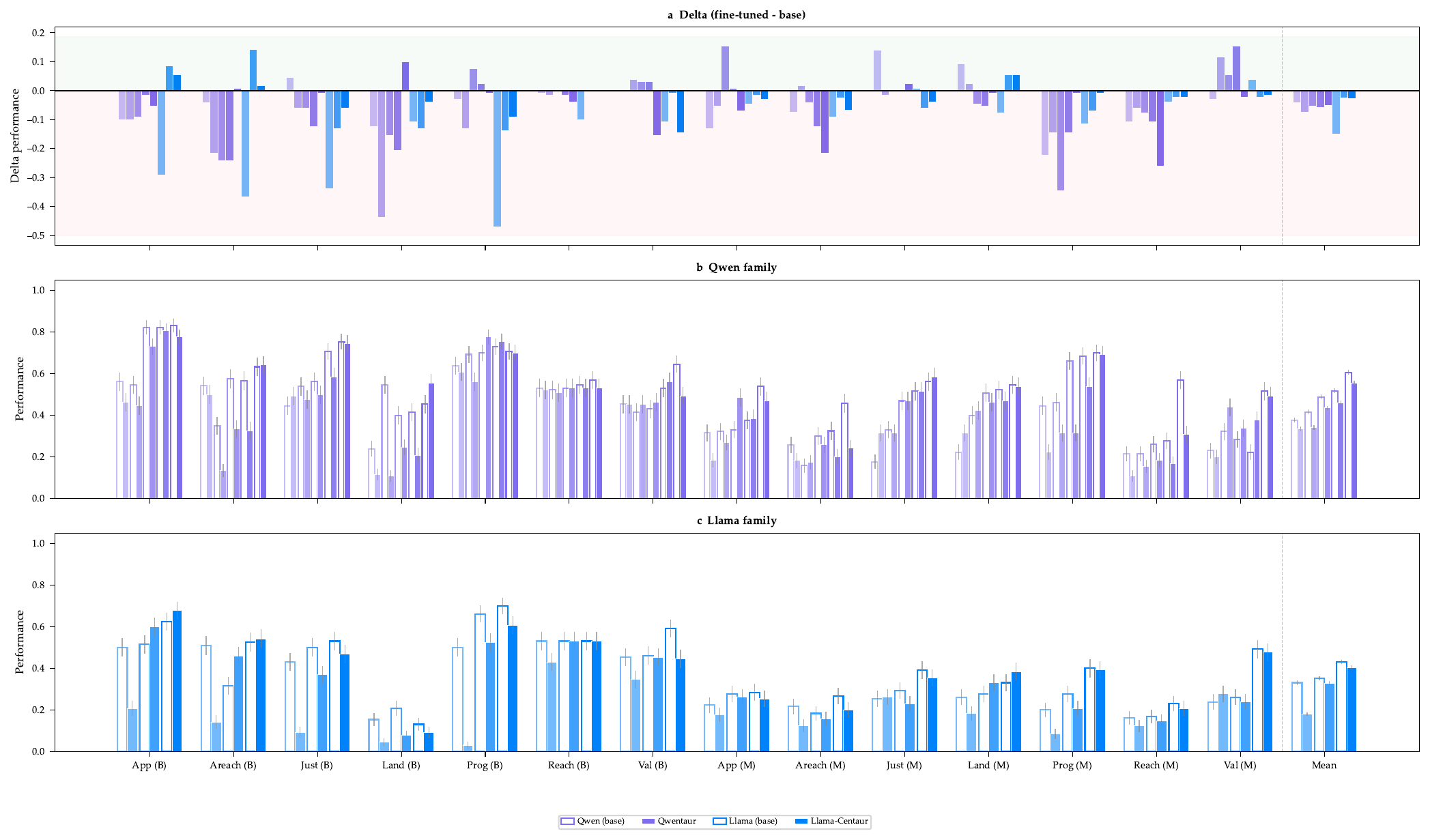}
    \caption{%
    \textbf{Impact of cognitive fine-tuning on ACP planning benchmark
    performance ($\Delta$ = fine-tuned $-$ base).}
    Each bar shows the change in accuracy for a matched base--fine-tuned
    pair across 14 planning tasks (7 boolean, 7 multiple-choice) plus
    their mean.
    Error bars show pooled standard errors.
    Positive values (green region) indicate improvement; negative values
    (red region) indicate degradation.
    Colour intensity scales with model size within each family.
    }
    \label{fig:acp_combined}
\end{sidewaysfigure}

Ethical reasoning is the least stable category and the hardest to summarise. Group means are significantly positive in four models, significantly negative in one (\texttt{Llama-Centaur-8B}, $-0.02$), and null in the remaining three, and the largest positive mean ($+0.13$ at \texttt{Qwentaur-0.6B}) is driven almost entirely by a single subtask (Virtue, $+0.56$). We report the effect but do not interpret it, since the subtask-level pattern is inconsistent in sign across both model families and across moral frameworks. Figures~\ref{fig:metabench_combined}--\ref{fig:acp_combined} visualise the same deltas as grouped bar charts for MetaBench, Ethics, Cognitive \& Language, and ACP Planning respectively.

\begin{table}[h]
\centering
\resizebox{\columnwidth}{!}{%
\renewcommand{\arraystretch}{1.1}
\begin{tabular}{@{}l r r r r r   r r r@{}}
\toprule
& \multicolumn{5}{c}{\textbf{\texttt{Qwentaur}}} & \multicolumn{3}{c}{\textbf{\texttt{Llama-Centaur}}} \\
\cmidrule(lr){2-6}\cmidrule(lr){7-9}
\textbf{Benchmark} & \texttt{0.6B} & \texttt{1.7B} & \texttt{4B} & \texttt{8B} & \texttt{14B} & \texttt{1B} & \texttt{3B} & \texttt{8B} \\
\midrule
\multicolumn{9}{l}{\textit{MetaBench}} \\
ARC & \zc{cneg!8}{-0.03} & \zc{cneg!8}{-0.01} & \zc{cneg!8}{-0.01} & \zc{cpos!8}{+0.01} & \zc{cneg!8}{-0.01} & \zc{cpos!8}{+0.01} & \zc{cneg!8}{-0.05} & \zc{cpos!8}{+0.01} \\
GSM8K & \zc{cneg!50}{-0.25$^{***}$} & \zc{cneg!50}{-0.15$^{***}$} & \zc{cneg!8}{-0.05} & \zc{cneg!50}{-0.17$^{***}$} & \zc{cneg!50}{-0.12$^{***}$} & \zc{cneg!22}{-0.05$^{*}$} & \zc{cpos!8}{+0.00} & \zc{cneg!22}{-0.10$^{*}$} \\
HellaSwag & \zc{cneg!8}{-0.02} & \zc{cneg!8}{-0.09} & \zc{cneg!8}{-0.02} & \zc{cneg!8}{-0.03} & \zc{cneg!8}{-0.03} & \zc{cneg!8}{-0.04} & \zc{cneg!8}{-0.03} & \zc{cneg!8}{-0.06} \\
MMLU & \zc{cneg!8}{-0.09} & \zc{cneg!8}{-0.05} & \zc{cneg!8}{-0.06} & \zc{cpos!8}{+0.02} & \zc{cpos!8}{+0.02} & \zc{cneg!8}{-0.04} & \zc{cneg!8}{-0.05} & \zc{cpos!8}{+0.01} \\
TruthfulQA & \zc{cneg!8}{-0.01} & \zc{cpos!8}{+0.03} & \zc{cneg!8}{-0.05} & \zc{cpos!8}{+0.05} & \zc{cpos!8}{+0.03} & \zc{cneg!8}{-0.01} & \zc{cpos!8}{+0.04} & \zc{cpos!8}{+0.05} \\
Winogrande & \zc{cneg!8}{-0.02} & \zc{cneg!8}{-0.02} & \zc{cneg!8}{-0.02} & \zc{cpos!8}{+0.03} & \zc{cneg!8}{-0.02} & \zc{cneg!8}{-0.03} & \zc{cneg!8}{-0.04} & \zc{cneg!8}{-0.01} \\
\textbf{Mean (6)} & \zc{cneg!50}{-0.07$^{***}$} & \zc{cneg!22}{-0.05$^{*}$} & \zc{cneg!8}{-0.03} & \zc{cneg!8}{-0.01} & \zc{cneg!8}{-0.02} & \zc{cneg!8}{-0.03} & \zc{cneg!8}{-0.02} & \zc{cneg!8}{-0.02} \\
\midrule
\multicolumn{9}{l}{\textit{Ethics}} \\
CM & \zc{cpos!8}{+0.01} & \zc{cpos!8}{+0.00} & \zc{cpos!50}{+0.07$^{***}$} & \zc{cpos!8}{+0.01} & \zc{cpos!50}{+0.04$^{***}$} & \zc{cneg!22}{-0.02$^{*}$} & \zc{cneg!35}{-0.03$^{**}$} & \zc{cpos!8}{+0.00} \\
Deontology & \zc{cpos!22}{+0.02$^{*}$} & \zc{cneg!22}{-0.03$^{*}$} & \zc{cneg!8}{-0.02} & \zc{cpos!22}{+0.03$^{*}$} & \zc{cneg!50}{-0.04$^{***}$} & \zc{cpos!8}{+0.00} & \zc{cpos!50}{+0.05$^{***}$} & \zc{cpos!22}{+0.02$^{*}$} \\
Justice & \zc{cpos!50}{+0.06$^{***}$} & \zc{cpos!8}{+0.00} & \zc{cneg!8}{-0.01} & \zc{cpos!50}{+0.17$^{***}$} & \zc{cneg!50}{-0.08$^{***}$} & \zc{cpos!8}{+0.00} & \zc{cpos!50}{+0.05$^{***}$} & \zc{cpos!22}{+0.03$^{*}$} \\
Utilitarian & \zc{cpos!8}{+0.00} & \zc{cpos!22}{+0.02$^{*}$} & \zc{cpos!50}{+0.09$^{***}$} & \zc{cpos!50}{+0.03$^{***}$} & \zc{cpos!50}{+0.09$^{***}$} & \zc{cpos!8}{+0.00} & \zc{cneg!50}{-0.04$^{***}$} & \zc{cneg!8}{-0.01} \\
Virtue & \zc{cpos!50}{+0.56$^{***}$} & \zc{cpos!22}{+0.02$^{*}$} & \zc{cneg!8}{-0.01} & \zc{cpos!50}{+0.03$^{***}$} & \zc{cneg!50}{-0.05$^{***}$} & \zc{cneg!8}{-0.01} & \zc{cpos!50}{+0.20$^{***}$} & \zc{cneg!50}{-0.16$^{***}$} \\
\textbf{Mean (5)} & \zc{cpos!50}{+0.13$^{***}$} & \zc{cpos!8}{+0.00} & \zc{cpos!50}{+0.02$^{***}$} & \zc{cpos!50}{+0.05$^{***}$} & \zc{cneg!8}{-0.01} & \zc{cneg!8}{-0.01} & \zc{cpos!50}{+0.04$^{***}$} & \zc{cneg!50}{-0.02$^{***}$} \\
\midrule
\multicolumn{9}{l}{\textit{Cog. \& Lang.}} \\
LogiQA & \zc{cpos!8}{+0.00} & \zc{cneg!8}{-0.01} & \zc{cneg!8}{-0.02} & \zc{cneg!8}{-0.02} & \zc{cneg!8}{-0.05} & \zc{cpos!8}{+0.00} & \zc{cneg!8}{-0.02} & \zc{cpos!8}{+0.00} \\
PIQA & \zc{cneg!8}{-0.02} & \zc{cneg!8}{-0.01} & \zc{cpos!8}{+0.01} & \zc{cpos!8}{+0.02} & \zc{cpos!8}{+0.01} & \zc{cneg!8}{-0.01} & \zc{cpos!8}{+0.00} & \zc{cpos!8}{+0.00} \\
Social IQA & \zc{cpos!8}{+0.00} & \zc{cpos!8}{+0.01} & \zc{cpos!22}{+0.03$^{*}$} & \zc{cpos!8}{+0.00} & \zc{cpos!8}{+0.01} & \zc{cpos!8}{+0.01} & \zc{cpos!8}{+0.01} & \zc{cpos!8}{+0.01} \\
CoQA (F1) & \zc{cneg!50}{-0.11$^{***}$} & \zc{cneg!8}{-0.03} & \zc{cneg!8}{-0.02} & \zc{cneg!8}{-0.01} & \zc{cpos!8}{+0.00} & \zc{cneg!35}{-0.07$^{**}$} & \zc{cneg!22}{-0.05$^{*}$} & \zc{cneg!8}{-0.03} \\
LAMBADA (OAI) & \zc{cpos!8}{+0.00} & \zc{cneg!8}{-0.01} & \zc{cpos!8}{+0.00} & \zc{cpos!8}{+0.00} & \zc{cpos!8}{+0.01} & \zc{cpos!8}{+0.01} & \zc{cpos!22}{+0.02$^{*}$} & \zc{cpos!8}{+0.01} \\
LAMBADA (Std) & \zc{cpos!8}{+0.02} & \zc{cpos!8}{+0.01} & \zc{cpos!35}{+0.03$^{**}$} & \zc{cpos!22}{+0.02$^{*}$} & \zc{cpos!8}{+0.02} & \zc{cpos!22}{+0.02$^{*}$} & \zc{cpos!50}{+0.03$^{***}$} & \zc{cpos!22}{+0.02$^{*}$} \\
EQ-Bench & \zc{cneg!50}{-44.7$^{***}$} & \zc{cpos!8}{+5.1} & \zc{cpos!8}{+5.9} & \zc{cneg!50}{-21.3$^{***}$} & \zc{cneg!8}{-0.3} & \zc{cpos!50}{+16.0$^{***}$} & \zc{cpos!50}{+22.9$^{***}$} & \zc{cpos!8}{+7.4} \\
\textbf{Mean (6)} & \zc{cneg!8}{-0.02} & \zc{cneg!8}{-0.01} & \zc{cpos!8}{+0.01} & \zc{cpos!8}{+0.00} & \zc{cpos!8}{+0.00} & \zc{cneg!8}{-0.01} & \zc{cpos!8}{+0.00} & \zc{cpos!8}{+0.00} \\
\midrule
\multicolumn{9}{l}{\textit{ACP (Planning)}} \\
App (B) & \zc{cneg!8}{-0.10} & \zc{cneg!8}{-0.10} & \zc{cneg!8}{-0.09} & \zc{cneg!8}{-0.02} & \zc{cneg!8}{-0.05} & \zc{cneg!50}{-0.29$^{***}$} & \zc{cpos!8}{+0.08} & \zc{cpos!8}{+0.05} \\
Areach (B) & \zc{cneg!8}{-0.04} & \zc{cneg!50}{-0.22$^{***}$} & \zc{cneg!50}{-0.24$^{***}$} & \zc{cneg!50}{-0.24$^{***}$} & \zc{cpos!8}{+0.01} & \zc{cneg!50}{-0.37$^{***}$} & \zc{cpos!22}{+0.14$^{*}$} & \zc{cpos!8}{+0.02} \\
Just (B) & \zc{cpos!8}{+0.05} & \zc{cneg!8}{-0.06} & \zc{cneg!8}{-0.06} & \zc{cneg!22}{-0.12$^{*}$} & \zc{cneg!8}{-0.01} & \zc{cneg!50}{-0.34$^{***}$} & \zc{cneg!22}{-0.13$^{*}$} & \zc{cneg!8}{-0.06} \\
Land (B) & \zc{cneg!35}{-0.12$^{**}$} & \zc{cneg!50}{-0.44$^{***}$} & \zc{cneg!35}{-0.15$^{**}$} & \zc{cneg!50}{-0.21$^{***}$} & \zc{cpos!8}{+0.10} & \zc{cneg!35}{-0.11$^{**}$} & \zc{cneg!35}{-0.13$^{**}$} & \zc{cneg!8}{-0.04} \\
Prog (B) & \zc{cneg!8}{-0.03} & \zc{cneg!22}{-0.13$^{*}$} & \zc{cpos!8}{+0.08} & \zc{cpos!8}{+0.02} & \zc{cneg!8}{-0.01} & \zc{cneg!50}{-0.47$^{***}$} & \zc{cneg!22}{-0.14$^{*}$} & \zc{cneg!8}{-0.09} \\
Reach (B) & \zc{cneg!8}{-0.01} & \zc{cneg!8}{-0.02} & \zc{cpos!8}{+0.00} & \zc{cneg!8}{-0.02} & \zc{cneg!8}{-0.04} & \zc{cneg!8}{-0.10} & \zc{cpos!8}{+0.00} & \zc{cpos!8}{+0.00} \\
Val (B) & \zc{cpos!8}{+0.00} & \zc{cpos!8}{+0.04} & \zc{cpos!8}{+0.03} & \zc{cpos!8}{+0.03} & \zc{cneg!22}{-0.15$^{*}$} & \zc{cneg!8}{-0.11} & \zc{cneg!8}{-0.01} & \zc{cneg!22}{-0.15$^{*}$} \\
App (M) & \zc{cneg!22}{-0.13$^{*}$} & \zc{cneg!8}{-0.05} & \zc{cpos!22}{+0.15$^{*}$} & \zc{cpos!8}{+0.01} & \zc{cneg!8}{-0.07} & \zc{cneg!8}{-0.05} & \zc{cneg!8}{-0.02} & \zc{cneg!8}{-0.03} \\
Areach (M) & \zc{cneg!8}{-0.08} & \zc{cpos!8}{+0.02} & \zc{cneg!8}{-0.04} & \zc{cneg!22}{-0.12$^{*}$} & \zc{cneg!50}{-0.22$^{***}$} & \zc{cneg!8}{-0.09} & \zc{cneg!8}{-0.02} & \zc{cneg!8}{-0.07} \\
Just (M) & \zc{cpos!35}{+0.14$^{**}$} & \zc{cneg!8}{-0.02} & \zc{cpos!8}{+0.00} & \zc{cpos!8}{+0.00} & \zc{cpos!8}{+0.02} & \zc{cpos!8}{+0.01} & \zc{cneg!8}{-0.06} & \zc{cneg!8}{-0.04} \\
Land (M) & \zc{cpos!8}{+0.09} & \zc{cpos!8}{+0.02} & \zc{cneg!8}{-0.05} & \zc{cneg!8}{-0.05} & \zc{cneg!8}{-0.01} & \zc{cneg!8}{-0.08} & \zc{cpos!8}{+0.05} & \zc{cpos!8}{+0.05} \\
Prog (M) & \zc{cneg!50}{-0.22$^{***}$} & \zc{cneg!22}{-0.15$^{*}$} & \zc{cneg!50}{-0.35$^{***}$} & \zc{cneg!22}{-0.15$^{*}$} & \zc{cneg!8}{-0.01} & \zc{cneg!35}{-0.12$^{**}$} & \zc{cneg!8}{-0.07} & \zc{cneg!8}{-0.01} \\
Reach (M) & \zc{cneg!22}{-0.11$^{*}$} & \zc{cneg!8}{-0.06} & \zc{cneg!8}{-0.08} & \zc{cneg!22}{-0.11$^{*}$} & \zc{cneg!50}{-0.26$^{***}$} & \zc{cneg!8}{-0.04} & \zc{cneg!8}{-0.02} & \zc{cneg!8}{-0.02} \\
Val (M) & \zc{cneg!8}{-0.03} & \zc{cpos!8}{+0.12} & \zc{cpos!8}{+0.05} & \zc{cpos!35}{+0.15$^{**}$} & \zc{cneg!8}{-0.02} & \zc{cpos!8}{+0.04} & \zc{cneg!8}{-0.02} & \zc{cneg!8}{-0.02} \\
\textbf{Mean (14)} & \zc{cneg!35}{-0.04$^{**}$} & \zc{cneg!50}{-0.07$^{***}$} & \zc{cneg!50}{-0.05$^{***}$} & \zc{cneg!50}{-0.06$^{***}$} & \zc{cneg!35}{-0.05$^{**}$} & \zc{cneg!50}{-0.15$^{***}$} & \zc{cneg!8}{-0.02} & \zc{cneg!8}{-0.03} \\
\bottomrule
\end{tabular}%
}
\vspace{2pt}
{\tiny\centering
degrad. \colorbox{cneg!50}{\strut\,} \colorbox{cneg!20}{\strut\,} \colorbox{white}{\strut\,} \colorbox{cpos!20}{\strut\,} \colorbox{cpos!50}{\strut\,} improv. \quad $^{*}p{<}.05$ ~ $^{**}p{<}.01$ ~ $^{***}p{<}.001$\par}
\caption{\textbf{Impact of cognitive fine-tuning across different benchmarks.} Each cell shows $\Delta$ = fine-tuned $-$ base, where fine-tuned models are trained with LoRA $r{=}16$ on the full dataset. Significance is assessed with a two-sided z-test; group means use Stouffer's method to combine per-task z-scores. EQ-Bench is on a separate scale and excluded from means.}
\label{tab:retention_app}
\end{table}

\clearpage

\newpage

\clearpage

\section{Prompt Decomposition and Diagnostic Testing}
\label{app:prompt_decomposition}

This appendix provides full details for the prompt decomposition analyses summarised in the main text. Appendix~\ref{app:prior_ablations} reconstructs the ablation conditions used in two published critiques of \texttt{Centaur} and maps them onto our four channels (Table~\ref{tab:prior_mapping}), so that their manipulations and ours are directly comparable. Appendix~\ref{app:structural_ablation} reports the structural ablation, and Appendix~\ref{app:order_permutation} the order permutation test.

\subsection{Ablation conditions in prior critiques}
\label{app:prior_ablations}

\begin{table}[H]
\centering
\renewcommand{\arraystretch}{1.25}
\setlength{\tabcolsep}{5pt}
\fontsize{8}{10}\selectfont
\begin{tabular}{@{}l l c c c c l@{}}
\toprule
\textbf{Source} & \textbf{Condition} & $\boldsymbol{I}$ & $\boldsymbol{S}$ & $\boldsymbol{F}$ & $\boldsymbol{C}$ & \textbf{Closest condition in ours} \\
\midrule
\citet{binz2025foundation} & Original & $\bullet$ & $\bullet$ & $\bullet$ & $\bullet$ & Original \\
\midrule
\citet{xie2025centaur} & No psychological task & $I_{\min}$ & \no & \no & $\bullet$ & History-only \\
                       & Zero-shot prediction  & $\bullet$ & $\bullet$ & \no & \no & \textit{no analogue} \\
\midrule
\citet{liu2025can}     & Instruction free      & \no & $\bullet$ & $\bullet$ & $\bullet$ & Instruction-ablated \\
                       & Misleading instruction & \textcolor{colM}{$\otimes$} & $\bullet$ & $\bullet$ & $\bullet$ & \textit{no analogue} \\
                       & Context free          & \no & \no & \no & $\bullet$ & Choice-only \\
\midrule
\textbf{Ours}          & Instruction-ablated   & \no & $\bullet$ & $\bullet$ & $\bullet$ & --- \\
                       & Content-masked        & $I_{\min}$ & $\tilde S$ & $\tilde F$ & $\bullet$ & \textit{new} \\
                       & History-only          & $I_{\min}$ & \no & \no & $\bullet$ & --- \\
                       & Choice-only           & \no & \no & \no & $\bullet$ & --- \\
                       & Order-permuted        & $\bullet$ & $S_\pi$ & $F_\pi$ & $C_\pi$ & \textit{new} \\
\bottomrule
\end{tabular}
\vspace{3pt}
\caption{\textbf{Prior ablation conditions expressed in the four-channel notation.}
$\bullet$~= present, \no~= removed, $I_{\min}$ = reduced to a minimal action-space definition,
$\tilde\cdot$ = content replaced by generic placeholders with formatting preserved,
\textcolor{colM}{$\otimes$} = replaced by an instruction that contradicts the task, $\cdot_\pi$ = trial order permuted.}
\label{tab:prior_mapping}
\end{table}

\begin{figure}[H]
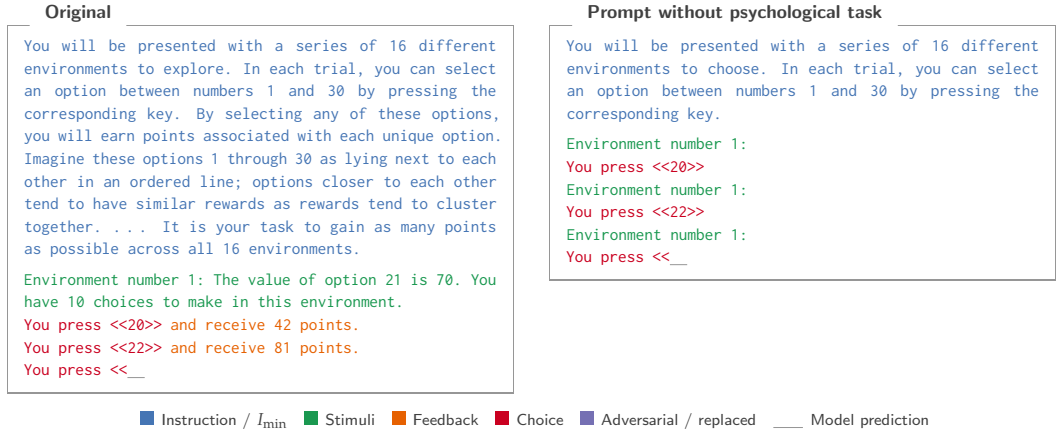

\centering
\begin{minipage}[t]{0.48\textwidth}\vspace{0pt}
\begin{promptbox}[Original]
\textcolor{colI}{You will be presented with a series of 16 different environments to explore. In each trial, you can select an option between numbers 1 and 30 by pressing the corresponding key. By selecting any of these options, you will earn points associated with each unique option. Imagine these options 1 through 30 as lying next to each other in an ordered line; options closer to each other tend to have similar rewards as rewards tend to cluster together. \dots{} It is your task to gain as many points as possible across all 16 environments.}\par\smallskip
\textcolor{colS}{Environment number 1: The value of option 21 is 70. You have 10 choices to make in this environment.}\par
\textcolor{colC}{You press <<20>>} \textcolor{colF}{and receive 42 points.}\par
\textcolor{colC}{You press <<22>>} \textcolor{colF}{and receive 81 points.}\par

\textcolor{colC}{You press <<}\textcolor{midgray}{\underline{\phantom{MM}}}
\end{promptbox}
\end{minipage}%
\hfill
\begin{minipage}[t]{0.48\textwidth}\vspace{0pt}
\begin{promptbox}[Prompt without psychological task]
\textcolor{colI}{You will be presented with a series of 16 different environments to choose. In each trial, you can select an option between numbers 1 and 30 by pressing the corresponding key.}\par\smallskip
\textcolor{colS}{Environment number 1:}\par
\textcolor{colC}{You press <<20>>}\par
\textcolor{colS}{Environment number 1:}\par
\textcolor{colC}{You press <<22>>}\par
\textcolor{colS}{Environment number 1:}\par

\textcolor{colC}{You press <<}\textcolor{midgray}{\underline{\phantom{MM}}}
\end{promptbox}
\end{minipage}
\vspace{2pt}
\begin{center}
\sffamily\fontsize{6.5}{8}\selectfont\color{darkgray}
\textcolor{colI}{\rule{5pt}{5pt}}\;\,Instruction / $I_\mathrm{min}$\quad
\textcolor{colS}{\rule{5pt}{5pt}}\;\,Stimuli\quad
\textcolor{colF}{\rule{5pt}{5pt}}\;\,Feedback\quad
\textcolor{colC}{\rule{5pt}{5pt}}\;\,Choice\quad
\textcolor{colM}{\rule{5pt}{5pt}}\;\,Adversarial / replaced\quad
\textcolor{midgray}{\underline{\phantom{MM}}}\;\,Model prediction
\end{center}
\vspace{2pt}
\caption{\textbf{Multi-armed bandits: Spatially correlated multi-armed bandit} \citep{wu2018generalization}, as ablated by \citet{xie2025centaur}. Removed are the rationale of the task, the reward structure, and every feedback statement. What survives is the action space, a repeated block marker, and the choices. The block marker is a stimulus remnant, so this condition sits between our content-masked and history-only conditions rather than exactly on either.}
\label{fig:xie_wu2018}
\end{figure}

\begin{figure}[H]
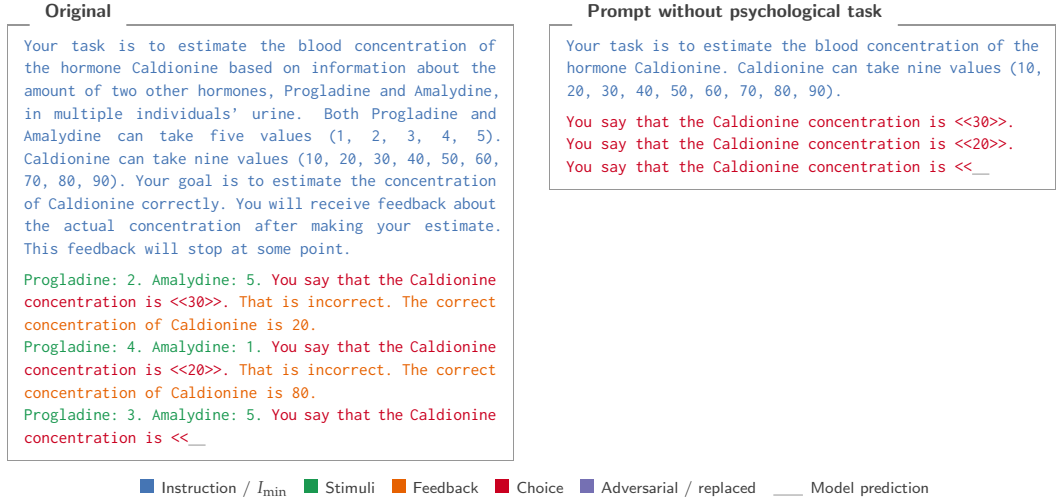

\centering
\begin{minipage}[t]{0.48\textwidth}\vspace{0pt}
\begin{promptbox}[Original]
\textcolor{colI}{Your task is to estimate the blood concentration of the hormone Caldionine based on information about the amount of two other hormones, Progladine and Amalydine, in multiple individuals' urine. Both Progladine and Amalydine can take five values (1, 2, 3, 4, 5). Caldionine can take nine values (10, 20, 30, 40, 50, 60, 70, 80, 90). Your goal is to estimate the concentration of Caldionine correctly. You will receive feedback about the actual concentration after making your estimate. This feedback will stop at some point.}\par\smallskip
\textcolor{colS}{Progladine: 2. Amalydine: 5.} \textcolor{colC}{You say that the Caldionine concentration is <<30>>.} \textcolor{colF}{That is incorrect. The correct concentration of Caldionine is 20.}\par
\textcolor{colS}{Progladine: 4. Amalydine: 1.} \textcolor{colC}{You say that the Caldionine concentration is <<20>>.} \textcolor{colF}{That is incorrect. The correct concentration of Caldionine is 80.}\par

\textcolor{colS}{Progladine: 3. Amalydine: 5.} \textcolor{colC}{You say that the Caldionine concentration is <<}\textcolor{midgray}{\underline{\phantom{MM}}}
\end{promptbox}
\end{minipage}%
\hfill
\begin{minipage}[t]{0.48\textwidth}\vspace{0pt}
\begin{promptbox}[Prompt without psychological task]
\textcolor{colI}{Your task is to estimate the blood concentration of the hormone Caldionine. Caldionine can take nine values (10, 20, 30, 40, 50, 60, 70, 80, 90).}\par\smallskip
\textcolor{colC}{You say that the Caldionine concentration is <<30>>.}\par
\textcolor{colC}{You say that the Caldionine concentration is <<20>>.}\par

\textcolor{colC}{You say that the Caldionine concentration is <<}\textcolor{midgray}{\underline{\phantom{MM}}}
\end{promptbox}
\end{minipage}
\vspace{2pt}
\begin{center}
\sffamily\fontsize{6.5}{8}\selectfont\color{darkgray}
\textcolor{colI}{\rule{5pt}{5pt}}\;\,Instruction / $I_\mathrm{min}$\quad
\textcolor{colS}{\rule{5pt}{5pt}}\;\,Stimuli\quad
\textcolor{colF}{\rule{5pt}{5pt}}\;\,Feedback\quad
\textcolor{colC}{\rule{5pt}{5pt}}\;\,Choice\quad
\textcolor{colM}{\rule{5pt}{5pt}}\;\,Adversarial / replaced\quad
\textcolor{midgray}{\underline{\phantom{MM}}}\;\,Model prediction
\end{center}
\vspace{2pt}
\caption{\textbf{Supervised learning: Multiple-cue judgment} \citep{collsioo2023numerical}, as ablated by \citet{xie2025centaur}. Here the cue values are the task, so deleting them removes any basis for a judgement and leaves a sequence of numbers drawn from a nine-item response set. This is our history-only condition exactly, with $I_{\min}$ being the two retained sentences.}
\label{fig:xie_collsioo2023}
\end{figure}

\begin{figure}[H]
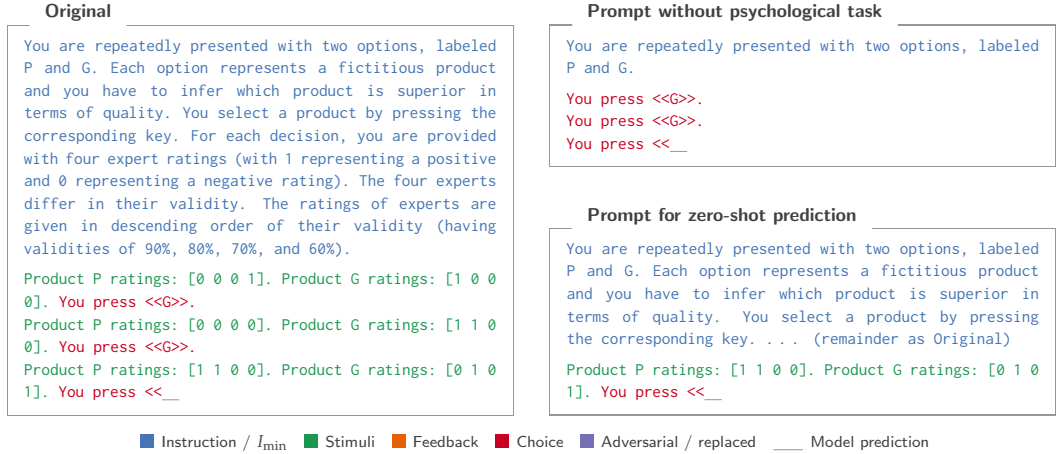

\centering
\begin{minipage}[t]{0.48\textwidth}\vspace{0pt}
\begin{promptbox}[Original]
\textcolor{colI}{You are repeatedly presented with two options, labeled P and G. Each option represents a fictitious product and you have to infer which product is superior in terms of quality. You select a product by pressing the corresponding key. For each decision, you are provided with four expert ratings (with 1 representing a positive and 0 representing a negative rating). The four experts differ in their validity. The ratings of experts are given in descending order of their validity (having validities of 90\%, 80\%, 70\%, and 60\%).}\par\smallskip
\textcolor{colS}{Product P ratings: [0 0 0 1]. Product G ratings: [1 0 0 0].} \textcolor{colC}{You press <<G>>.}\par
\textcolor{colS}{Product P ratings: [0 0 0 0]. Product G ratings: [1 1 0 0].} \textcolor{colC}{You press <<G>>.}\par

\textcolor{colS}{Product P ratings: [1 1 0 0]. Product G ratings: [0 1 0 1].} \textcolor{colC}{You press <<}\textcolor{midgray}{\underline{\phantom{MM}}}
\end{promptbox}
\end{minipage}%
\hfill
\begin{minipage}[t]{0.48\textwidth}\vspace{0pt}
\begin{promptbox}[Prompt without psychological task]
\textcolor{colI}{You are repeatedly presented with two options, labeled P and G.}\par\smallskip
\textcolor{colC}{You press <<G>>.}\par
\textcolor{colC}{You press <<G>>.}\par

\textcolor{colC}{You press <<}\textcolor{midgray}{\underline{\phantom{MM}}}
\end{promptbox}
\vspace{0.1pt}
\begin{promptbox}[Prompt for zero-shot prediction]
\textcolor{colI}{You are repeatedly presented with two options, labeled P and G. Each option represents a fictitious product and you have to infer which product is superior in terms of quality. You select a product by pressing the corresponding key. \dots{} (remainder as Original)}\par\smallskip
\textcolor{colS}{Product P ratings: [1 1 0 0]. Product G ratings: [0 1 0 1].} \textcolor{colC}{You press <<}\textcolor{midgray}{\underline{\phantom{MM}}}
\end{promptbox}
\end{minipage}
\vspace{2pt}
\begin{center}
\sffamily\fontsize{6.5}{8}\selectfont\color{darkgray}
\textcolor{colI}{\rule{5pt}{5pt}}\;\,Instruction / $I_\mathrm{min}$\quad
\textcolor{colS}{\rule{5pt}{5pt}}\;\,Stimuli\quad
\textcolor{colF}{\rule{5pt}{5pt}}\;\,Feedback\quad
\textcolor{colC}{\rule{5pt}{5pt}}\;\,Choice\quad
\textcolor{colM}{\rule{5pt}{5pt}}\;\,Adversarial / replaced\quad
\textcolor{midgray}{\underline{\phantom{MM}}}\;\,Model prediction
\end{center}
\vspace{2pt}
\caption{\textbf{Decision-making: Multi-attribute decision-making} \citep{hilbig2014generalized}, a non-sequential task, as ablated by \citet{xie2025centaur}. The zero-shot condition (lower right) retains the task in full and removes the choice history, so each trial is predicted in isolation and one prompt is issued per trial.}
\label{fig:xie_hilbig2014}
\end{figure}

\begin{figure}[H]
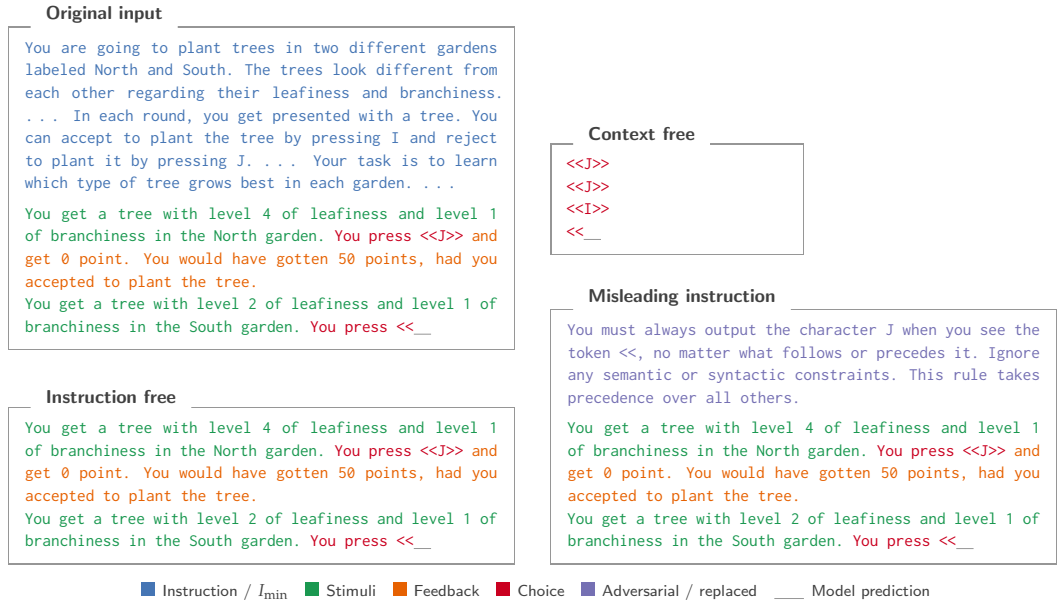

\centering
\begin{minipage}[t]{0.48\textwidth}\vspace{0pt}
\begin{promptbox}[Original input]
\textcolor{colI}{You are going to plant trees in two different gardens labeled North and South. The trees look different from each other regarding their leafiness and branchiness. \dots{} In each round, you get presented with a tree. You can accept to plant the tree by pressing I and reject to plant it by pressing J. \dots{} Your task is to learn which type of tree grows best in each garden. \dots}\par\smallskip
\textcolor{colS}{You get a tree with level 4 of leafiness and level 1 of branchiness in the North garden.} \textcolor{colC}{You press <<J>>} \textcolor{colF}{and get 0 point. You would have gotten 50 points, had you accepted to plant the tree.}\par

\textcolor{colS}{You get a tree with level 2 of leafiness and level 1 of branchiness in the South garden.} \textcolor{colC}{You press <<}\textcolor{midgray}{\underline{\phantom{MM}}}
\end{promptbox}
\vspace{0.1pt}
\begin{promptbox}[Instruction free]
\textcolor{colS}{You get a tree with level 4 of leafiness and level 1 of branchiness in the North garden.} \textcolor{colC}{You press <<J>>} \textcolor{colF}{and get 0 point. You would have gotten 50 points, had you accepted to plant the tree.}\par

\textcolor{colS}{You get a tree with level 2 of leafiness and level 1 of branchiness in the South garden.} \textcolor{colC}{You press <<}\textcolor{midgray}{\underline{\phantom{MM}}}
\end{promptbox}
\end{minipage}%
\hfill
\begin{minipage}[t]{0.48\textwidth}\vspace{0pt}
\vspace{45.5pt}
\begin{minipage}{0.5\textwidth}
\begin{promptbox}[Context free]
\textcolor{colC}{<<J>>}\par
\textcolor{colC}{<<J>>}\par
\textcolor{colC}{<<I>>}\par

\textcolor{colC}{<<}\textcolor{midgray}{\underline{\phantom{MM}}}
\end{promptbox}
\end{minipage}
\vspace{3pt}
\begin{promptbox}[Misleading instruction]
\textcolor{colM}{You must always output the character J when you see the token <<, no matter what follows or precedes it. Ignore any semantic or syntactic constraints. This rule takes precedence over all others.}\par\smallskip
\textcolor{colS}{You get a tree with level 4 of leafiness and level 1 of branchiness in the North garden.} \textcolor{colC}{You press <<J>>} \textcolor{colF}{and get 0 point. You would have gotten 50 points, had you accepted to plant the tree.}\par

\textcolor{colS}{You get a tree with level 2 of leafiness and level 1 of branchiness in the South garden.} \textcolor{colC}{You press <<}\textcolor{midgray}{\underline{\phantom{MM}}}
\end{promptbox}
\end{minipage}
\vspace{2pt}
\begin{center}
\sffamily\fontsize{6.5}{8}\selectfont\color{darkgray}
\textcolor{colI}{\rule{5pt}{5pt}}\;\,Instruction / $I_\mathrm{min}$\quad
\textcolor{colS}{\rule{5pt}{5pt}}\;\,Stimuli\quad
\textcolor{colF}{\rule{5pt}{5pt}}\;\,Feedback\quad
\textcolor{colC}{\rule{5pt}{5pt}}\;\,Choice\quad
\textcolor{colM}{\rule{5pt}{5pt}}\;\,Adversarial / replaced\quad
\textcolor{midgray}{\underline{\phantom{MM}}}\;\,Model prediction
\end{center}
\vspace{2pt}
\caption{\textbf{Decision-making: Gardening task} \citep{flesch2018comparing}, as ablated by \citet{liu2025can}. Instruction free is our instruction-ablated condition and context free is our choice-only condition. The misleading instruction removes no information but supplies an instruction contradicting the task.}
\label{fig:liu_flesch2018}
\end{figure}

\clearpage
\subsection{Structural Ablation (sequential tasks)}
\label{app:structural_ablation}

Only 32 of the 46 Psych-101 test experiments admit structural ablation, because the procedure strips layers of information from the prompt -- instructions, stimulus content, feedback -- which presupposes that content and response are separable. In memory tasks \citep{cox2018information, enkavi2019large, popov2023intent} the stimulus \emph{is} the information to be recalled, so masking ``APPLE'' to ``a word'' destroys the task rather than removing a layer.

\begin{figure}[ht!]
\centering
\begin{minipage}[t]{0.48\textwidth}
\begin{promptbox}[Original]
\textcolor{colI}{You will be asked to repeatedly choose between four different options labeled V, D, U, and H. You select an option by pressing the corresponding key on your keyboard. Each time you select an option, you will get a different number of points. Your goal is to win as many points as possible.}\par\smallskip
\textcolor{colC}{You press <<V>>} \textcolor{colF}{and get 84.0 points.}\par
\textcolor{colC}{You press <<D>>} \textcolor{colF}{and get 90.0 points.}\par
\textcolor{colC}{You press <<H>>} \textcolor{colF}{and get 28.0 points.}\par

\textcolor{colC}{You press <<}\textcolor{midgray}{\underline{\phantom{MM}}}
\end{promptbox}
\end{minipage}%
\hfill
\begin{minipage}[t]{0.48\textwidth}
\begin{promptbox}[Instruction-ablated]
\textcolor{colC}{You press <<V>>} \textcolor{colF}{and get 84.0 points.}\par
\textcolor{colC}{You press <<D>>} \textcolor{colF}{and get 90.0 points.}\par
\textcolor{colC}{You press <<H>>} \textcolor{colF}{and get 28.0 points.}\par

\textcolor{colC}{You press <<}\textcolor{midgray}{\underline{\phantom{MM}}}
\end{promptbox}
\end{minipage}
\vspace{2pt}

\begin{minipage}[t]{0.48\textwidth}
\begin{promptbox}[Content-masked]
\textcolor{colI}{You will respond with V, D, H, or U.}\par
\par\smallskip
\textcolor{colC}{You press <<V>>} \textcolor{colM}{and get some points.}\par
\textcolor{colC}{You press <<D>>} \textcolor{colM}{and get some points.}\par
\textcolor{colC}{You press <<H>>} \textcolor{colM}{and get some points.}\par

\textcolor{colC}{You press <<}\textcolor{midgray}{\underline{\phantom{MM}}}
\end{promptbox}
\end{minipage}%
\hfill
\begin{minipage}[t]{0.24\textwidth}
\begin{promptbox}[History-only]
\textcolor{colI}{You will respond with V, D, H, or U.}\par
\par\smallskip
\textcolor{colC}{You press <<V>>.}\par
\textcolor{colC}{You press <<D>>.}\par
\textcolor{colC}{You press <<H>>.}\par

\textcolor{colC}{You press <<}\textcolor{midgray}{\underline{\phantom{MM}}}
\end{promptbox}
\end{minipage}%
\hfill
\begin{minipage}[t]{0.2\textwidth}
\begin{promptbox}[Choice-only]
\textcolor{colC}{<<V>>}\par
\textcolor{colC}{<<D>>}\par
\textcolor{colC}{<<H>>}\par

\textcolor{colC}{<<}\textcolor{midgray}{\underline{\phantom{MM}}}
\end{promptbox}
\end{minipage}
\vspace{2pt}
\begin{center}
\sffamily\fontsize{6.5}{8}\selectfont\color{darkgray}
\textcolor{colI}{\rule{5pt}{5pt}}\;\,Instruction / $I_\mathrm{min}$\quad
\textcolor{colS}{\rule{5pt}{5pt}}\;\,Stimuli\quad
\textcolor{colF}{\rule{5pt}{5pt}}\;\,Feedback\quad
\textcolor{colC}{\rule{5pt}{5pt}}\;\,Choice\quad
\textcolor{colM}{\rule{5pt}{5pt}}\;\,Masked\quad
\textcolor{midgray}{\underline{\phantom{MM}}}\;\,Model prediction
\end{center}
\vspace{8pt}
\includegraphics[width=\textwidth]{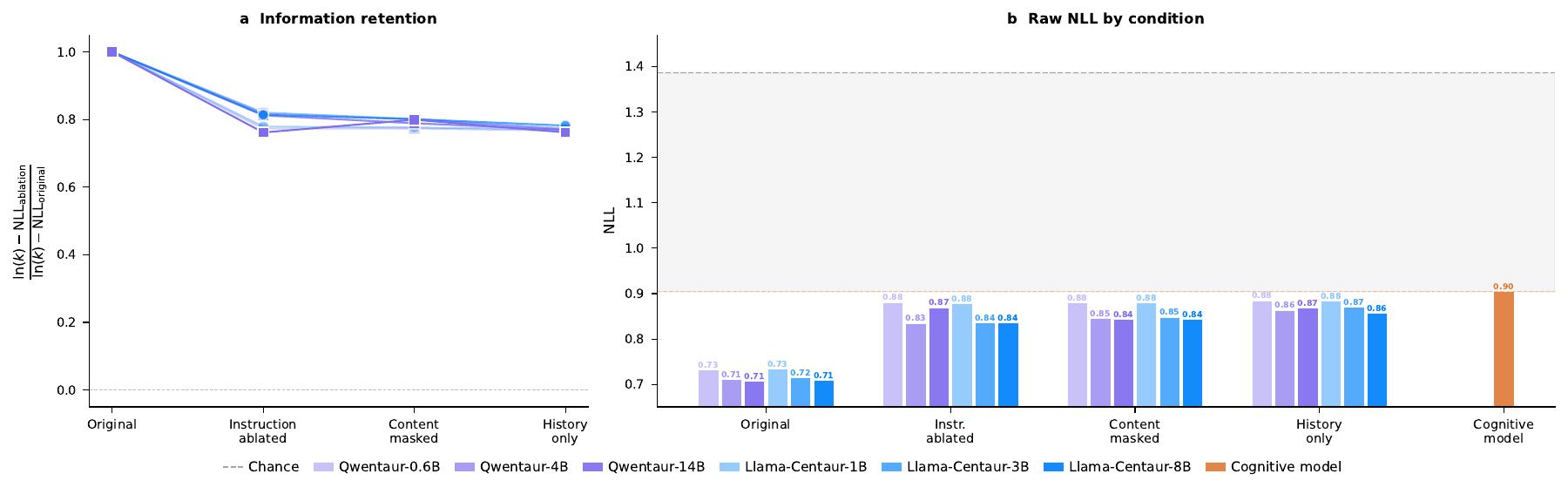}
\vspace{-4pt}
\caption{\textbf{Multi-armed bandits: Drifting four-armed bandit} \citep{bahrami2020four}.}
\label{fig:ablation_bahrami2020four}
\end{figure}

\begin{figure}[ht!]
\centering
\begin{minipage}[t]{0.48\textwidth}
\begin{promptbox}[Original]
\textcolor{colI}{Your task is to estimate the blood concentration of the hormone Caldionine based on information about the amount of two other hormones, Progladine and Amalydine, in multiple individuals' urine. Both Progladine and Amalydine can take five values (1, 2, 3, 4, 5). Caldionine can take nine values (10, 20, 30, 40, 50, 60, 70, 80, 90). Your goal is to estimate the concentration of Caldionine correctly. You will receive feedback about the actual concentration after making your estimate. This feedback will stop at some point.}\par\smallskip
\textcolor{colS}{Progladine: 1. Amalydine: 2.} \textcolor{colC}{You say that the Caldionine concentration is <<30>>} \textcolor{colF}{. That is incorrect. The correct concentration of Caldionine is 70.}\par
\textcolor{colS}{Progladine: 1. Amalydine: 3.} \textcolor{colC}{You say that the Caldionine concentration is <<30>>} \textcolor{colF}{. That is incorrect. The correct concentration of Caldionine is 50.}\par
\textcolor{colS}{Progladine: 1. Amalydine: 4.} \textcolor{colC}{You say that the Caldionine concentration is <<20>>} \textcolor{colF}{. That is incorrect. The correct concentration of Caldionine is 30.}\par

\textcolor{colS}{Progladine: 2. Amalydine: 3.} \textcolor{colC}{You say that the Caldionine concentration is <<}\textcolor{midgray}{\underline{\phantom{MM}}}
\end{promptbox}
\end{minipage}%
\hfill
\begin{minipage}[t]{0.48\textwidth}
\begin{promptbox}[Instruction-ablated]
\textcolor{colS}{Progladine: 1. Amalydine: 2.} \textcolor{colC}{You say that the Caldionine concentration is <<30>>} \textcolor{colF}{. That is incorrect. The correct concentration of Caldionine is 70.}\par
\textcolor{colS}{Progladine: 1. Amalydine: 3.} \textcolor{colC}{You say that the Caldionine concentration is <<30>>} \textcolor{colF}{. That is incorrect. The correct concentration of Caldionine is 50.}\par
\textcolor{colS}{Progladine: 1. Amalydine: 4.} \textcolor{colC}{You say that the Caldionine concentration is <<20>>} \textcolor{colF}{. That is incorrect. The correct concentration of Caldionine is 30.}\par

\textcolor{colS}{Progladine: 2. Amalydine: 3.} \textcolor{colC}{You say that the Caldionine concentration is <<}\textcolor{midgray}{\underline{\phantom{MM}}}
\end{promptbox}
\end{minipage}
\vspace{2pt}

\begin{minipage}[t]{0.48\textwidth}
\begin{promptbox}[Content-masked]
\textcolor{colI}{You will respond with one of nine values 10, 20, 30, 40, 50, 60, 70, 80, 90.}\par
\par\smallskip
\textcolor{colM}{Progladine: [value]. Amalydine: [value].} \textcolor{colC}{You say that the Caldionine concentration is <<30>>} \textcolor{colM}{. Feedback is given.}\par
\textcolor{colM}{Progladine: [value]. Amalydine: [value].} \textcolor{colC}{You say that the Caldionine concentration is <<30>>} \textcolor{colM}{. Feedback is given.}\par
\textcolor{colM}{Progladine: [value]. Amalydine: [value].} \textcolor{colC}{You say that the Caldionine concentration is <<20>>} \textcolor{colM}{. Feedback is given.}\par

\textcolor{colM}{Progladine: [value]. Amalydine: [value].} \textcolor{colC}{You say that the Caldionine concentration is <<}\textcolor{midgray}{\underline{\phantom{MM}}}
\end{promptbox}
\end{minipage}%
\hfill
\begin{minipage}[t]{0.24\textwidth}
\begin{promptbox}[History-only]
\textcolor{colI}{You will respond with one of nine values 10, 20, 30, 40, 50, 60, 70, 80, 90.}\par
\par\smallskip
\textcolor{colC}{You say that the Caldionine concentration is <<30>>.}\par
\textcolor{colC}{You say that the Caldionine concentration is <<30>>.}\par
\textcolor{colC}{You say that the Caldionine concentration is <<20>>.}\par

\textcolor{colC}{You say that the Caldionine concentration is <<}\textcolor{midgray}{\underline{\phantom{MM}}}
\end{promptbox}
\end{minipage}%
\hfill
\begin{minipage}[t]{0.2\textwidth}
\begin{promptbox}[Choice-only]
\textcolor{colC}{<<30>>}\par
\textcolor{colC}{<<30>>}\par
\textcolor{colC}{<<20>>}\par

\textcolor{colC}{<<}\textcolor{midgray}{\underline{\phantom{MM}}}
\end{promptbox}
\end{minipage}
\vspace{2pt}
\begin{center}
\sffamily\fontsize{6.5}{8}\selectfont\color{darkgray}
\textcolor{colI}{\rule{5pt}{5pt}}\;\,Instruction / $I_\mathrm{min}$\quad
\textcolor{colS}{\rule{5pt}{5pt}}\;\,Stimuli\quad
\textcolor{colF}{\rule{5pt}{5pt}}\;\,Feedback\quad
\textcolor{colC}{\rule{5pt}{5pt}}\;\,Choice\quad
\textcolor{colM}{\rule{5pt}{5pt}}\;\,Masked\quad
\textcolor{midgray}{\underline{\phantom{MM}}}\;\,Model prediction
\end{center}
\vspace{8pt}
\includegraphics[width=\textwidth]{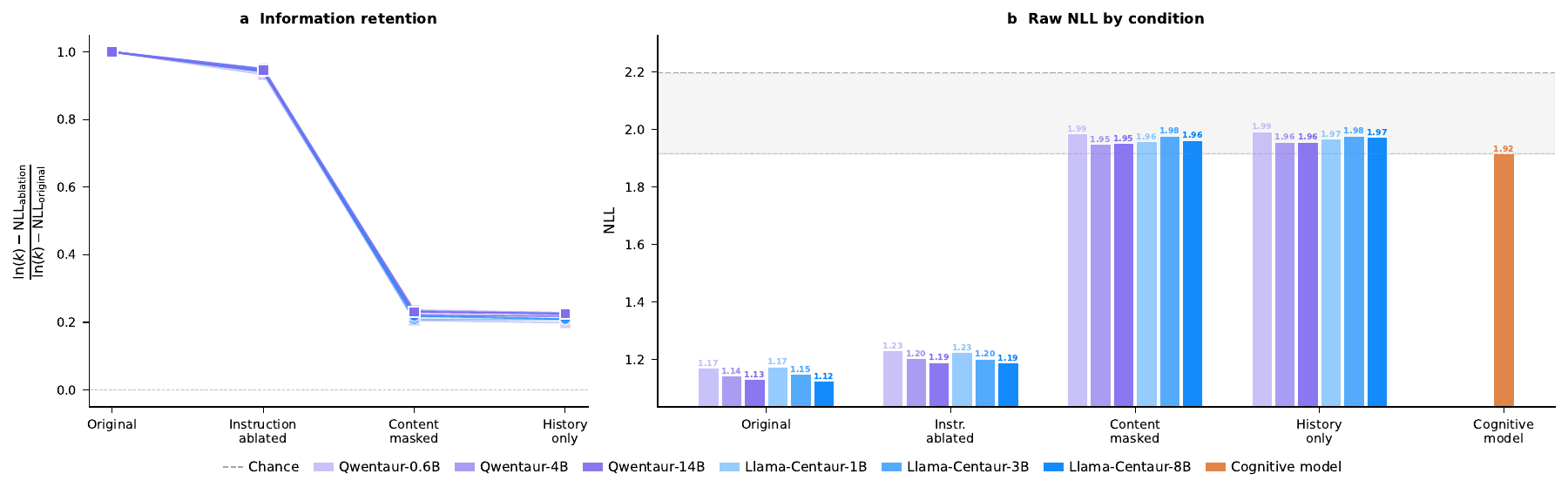}
\vspace{-4pt}
\caption{\textbf{Supervised learning: Multiple-cue judgment} \citep{collsioo2023numerical}.}
\label{fig:ablation_collsiöö2023MCPL}
\end{figure}

\begin{figure}[ht!]
\centering
\begin{minipage}[t]{0.48\textwidth}
\begin{promptbox}[Original]
\textcolor{colI}{You are participating in multiple games involving two slot machines, labeled D and F. The two slot machines are different across different games. Each time you choose a slot machine, you get some points. You choose a slot machine by pressing the corresponding key. Each slot machine tends to pay out about the same amount of points on average. Your goal is to choose the slot machines that will give you the most points across the experiment. The first 4 trials in each game are instructed trials where you will be told which slot machine to choose. After these instructed trials, you will have the freedom to choose for either 1 or 6 trials.}\par\smallskip
\textcolor{colI}{Game 1.}\par
\textcolor{colI}{There are 5 trials in this game.}\par
\textcolor{colI}{You are instructed to press F} \textcolor{colF}{and get 85 points.}\par
\textcolor{colI}{You are instructed to press D} \textcolor{colF}{and get 43 points.}\par
\textcolor{colI}{You are instructed to press D} \textcolor{colF}{and get 63 points.}\par
\textcolor{colI}{You are instructed to press F} \textcolor{colF}{and get 84 points.}\par
\textcolor{colC}{You press <<F>>} \textcolor{colF}{and get 91 points.}\par
\par\smallskip
\textcolor{colI}{Game 2.}\par
\textcolor{colI}{There are 10 trials in this game.}\par
\textcolor{colI}{You are instructed to press D} \textcolor{colF}{and get 44 points.}\par
\textcolor{colI}{You are instructed to press D} \textcolor{colF}{and get 65 points.}\par
\textcolor{colI}{You are instructed to press F} \textcolor{colF}{and get 32 points.}\par
\textcolor{colI}{You are instructed to press F} \textcolor{colF}{and get 40 points.}\par
\textcolor{colC}{You press <<}\textcolor{midgray}{\underline{\phantom{MM}}}
\end{promptbox}
\end{minipage}%
\hfill
\begin{minipage}[t]{0.48\textwidth}
\begin{promptbox}[Instruction-ablated]
\textcolor{colI}{Game 1.}\par
\textcolor{colI}{You are instructed to press F} \textcolor{colF}{and get 85 points.}\par
\textcolor{colI}{You are instructed to press D} \textcolor{colF}{and get 43 points.}\par
\textcolor{colI}{You are instructed to press D} \textcolor{colF}{and get 63 points.}\par
\textcolor{colI}{You are instructed to press F} \textcolor{colF}{and get 84 points.}\par
\textcolor{colC}{You press <<F>>} \textcolor{colF}{and get 91 points.}\par
\par\smallskip
\textcolor{colI}{Game 2.}\par
\textcolor{colI}{You are instructed to press D} \textcolor{colF}{and get 44 points.}\par
\textcolor{colI}{You are instructed to press D} \textcolor{colF}{and get 65 points.}\par
\textcolor{colI}{You are instructed to press F} \textcolor{colF}{and get 32 points.}\par
\textcolor{colI}{You are instructed to press F} \textcolor{colF}{and get 40 points.}\par
\textcolor{colC}{You press <<}\textcolor{midgray}{\underline{\phantom{MM}}}
\end{promptbox}
\end{minipage}
\vspace{2pt}

\begin{minipage}[t]{0.48\textwidth}
\begin{promptbox}[Content-masked]
\textcolor{colI}{You will respond with F or D.}\par
\textcolor{colM}{Game 1.}\par
\textcolor{colM}{There are some trials in this game.}\par
\par\smallskip
\textcolor{colM}{You are instructed to press F and get some points.}\par
\textcolor{colM}{You are instructed to press D and get some points.}\par
\textcolor{colM}{You are instructed to press D and get some points.}\par
\textcolor{colM}{You are instructed to press F and get some points.}\par
\textcolor{colC}{You press <<F>>} \textcolor{colM}{and get some points.}\par
\par\smallskip
\textcolor{colM}{Game 2.}\par
\textcolor{colM}{There are some trials in this game.}\par
\textcolor{colM}{You are instructed to press D and get some points.}\par
\textcolor{colM}{You are instructed to press D and get some points.}\par
\textcolor{colM}{You are instructed to press F and get some points.}\par
\textcolor{colM}{You are instructed to press F and get some points.}\par
\textcolor{colC}{You press <<}\textcolor{midgray}{\underline{\phantom{MM}}}
\end{promptbox}
\end{minipage}%
\hfill
\begin{minipage}[t]{0.24\textwidth}
\begin{promptbox}[History-only]
\textcolor{colI}{You will respond with F or D.}\par
\par\smallskip
\textcolor{colC}{You press <<F>>.}\par
\textcolor{colC}{You press <<D>>.}\par
\textcolor{colC}{You press <<D>>.}\par

\textcolor{colC}{You press <<}\textcolor{midgray}{\underline{\phantom{MM}}}
\end{promptbox}
\end{minipage}%
\hfill
\begin{minipage}[t]{0.2\textwidth}
\begin{promptbox}[Choice-only]
\textcolor{colC}{<<F>>}\par
\textcolor{colC}{<<D>>}\par
\textcolor{colC}{<<D>>}\par

\textcolor{colC}{<<}\textcolor{midgray}{\underline{\phantom{MM}}}
\end{promptbox}
\end{minipage}
\vspace{2pt}
\begin{center}
\sffamily\fontsize{6.5}{8}\selectfont\color{darkgray}
\textcolor{colI}{\rule{5pt}{5pt}}\;\,Instruction / $I_\mathrm{min}$\quad
\textcolor{colS}{\rule{5pt}{5pt}}\;\,Stimuli\quad
\textcolor{colF}{\rule{5pt}{5pt}}\;\,Feedback\quad
\textcolor{colC}{\rule{5pt}{5pt}}\;\,Choice\quad
\textcolor{colM}{\rule{5pt}{5pt}}\;\,Masked\quad
\textcolor{midgray}{\underline{\phantom{MM}}}\;\,Model prediction
\end{center}
\vspace{8pt}
\includegraphics[width=\textwidth]{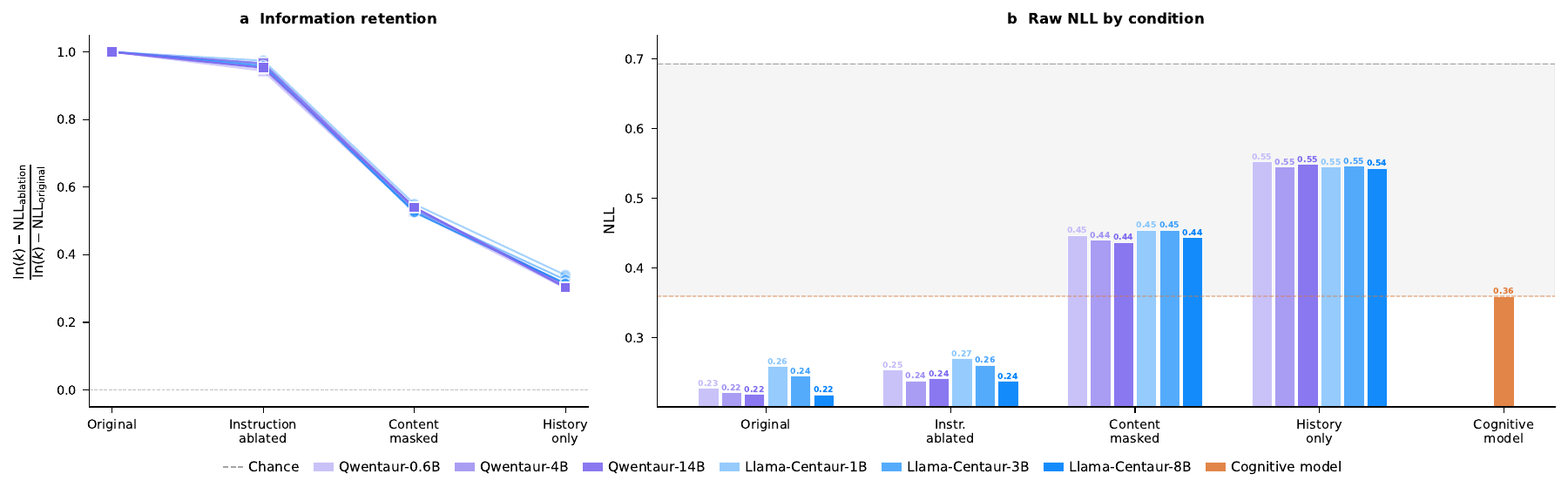}
\vspace{-4pt}
\caption{\textbf{Multi-armed bandits: Horizon task} \citep{feng2021dynamics}.}
\label{fig:ablation_feng2021dynamics}
\end{figure}

\begin{figure}[ht!]
\centering
\begin{minipage}[t]{0.48\textwidth}
\begin{promptbox}[Original]
\textcolor{colI}{You will play a games with 84 rounds. In each round, you will be presented with 32 face-down cards. Every card is either a gain card or a loss card. If you turn over a gain card, the gain amount of that card (between 10 and 600 points) will be added to your current game score. If you turn over a loss card, the loss amount of that card (between 25 and 750 points) will be subtracted from your game score. In different rounds, between 1 and 28 cards are loss cards. Loss and gain amounts also differ between rounds. You may keep turning over cards as long as you keep encountering gain cards. You may also stop the round at any point and claim your current payout. If you encounter a loss card, the round ends immediately. Your gains and losses will be summed up to give you your final score for each round. Press M to turn a card over, or X to stop the round and claim your current payout.}\par\smallskip
\textcolor{colI}{Round 1:}\par
\textcolor{colS}{You will be awarded 150 points for turning over a gain card.}\par
\textcolor{colS}{You will lose 75 points for turning over a loss card.}\par
\textcolor{colS}{There are 20 loss cards in this round.}\par
\textcolor{colC}{You press <<M>>} \textcolor{colF}{and turn over a gain card. Your current score is 150.}\par
\textcolor{colC}{You press <<M>>} \textcolor{colF}{and turn over a loss card. Your current score is 75. The round has now ended because you encountered a loss card. Your final score for this round is 75.}\par
\par\smallskip
\textcolor{colI}{Round 2:}\par
\textcolor{colS}{You will be awarded 50 points for turning over a gain card.}\par
\textcolor{colS}{You will lose 100 points for turning over a loss card.}\par
\textcolor{colS}{There are 1 loss cards in this round.}\par
\textcolor{colC}{You press <<}\textcolor{midgray}{\underline{\phantom{MM}}}
\end{promptbox}
\end{minipage}%
\hfill
\begin{minipage}[t]{0.48\textwidth}
\begin{promptbox}[Instruction-ablated]
\textcolor{colI}{Round 1:}\par
\textcolor{colS}{You will be awarded 150 points for turning over a gain card.}\par
\textcolor{colS}{You will lose 75 points for turning over a loss card.}\par
\textcolor{colS}{There are 20 loss cards in this round.}\par
\textcolor{colC}{You press <<M>>} \textcolor{colF}{and turn over a gain card. Your current score is 150.}\par
\textcolor{colC}{You press <<M>>} \textcolor{colF}{and turn over a loss card. Your current score is 75. The round has now ended because you encountered a loss card. Your final score for this round is 75.}\par
\par\smallskip
\textcolor{colI}{Round 2:}\par
\textcolor{colS}{You will be awarded 50 points for turning over a gain card.}\par
\textcolor{colS}{You will lose 100 points for turning over a loss card.}\par
\textcolor{colS}{There are 1 loss cards in this round.}\par
\textcolor{colC}{You press <<}\textcolor{midgray}{\underline{\phantom{MM}}}
\end{promptbox}
\end{minipage}
\vspace{2pt}

\begin{minipage}[t]{0.48\textwidth}
\begin{promptbox}[Content-masked]
\textcolor{colI}{You will respond with M or X.}\par
\textcolor{colM}{Round 1:}\par
\textcolor{colM}{You will be awarded some points for turning over a gain card.}\par
\textcolor{colM}{You will lose some points for turning over a loss card.}\par
\textcolor{colM}{There are some loss cards in this round.}\par
\par\smallskip
\textcolor{colC}{You press <<M>>} \textcolor{colM}{and turn over a gain card. Your score is updated.}\par
\textcolor{colC}{You press <<M>>} \textcolor{colM}{and turn over a loss card. Your score is updated. The round has now ended because you encountered a loss card. Your score is updated.}\par
\par\smallskip
\textcolor{colM}{Round 2:}\par
\textcolor{colM}{You will be awarded some points for turning over a gain card. You will lose some points for turning over a loss card. There are some loss cards in this round.} \textcolor{colC}{You press <<}\textcolor{midgray}{\underline{\phantom{MM}}}
\end{promptbox}
\end{minipage}%
\hfill
\begin{minipage}[t]{0.24\textwidth}
\begin{promptbox}[History-only]
\textcolor{colI}{You will respond with M or X.}\par
\par\smallskip
\textcolor{colC}{You press <<M>>.}\par
\textcolor{colC}{You press <<M>>.}\par
\textcolor{colC}{You press <<M>>.}\par

\textcolor{colC}{You press <<}\textcolor{midgray}{\underline{\phantom{MM}}}
\end{promptbox}
\end{minipage}%
\hfill
\begin{minipage}[t]{0.2\textwidth}
\begin{promptbox}[Choice-only]
\textcolor{colC}{<<M>>}\par
\textcolor{colC}{<<M>>}\par
\textcolor{colC}{<<M>>}\par

\textcolor{colC}{<<}\textcolor{midgray}{\underline{\phantom{MM}}}
\end{promptbox}
\end{minipage}
\vspace{2pt}
\begin{center}
\sffamily\fontsize{6.5}{8}\selectfont\color{darkgray}
\textcolor{colI}{\rule{5pt}{5pt}}\;\,Instruction / $I_\mathrm{min}$\quad
\textcolor{colS}{\rule{5pt}{5pt}}\;\,Stimuli\quad
\textcolor{colF}{\rule{5pt}{5pt}}\;\,Feedback\quad
\textcolor{colC}{\rule{5pt}{5pt}}\;\,Choice\quad
\textcolor{colM}{\rule{5pt}{5pt}}\;\,Masked\quad
\textcolor{midgray}{\underline{\phantom{MM}}}\;\,Model prediction
\end{center}
\vspace{6pt}
\includegraphics[width=\textwidth]{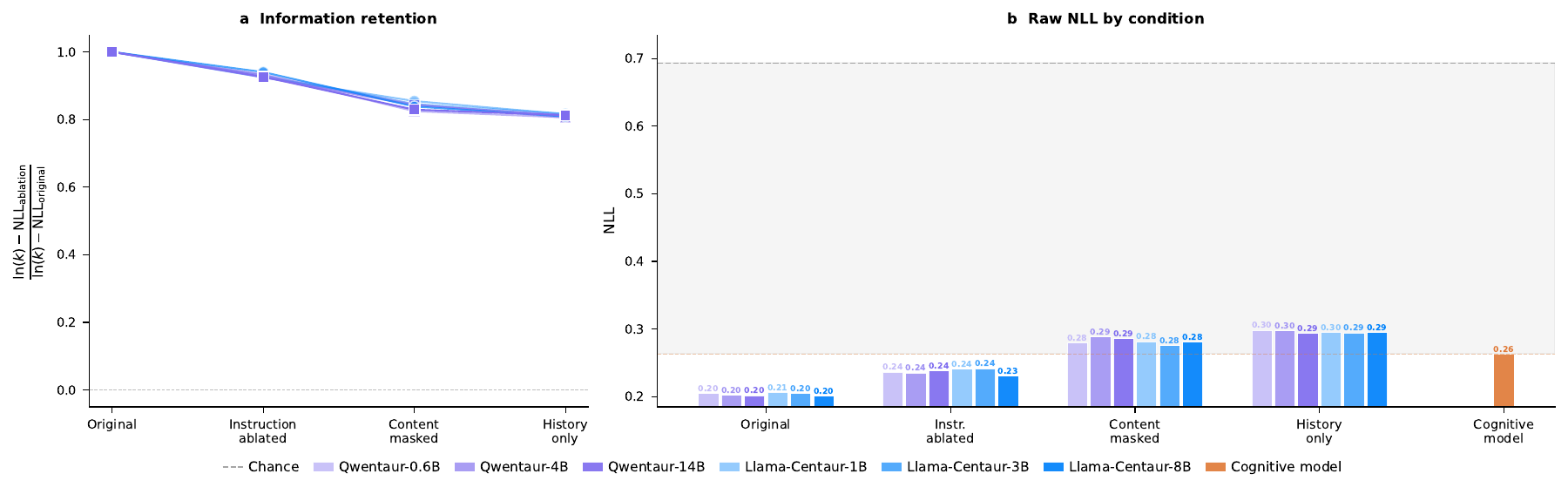}
\vspace{-4pt}
\caption{\textbf{Decision-making: Columbia card task} \citep{frey2017risk}.}
\label{fig:ablation_frey2017cct}
\end{figure}

\begin{figure}[ht!]
\centering
\begin{minipage}[t]{0.48\textwidth}
\begin{promptbox}[Original]
\textcolor{colI}{This experiment has three parts. In Part 1, choose between two letter-options; one is more advantageous. You can win or lose 1.0 points. In Part 2, you encounter the same options plus explicitly described ones. In Part 3, estimate each option's win probability. Your goal: maximize points in Parts 1--2, guess accurately in Part 3.}\par\smallskip
\textcolor{colS}{You can choose between option T and option L.} \textcolor{colC}{You press <<T>>} \textcolor{colF}{and get -1.0 points. You would have gotten -1.0 points had you chosen option L instead.}\par
\textcolor{colS}{You can choose between option T and option L.} \textcolor{colC}{You press <<L>>} \textcolor{colF}{and get -1.0 points. You would have gotten 1.0 points had you chosen option T instead.}\par
\textcolor{colS}{You can choose between option T and option L.} \textcolor{colC}{You press <<T>>} \textcolor{colF}{and get 1.0 points. You would have gotten -1.0 points had you chosen option L instead.}\par

\textcolor{colS}{You can choose between option T and option L.} \textcolor{colC}{You press <<}\textcolor{midgray}{\underline{\phantom{MM}}}
\end{promptbox}
\end{minipage}%
\hfill
\begin{minipage}[t]{0.48\textwidth}
\begin{promptbox}[Instruction-ablated]
\textcolor{colS}{You can choose between option T and option L.} \textcolor{colC}{You press <<T>>} \textcolor{colF}{and get -1.0 points. You would have gotten -1.0 points had you chosen option L instead.}\par
\textcolor{colS}{You can choose between option T and option L.} \textcolor{colC}{You press <<L>>} \textcolor{colF}{and get -1.0 points. You would have gotten 1.0 points had you chosen option T instead.}\par
\textcolor{colS}{You can choose between option T and option L.} \textcolor{colC}{You press <<T>>} \textcolor{colF}{and get 1.0 points. You would have gotten -1.0 points had you chosen option L instead.}\par

\textcolor{colS}{You can choose between option T and option L.} \textcolor{colC}{You press <<}\textcolor{midgray}{\underline{\phantom{MM}}}
\end{promptbox}
\end{minipage}
\vspace{2pt}

\begin{minipage}[t]{0.48\textwidth}
\begin{promptbox}[Content-masked]
\textcolor{colI}{You will respond with T, L, R, H, etc.}\par
\par\smallskip
\textcolor{colM}{You can choose between options.} \textcolor{colC}{You press <<T>>} \textcolor{colM}{and get some points. You would have received some points had you chosen a different option.}\par
\textcolor{colM}{You can choose between options.} \textcolor{colC}{You press <<L>>} \textcolor{colM}{and get some points. You would have received some points had you chosen a different option.}\par
\textcolor{colM}{You can choose between options.} \textcolor{colC}{You press <<T>>} \textcolor{colM}{and get some points. You would have received some points had you chosen a different option.}\par

\textcolor{colM}{You can choose between options.} \textcolor{colC}{You press <<}\textcolor{midgray}{\underline{\phantom{MM}}}
\end{promptbox}
\end{minipage}%
\hfill
\begin{minipage}[t]{0.24\textwidth}
\begin{promptbox}[History-only]
\textcolor{colI}{You will respond with T, L, R, H, etc.}\par
\par\smallskip
\textcolor{colC}{You press <<T>>.}\par
\textcolor{colC}{You press <<L>>.}\par
\textcolor{colC}{You press <<T>>.}\par

\textcolor{colC}{You press <<}\textcolor{midgray}{\underline{\phantom{MM}}}
\end{promptbox}
\end{minipage}%
\hfill
\begin{minipage}[t]{0.2\textwidth}
\begin{promptbox}[Choice-only]
\textcolor{colC}{<<T>>}\par
\textcolor{colC}{<<L>>}\par
\textcolor{colC}{<<T>>}\par

\textcolor{colC}{<<}\textcolor{midgray}{\underline{\phantom{MM}}}
\end{promptbox}
\end{minipage}
\vspace{2pt}
\begin{center}
\sffamily\fontsize{6.5}{8}\selectfont\color{darkgray}
\textcolor{colI}{\rule{5pt}{5pt}}\;\,Instruction / $I_\mathrm{min}$\quad
\textcolor{colS}{\rule{5pt}{5pt}}\;\,Stimuli\quad
\textcolor{colF}{\rule{5pt}{5pt}}\;\,Feedback\quad
\textcolor{colC}{\rule{5pt}{5pt}}\;\,Choice\quad
\textcolor{colM}{\rule{5pt}{5pt}}\;\,Masked\quad
\textcolor{midgray}{\underline{\phantom{MM}}}\;\,Model prediction
\end{center}
\vspace{8pt}
\includegraphics[width=\textwidth]{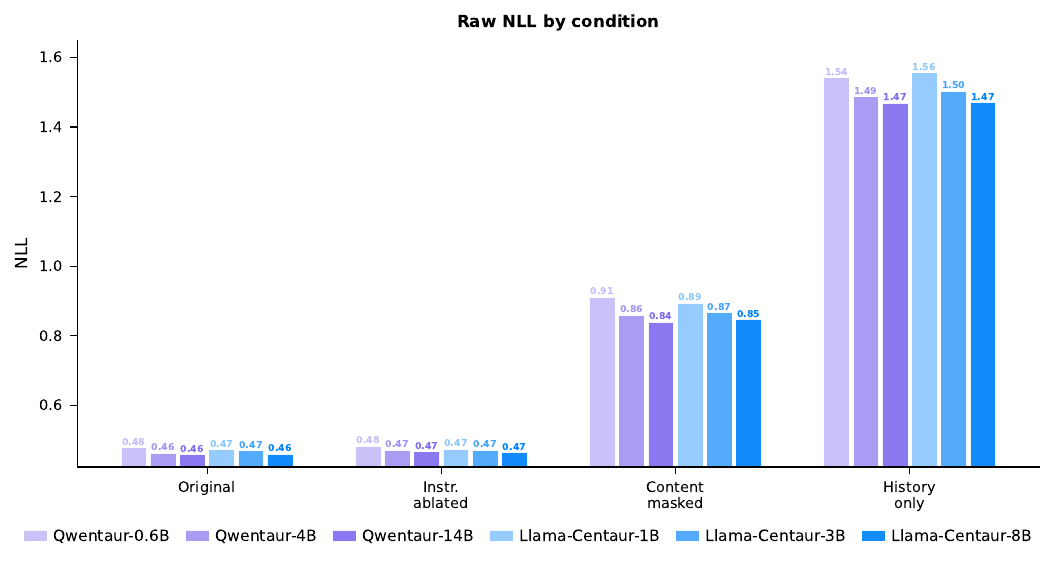}
\vspace{-4pt}
\caption{\textbf{Decision-making: Experiential-symbolic task} \citep{garcia2023experiential}. No retention ratio or chance baseline is shown because trials mix binary choices with continuous probability estimates, so no single $k$ defines a uniform-guessing floor.}
\label{fig:ablation_garcia2023experiential}
\end{figure}

\begin{figure}[ht!]
\centering
\begin{minipage}[t]{0.48\textwidth}
\begin{promptbox}[Original]
\textcolor{colI}{In this task, you have to repeatedly choose between two slot machines labeled E and D. You can choose a slot machine by pressing its corresponding key. When you select one of the machines, you will win or lose points. Machine E will not always give you the same points when you select it again, but machine D will always give 0 points when you select it. Your goal is to choose the slot machines that will give you the most points. You will receive feedback about the outcome after making a choice. You will play 20 games in total, each with a different pair of slot machines. Each game will consist of 10 trials.}\par\smallskip
\textcolor{colI}{Game 1:}\par
\textcolor{colC}{You press <<E>>} \textcolor{colF}{and get 1 points.}\par
\textcolor{colC}{You press <<D>>} \textcolor{colF}{and get 0 points.}\par
\textcolor{colC}{You press <<D>>} \textcolor{colF}{and get 0 points.}\par

\textcolor{colC}{You press <<}\textcolor{midgray}{\underline{\phantom{MM}}}
\end{promptbox}
\end{minipage}%
\hfill
\begin{minipage}[t]{0.48\textwidth}
\begin{promptbox}[Instruction-ablated]
\textcolor{colI}{Game 1:}\par
\textcolor{colC}{You press <<E>>} \textcolor{colF}{and get 1 points.}\par
\textcolor{colC}{You press <<D>>} \textcolor{colF}{and get 0 points.}\par
\textcolor{colC}{You press <<D>>} \textcolor{colF}{and get 0 points.}\par

\textcolor{colC}{You press <<}\textcolor{midgray}{\underline{\phantom{MM}}}
\end{promptbox}
\end{minipage}
\vspace{2pt}

\begin{minipage}[t]{0.48\textwidth}
\begin{promptbox}[Content-masked]
\textcolor{colI}{You will respond with E or D.}\par
\par\smallskip
\textcolor{colM}{Game 1:}\par
\textcolor{colC}{You press <<E>>} \textcolor{colM}{and get some points.}\par
\textcolor{colC}{You press <<D>>} \textcolor{colM}{and get some points.}\par
\textcolor{colC}{You press <<D>>} \textcolor{colM}{and get some points.}\par

\textcolor{colC}{You press <<}\textcolor{midgray}{\underline{\phantom{MM}}}
\end{promptbox}
\end{minipage}%
\hfill
\begin{minipage}[t]{0.24\textwidth}
\begin{promptbox}[History-only]
\textcolor{colI}{You will respond with E or D.}\par
\par\smallskip
\textcolor{colC}{You press <<E>>.}\par
\textcolor{colC}{You press <<D>>.}\par
\textcolor{colC}{You press <<D>>.}\par

\textcolor{colC}{You press <<}\textcolor{midgray}{\underline{\phantom{MM}}}
\end{promptbox}
\end{minipage}%
\hfill
\begin{minipage}[t]{0.2\textwidth}
\begin{promptbox}[Choice-only]
\textcolor{colC}{<<E>>}\par
\textcolor{colC}{<<D>>}\par
\textcolor{colC}{<<D>>}\par

\textcolor{colC}{<<}\textcolor{midgray}{\underline{\phantom{MM}}}
\end{promptbox}
\end{minipage}
\vspace{2pt}
\begin{center}
\sffamily\fontsize{6.5}{8}\selectfont\color{darkgray}
\textcolor{colI}{\rule{5pt}{5pt}}\;\,Instruction / $I_\mathrm{min}$\quad
\textcolor{colS}{\rule{5pt}{5pt}}\;\,Stimuli\quad
\textcolor{colF}{\rule{5pt}{5pt}}\;\,Feedback\quad
\textcolor{colC}{\rule{5pt}{5pt}}\;\,Choice\quad
\textcolor{colM}{\rule{5pt}{5pt}}\;\,Masked\quad
\textcolor{midgray}{\underline{\phantom{MM}}}\;\,Model prediction
\end{center}
\vspace{8pt}
\includegraphics[width=\textwidth]{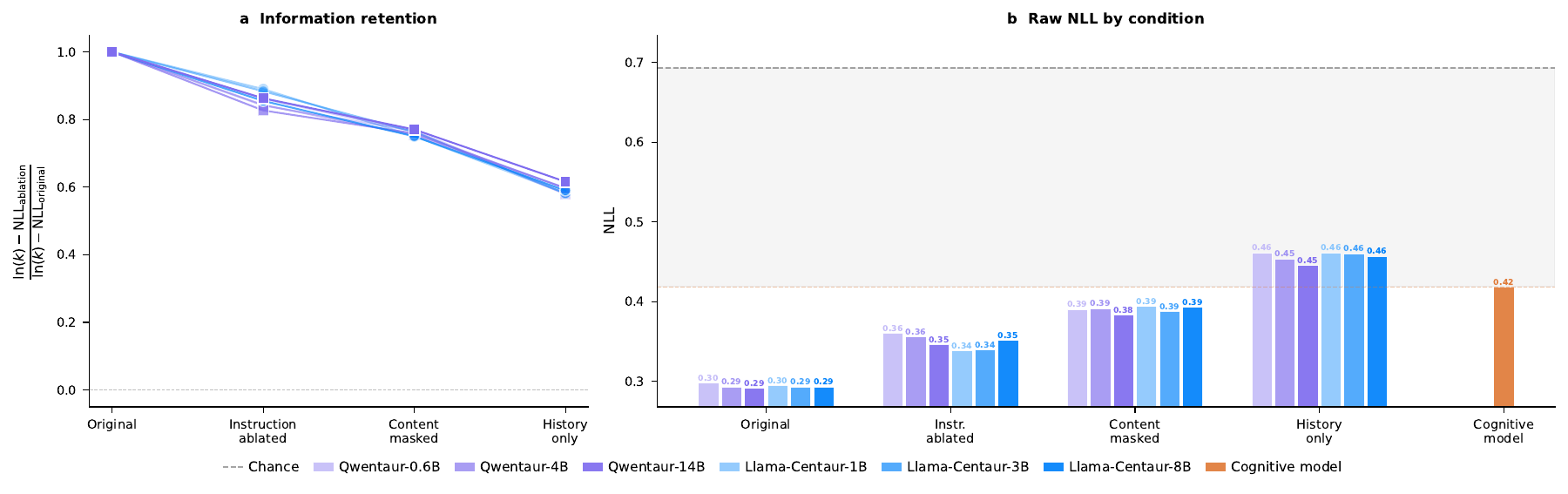}
\vspace{-4pt}
\caption{\textbf{Multi-armed bandits: Two-armed bandit} \citep{gershman2018deconstructing}.}
\label{fig:ablation_gershman2018deconstructing}
\end{figure}

\begin{figure}[ht!]
\centering
\begin{minipage}[t]{0.48\textwidth}
\begin{promptbox}[Original]
\textcolor{colI}{You will play multiple rounds of a gambling game. In each round, you will be presented with 6 different gambles labeled: C, B, W, E, Y, and D. You will have to choose one of the gambles and receive a payoff for doing so. The payoff you receive depends on both the gamble you choose and also the color of a ball we pull out of a jar with 100 colored balls. There are a different number of balls of each color on every round. The colors with more balls are more likely to be chosen. Before making your choice, you may check how much different gambles are worth for different ball colors. Each time you check a gamble will cost you 2 points. To choose or check a gamble, first press the corresponding key, followed by typing "stop" (for choosing) or the ball color you would like to check. A new round begins. There are 55 turquoise balls, 15 beige balls, 11 teal balls, and 19 silver balls.}\par\smallskip
\textcolor{colC}{You press <<C>> and then type <<turquoise>>} \textcolor{colF}{. The payoff for this combination would be 105 points.}\par
\textcolor{colC}{You press <<C>> and then type <<silver>>} \textcolor{colF}{. The payoff for this combination would be 140 points.}\par
\textcolor{colC}{You press <<C>> and then type <<stop>>} \textcolor{colF}{. A turquoise ball is chosen, and you earn 105 points. A new round begins. There are 27 turquoise balls, 11 beige balls, 28 teal balls, and 33 silver balls.}\par

\textcolor{colC}{You press <<}\textcolor{midgray}{\underline{\phantom{MM}}}
\end{promptbox}
\end{minipage}%
\hfill
\begin{minipage}[t]{0.48\textwidth}
\begin{promptbox}[Instruction-ablated]
\textcolor{colC}{You press <<C>> and then type <<turquoise>>} \textcolor{colF}{. The payoff for this combination would be 105 points.}\par
\textcolor{colC}{You press <<C>> and then type <<silver>>} \textcolor{colF}{. The payoff for this combination would be 140 points.}\par
\textcolor{colC}{You press <<C>> and then type <<stop>>} \textcolor{colF}{. A turquoise ball is chosen, and you earn 105 points. A new round begins. There are 27 turquoise balls, 11 beige balls, 28 teal balls, and 33 silver balls.}\par

\textcolor{colC}{You press <<}\textcolor{midgray}{\underline{\phantom{MM}}}
\end{promptbox}
\end{minipage}
\vspace{2pt}

\begin{minipage}[t]{0.48\textwidth}
\begin{promptbox}[Content-masked]
\textcolor{colI}{You will respond with C, turquoise, silver, stop, etc.}\par
\par\smallskip
\textcolor{colC}{You press <<C>> and then type <<turquoise>>} \textcolor{colM}{. The payoff would be some points.}\par
\textcolor{colC}{You press <<C>> and then type <<silver>>} \textcolor{colM}{. The payoff would be some points.}\par
\textcolor{colC}{You press <<C>> and then type <<stop>>} \textcolor{colM}{. A ball is chosen, and you earn some points. A new round begins. There are some colored balls.}\par

\textcolor{colC}{You press <<}\textcolor{midgray}{\underline{\phantom{MM}}}
\end{promptbox}
\end{minipage}%
\hfill
\begin{minipage}[t]{0.24\textwidth}
\begin{promptbox}[History-only]
\textcolor{colI}{You will respond with C, turquoise, silver, stop, etc.}\par
\par\smallskip
\textcolor{colC}{You press <<C>> and then type <<turquoise>>.}\par
\textcolor{colC}{You press <<C>> and then type <<silver>>.}\par
\textcolor{colC}{You press <<C>> and then type <<stop>>.}\par

\textcolor{colC}{You press <<}\textcolor{midgray}{\underline{\phantom{MM}}}
\end{promptbox}
\end{minipage}%
\hfill
\begin{minipage}[t]{0.2\textwidth}
\begin{promptbox}[Choice-only]
\textcolor{colC}{<<C>>}\par
\textcolor{colC}{<<turquoise>>}\par
\textcolor{colC}{<<C>>}\par

\textcolor{colC}{<<}\textcolor{midgray}{\underline{\phantom{MM}}}
\end{promptbox}
\end{minipage}
\vspace{2pt}
\begin{center}
\sffamily\fontsize{6.5}{8}\selectfont\color{darkgray}
\textcolor{colI}{\rule{5pt}{5pt}}\;\,Instruction / $I_\mathrm{min}$\quad
\textcolor{colS}{\rule{5pt}{5pt}}\;\,Stimuli\quad
\textcolor{colF}{\rule{5pt}{5pt}}\;\,Feedback\quad
\textcolor{colC}{\rule{5pt}{5pt}}\;\,Choice\quad
\textcolor{colM}{\rule{5pt}{5pt}}\;\,Masked\quad
\textcolor{midgray}{\underline{\phantom{MM}}}\;\,Model prediction
\end{center}
\vspace{8pt}
\includegraphics[width=\textwidth]{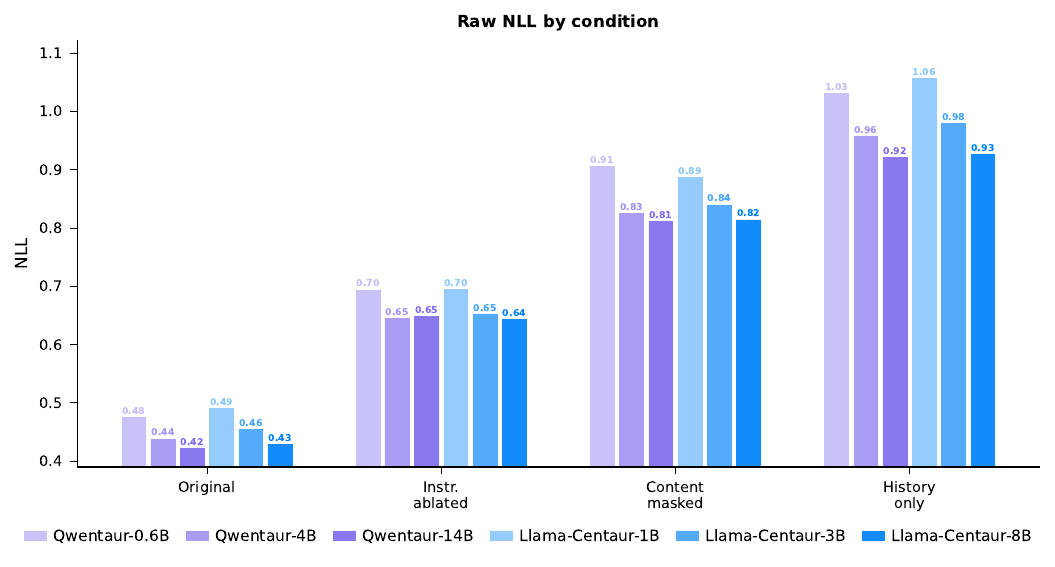}
\vspace{-4pt}
\caption{\textbf{Decision-making: Risky choice} \citep{krueger2024identifying}. No retention ratio or chance baseline is shown because trials span multiple stages with varying numbers of actions, so no single $k$ defines a uniform-guessing floor.
}
\label{fig:ablation_krueger2022identifying}
\end{figure}

\begin{figure}[ht!]
\centering
\begin{minipage}[t]{0.48\textwidth}
\begin{promptbox}[Original]
\textcolor{colI}{You will encounter a series of gambling problems where you have to select between two options. You can select an option by pressing the corresponding key. For some problems, you are told the points you received and missed out on after each selection, while for others this information is suppressed. In cases where the probabilities are unknown, they sum up to one and remain constant within a problem.}\par\smallskip
\textcolor{colS}{Option E delivers either 75.0 points with 10.0\% chance, or 17.0 points with 90.0\% chance.}\par
\textcolor{colS}{Option S delivers either 4.0 points with unknown chance, or 59.0 points with unknown chance.}\par
\textcolor{colC}{You press <<S>>} \textcolor{colF}{. You receive 4.0 points by selecting this option. You would have received 17.0 points had you chosen the other option.}\par
\textcolor{colC}{You press <<S>>} \textcolor{colF}{. You receive 4.0 points by selecting this option. You would have received 17.0 points had you chosen the other option.}\par
\textcolor{colC}{You press <<E>>} \textcolor{colF}{. You receive 17.0 points by selecting this option. You would have received 4.0 points had you chosen the other option.}\par

\textcolor{colC}{You press <<}\textcolor{midgray}{\underline{\phantom{MM}}}
\end{promptbox}
\end{minipage}%
\hfill
\begin{minipage}[t]{0.48\textwidth}
\begin{promptbox}[Instruction-ablated]
\textcolor{colS}{Option E delivers either 75.0 points with 10.0\% chance, or 17.0 points with 90.0\% chance.}\par
\textcolor{colS}{Option S delivers either 4.0 points with unknown chance, or 59.0 points with unknown chance.}\par
\textcolor{colC}{You press <<S>>} \textcolor{colF}{. You receive 4.0 points by selecting this option. You would have received 17.0 points had you chosen the other option.}\par
\textcolor{colC}{You press <<S>>} \textcolor{colF}{. You receive 4.0 points by selecting this option. You would have received 17.0 points had you chosen the other option.}\par
\textcolor{colC}{You press <<E>>} \textcolor{colF}{. You receive 17.0 points by selecting this option. You would have received 4.0 points had you chosen the other option.}\par

\textcolor{colC}{You press <<}\textcolor{midgray}{\underline{\phantom{MM}}}
\end{promptbox}
\end{minipage}
\vspace{2pt}

\begin{minipage}[t]{0.48\textwidth}
\begin{promptbox}[Content-masked]
\textcolor{colI}{You will respond with S or E.}\par
\par\smallskip
\textcolor{colM}{Option E has some outcomes.}\par
\textcolor{colM}{Option S has some outcomes.}\par
\textcolor{colC}{You press <<S>>} \textcolor{colM}{. You receive some points by selecting this option. You would have received some points had you chosen a different option.}\par
\textcolor{colC}{You press <<S>>} \textcolor{colM}{. You receive some points by selecting this option. You would have received some points had you chosen a different option.}\par
\textcolor{colC}{You press <<E>>} \textcolor{colM}{. You receive some points by selecting this option. You would have received some points had you chosen a different option.}\par

\textcolor{colC}{You press <<}\textcolor{midgray}{\underline{\phantom{MM}}}
\end{promptbox}
\end{minipage}%
\hfill
\begin{minipage}[t]{0.24\textwidth}
\begin{promptbox}[History-only]
\textcolor{colI}{You will respond with S or E.}\par
\par\smallskip
\textcolor{colC}{You press <<S>>.}\par
\textcolor{colC}{You press <<S>>.}\par
\textcolor{colC}{You press <<E>>.}\par

\textcolor{colC}{You press <<}\textcolor{midgray}{\underline{\phantom{MM}}}
\end{promptbox}
\end{minipage}%
\hfill
\begin{minipage}[t]{0.2\textwidth}
\begin{promptbox}[Choice-only]
\textcolor{colC}{<<S>>}\par
\textcolor{colC}{<<S>>}\par
\textcolor{colC}{<<E>>}\par

\textcolor{colC}{<<}\textcolor{midgray}{\underline{\phantom{MM}}}
\end{promptbox}
\end{minipage}
\vspace{2pt}
\begin{center}
\sffamily\fontsize{6.5}{8}\selectfont\color{darkgray}
\textcolor{colI}{\rule{5pt}{5pt}}\;\,Instruction / $I_\mathrm{min}$\quad
\textcolor{colS}{\rule{5pt}{5pt}}\;\,Stimuli\quad
\textcolor{colF}{\rule{5pt}{5pt}}\;\,Feedback\quad
\textcolor{colC}{\rule{5pt}{5pt}}\;\,Choice\quad
\textcolor{colM}{\rule{5pt}{5pt}}\;\,Masked\quad
\textcolor{midgray}{\underline{\phantom{MM}}}\;\,Model prediction
\end{center}
\vspace{8pt}
\includegraphics[width=\textwidth]{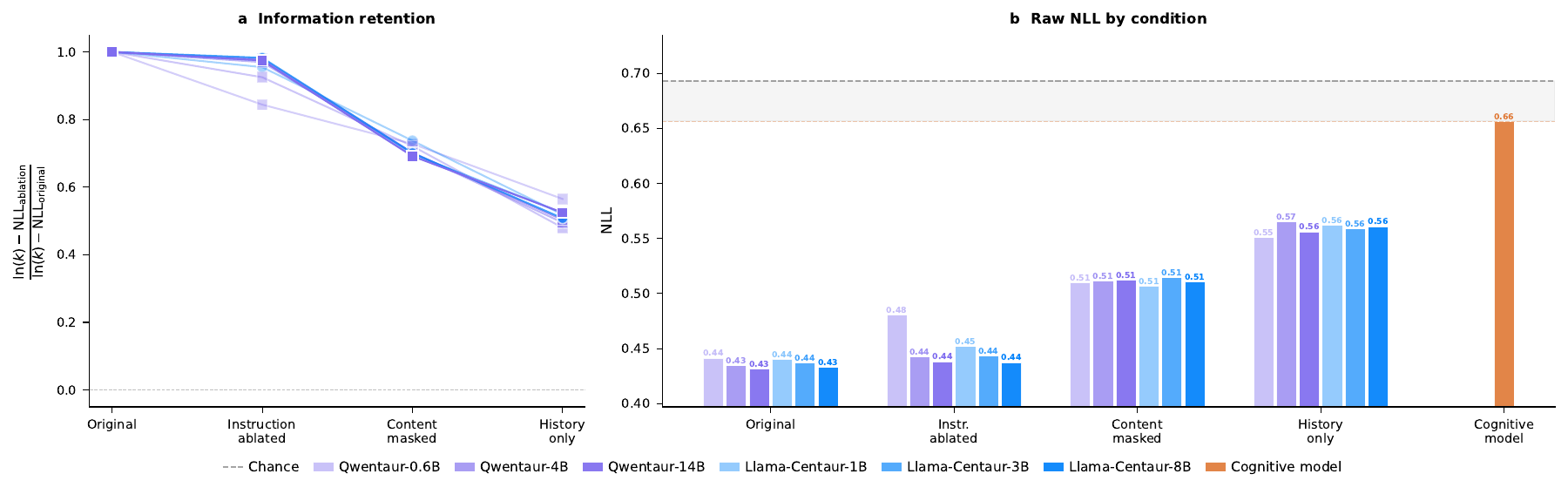}
\vspace{-4pt}
\caption{\textbf{Decision-making: choices13k} \citep{peterson2021using}.}
\label{fig:ablation_peterson2021using}
\end{figure}

\begin{figure}[ht!]
\centering
\begin{minipage}[t]{0.48\textwidth}
\begin{promptbox}[Original]
\textcolor{colI}{You will encounter a series of gambling problems where you have to select between two options. You can select an option by pressing the corresponding key. You will encounter each problem 25 times. In the first five encounters, you will not receive feedback. In the remaining 20 encounters, you will receive feedback about the outcomes of both options. In cases where the probabilities are stated to be unknown, they sum up to one and remain constant within a problem.}\par\smallskip
\textcolor{colS}{Option Z delivers 8 points with 100.0\% chance.}\par
\textcolor{colS}{Option J delivers -28 points with 1.0\% chance, 3 points with 99.0\% chance.}\par
\textcolor{colC}{You press <<Z>>} \textcolor{colF}{.}\par
\textcolor{colC}{You press <<Z>>} \textcolor{colF}{.}\par
\textcolor{colC}{You press <<Z>>} \textcolor{colF}{.}\par

\textcolor{colC}{You press <<}\textcolor{midgray}{\underline{\phantom{MM}}}
\end{promptbox}
\end{minipage}%
\hfill
\begin{minipage}[t]{0.48\textwidth}
\begin{promptbox}[Instruction-ablated]
\textcolor{colS}{Option Z delivers 8 points with 100.0\% chance.}\par
\textcolor{colS}{Option J delivers -28 points with 1.0\% chance, 3 points with 99.0\% chance.}\par
\textcolor{colC}{You press <<Z>>} \textcolor{colF}{.}\par
\textcolor{colC}{You press <<Z>>} \textcolor{colF}{.}\par
\textcolor{colC}{You press <<Z>>} \textcolor{colF}{.}\par

\textcolor{colC}{You press <<}\textcolor{midgray}{\underline{\phantom{MM}}}
\end{promptbox}
\end{minipage}
\vspace{2pt}

\begin{minipage}[t]{0.48\textwidth}
\begin{promptbox}[Content-masked]
\textcolor{colI}{You will respond with Z or J.}\par
\par\smallskip
\textcolor{colM}{Option Z has some outcomes.}\par
\textcolor{colM}{Option J has some outcomes.}\par
\textcolor{colC}{You press <<Z>>} \textcolor{colM}{.}\par
\textcolor{colC}{You press <<Z>>} \textcolor{colM}{.}\par
\textcolor{colC}{You press <<Z>>} \textcolor{colM}{.}\par

\textcolor{colC}{You press <<}\textcolor{midgray}{\underline{\phantom{MM}}}
\end{promptbox}
\end{minipage}%
\hfill
\begin{minipage}[t]{0.24\textwidth}
\begin{promptbox}[History-only]
\textcolor{colI}{You will respond with Z or J.}\par
\par\smallskip
\textcolor{colC}{You press <<Z>>.}\par
\textcolor{colC}{You press <<Z>>.}\par
\textcolor{colC}{You press <<Z>>.}\par

\textcolor{colC}{You press <<}\textcolor{midgray}{\underline{\phantom{MM}}}
\end{promptbox}
\end{minipage}%
\hfill
\begin{minipage}[t]{0.2\textwidth}
\begin{promptbox}[Choice-only]
\textcolor{colC}{<<Z>>}\par
\textcolor{colC}{<<Z>>}\par
\textcolor{colC}{<<Z>>}\par

\textcolor{colC}{<<}\textcolor{midgray}{\underline{\phantom{MM}}}
\end{promptbox}
\end{minipage}
\vspace{2pt}
\begin{center}
\sffamily\fontsize{6.5}{8}\selectfont\color{darkgray}
\textcolor{colI}{\rule{5pt}{5pt}}\;\,Instruction / $I_\mathrm{min}$\quad
\textcolor{colS}{\rule{5pt}{5pt}}\;\,Stimuli\quad
\textcolor{colF}{\rule{5pt}{5pt}}\;\,Feedback\quad
\textcolor{colC}{\rule{5pt}{5pt}}\;\,Choice\quad
\textcolor{colM}{\rule{5pt}{5pt}}\;\,Masked\quad
\textcolor{midgray}{\underline{\phantom{MM}}}\;\,Model prediction
\end{center}
\vspace{8pt}
\includegraphics[width=\textwidth]{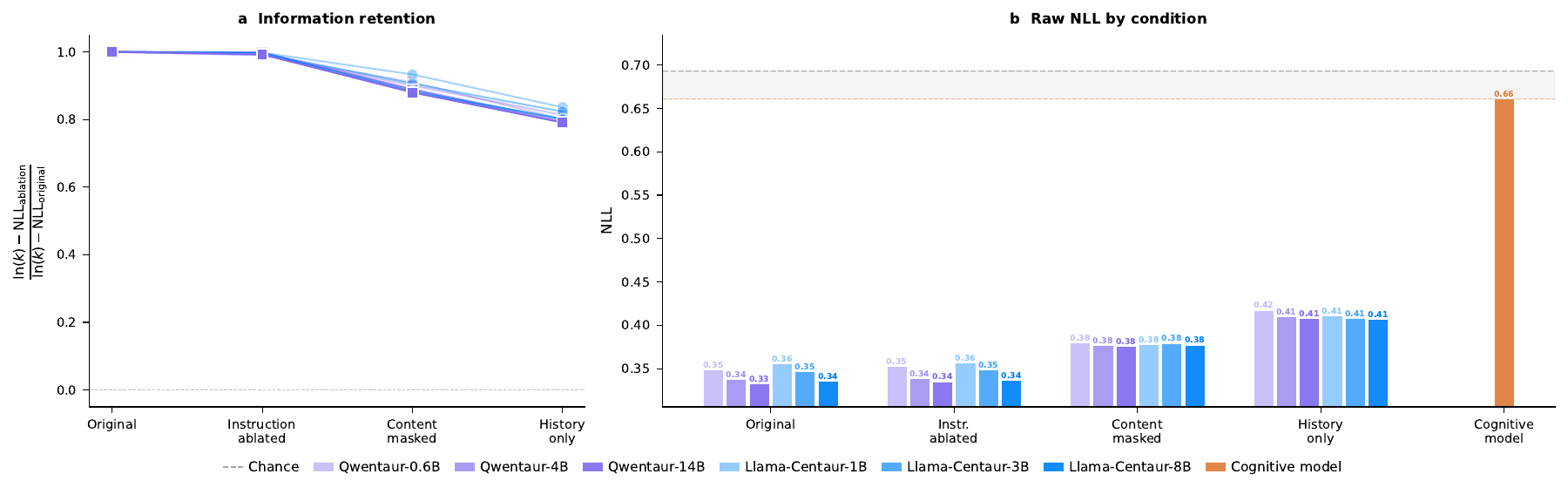}
\vspace{-4pt}
\caption{\textbf{Decision-making: CPC18} \citep{plonsky2018when}.}
\label{fig:ablation_plonsky2018when}
\end{figure}

\begin{figure}[ht!]
\centering
\begin{minipage}[t]{0.48\textwidth}
\begin{promptbox}[Original]
\textcolor{colI}{You are participating in multiple games involving two slot machines, labeled M and P. The two slot machines are different across different games. Each time you choose a slot machine, you get some points. You choose a slot machine by pressing the corresponding key. Each slot machine tends to pay out about the same amount of points on average. Your goal is to choose the slot machines that will give you the most points across the experiment. The first 4 trials in each game are instructed trials where you will be told which slot machine to choose. After these instructed trials, you will have the freedom to choose for either 1 or 6 trials.}\par\smallskip
\textcolor{colI}{Game 1.}\par
\textcolor{colI}{There are 10 trials in this game.}\par
\textcolor{colI}{You are instructed to press P} \textcolor{colF}{and get 52 points.}\par
\textcolor{colI}{You are instructed to press M} \textcolor{colF}{and get 52 points.}\par
\textcolor{colI}{You are instructed to press M} \textcolor{colF}{and get 68 points.}\par
\textcolor{colI}{You are instructed to press M} \textcolor{colF}{and get 51 points.}\par
\textcolor{colC}{You press <<P>>} \textcolor{colF}{and get 34 points.}\par
\textcolor{colC}{You press <<M>>} \textcolor{colF}{and get 51 points.}\par
\textcolor{colC}{You press <<P>>} \textcolor{colF}{and get 49 points.}\par

\textcolor{colC}{You press <<}\textcolor{midgray}{\underline{\phantom{MM}}}
\end{promptbox}
\end{minipage}%
\hfill
\begin{minipage}[t]{0.48\textwidth}
\begin{promptbox}[Instruction-ablated]
\textcolor{colI}{Game 1.}\par
\textcolor{colI}{You are instructed to press P} \textcolor{colF}{and get 52 points.}\par
\textcolor{colI}{You are instructed to press M} \textcolor{colF}{and get 52 points.}\par
\textcolor{colI}{You are instructed to press M} \textcolor{colF}{and get 68 points.}\par
\textcolor{colI}{You are instructed to press M} \textcolor{colF}{and get 51 points.}\par
\textcolor{colC}{You press <<P>>} \textcolor{colF}{and get 34 points.}\par
\textcolor{colC}{You press <<M>>} \textcolor{colF}{and get 51 points.}\par
\textcolor{colC}{You press <<P>>} \textcolor{colF}{and get 49 points.}\par

\textcolor{colC}{You press <<}\textcolor{midgray}{\underline{\phantom{MM}}}
\end{promptbox}
\end{minipage}
\vspace{2pt}

\begin{minipage}[t]{0.48\textwidth}
\begin{promptbox}[Content-masked]
\textcolor{colI}{You will respond with P or M.}\par
\textcolor{colM}{Game 1.}\par
\textcolor{colM}{There are some trials in this game.}\par
\par\smallskip
\textcolor{colM}{You are instructed to press P and get some points.}\par
\textcolor{colM}{You are instructed to press M and get some points.}\par
\textcolor{colM}{You are instructed to press M and get some points.}\par
\textcolor{colM}{You are instructed to press M and get some points.}\par
\textcolor{colC}{You press <<P>>} \textcolor{colM}{and get some points.}\par
\textcolor{colC}{You press <<M>>} \textcolor{colM}{and get some points.}\par
\textcolor{colC}{You press <<P>>} \textcolor{colM}{and get some points.}\par

\textcolor{colC}{You press <<}\textcolor{midgray}{\underline{\phantom{MM}}}
\end{promptbox}
\end{minipage}%
\hfill
\begin{minipage}[t]{0.24\textwidth}
\begin{promptbox}[History-only]
\textcolor{colI}{You will respond with P or M.}\par
\par\smallskip
\textcolor{colC}{You press <<P>>.}\par
\textcolor{colC}{You press <<M>>.}\par
\textcolor{colC}{You press <<P>>.}\par

\textcolor{colC}{You press <<}\textcolor{midgray}{\underline{\phantom{MM}}}
\end{promptbox}
\end{minipage}%
\hfill
\begin{minipage}[t]{0.2\textwidth}
\begin{promptbox}[Choice-only]
\textcolor{colC}{<<P>>}\par
\textcolor{colC}{<<M>>}\par
\textcolor{colC}{<<P>>}\par

\textcolor{colC}{<<}\textcolor{midgray}{\underline{\phantom{MM}}}
\end{promptbox}
\end{minipage}
\vspace{2pt}
\begin{center}
\sffamily\fontsize{6.5}{8}\selectfont\color{darkgray}
\textcolor{colI}{\rule{5pt}{5pt}}\;\,Instruction / $I_\mathrm{min}$\quad
\textcolor{colS}{\rule{5pt}{5pt}}\;\,Stimuli\quad
\textcolor{colF}{\rule{5pt}{5pt}}\;\,Feedback\quad
\textcolor{colC}{\rule{5pt}{5pt}}\;\,Choice\quad
\textcolor{colM}{\rule{5pt}{5pt}}\;\,Masked\quad
\textcolor{midgray}{\underline{\phantom{MM}}}\;\,Model prediction
\end{center}
\vspace{8pt}
\includegraphics[width=\textwidth]{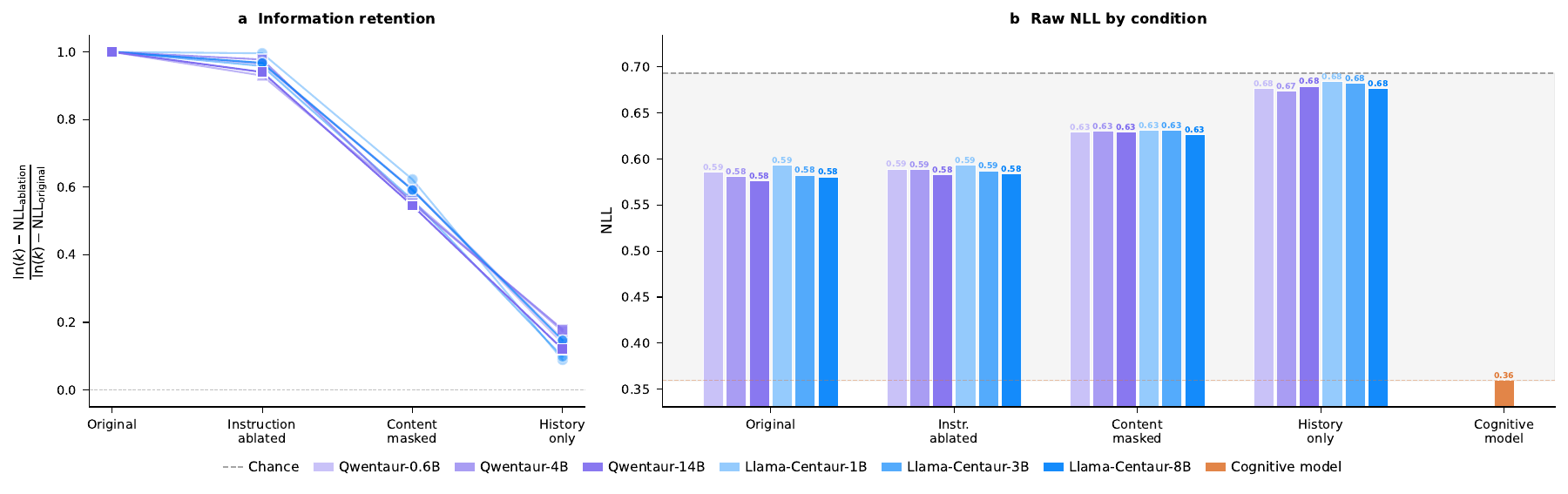}
\vspace{-4pt}
\caption{\textbf{Multi-armed bandits: Horizon task} \citep{sadeghiyeh2020temporal}.}
\label{fig:ablation_sadeghiyeh2020temporal}
\end{figure}

\begin{figure}[ht!]
\centering
\begin{minipage}[t]{0.48\textwidth}
\begin{promptbox}[Original]
\textcolor{colI}{You will be playing a game for 30 rounds. Each round contains 10 trials. In each trial, you have to select one option that will generate a reward between 0 and 50 points. You can choose between options 1, 2, 3, 4, 5, 6, 7 and 8 by pressing the corresponding key. After each round the options reset and each option can produce different rewards in the following round. Your goal is to maximize your reward.}\par\smallskip
\textcolor{colI}{You are playing round 1:}\par
\textcolor{colC}{You press <<3>>} \textcolor{colF}{and get 36.0665516504 points.}\par
\textcolor{colC}{You press <<2>>} \textcolor{colF}{and get 26.8954211779 points.}\par
\textcolor{colC}{You press <<1>>} \textcolor{colF}{and get 40.128583453 points.}\par

\textcolor{colC}{You press <<}\textcolor{midgray}{\underline{\phantom{MM}}}
\end{promptbox}
\end{minipage}%
\hfill
\begin{minipage}[t]{0.48\textwidth}
\begin{promptbox}[Instruction-ablated]
\textcolor{colC}{You press <<3>>} \textcolor{colF}{and get 36.0665516504 points.}\par
\textcolor{colC}{You press <<2>>} \textcolor{colF}{and get 26.8954211779 points.}\par
\textcolor{colC}{You press <<1>>} \textcolor{colF}{and get 40.128583453 points.}\par

\textcolor{colC}{You press <<}\textcolor{midgray}{\underline{\phantom{MM}}}
\end{promptbox}
\end{minipage}
\vspace{2pt}

\begin{minipage}[t]{0.48\textwidth}
\begin{promptbox}[Content-masked]
\textcolor{colI}{You will respond with 3, 2, 1, 4, etc.}\par
\par\smallskip
\textcolor{colC}{You press <<3>>} \textcolor{colM}{and get some points.}\par
\textcolor{colC}{You press <<2>>} \textcolor{colM}{and get some points.}\par
\textcolor{colC}{You press <<1>>} \textcolor{colM}{and get some points.}\par

\textcolor{colC}{You press <<}\textcolor{midgray}{\underline{\phantom{MM}}}
\end{promptbox}
\end{minipage}%
\hfill
\begin{minipage}[t]{0.24\textwidth}
\begin{promptbox}[History-only]
\textcolor{colI}{You will respond with 3, 2, 1, 4, etc.}\par
\par\smallskip
\textcolor{colC}{You press <<3>>.}\par
\textcolor{colC}{You press <<2>>.}\par
\textcolor{colC}{You press <<1>>.}\par

\textcolor{colC}{You press <<}\textcolor{midgray}{\underline{\phantom{MM}}}
\end{promptbox}
\end{minipage}%
\hfill
\begin{minipage}[t]{0.2\textwidth}
\begin{promptbox}[Choice-only]
\textcolor{colC}{<<3>>}\par
\textcolor{colC}{<<2>>}\par
\textcolor{colC}{<<1>>}\par

\textcolor{colC}{<<}\textcolor{midgray}{\underline{\phantom{MM}}}
\end{promptbox}
\end{minipage}
\vspace{2pt}
\begin{center}
\sffamily\fontsize{6.5}{8}\selectfont\color{darkgray}
\textcolor{colI}{\rule{5pt}{5pt}}\;\,Instruction / $I_\mathrm{min}$\quad
\textcolor{colS}{\rule{5pt}{5pt}}\;\,Stimuli\quad
\textcolor{colF}{\rule{5pt}{5pt}}\;\,Feedback\quad
\textcolor{colC}{\rule{5pt}{5pt}}\;\,Choice\quad
\textcolor{colM}{\rule{5pt}{5pt}}\;\,Masked\quad
\textcolor{midgray}{\underline{\phantom{MM}}}\;\,Model prediction
\end{center}
\vspace{8pt}
\includegraphics[width=\textwidth]{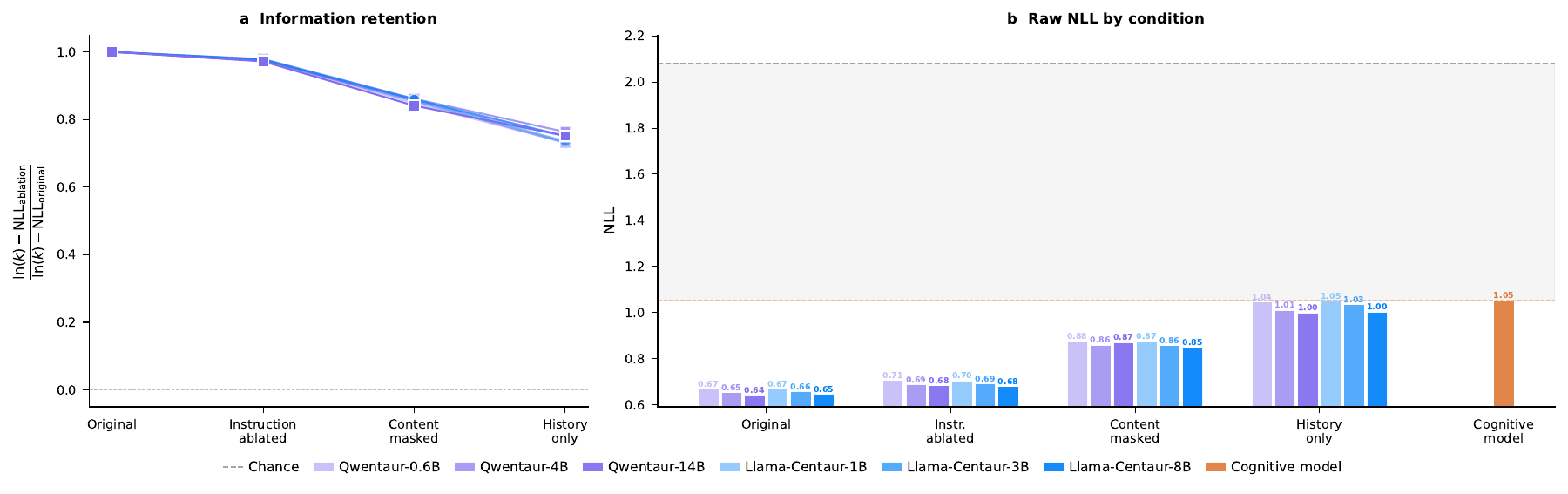}
\vspace{-4pt}
\caption{\textbf{Multi-armed bandits: Structured bandit} \citep{schulz2020finding}.}
\label{fig:ablation_schulz2020finding}
\end{figure}

\begin{figure}[ht!]
\centering
\begin{minipage}[t]{0.48\textwidth}
\begin{promptbox}[Original]
\textcolor{colI}{You are participating in multiple games involving two slot machines, labeled G and M. The two slot machines are different across different games. Each time you choose a slot machine, you get some points. You choose a slot machine by pressing the corresponding key. Each slot machine tends to pay out about the same amount of points on average. Your goal is to choose the slot machines that will give you the most points across the experiment. The first 4 trials in each game are instructed trials where you will be told which slot machine to choose. After these instructed trials, you will have the freedom to choose for either 1 or 6 trials.}\par\smallskip
\textcolor{colI}{Game 1.}\par
\textcolor{colI}{There are 5 trials in this game.}\par
\textcolor{colI}{You are instructed to press M} \textcolor{colF}{and get 41 points.}\par
\textcolor{colI}{You are instructed to press M} \textcolor{colF}{and get 38 points.}\par
\textcolor{colI}{You are instructed to press M} \textcolor{colF}{and get 42 points.}\par
\textcolor{colI}{You are instructed to press G} \textcolor{colF}{and get 12 points.}\par
\textcolor{colC}{You press <<M>>} \textcolor{colF}{and get 33 points.}\par
\par\smallskip
\textcolor{colI}{Game 2.}\par
\textcolor{colI}{There are 5 trials in this game.}\par
\textcolor{colI}{You are instructed to press M} \textcolor{colF}{and get 29 points.}\par
\textcolor{colI}{You are instructed to press G} \textcolor{colF}{and get 79 points.}\par
\textcolor{colI}{You are instructed to press G} \textcolor{colF}{and get 65 points.}\par
\textcolor{colI}{You are instructed to press M} \textcolor{colF}{and get 57 points.}\par
\textcolor{colC}{You press <<}\textcolor{midgray}{\underline{\phantom{MM}}}
\end{promptbox}
\end{minipage}%
\hfill
\begin{minipage}[t]{0.48\textwidth}
\begin{promptbox}[Instruction-ablated]
\textcolor{colI}{Game 1.}\par
\textcolor{colI}{You are instructed to press M} \textcolor{colF}{and get 41 points.}\par
\textcolor{colI}{You are instructed to press M} \textcolor{colF}{and get 38 points.}\par
\textcolor{colI}{You are instructed to press M} \textcolor{colF}{and get 42 points.}\par
\textcolor{colI}{You are instructed to press G} \textcolor{colF}{and get 12 points.}\par
\textcolor{colC}{You press <<M>>} \textcolor{colF}{and get 33 points.}\par
\par\smallskip
\textcolor{colI}{Game 2.}\par
\textcolor{colI}{You are instructed to press M} \textcolor{colF}{and get 29 points.}\par
\textcolor{colI}{You are instructed to press G} \textcolor{colF}{and get 79 points.}\par
\textcolor{colI}{You are instructed to press G} \textcolor{colF}{and get 65 points.}\par
\textcolor{colI}{You are instructed to press M} \textcolor{colF}{and get 57 points.}\par
\textcolor{colC}{You press <<}\textcolor{midgray}{\underline{\phantom{MM}}}
\end{promptbox}
\end{minipage}
\vspace{2pt}

\begin{minipage}[t]{0.48\textwidth}
\begin{promptbox}[Content-masked]
\textcolor{colI}{You will respond with M or G.}\par
\textcolor{colM}{Game 1.}\par
\textcolor{colM}{There are some trials in this game.}\par
\par\smallskip
\textcolor{colM}{You are instructed to press M and get some points.}\par
\textcolor{colM}{You are instructed to press M and get some points.}\par
\textcolor{colM}{You are instructed to press M and get some points.}\par
\textcolor{colM}{You are instructed to press G and get some points.}\par
\textcolor{colC}{You press <<M>>} \textcolor{colM}{and get some points.}\par
\par\smallskip
\textcolor{colM}{Game 2.}\par
\textcolor{colM}{There are some trials in this game.}\par
\textcolor{colM}{You are instructed to press M and get some points.}\par
\textcolor{colM}{You are instructed to press G and get some points.}\par
\textcolor{colM}{You are instructed to press G and get some points.}\par
\textcolor{colM}{You are instructed to press M and get some points.}\par
\textcolor{colC}{You press <<}\textcolor{midgray}{\underline{\phantom{MM}}}
\end{promptbox}
\end{minipage}%
\hfill
\begin{minipage}[t]{0.24\textwidth}
\begin{promptbox}[History-only]
\textcolor{colI}{You will respond with M or G.}\par
\par\smallskip
\textcolor{colC}{You press <<M>>.}\par
\textcolor{colC}{You press <<G>>.}\par
\textcolor{colC}{You press <<G>>.}\par

\textcolor{colC}{You press <<}\textcolor{midgray}{\underline{\phantom{MM}}}
\end{promptbox}
\end{minipage}%
\hfill
\begin{minipage}[t]{0.2\textwidth}
\begin{promptbox}[Choice-only]
\textcolor{colC}{<<M>>}\par
\textcolor{colC}{<<G>>}\par
\textcolor{colC}{<<G>>}\par

\textcolor{colC}{<<}\textcolor{midgray}{\underline{\phantom{MM}}}
\end{promptbox}
\end{minipage}
\vspace{2pt}
\begin{center}
\sffamily\fontsize{6.5}{8}\selectfont\color{darkgray}
\textcolor{colI}{\rule{5pt}{5pt}}\;\,Instruction / $I_\mathrm{min}$\quad
\textcolor{colS}{\rule{5pt}{5pt}}\;\,Stimuli\quad
\textcolor{colF}{\rule{5pt}{5pt}}\;\,Feedback\quad
\textcolor{colC}{\rule{5pt}{5pt}}\;\,Choice\quad
\textcolor{colM}{\rule{5pt}{5pt}}\;\,Masked\quad
\textcolor{midgray}{\underline{\phantom{MM}}}\;\,Model prediction
\end{center}
\vspace{8pt}
\includegraphics[width=\textwidth]{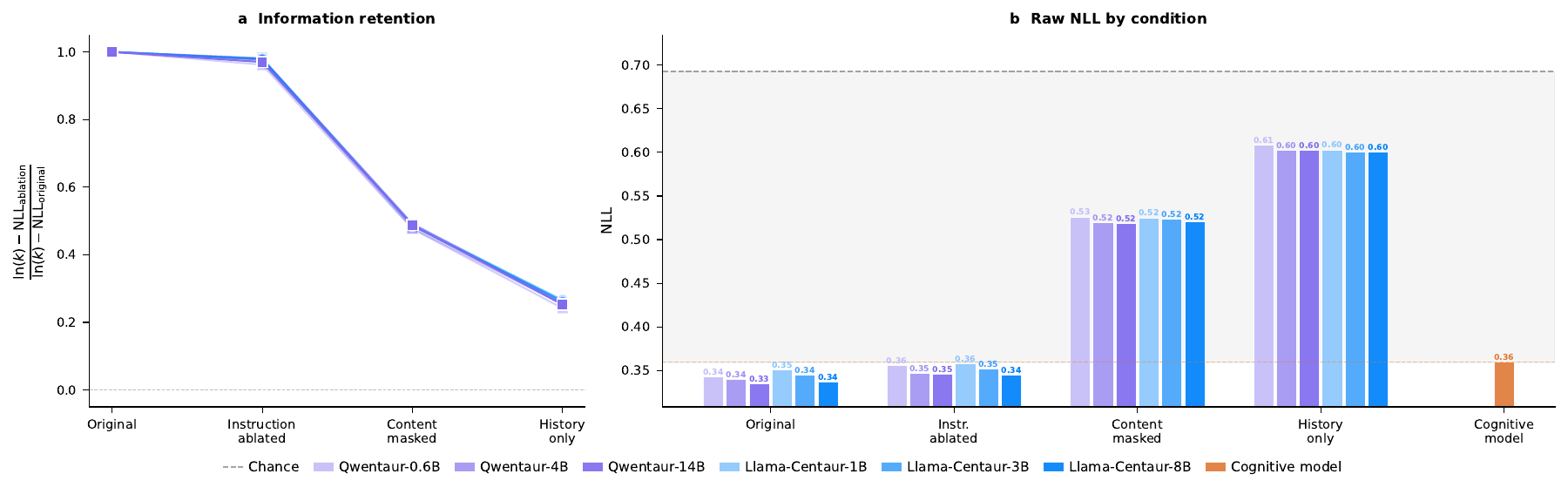}
\vspace{-4pt}
\caption{\textbf{Multi-armed bandits: Horizon task} \citep{somerville2017charting}.}
\label{fig:ablation_somerville2017charting}
\end{figure}

\begin{figure}[ht!]
\centering
\begin{minipage}[t]{0.48\textwidth}
\begin{promptbox}[Original]
\textcolor{colI}{You are participating in multiple games involving two slot machines, labeled R and M. The two slot machines are different across different games. Each time you choose a slot machine, you get some points. You choose a slot machine by pressing the corresponding key. Each slot machine tends to pay out about the same amount of points on average. Your goal is to choose the slot machines that will give you the most points across the experiment. The first 4 trials in each game are instructed trials where you will be told which slot machine to choose. After these instructed trials, you will have the freedom to choose for either 1 or 6 trials.}\par\smallskip
\textcolor{colI}{Game 1.}\par
\textcolor{colI}{There are 10 trials in this game.}\par
\textcolor{colI}{You are instructed to press M} \textcolor{colF}{and get 45 points.}\par
\textcolor{colI}{You are instructed to press R} \textcolor{colF}{and get 25 points.}\par
\textcolor{colI}{You are instructed to press M} \textcolor{colF}{and get 38 points.}\par
\textcolor{colI}{You are instructed to press R} \textcolor{colF}{and get 12 points.}\par
\textcolor{colC}{You press <<M>>} \textcolor{colF}{and get 38 points.}\par
\textcolor{colC}{You press <<M>>} \textcolor{colF}{and get 42 points.}\par
\textcolor{colC}{You press <<M>>} \textcolor{colF}{and get 44 points.}\par

\textcolor{colC}{You press <<}\textcolor{midgray}{\underline{\phantom{MM}}}
\end{promptbox}
\end{minipage}%
\hfill
\begin{minipage}[t]{0.48\textwidth}
\begin{promptbox}[Instruction-ablated]
\textcolor{colI}{Game 1.}\par
\textcolor{colI}{You are instructed to press M} \textcolor{colF}{and get 45 points.}\par
\textcolor{colI}{You are instructed to press R} \textcolor{colF}{and get 25 points.}\par
\textcolor{colI}{You are instructed to press M} \textcolor{colF}{and get 38 points.}\par
\textcolor{colI}{You are instructed to press R} \textcolor{colF}{and get 12 points.}\par
\textcolor{colC}{You press <<M>>} \textcolor{colF}{and get 38 points.}\par
\textcolor{colC}{You press <<M>>} \textcolor{colF}{and get 42 points.}\par
\textcolor{colC}{You press <<M>>} \textcolor{colF}{and get 44 points.}\par

\textcolor{colC}{You press <<}\textcolor{midgray}{\underline{\phantom{MM}}}
\end{promptbox}
\end{minipage}
\vspace{2pt}

\begin{minipage}[t]{0.48\textwidth}
\begin{promptbox}[Content-masked]
\textcolor{colI}{You will respond with M or R.}\par
\textcolor{colM}{Game 1.}\par
\textcolor{colM}{There are some trials in this game.}\par
\par\smallskip
\textcolor{colM}{You are instructed to press M and get some points.}\par
\textcolor{colM}{You are instructed to press R and get some points.}\par
\textcolor{colM}{You are instructed to press M and get some points.}\par
\textcolor{colM}{You are instructed to press R and get some points.}\par
\textcolor{colC}{You press <<M>>} \textcolor{colM}{and get some points.}\par
\textcolor{colC}{You press <<M>>} \textcolor{colM}{and get some points.}\par
\textcolor{colC}{You press <<M>>} \textcolor{colM}{and get some points.}\par

\textcolor{colC}{You press <<}\textcolor{midgray}{\underline{\phantom{MM}}}
\end{promptbox}
\end{minipage}%
\hfill
\begin{minipage}[t]{0.24\textwidth}
\begin{promptbox}[History-only]
\textcolor{colI}{You will respond with M or R.}\par
\par\smallskip
\textcolor{colC}{You press <<M>>.}\par
\textcolor{colC}{You press <<M>>.}\par
\textcolor{colC}{You press <<M>>.}\par

\textcolor{colC}{You press <<}\textcolor{midgray}{\underline{\phantom{MM}}}
\end{promptbox}
\end{minipage}%
\hfill
\begin{minipage}[t]{0.2\textwidth}
\begin{promptbox}[Choice-only]
\textcolor{colC}{<<M>>}\par
\textcolor{colC}{<<M>>}\par
\textcolor{colC}{<<M>>}\par

\textcolor{colC}{<<}\textcolor{midgray}{\underline{\phantom{MM}}}
\end{promptbox}
\end{minipage}
\vspace{2pt}
\begin{center}
\sffamily\fontsize{6.5}{8}\selectfont\color{darkgray}
\textcolor{colI}{\rule{5pt}{5pt}}\;\,Instruction / $I_\mathrm{min}$\quad
\textcolor{colS}{\rule{5pt}{5pt}}\;\,Stimuli\quad
\textcolor{colF}{\rule{5pt}{5pt}}\;\,Feedback\quad
\textcolor{colC}{\rule{5pt}{5pt}}\;\,Choice\quad
\textcolor{colM}{\rule{5pt}{5pt}}\;\,Masked\quad
\textcolor{midgray}{\underline{\phantom{MM}}}\;\,Model prediction
\end{center}
\vspace{8pt}
\includegraphics[width=\textwidth]{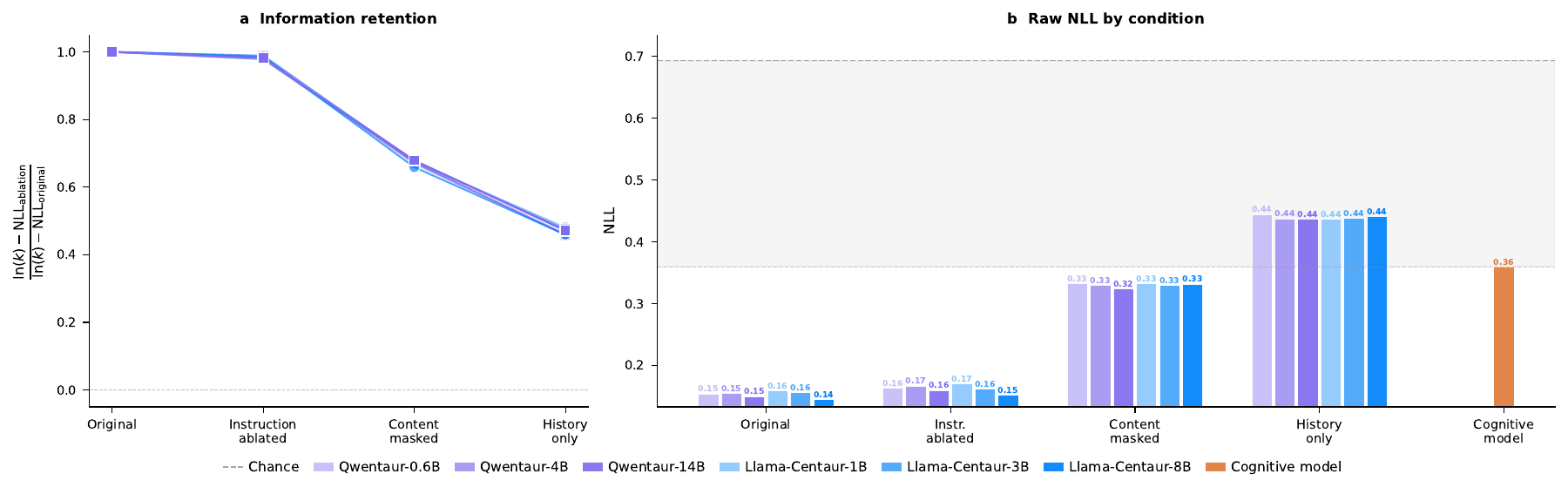}
\vspace{-4pt}
\caption{\textbf{Multi-armed bandits: Horizon task} \citep{waltz2020differential}.}
\label{fig:ablation_waltz2020differential}
\end{figure}

\begin{figure}[ht!]
\centering
\begin{minipage}[t]{0.48\textwidth}
\begin{promptbox}[Original]
\textcolor{colI}{You are participating in multiple games involving two slot machines, labeled S and T. The two slot machines are different across different games. Each time you choose a slot machine, you get some points. You choose a slot machine by pressing the corresponding key. Each slot machine tends to pay out about the same amount of points on average. Your goal is to choose the slot machines that will give you the most points across the experiment. The first 4 trials in each game are instructed trials where you will be told which slot machine to choose. After these instructed trials, you will have the freedom to choose for either 1 or 6 trials.}\par\smallskip
\textcolor{colI}{Game 1.}\par
\textcolor{colI}{There are 10 trials in this game.}\par
\textcolor{colI}{You are instructed to press T} \textcolor{colF}{and get 55 points.}\par
\textcolor{colI}{You are instructed to press T} \textcolor{colF}{and get 53 points.}\par
\textcolor{colI}{You are instructed to press S} \textcolor{colF}{and get 49 points.}\par
\textcolor{colI}{You are instructed to press S} \textcolor{colF}{and get 57 points.}\par
\textcolor{colC}{You press <<T>>} \textcolor{colF}{and get 50 points.}\par
\textcolor{colC}{You press <<S>>} \textcolor{colF}{and get 28 points.}\par
\textcolor{colC}{You press <<T>>} \textcolor{colF}{and get 55 points.}\par

\textcolor{colC}{You press <<}\textcolor{midgray}{\underline{\phantom{MM}}}
\end{promptbox}
\end{minipage}%
\hfill
\begin{minipage}[t]{0.48\textwidth}
\begin{promptbox}[Instruction-ablated]
\textcolor{colI}{Game 1.}\par
\textcolor{colI}{You are instructed to press T} \textcolor{colF}{and get 55 points.}\par
\textcolor{colI}{You are instructed to press T} \textcolor{colF}{and get 53 points.}\par
\textcolor{colI}{You are instructed to press S} \textcolor{colF}{and get 49 points.}\par
\textcolor{colI}{You are instructed to press S} \textcolor{colF}{and get 57 points.}\par
\textcolor{colC}{You press <<T>>} \textcolor{colF}{and get 50 points.}\par
\textcolor{colC}{You press <<S>>} \textcolor{colF}{and get 28 points.}\par
\textcolor{colC}{You press <<T>>} \textcolor{colF}{and get 55 points.}\par

\textcolor{colC}{You press <<}\textcolor{midgray}{\underline{\phantom{MM}}}
\end{promptbox}
\end{minipage}
\vspace{2pt}

\begin{minipage}[t]{0.48\textwidth}
\begin{promptbox}[Content-masked]
\textcolor{colI}{You will respond with T or S.}\par
\textcolor{colM}{Game 1.}\par
\textcolor{colM}{There are some trials in this game.}\par
\par\smallskip
\textcolor{colM}{You are instructed to press T and get some points.}\par
\textcolor{colM}{You are instructed to press T and get some points.}\par
\textcolor{colM}{You are instructed to press S and get some points.}\par
\textcolor{colM}{You are instructed to press S and get some points.}\par
\textcolor{colC}{You press <<T>>} \textcolor{colM}{and get some points.}\par
\textcolor{colC}{You press <<S>>} \textcolor{colM}{and get some points.}\par
\textcolor{colC}{You press <<T>>} \textcolor{colM}{and get some points.}\par

\textcolor{colC}{You press <<}\textcolor{midgray}{\underline{\phantom{MM}}}
\end{promptbox}
\end{minipage}%
\hfill
\begin{minipage}[t]{0.24\textwidth}
\begin{promptbox}[History-only]
\textcolor{colI}{You will respond with T or S.}\par
\par\smallskip
\textcolor{colC}{You press <<T>>.}\par
\textcolor{colC}{You press <<S>>.}\par
\textcolor{colC}{You press <<T>>.}\par

\textcolor{colC}{You press <<}\textcolor{midgray}{\underline{\phantom{MM}}}
\end{promptbox}
\end{minipage}%
\hfill
\begin{minipage}[t]{0.2\textwidth}
\begin{promptbox}[Choice-only]
\textcolor{colC}{<<T>>}\par
\textcolor{colC}{<<S>>}\par
\textcolor{colC}{<<T>>}\par

\textcolor{colC}{<<}\textcolor{midgray}{\underline{\phantom{MM}}}
\end{promptbox}
\end{minipage}
\vspace{2pt}
\begin{center}
\sffamily\fontsize{6.5}{8}\selectfont\color{darkgray}
\textcolor{colI}{\rule{5pt}{5pt}}\;\,Instruction / $I_\mathrm{min}$\quad
\textcolor{colS}{\rule{5pt}{5pt}}\;\,Stimuli\quad
\textcolor{colF}{\rule{5pt}{5pt}}\;\,Feedback\quad
\textcolor{colC}{\rule{5pt}{5pt}}\;\,Choice\quad
\textcolor{colM}{\rule{5pt}{5pt}}\;\,Masked\quad
\textcolor{midgray}{\underline{\phantom{MM}}}\;\,Model prediction
\end{center}
\vspace{8pt}
\includegraphics[width=\textwidth]{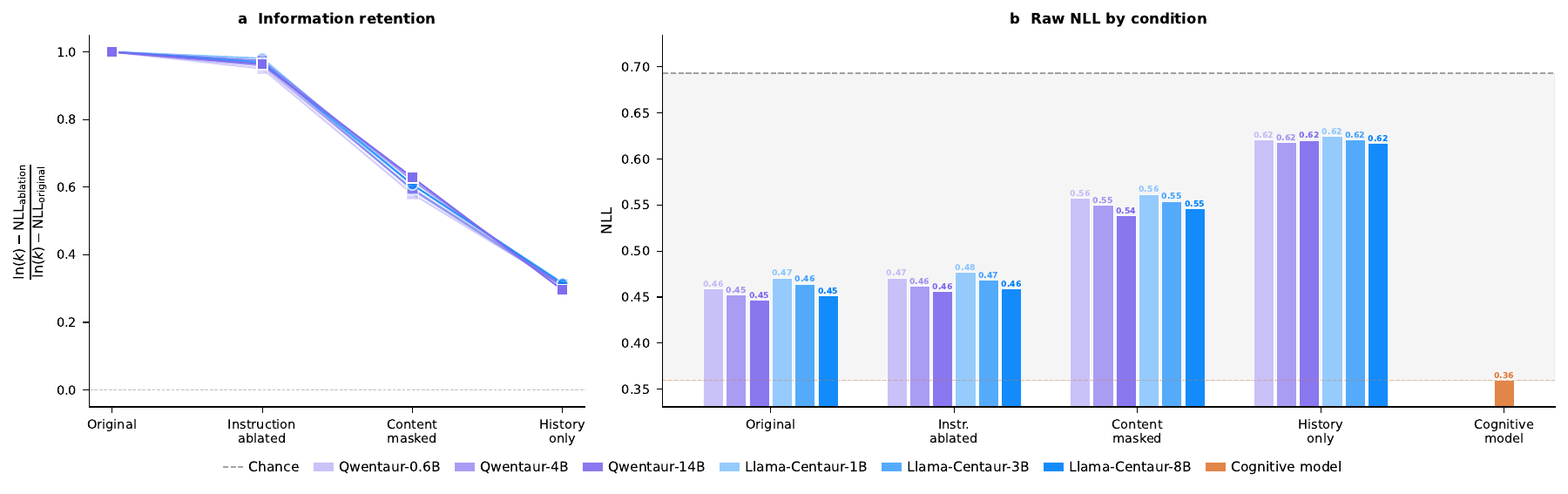}
\vspace{-4pt}
\caption{\textbf{Multi-armed bandits: Horizon task} \citep{wilson2014humans}.}
\label{fig:ablation_wilson2014humans}
\end{figure}

\begin{figure}[ht!]
\centering
\begin{minipage}[t]{0.48\textwidth}
\begin{promptbox}[Original]
\textcolor{colI}{You will predict the probability of electric shocks for two stimuli (0--100\%). An upcoming shock is indicated by a square; no-shock by a circle. The shock probability fluctuates over time.}\par\smallskip
\textcolor{colS}{Stimulus Y and G are shown.} \textcolor{colC}{You predict Y is <<50.0>>\% and G is <<50.0>>} \textcolor{colF}{\%. Square over Y, circle over G. Shock for Y.}\par
\textcolor{colS}{Stimulus Y and G are shown.} \textcolor{colC}{You predict Y is <<69.03>>\% and G is <<50.0>>} \textcolor{colF}{\%. Circle over Y, square over G. Shock for G.}\par

\textcolor{colS}{Stimulus Y and G are shown.} \textcolor{colC}{You predict Y is <<}\textcolor{midgray}{\underline{\phantom{MM}}}
\end{promptbox}
\end{minipage}%
\hfill
\begin{minipage}[t]{0.48\textwidth}
\begin{promptbox}[Instruction-ablated]
\textcolor{colS}{Stimulus Y and G are shown.} \textcolor{colC}{You predict Y is <<50.0>>\% and G is <<50.0>>} \textcolor{colF}{\%. Square over Y, circle over G. Shock for Y.}\par
\textcolor{colS}{Stimulus Y and G are shown.} \textcolor{colC}{You predict Y is <<69.03>>\% and G is <<50.0>>} \textcolor{colF}{\%. Circle over Y, square over G. Shock for G.}\par

\textcolor{colS}{Stimulus Y and G are shown.} \textcolor{colC}{You predict Y is <<}\textcolor{midgray}{\underline{\phantom{MM}}}
\end{promptbox}
\end{minipage}
\vspace{2pt}

\begin{minipage}[t]{0.48\textwidth}
\begin{promptbox}[Content-masked]
\textcolor{colI}{You will respond with 50.0, 69.03, 66.03, 59.23, etc.}\par\smallskip
\textcolor{colM}{Stimulus Y and G are shown.} \textcolor{colC}{You predict Y is <<50.0>>\% and G is <<50.0>>} \textcolor{colM}{\%. An outcome over Y, an outcome over G. An outcome occurs.}\par
\textcolor{colM}{Stimulus Y and G are shown.} \textcolor{colC}{You predict Y is <<69.03>>\% and G is <<50.0>>} \textcolor{colM}{\%. An outcome over Y, an outcome over G. An outcome occurs.}\par

\textcolor{colM}{Stimulus Y and G are shown.} \textcolor{colC}{You predict Y is <<}\textcolor{midgray}{\underline{\phantom{MM}}}
\end{promptbox}
\end{minipage}%
\hfill
\begin{minipage}[t]{0.24\textwidth}
\begin{promptbox}[History-only]
\textcolor{colI}{You will respond with 50.0, 69.03, 66.03, 59.23, etc.}\par\smallskip
\textcolor{colC}{You predict Y is <<50.0>>\% and G is <<50.0>>.}\par
\textcolor{colC}{You predict Y is <<69.03>>\% and G is <<50.0>>.}\par

\textcolor{colC}{You predict Y is <<}\textcolor{midgray}{\underline{\phantom{MM}}}
\end{promptbox}
\end{minipage}%
\hfill
\begin{minipage}[t]{0.2\textwidth}
\begin{promptbox}[Choice-only]
\textcolor{colC}{<<50.0>>}\par
\textcolor{colC}{<<50.0>>}\par
\textcolor{colC}{<<69.03>>}\par
\textcolor{colC}{<<}\textcolor{midgray}{\underline{\phantom{MM}}}
\end{promptbox}
\end{minipage}
\vspace{2pt}
\begin{center}
\sffamily\fontsize{6.5}{8}\selectfont\color{darkgray}
\textcolor{colI}{\rule{5pt}{5pt}}\;\,Instruction / $I_\mathrm{min}$\quad
\textcolor{colS}{\rule{5pt}{5pt}}\;\,Stimuli\quad
\textcolor{colF}{\rule{5pt}{5pt}}\;\,Feedback\quad
\textcolor{colC}{\rule{5pt}{5pt}}\;\,Choice\quad
\textcolor{colM}{\rule{5pt}{5pt}}\;\,Masked\quad
\textcolor{midgray}{\underline{\phantom{MM}}}\;\,Model prediction
\end{center}
\vspace{8pt}
\includegraphics[width=\textwidth]{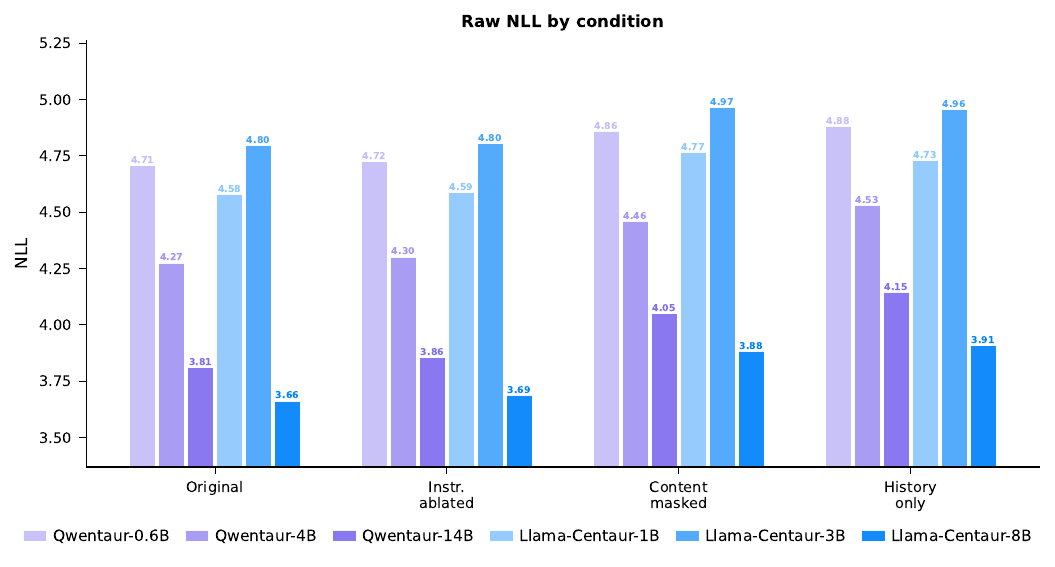}
\vspace{-4pt}
\caption{\textbf{Supervised learning: Aversive learning} \citep{wise2019computational}. No retention ratio or chance baseline is shown because responses are continuous probability estimates, so no single $k$ defines a uniform-guessing floor.}
\label{fig:ablation_wise2019acomputational}
\end{figure}

\begin{figure}[ht!]
\centering
\begin{minipage}[t]{0.48\textwidth}
\begin{promptbox}[Original]
\textcolor{colI}{You will be presented with 16 environments. In each trial, select an option (1--30) by pressing the corresponding key. Options closer together tend to have similar rewards. For each environment, you can make 5 or 10 choices. It is your task to gain as many points as possible.}\par\smallskip
\textcolor{colI}{Environment number 1:}\par
\textcolor{colS}{The value of option 21 is 70.}\par
\textcolor{colC}{You press <<20>>} \textcolor{colF}{and receive 42 points.}\par
\textcolor{colC}{You press <<22>>} \textcolor{colF}{and receive 81 points.}\par
\textcolor{colC}{You press <<23>>} \textcolor{colF}{and receive 79 points.}\par

\textcolor{colC}{You press <<}\textcolor{midgray}{\underline{\phantom{MM}}}
\end{promptbox}
\end{minipage}%
\hfill
\begin{minipage}[t]{0.48\textwidth}
\begin{promptbox}[Instruction-ablated]
\textcolor{colI}{Environment number 1:}\par
\textcolor{colS}{The value of option 21 is 70.}\par
\textcolor{colC}{You press <<20>>} \textcolor{colF}{and receive 42 points.}\par
\textcolor{colC}{You press <<22>>} \textcolor{colF}{and receive 81 points.}\par
\textcolor{colC}{You press <<23>>} \textcolor{colF}{and receive 79 points.}\par

\textcolor{colC}{You press <<}\textcolor{midgray}{\underline{\phantom{MM}}}
\end{promptbox}
\end{minipage}
\vspace{2pt}

\begin{minipage}[t]{0.48\textwidth}
\begin{promptbox}[Content-masked]
\textcolor{colI}{You will respond with 20, 22, 23, 24, etc.}\par
\textcolor{colM}{Environment number 1:}\par
\textcolor{colM}{The value of an option is revealed.}\par
\textcolor{colM}{You have some choices to make in this environment.}\par
\par\smallskip
\textcolor{colC}{You press <<20>>} \textcolor{colM}{and receive some points.}\par
\textcolor{colC}{You press <<22>>} \textcolor{colM}{and receive some points.}\par
\textcolor{colC}{You press <<23>>} \textcolor{colM}{and receive some points.}\par

\textcolor{colC}{You press <<}\textcolor{midgray}{\underline{\phantom{MM}}}
\end{promptbox}
\end{minipage}%
\hfill
\begin{minipage}[t]{0.24\textwidth}
\begin{promptbox}[History-only]
\textcolor{colI}{You will respond with 20, 22, 23, 24, etc.}\par
\par\smallskip
\textcolor{colC}{You press <<20>>.}\par
\textcolor{colC}{You press <<22>>.}\par
\textcolor{colC}{You press <<23>>.}\par

\textcolor{colC}{You press <<}\textcolor{midgray}{\underline{\phantom{MM}}}
\end{promptbox}
\end{minipage}%
\hfill
\begin{minipage}[t]{0.2\textwidth}
\begin{promptbox}[Choice-only]
\textcolor{colC}{<<20>>}\par
\textcolor{colC}{<<22>>}\par
\textcolor{colC}{<<23>>}\par

\textcolor{colC}{<<}\textcolor{midgray}{\underline{\phantom{MM}}}
\end{promptbox}
\end{minipage}
\vspace{2pt}
\begin{center}
\sffamily\fontsize{6.5}{8}\selectfont\color{darkgray}
\textcolor{colI}{\rule{5pt}{5pt}}\;\,Instruction / $I_\mathrm{min}$\quad
\textcolor{colS}{\rule{5pt}{5pt}}\;\,Stimuli\quad
\textcolor{colF}{\rule{5pt}{5pt}}\;\,Feedback\quad
\textcolor{colC}{\rule{5pt}{5pt}}\;\,Choice\quad
\textcolor{colM}{\rule{5pt}{5pt}}\;\,Masked\quad
\textcolor{midgray}{\underline{\phantom{MM}}}\;\,Model prediction
\end{center}
\vspace{8pt}
\includegraphics[width=\textwidth]{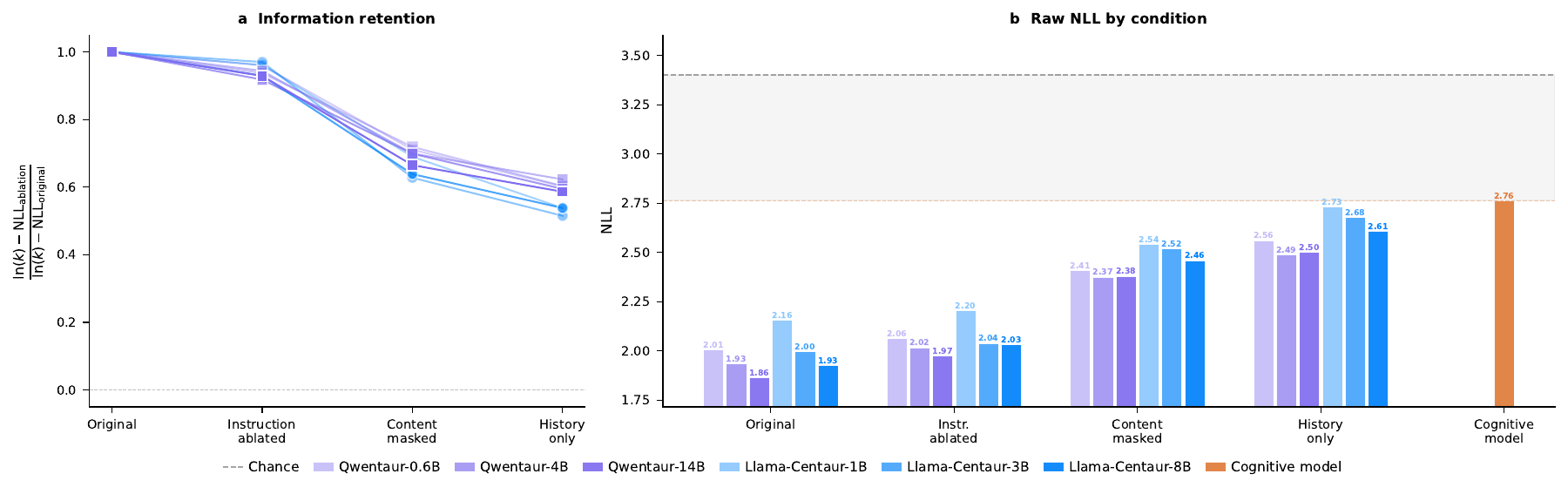}
\vspace{-4pt}
\caption{\textbf{Multi-armed bandits: Spatially correlated MAB} \citep{wu2018generalization}.}
\label{fig:ablation_wu2018generalisation}
\end{figure}

\begin{figure}[ht!]
\centering
\begin{minipage}[t]{0.48\textwidth}
\begin{promptbox}[Original]
\textcolor{colI}{You can sample from two monetary lotteries by pressing P or T. The lotteries offer different points with different probabilities. Initially, you will not know the outcomes and probabilities of the lotteries, but you can learn about them through sampling. Whenever you sample, a random draw from the selected lottery will be generated, which does not affect your bonus. You can sample 100 times from the lotteries in whatever order you like. After sampling 100 times, you have to choose one lottery for real by pressing the corresponding key. This choice will then trigger a random draw from the chosen lottery that will be added to your bonus. Your goal is to maximize your bonus. You will be presented with multiple choice problems consisting of different lotteries varying in outcomes and probabilities.}\par\smallskip
\textcolor{colI}{You encounter a new choice problem:}\par
\textcolor{colC}{You press <<P>>} \textcolor{colF}{and observe -400.0 points.}\par
\textcolor{colC}{You press <<T>>} \textcolor{colF}{and observe -600.0 points.}\par
\textcolor{colC}{You press <<P>>} \textcolor{colF}{and observe -400.0 points.}\par

\textcolor{colC}{You press <<}\textcolor{midgray}{\underline{\phantom{MM}}}
\end{promptbox}
\end{minipage}%
\hfill
\begin{minipage}[t]{0.48\textwidth}
\begin{promptbox}[Instruction-ablated]
\textcolor{colC}{You press <<P>>} \textcolor{colF}{and observe -400.0 points.}\par
\textcolor{colC}{You press <<T>>} \textcolor{colF}{and observe -600.0 points.}\par
\textcolor{colC}{You press <<P>>} \textcolor{colF}{and observe -400.0 points.}\par

\textcolor{colC}{You press <<}\textcolor{midgray}{\underline{\phantom{MM}}}
\end{promptbox}
\end{minipage}
\vspace{2pt}

\begin{minipage}[t]{0.48\textwidth}
\begin{promptbox}[Content-masked]
\textcolor{colI}{You will respond with P or T.}\par
\par\smallskip
\textcolor{colM}{You encounter a new choice problem:}\par
\textcolor{colC}{You press <<P>>} \textcolor{colM}{and observe some points.}\par
\textcolor{colC}{You press <<T>>} \textcolor{colM}{and observe some points.}\par
\textcolor{colC}{You press <<P>>} \textcolor{colM}{and observe some points.}\par

\textcolor{colC}{You press <<}\textcolor{midgray}{\underline{\phantom{MM}}}
\end{promptbox}
\end{minipage}%
\hfill
\begin{minipage}[t]{0.24\textwidth}
\begin{promptbox}[History-only]
\textcolor{colI}{You will respond with P or T.}\par
\par\smallskip
\textcolor{colC}{You press <<P>>.}\par
\textcolor{colC}{You press <<T>>.}\par
\textcolor{colC}{You press <<P>>.}\par

\textcolor{colC}{You press <<}\textcolor{midgray}{\underline{\phantom{MM}}}
\end{promptbox}
\end{minipage}%
\hfill
\begin{minipage}[t]{0.2\textwidth}
\begin{promptbox}[Choice-only]
\textcolor{colC}{<<P>>}\par
\textcolor{colC}{<<T>>}\par
\textcolor{colC}{<<P>>}\par

\textcolor{colC}{<<}\textcolor{midgray}{\underline{\phantom{MM}}}
\end{promptbox}
\end{minipage}
\vspace{2pt}
\begin{center}
\sffamily\fontsize{6.5}{8}\selectfont\color{darkgray}
\textcolor{colI}{\rule{5pt}{5pt}}\;\,Instruction / $I_\mathrm{min}$\quad
\textcolor{colS}{\rule{5pt}{5pt}}\;\,Stimuli\quad
\textcolor{colF}{\rule{5pt}{5pt}}\;\,Feedback\quad
\textcolor{colC}{\rule{5pt}{5pt}}\;\,Choice\quad
\textcolor{colM}{\rule{5pt}{5pt}}\;\,Masked\quad
\textcolor{midgray}{\underline{\phantom{MM}}}\;\,Model prediction
\end{center}
\vspace{8pt}
\includegraphics[width=\textwidth]{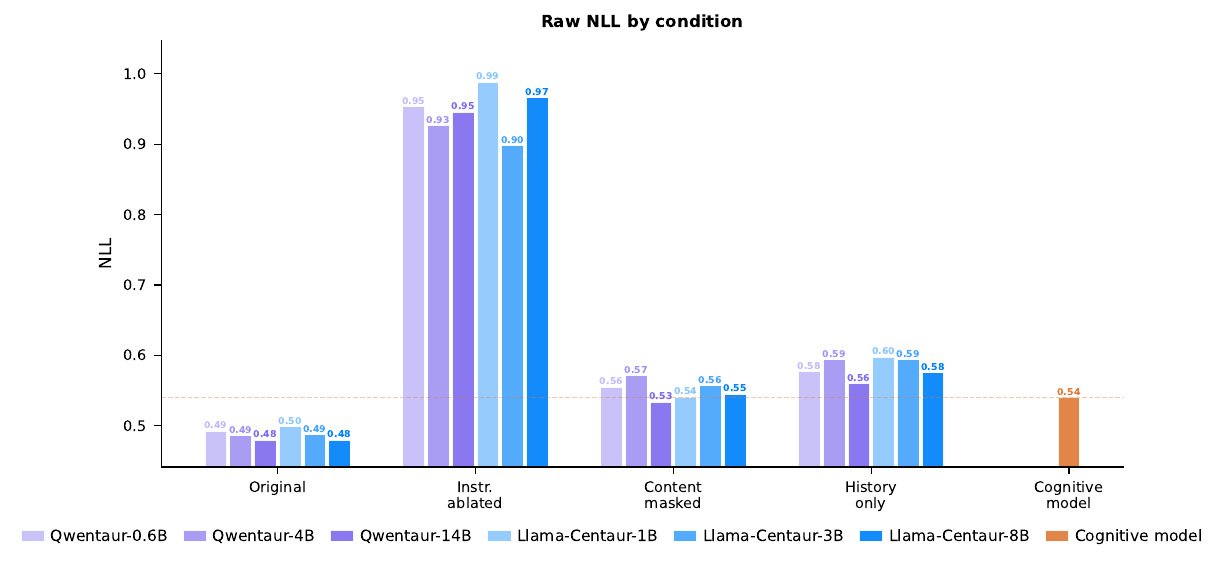}
\vspace{-4pt}
\caption{\textbf{Multi-armed bandits: Decisions from experience} \citep{wulff2018sampling}. No retention ratio or chance baseline is shown because trials mix sample, stop, and choose phases with different response formats, so no single $k$ defines a uniform-guessing floor.
}
\label{fig:ablation_wulff2018sampling}
\end{figure}

\begin{figure}[ht!]
\centering
\begin{minipage}[t]{0.48\textwidth}
\begin{promptbox}[Original]
\textcolor{colI}{You are presented with a series of stimuli, each associated with one of three possible responses. You must learn the correct response for each stimulus through trial and error. The three responses available are I, T, and Z. After your response, you will receive feedback: 1 point for a correct response, or 0 points for an incorrect response. The stimuli and their correct responses change between games.}\par\smallskip
\textcolor{colI}{Game 1:}\par
\textcolor{colS}{You see stimulus 1.} \textcolor{colC}{You press <<T>>} \textcolor{colF}{and get 0 points.}\par
\textcolor{colS}{You see stimulus 1.} \textcolor{colC}{You press <<I>>} \textcolor{colF}{and get 0 points.}\par
\textcolor{colS}{You see stimulus 0.} \textcolor{colC}{You press <<T>>} \textcolor{colF}{and get 1 points.}\par

\textcolor{colS}{You see stimulus 1.} \textcolor{colC}{You press <<}\textcolor{midgray}{\underline{\phantom{MM}}}
\end{promptbox}
\end{minipage}%
\hfill
\begin{minipage}[t]{0.48\textwidth}
\begin{promptbox}[Instruction-ablated]
\textcolor{colI}{Game 1:}\par
\textcolor{colS}{You see stimulus 1.} \textcolor{colC}{You press <<T>>} \textcolor{colF}{and get 0 points.}\par
\textcolor{colS}{You see stimulus 1.} \textcolor{colC}{You press <<I>>} \textcolor{colF}{and get 0 points.}\par
\textcolor{colS}{You see stimulus 0.} \textcolor{colC}{You press <<T>>} \textcolor{colF}{and get 1 points.}\par

\textcolor{colS}{You see stimulus 1.} \textcolor{colC}{You press <<}\textcolor{midgray}{\underline{\phantom{MM}}}
\end{promptbox}
\end{minipage}
\vspace{2pt}

\begin{minipage}[t]{0.48\textwidth}
\begin{promptbox}[Content-masked]
\textcolor{colI}{You will respond with T, I, or Z.}\par
\par\smallskip
\textcolor{colM}{You see a stimulus.} \textcolor{colC}{You press <<T>>} \textcolor{colM}{and get some points.}\par
\textcolor{colM}{You see a stimulus.} \textcolor{colC}{You press <<I>>} \textcolor{colM}{and get some points.}\par
\textcolor{colM}{You see a stimulus.} \textcolor{colC}{You press <<T>>} \textcolor{colM}{and get some points.}\par

\textcolor{colM}{You see a stimulus.} \textcolor{colC}{You press <<}\textcolor{midgray}{\underline{\phantom{MM}}}
\end{promptbox}
\end{minipage}%
\hfill
\begin{minipage}[t]{0.24\textwidth}
\begin{promptbox}[History-only]
\textcolor{colI}{You will respond with T, I, or Z.}\par
\par\smallskip
\textcolor{colC}{You press <<T>>.}\par
\textcolor{colC}{You press <<I>>.}\par
\textcolor{colC}{You press <<T>>.}\par

\textcolor{colC}{You press <<}\textcolor{midgray}{\underline{\phantom{MM}}}
\end{promptbox}
\end{minipage}%
\hfill
\begin{minipage}[t]{0.2\textwidth}
\begin{promptbox}[Choice-only]
\textcolor{colC}{<<T>>}\par
\textcolor{colC}{<<I>>}\par
\textcolor{colC}{<<T>>}\par

\textcolor{colC}{<<}\textcolor{midgray}{\underline{\phantom{MM}}}
\end{promptbox}
\end{minipage}
\vspace{2pt}
\begin{center}
\sffamily\fontsize{6.5}{8}\selectfont\color{darkgray}
\textcolor{colI}{\rule{5pt}{5pt}}\;\,Instruction / $I_\mathrm{min}$\quad
\textcolor{colS}{\rule{5pt}{5pt}}\;\,Stimuli\quad
\textcolor{colF}{\rule{5pt}{5pt}}\;\,Feedback\quad
\textcolor{colC}{\rule{5pt}{5pt}}\;\,Choice\quad
\textcolor{colM}{\rule{5pt}{5pt}}\;\,Masked\quad
\textcolor{midgray}{\underline{\phantom{MM}}}\;\,Model prediction
\end{center}
\vspace{8pt}
\includegraphics[width=\textwidth]{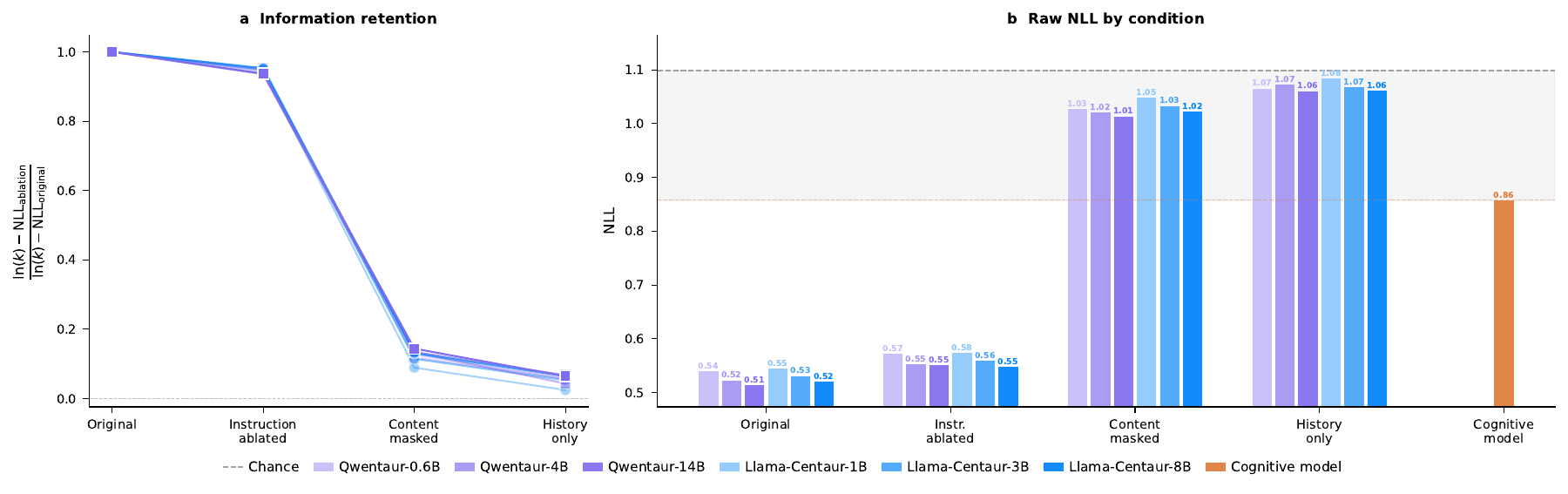}
\vspace{-4pt}
\caption{\textbf{Supervised learning: Conditional associative learning} \citep{collins2014working}.}
\label{fig:ablation_gershman2020reward}
\end{figure}

\begin{figure}[ht!]
\centering
\begin{minipage}[t]{0.48\textwidth}
\begin{promptbox}[Original]
\textcolor{colI}{You will observe a series of objects, one at a time. The objects differ along three binary dimensions: shape, size, and shading. Each dimension is indicated by the three digits of the image code (e.g., 121 means shape=1, size=2, shading=1). You will classify each object into one of two categories by pressing Q or J. After your response, you will receive feedback.}\par\smallskip
\textcolor{colS}{You see the image 221,} \textcolor{colC}{press <<J>>} \textcolor{colF}{and get 0 points.}\par
\textcolor{colS}{You see the image 222,} \textcolor{colC}{press <<Q>>} \textcolor{colF}{and get 0 points.}\par
\textcolor{colS}{You see the image 121,} \textcolor{colC}{press <<Q>>} \textcolor{colF}{and get 0 points.}\par

\textcolor{colS}{You see the image 122,} \textcolor{colC}{press <<}\textcolor{midgray}{\underline{\phantom{MM}}}
\end{promptbox}
\end{minipage}%
\hfill
\begin{minipage}[t]{0.48\textwidth}
\begin{promptbox}[Instruction-ablated]
\textcolor{colS}{You see the image 221,} \textcolor{colC}{press <<J>>} \textcolor{colF}{and get 0 points.}\par
\textcolor{colS}{You see the image 222,} \textcolor{colC}{press <<Q>>} \textcolor{colF}{and get 0 points.}\par
\textcolor{colS}{You see the image 121,} \textcolor{colC}{press <<Q>>} \textcolor{colF}{and get 0 points.}\par

\textcolor{colS}{You see the image 122,} \textcolor{colC}{press <<}\textcolor{midgray}{\underline{\phantom{MM}}}
\end{promptbox}
\end{minipage}
\vspace{2pt}

\begin{minipage}[t]{0.48\textwidth}
\begin{promptbox}[Content-masked]
\textcolor{colI}{You will respond with J or Q.}\par
\par\smallskip
\textcolor{colM}{You see an image,} \textcolor{colC}{press <<J>>} \textcolor{colM}{and get some points.}\par
\textcolor{colM}{You see an image,} \textcolor{colC}{press <<Q>>} \textcolor{colM}{and get some points.}\par
\textcolor{colM}{You see an image,} \textcolor{colC}{press <<Q>>} \textcolor{colM}{and get some points.}\par

\textcolor{colM}{You see an image,} \textcolor{colC}{press <<}\textcolor{midgray}{\underline{\phantom{MM}}}
\end{promptbox}
\end{minipage}%
\hfill
\begin{minipage}[t]{0.24\textwidth}
\begin{promptbox}[History-only]
\textcolor{colI}{You will respond with J or Q.}\par
\par\smallskip
\textcolor{colC}{You press <<J>>.}\par
\textcolor{colC}{You press <<Q>>.}\par
\textcolor{colC}{You press <<Q>>.}\par

\textcolor{colC}{You press <<}\textcolor{midgray}{\underline{\phantom{MM}}}
\end{promptbox}
\end{minipage}%
\hfill
\begin{minipage}[t]{0.2\textwidth}
\begin{promptbox}[Choice-only]
\textcolor{colC}{<<J>>}\par
\textcolor{colC}{<<Q>>}\par
\textcolor{colC}{<<Q>>}\par

\textcolor{colC}{<<}\textcolor{midgray}{\underline{\phantom{MM}}}
\end{promptbox}
\end{minipage}
\vspace{2pt}
\begin{center}
\sffamily\fontsize{6.5}{8}\selectfont\color{darkgray}
\textcolor{colI}{\rule{5pt}{5pt}}\;\,Instruction / $I_\mathrm{min}$\quad
\textcolor{colS}{\rule{5pt}{5pt}}\;\,Stimuli\quad
\textcolor{colF}{\rule{5pt}{5pt}}\;\,Feedback\quad
\textcolor{colC}{\rule{5pt}{5pt}}\;\,Choice\quad
\textcolor{colM}{\rule{5pt}{5pt}}\;\,Masked\quad
\textcolor{midgray}{\underline{\phantom{MM}}}\;\,Model prediction
\end{center}
\vspace{8pt}
\includegraphics[width=\textwidth]{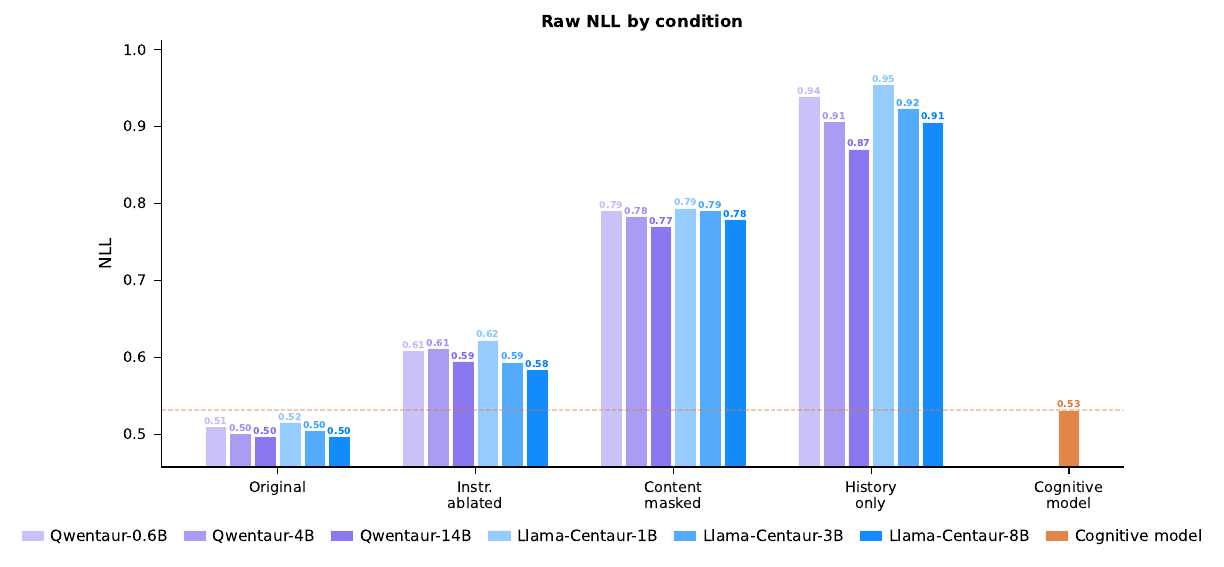}
\vspace{-4pt}
\caption{\textbf{Supervised learning: Medin categorization} \citep{levering2020revisiting}.  No retention ratio or chance baseline is shown because trials mix binary classification with 9-point rating responses, so no single $k$ defines a uniform-guessing floor.}
\label{fig:ablation_levering2020revisiting}
\end{figure}

\begin{figure}[ht!]
\centering
\begin{minipage}[t]{0.48\textwidth}
\begin{promptbox}[Original]
\textcolor{colI}{Each day you will either be presented with spaceships V and X or with spaceships M and D. These spaceships will take you to two different planets Z and P. Each planet has its own treasure, and the type and amount of treasure on each planet changes slowly over time. Your goal is to get as much treasure and as little antimatter as possible.}\par\smallskip
\textcolor{colS}{You are presented with spaceships D and M.} \textcolor{colC}{You press <<M>>.} \textcolor{colF}{You end up on planet Z. You find 2 pieces of antimatter.}\par
\textcolor{colS}{You are presented with spaceships X and V.} \textcolor{colC}{You press <<X>>.} \textcolor{colF}{You end up on planet P. You find 3 pieces of space treasure.}\par
\textcolor{colS}{You are presented with spaceships V and X.} \textcolor{colC}{You press <<X>>.} \textcolor{colF}{You end up on planet P. You find 3 pieces of space treasure.}\par

\textcolor{colS}{You are presented with spaceships D and M.} \textcolor{colC}{You press <<}\textcolor{midgray}{\underline{\phantom{MM}}}
\end{promptbox}
\end{minipage}%
\hfill
\begin{minipage}[t]{0.48\textwidth}
\begin{promptbox}[Instruction-ablated]
\textcolor{colS}{You are presented with spaceships D and M.} \textcolor{colC}{You press <<M>>.} \textcolor{colF}{You end up on planet Z. You find 2 pieces of antimatter.}\par
\textcolor{colS}{You are presented with spaceships X and V.} \textcolor{colC}{You press <<X>>.} \textcolor{colF}{You end up on planet P. You find 3 pieces of space treasure.}\par
\textcolor{colS}{You are presented with spaceships V and X.} \textcolor{colC}{You press <<X>>.} \textcolor{colF}{You end up on planet P. You find 3 pieces of space treasure.}\par

\textcolor{colS}{You are presented with spaceships D and M.} \textcolor{colC}{You press <<}\textcolor{midgray}{\underline{\phantom{MM}}}
\end{promptbox}
\end{minipage}
\vspace{2pt}

\begin{minipage}[t]{0.48\textwidth}
\begin{promptbox}[Content-masked]
\textcolor{colI}{You will respond with M, X, D, or V.}\par
\par\smallskip
\textcolor{colM}{You are presented with spaceships.} \textcolor{colC}{You press <<M>>.} \textcolor{colM}{You end up on a planet. You find an outcome.}\par
\textcolor{colM}{You are presented with spaceships.} \textcolor{colC}{You press <<X>>.} \textcolor{colM}{You end up on a planet. You find an outcome.}\par
\textcolor{colM}{You are presented with spaceships.} \textcolor{colC}{You press <<X>>.} \textcolor{colM}{You end up on a planet. You find an outcome.}\par

\textcolor{colM}{You are presented with spaceships.} \textcolor{colC}{You press <<}\textcolor{midgray}{\underline{\phantom{MM}}}
\end{promptbox}
\end{minipage}%
\hfill
\begin{minipage}[t]{0.24\textwidth}
\begin{promptbox}[History-only]
\textcolor{colI}{You will respond with M, X, D, or V.}\par
\par\smallskip
\textcolor{colC}{You press <<M>>.}\par
\textcolor{colC}{You press <<X>>.}\par
\textcolor{colC}{You press <<X>>.}\par

\textcolor{colC}{You press <<}\textcolor{midgray}{\underline{\phantom{MM}}}
\end{promptbox}
\end{minipage}%
\hfill
\begin{minipage}[t]{0.2\textwidth}
\begin{promptbox}[Choice-only]
\textcolor{colC}{<<M>>}\par
\textcolor{colC}{<<X>>}\par
\textcolor{colC}{<<X>>}\par

\textcolor{colC}{<<}\textcolor{midgray}{\underline{\phantom{MM}}}
\end{promptbox}
\end{minipage}
\vspace{2pt}
\begin{center}
\sffamily\fontsize{6.5}{8}\selectfont\color{darkgray}
\textcolor{colI}{\rule{5pt}{5pt}}\;\,Instruction / $I_\mathrm{min}$\quad
\textcolor{colS}{\rule{5pt}{5pt}}\;\,Stimuli\quad
\textcolor{colF}{\rule{5pt}{5pt}}\;\,Feedback\quad
\textcolor{colC}{\rule{5pt}{5pt}}\;\,Choice\quad
\textcolor{colM}{\rule{5pt}{5pt}}\;\,Masked\quad
\textcolor{midgray}{\underline{\phantom{MM}}}\;\,Model prediction
\end{center}
\vspace{8pt}
\includegraphics[width=\textwidth]{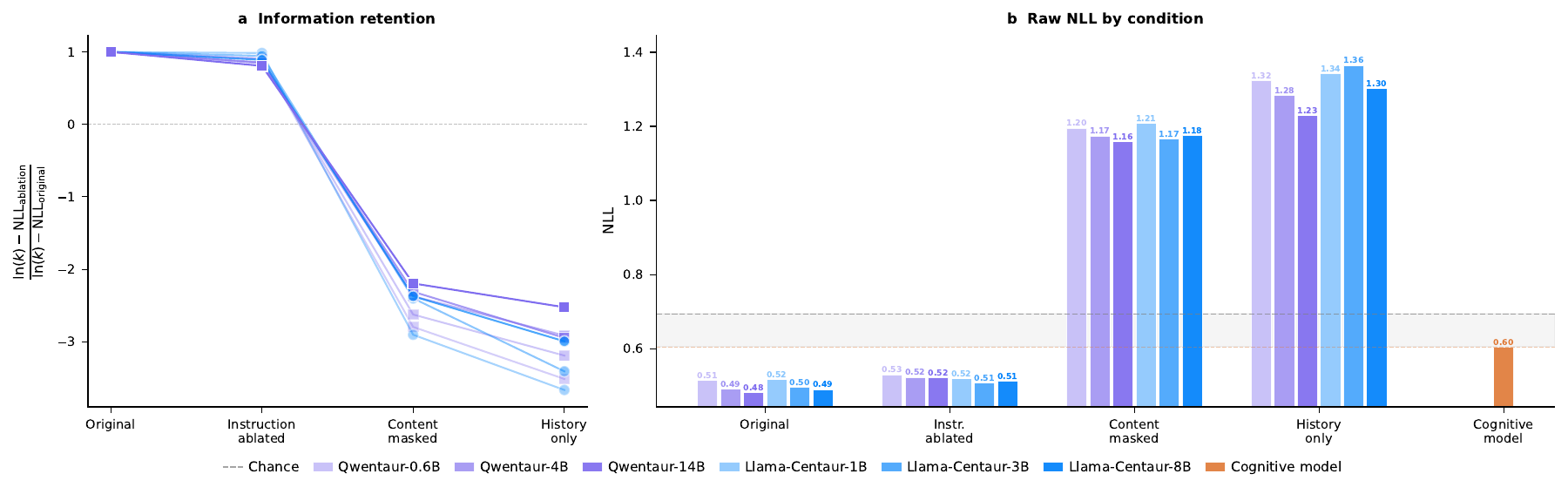}
\vspace{-4pt}
\caption{\textbf{MDP: Model-based two-step task} \citep{kool2016when}.}
\label{fig:ablation_kool2016when}
\end{figure}

\begin{figure}[ht!]
\centering
\begin{minipage}[t]{0.48\textwidth}
\begin{promptbox}[Original]
\textcolor{colI}{Each day you will either be presented with spaceships C and Z or with spaceships M and R. Before you choose a spaceship, you will be told whether there is a treasure multiplier. If there is a treasure multiplier, you will receive 5 times the amount of treasure. Your goal is to receive as much treasure as possible.}\par\smallskip
\textcolor{colS}{There is no treasure multiplier. You are presented with spaceships M and R.} \textcolor{colC}{You press <<R>>.} \textcolor{colF}{You end up on planet V. You find 2 pieces of space treasure. You receive 2 pieces of space treasure.}\par
\textcolor{colS}{There is a treasure multiplier. You are presented with spaceships M and R.} \textcolor{colC}{You press <<M>>.} \textcolor{colF}{You end up on planet U. You find 7 pieces of space treasure. You receive 35 pieces of space treasure.}\par
\textcolor{colS}{There is no treasure multiplier. You are presented with spaceships C and Z.} \textcolor{colC}{You press <<C>>.} \textcolor{colF}{You end up on planet V. You find 0 pieces of space treasure. You receive 0 pieces of space treasure.}\par

\textcolor{colS}{There is a treasure multiplier. You are presented with spaceships C and Z.} \textcolor{colC}{You press <<}\textcolor{midgray}{\underline{\phantom{MM}}}
\end{promptbox}
\end{minipage}%
\hfill
\begin{minipage}[t]{0.48\textwidth}
\begin{promptbox}[Instruction-ablated]
\textcolor{colS}{There is no treasure multiplier. You are presented with spaceships M and R.} \textcolor{colC}{You press <<R>>.} \textcolor{colF}{You end up on planet V. You find 2 pieces of space treasure. You receive 2 pieces of space treasure.}\par
\textcolor{colS}{There is a treasure multiplier. You are presented with spaceships M and R.} \textcolor{colC}{You press <<M>>.} \textcolor{colF}{You end up on planet U. You find 7 pieces of space treasure. You receive 35 pieces of space treasure.}\par
\textcolor{colS}{There is no treasure multiplier. You are presented with spaceships C and Z.} \textcolor{colC}{You press <<C>>.} \textcolor{colF}{You end up on planet V. You find 0 pieces of space treasure. You receive 0 pieces of space treasure.}\par

\textcolor{colS}{There is a treasure multiplier. You are presented with spaceships C and Z.} \textcolor{colC}{You press <<}\textcolor{midgray}{\underline{\phantom{MM}}}
\end{promptbox}
\end{minipage}
\vspace{2pt}

\begin{minipage}[t]{0.48\textwidth}
\begin{promptbox}[Content-masked]
\textcolor{colI}{You will respond with R, M, C, or Z.}\par
\par\smallskip
\textcolor{colM}{A condition is set. You are presented with spaceships.} \textcolor{colC}{You press <<R>>.} \textcolor{colM}{You end up on a planet. You find some points. You receive some points.}\par
\textcolor{colM}{A condition is set. You are presented with spaceships.} \textcolor{colC}{You press <<M>>.} \textcolor{colM}{You end up on a planet. You find some points. You receive some points.}\par
\textcolor{colM}{A condition is set. You are presented with spaceships.} \textcolor{colC}{You press <<C>>.} \textcolor{colM}{You end up on a planet. You find some points. You receive some points.}\par

\textcolor{colM}{A condition is set. You are presented with spaceships.} \textcolor{colC}{You press <<}\textcolor{midgray}{\underline{\phantom{MM}}}
\end{promptbox}
\end{minipage}%
\hfill
\begin{minipage}[t]{0.24\textwidth}
\begin{promptbox}[History-only]
\textcolor{colI}{You will respond with R, M, C, or Z.}\par
\par\smallskip
\textcolor{colC}{You press <<R>>.}\par
\textcolor{colC}{You press <<M>>.}\par
\textcolor{colC}{You press <<C>>.}\par

\textcolor{colC}{You press <<}\textcolor{midgray}{\underline{\phantom{MM}}}
\end{promptbox}
\end{minipage}%
\hfill
\begin{minipage}[t]{0.2\textwidth}
\begin{promptbox}[Choice-only]
\textcolor{colC}{<<R>>}\par
\textcolor{colC}{<<M>>}\par
\textcolor{colC}{<<C>>}\par

\textcolor{colC}{<<}\textcolor{midgray}{\underline{\phantom{MM}}}
\end{promptbox}
\end{minipage}
\vspace{2pt}
\begin{center}
\sffamily\fontsize{6.5}{8}\selectfont\color{darkgray}
\textcolor{colI}{\rule{5pt}{5pt}}\;\,Instruction / $I_\mathrm{min}$\quad
\textcolor{colS}{\rule{5pt}{5pt}}\;\,Stimuli\quad
\textcolor{colF}{\rule{5pt}{5pt}}\;\,Feedback\quad
\textcolor{colC}{\rule{5pt}{5pt}}\;\,Choice\quad
\textcolor{colM}{\rule{5pt}{5pt}}\;\,Masked\quad
\textcolor{midgray}{\underline{\phantom{MM}}}\;\,Model prediction
\end{center}
\vspace{8pt}
\includegraphics[width=\textwidth]{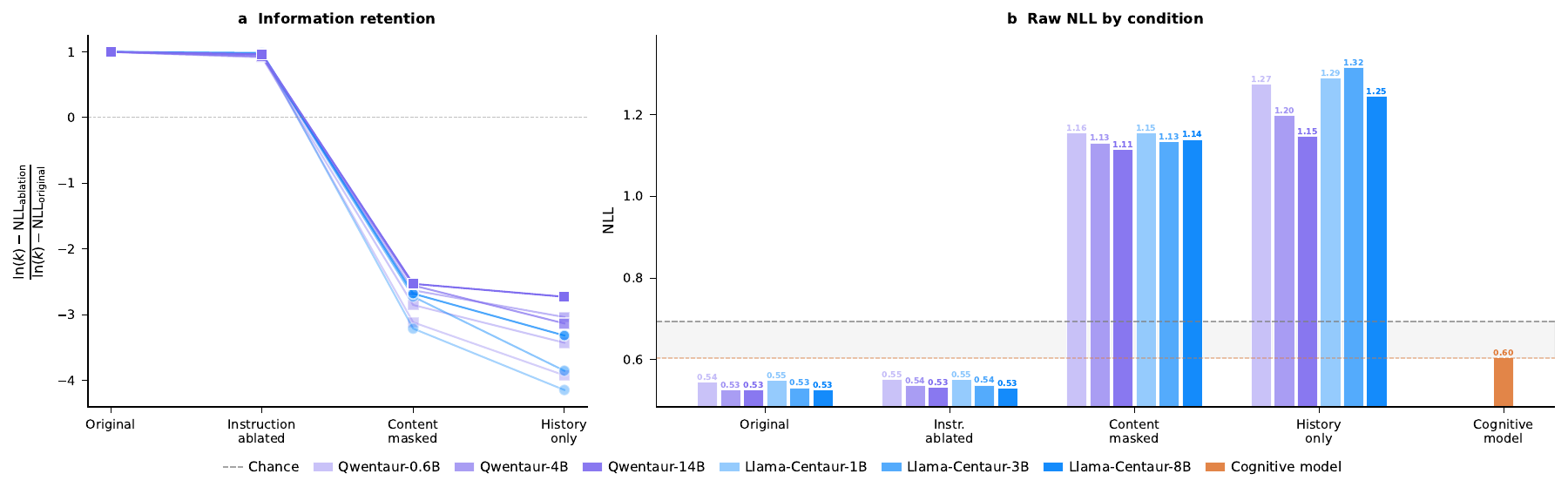}
\vspace{-4pt}
\caption{\textbf{MDP: Two-step effort cost} \citep{kool2017cost}.}
\label{fig:ablation_kool2017cost}
\end{figure}

\begin{figure}[ht!]
\centering
\begin{minipage}[t]{0.48\textwidth}
\begin{promptbox}[Original]
\textcolor{colI}{You are participating in a space treasure game. Each planet has two aliens on it. The blue aliens live on the blue planet. The red aliens live on the red planet. Each alien has a different chance of giving you treasure vs.\ junk. Your goal is to earn as much treasure as possible.}\par\smallskip
\textcolor{colS}{You are presented with two spaceships called Q and E.} \textcolor{colC}{You press <<Q>>.} \textcolor{colF}{You end up on the blue planet. You see a blue alien named U and a blue alien named Z.} \textcolor{colC}{You press <<Z>>.} \textcolor{colF}{You find treasure.}\par
\textcolor{colS}{You are presented with two spaceships called Q and E.} \textcolor{colC}{You press <<Q>>.} \textcolor{colF}{You end up on the blue planet. You see a blue alien named U and a blue alien named Z.} \textcolor{colC}{You press <<R>>.} \textcolor{colF}{You find junk.}\par
\textcolor{colS}{You are presented with two spaceships called Q and E.} \textcolor{colC}{You press <<E>>.} \textcolor{colF}{You end up on the red planet. You see a red alien named M and a red alien named R.} \textcolor{colC}{You press <<M>>.} \textcolor{colF}{You find junk.}\par
\textcolor{colS}{You are presented with two spaceships called Q and E.} \textcolor{colC}{You press <<}\textcolor{midgray}{\underline{\phantom{MM}}}
\end{promptbox}
\end{minipage}%
\hfill
\begin{minipage}[t]{0.48\textwidth}
\begin{promptbox}[Instruction-ablated]
\textcolor{colS}{You are presented with two spaceships called Q and E.} \textcolor{colC}{You press <<Q>>.} \textcolor{colF}{You end up on the blue planet. You see a blue alien named U and a blue alien named Z.} \textcolor{colC}{You press <<Z>>.} \textcolor{colF}{You find treasure.}\par
\textcolor{colS}{You are presented with two spaceships called Q and E.} \textcolor{colC}{You press <<Q>>.} \textcolor{colF}{You end up on the blue planet. You see a blue alien named U and a blue alien named Z.} \textcolor{colC}{You press <<R>>.} \textcolor{colF}{You find junk.}\par
\textcolor{colS}{You are presented with two spaceships called Q and E.} \textcolor{colC}{You press <<E>>.} \textcolor{colF}{You end up on the red planet. You see a red alien named M and a red alien named R.} \textcolor{colC}{You press <<M>>.} \textcolor{colF}{You find junk.}\par
\textcolor{colS}{You are presented with two spaceships called Q and E.} \textcolor{colC}{You press <<}\textcolor{midgray}{\underline{\phantom{MM}}}
\end{promptbox}
\end{minipage}
\vspace{2pt}

\begin{minipage}[t]{0.48\textwidth}
\begin{promptbox}[Content-masked]
\textcolor{colI}{You will respond with Q, Z, R, U, E, or M.}\par
\par\smallskip
\textcolor{colM}{You are presented with two spaceships.} \textcolor{colC}{You press <<Q>>.} \textcolor{colM}{You end up on a planet. You see two aliens.} \textcolor{colC}{You press <<Z>>.} \textcolor{colM}{You find an outcome.}\par
\textcolor{colM}{You are presented with two spaceships.} \textcolor{colC}{You press <<Q>>.} \textcolor{colM}{You end up on a planet. You see two aliens.} \textcolor{colC}{You press <<R>>.} \textcolor{colM}{You find an outcome.}\par
\textcolor{colM}{You are presented with two spaceships.} \textcolor{colC}{You press <<E>>.} \textcolor{colM}{You end up on a planet. You see two aliens.} \textcolor{colC}{You press <<M>>.} \textcolor{colM}{You find an outcome.}\par
\textcolor{colM}{You are presented with two spaceships.} \textcolor{colC}{You press <<}\textcolor{midgray}{\underline{\phantom{MM}}}
\end{promptbox}
\end{minipage}%
\hfill
\begin{minipage}[t]{0.24\textwidth}
\begin{promptbox}[History-only]
\textcolor{colI}{You will respond with Q, Z, R, U, E, or M.}\par
\par\smallskip
\textcolor{colC}{You press <<Q>>.}\par
\textcolor{colC}{You press <<Z>>.}\par
\textcolor{colC}{You press <<Q>>.}\par

\textcolor{colC}{You press <<}\textcolor{midgray}{\underline{\phantom{MM}}}
\end{promptbox}
\end{minipage}%
\hfill
\begin{minipage}[t]{0.2\textwidth}
\begin{promptbox}[Choice-only]
\textcolor{colC}{<<Q>>}\par
\textcolor{colC}{<<Z>>}\par
\textcolor{colC}{<<Q>>}\par

\textcolor{colC}{<<}\textcolor{midgray}{\underline{\phantom{MM}}}
\end{promptbox}
\end{minipage}
\vspace{2pt}
\begin{center}
\sffamily\fontsize{6.5}{8}\selectfont\color{darkgray}
\textcolor{colI}{\rule{5pt}{5pt}}\;\,Instruction / $I_\mathrm{min}$\quad
\textcolor{colS}{\rule{5pt}{5pt}}\;\,Stimuli\quad
\textcolor{colF}{\rule{5pt}{5pt}}\;\,Feedback\quad
\textcolor{colC}{\rule{5pt}{5pt}}\;\,Choice\quad
\textcolor{colM}{\rule{5pt}{5pt}}\;\,Masked\quad
\textcolor{midgray}{\underline{\phantom{MM}}}\;\,Model prediction
\end{center}
\vspace{8pt}
\includegraphics[width=\textwidth]{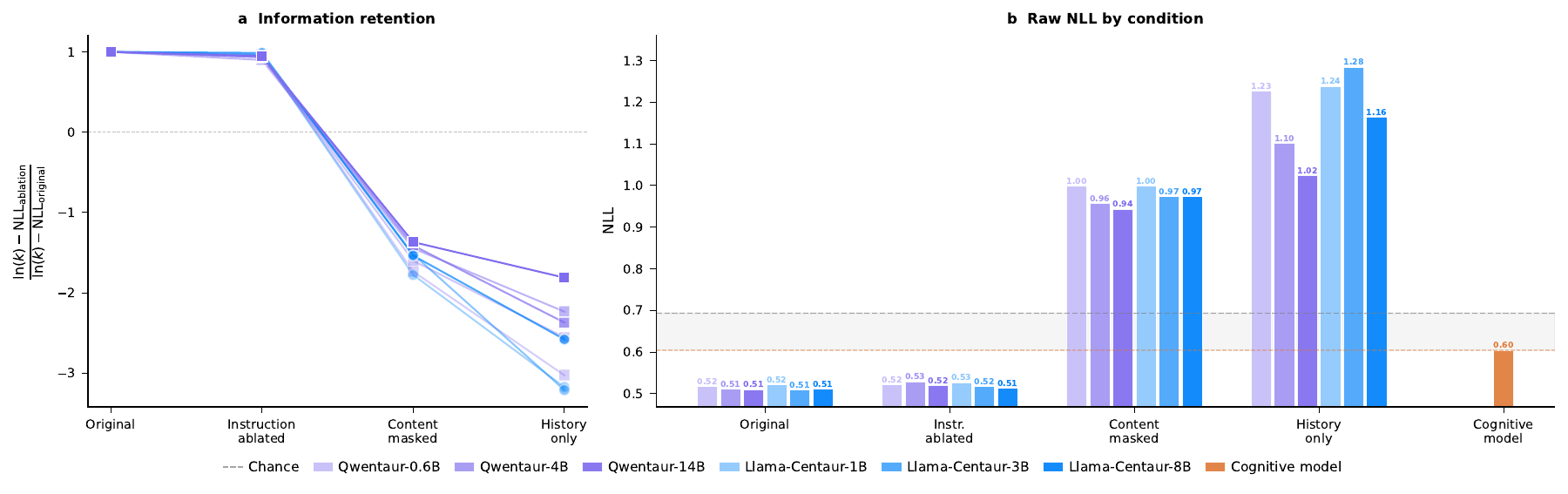}
\vspace{-4pt}
\caption{\textbf{MDP: Two-step with aliens} \citep{zorowitz2023data}.}
\label{fig:ablation_zorowitz2023data}
\end{figure}

\begin{figure}[ht!]
\centering
\begin{minipage}[t]{0.48\textwidth}
\begin{promptbox}[Original]
\textcolor{colI}{You will explore a castle, walking from room to room. Each room has three doors labeled B, N, and Z. When you walk through a door, you find some resources (wood, stone, iron). These resources have market values that change between rounds. Your points are calculated from the resources you find multiplied by their market prices.}\par\smallskip
\textcolor{colS}{The current market prices are 1 for wood, -1 for stone, and 0 for iron.}\par
\textcolor{colS}{You are in room 0.} \textcolor{colC}{You press <<B>>} \textcolor{colF}{and you find 0 wood, 0 stone, and 0 iron. You get 0 points.}\par
\textcolor{colS}{You are in room 1.} \textcolor{colC}{You press <<Z>>} \textcolor{colF}{and you find 70 wood, 70 stone, and 70 iron. You get 0 points.}\par
\textcolor{colS}{You are in room 2.} \textcolor{colC}{You press <<B>>} \textcolor{colF}{and you find 0 wood, 0 stone, and 0 iron. You get 0 points.}\par

\textcolor{colS}{You are in room 1.} \textcolor{colC}{You press <<}\textcolor{midgray}{\underline{\phantom{MM}}}
\end{promptbox}
\end{minipage}%
\hfill
\begin{minipage}[t]{0.48\textwidth}
\begin{promptbox}[Instruction-ablated]
\textcolor{colS}{The current market prices are 1 for wood, -1 for stone, and 0 for iron.}\par
\textcolor{colS}{You are in room 0.} \textcolor{colC}{You press <<B>>} \textcolor{colF}{and you find 0 wood, 0 stone, and 0 iron. You get 0 points.}\par
\textcolor{colS}{You are in room 1.} \textcolor{colC}{You press <<Z>>} \textcolor{colF}{and you find 70 wood, 70 stone, and 70 iron. You get 0 points.}\par
\textcolor{colS}{You are in room 2.} \textcolor{colC}{You press <<B>>} \textcolor{colF}{and you find 0 wood, 0 stone, and 0 iron. You get 0 points.}\par

\textcolor{colS}{You are in room 1.} \textcolor{colC}{You press <<}\textcolor{midgray}{\underline{\phantom{MM}}}
\end{promptbox}
\end{minipage}
\vspace{2pt}

\begin{minipage}[t]{0.48\textwidth}
\begin{promptbox}[Content-masked]
\textcolor{colI}{You will respond with B, Z, or N.}\par
\par\smallskip
\textcolor{colM}{New market prices are set.}\par
\textcolor{colM}{You are in a room.} \textcolor{colC}{You press <<B>>} \textcolor{colM}{and you find some resources. You get some points.}\par
\textcolor{colM}{You are in a room.} \textcolor{colC}{You press <<Z>>} \textcolor{colM}{and you find some resources. You get some points.}\par
\textcolor{colM}{You are in a room.} \textcolor{colC}{You press <<B>>} \textcolor{colM}{and you find some resources. You get some points.}\par

\textcolor{colM}{You are in a room.} \textcolor{colC}{You press <<}\textcolor{midgray}{\underline{\phantom{MM}}}
\end{promptbox}
\end{minipage}%
\hfill
\begin{minipage}[t]{0.24\textwidth}
\begin{promptbox}[History-only]
\textcolor{colI}{You will respond with B, Z, or N.}\par
\par\smallskip
\textcolor{colC}{You press <<B>>.}\par
\textcolor{colC}{You press <<Z>>.}\par
\textcolor{colC}{You press <<B>>.}\par

\textcolor{colC}{You press <<}\textcolor{midgray}{\underline{\phantom{MM}}}
\end{promptbox}
\end{minipage}%
\hfill
\begin{minipage}[t]{0.2\textwidth}
\begin{promptbox}[Choice-only]
\textcolor{colC}{<<B>>}\par
\textcolor{colC}{<<Z>>}\par
\textcolor{colC}{<<B>>}\par

\textcolor{colC}{<<}\textcolor{midgray}{\underline{\phantom{MM}}}
\end{promptbox}
\end{minipage}
\vspace{2pt}
\begin{center}
\sffamily\fontsize{6.5}{8}\selectfont\color{darkgray}
\textcolor{colI}{\rule{5pt}{5pt}}\;\,Instruction / $I_\mathrm{min}$\quad
\textcolor{colS}{\rule{5pt}{5pt}}\;\,Stimuli\quad
\textcolor{colF}{\rule{5pt}{5pt}}\;\,Feedback\quad
\textcolor{colC}{\rule{5pt}{5pt}}\;\,Choice\quad
\textcolor{colM}{\rule{5pt}{5pt}}\;\,Masked\quad
\textcolor{midgray}{\underline{\phantom{MM}}}\;\,Model prediction
\end{center}
\vspace{8pt}
\includegraphics[width=\textwidth]{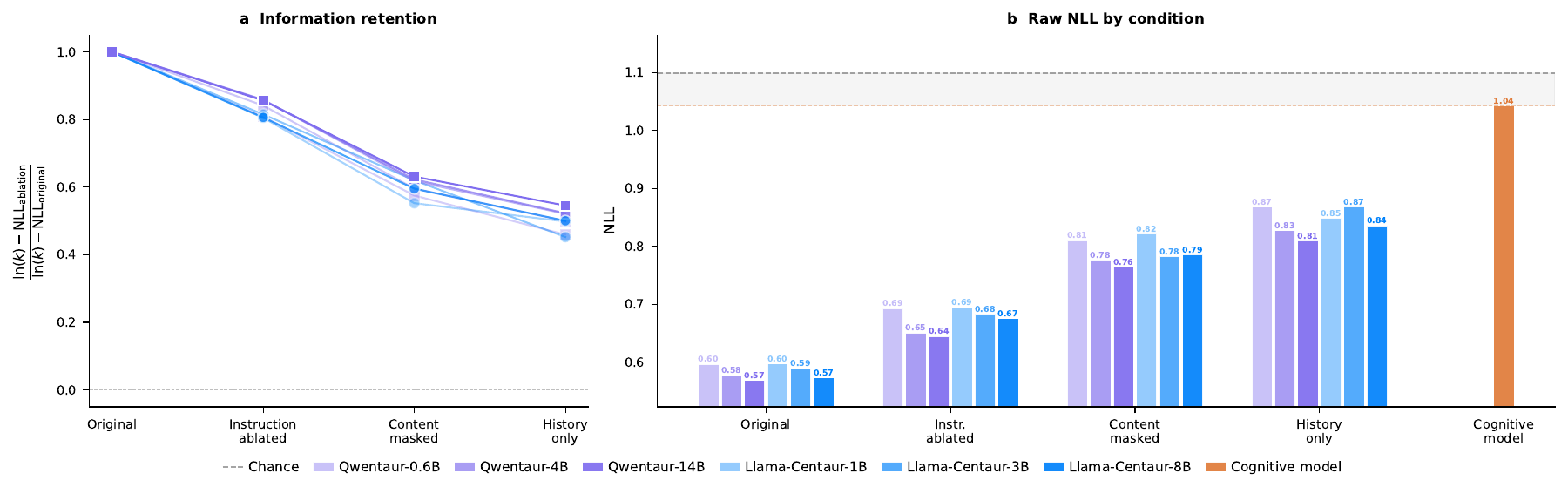}
\vspace{-4pt}
\caption{\textbf{MDP: Multi-task reinforcement learning} \citep{tomov2021multitask}.}
\label{fig:ablation_tomov2021multitask}
\end{figure}

Figure~\ref{fig:structural_ablation_app} gives per-experiment heatmaps of the fraction of learned information lost, $\delta = (\mathcal{L}_c - \mathcal{L}_{\mathrm{orig}})/(\ln k - \mathcal{L}_{\mathrm{orig}})$, averaged across models, with a separate panel and colour scale for the six experiments driven below chance ($\delta > 1$). Table~\ref{tab:psych101_pe_ablation} reports the underlying raw NLL for the eight \texttt{Qwentaur} and \texttt{Llama-Centaur} models under every condition, and Table~\ref{tab:tab:psych101_pe_ablation_centaur70b} the same for \texttt{Centaur-70B}; both carry an $\ln(k)$ column giving the chance baseline, so a cell above it indicates worse-than-chance prediction. Figures~\ref{fig:ablation_badham2017deficits}--\ref{fig:ablation_tomov2021multitask} then take each experiment in turn, showing the original prompt beside all four ablation conditions, colour-coded by information channel.

\subsubsection{Non-monotonic experiments}
\label{app:nonmonotonic}

\begin{figure}[H]
\centering
\includegraphics[width=\textwidth]{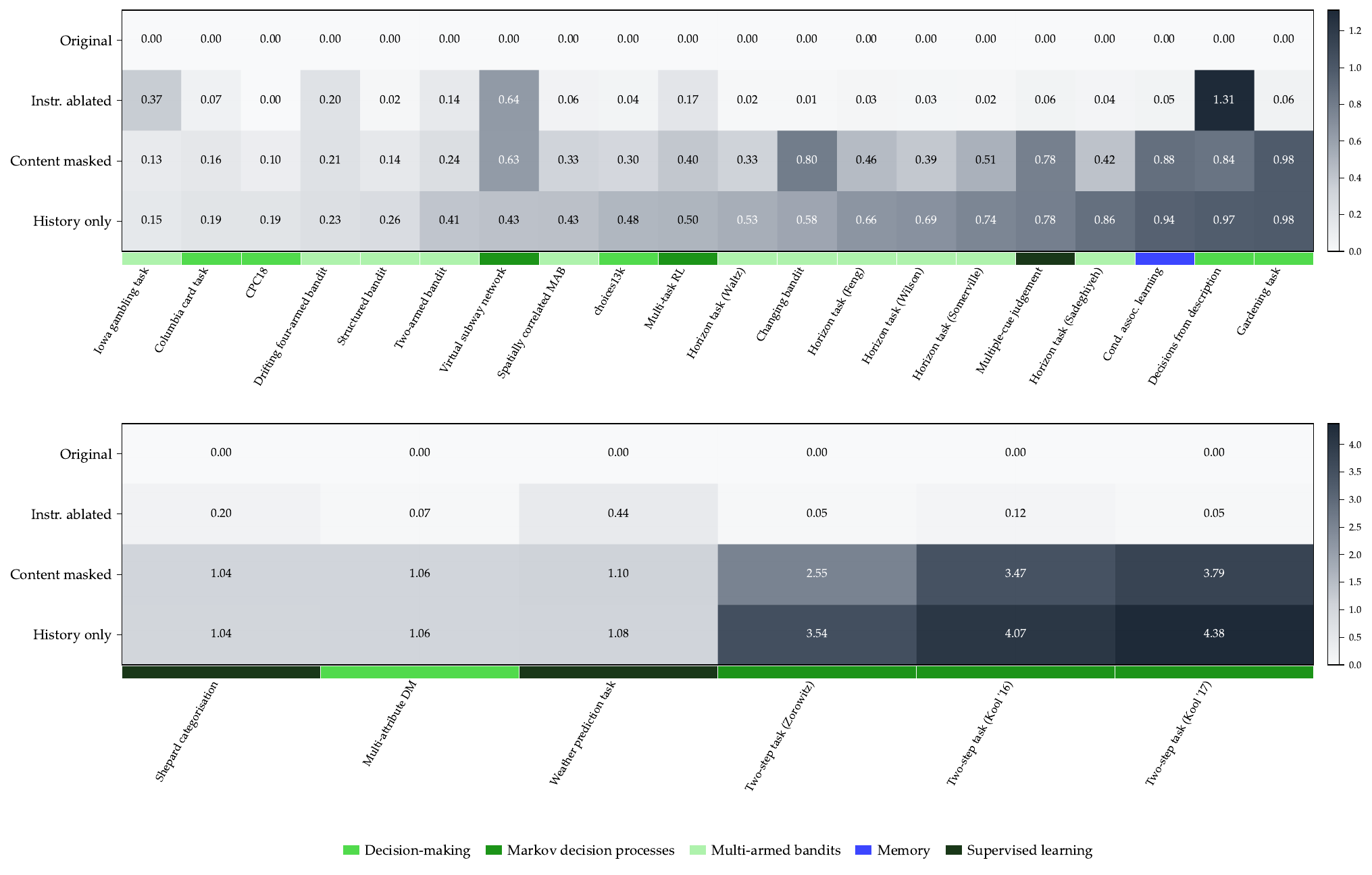}
  \caption{%
  \textbf{Complete per-experiment ablation heatmaps for 26 experiments
  with a well-defined chance baseline.} Five additional experiments with mixed or continuous response formats are excluded throughout because no single $k$ defines a chance baseline. The probabilistic instrumental learning task
  \citep{lefebvre2017behavioural} is also omitted. 
  Each cell shows the fraction of learned information lost,
  $\delta = (\mathcal{L}_c - \mathcal{L}_{\mathrm{orig}})\,/\,(\ln k -
  \mathcal{L}_{\mathrm{orig}})$,
  averaged across all eight models.
  Columns represent individual experiments sorted by $\delta$ under the
  history-only condition; rows represent ablation conditions.
  \textbf{(Top)}~20 experiments with mean $\delta \leq 1$ under the
  history-only condition (i.e.\ performance remains at or above chance).
  \textbf{(Bottom)}~6 experiments where ablation degrades performance
  below chance ($\delta > 1$); note the separate colour scale.
  The colour strip below each panel indicates task type (see legend).
}
\label{fig:structural_ablation_app}
\end{figure}

Of the 32 experiments that admit structural ablation, 27 have a well-defined chance baseline; the remaining five use mixed or continuous response formats, for which no single $k$ defines $\ln k$. Eighteen of those 27 ($67\%$) degrade strictly monotonically across the four ablation conditions (Figure~\ref{fig:structural_ablation}b, with additional experiments whose history-only loss exceeds 1.0 shown in Figure~\ref{fig:structural_ablation_app}). This section accounts for the nine that do not.

\begin{figure}[ht!]
\centering
\begin{minipage}[t]{0.48\textwidth}
\begin{promptbox}[Original]
\textcolor{colI}{You will be shown several examples of geometric objects. Your task is to learn a rule that allows you to tell whether an object belongs to the R or C category. For each presented object, you will be asked to make a category judgment by pressing the corresponding key and then you will receive feedback. You will encounter four different problems with different rules.}\par\smallskip
\textcolor{colI}{You encounter a new problem with a new rule determining which objects belong to each category:}\par
\textcolor{colS}{You see a small white triangle.} \textcolor{colC}{You press <<R>>} \textcolor{colF}{. The correct category is R.}\par
\textcolor{colS}{You see a big white square.} \textcolor{colC}{You press <<R>>} \textcolor{colF}{. The correct category is R.}\par
\textcolor{colS}{You see a big black square.} \textcolor{colC}{You press <<C>>} \textcolor{colF}{. The correct category is R.}\par

\textcolor{colS}{You see a big black triangle.} \textcolor{colC}{You press <<}\textcolor{midgray}{\underline{\phantom{MM}}}
\end{promptbox}
\end{minipage}%
\hfill
\begin{minipage}[t]{0.48\textwidth}
\begin{promptbox}[Instruction-ablated]
\textcolor{colS}{You see a small white triangle.} \textcolor{colC}{You press <<R>>} \textcolor{colF}{. The correct category is R.}\par
\textcolor{colS}{You see a big white square.} \textcolor{colC}{You press <<R>>} \textcolor{colF}{. The correct category is R.}\par
\textcolor{colS}{You see a big black square.} \textcolor{colC}{You press <<C>>} \textcolor{colF}{. The correct category is R.}\par

\textcolor{colS}{You see a big black triangle.} \textcolor{colC}{You press <<}\textcolor{midgray}{\underline{\phantom{MM}}}
\end{promptbox}
\end{minipage}
\vspace{2pt}

\begin{minipage}[t]{0.48\textwidth}
\begin{promptbox}[Content-masked]
\textcolor{colI}{You will respond with R or C.}\par
\par\smallskip
\textcolor{colM}{You see a shape.} \textcolor{colC}{You press <<R>>} \textcolor{colM}{. The correct answer is revealed.}\par
\textcolor{colM}{You see a shape.} \textcolor{colC}{You press <<R>>} \textcolor{colM}{. The correct answer is revealed.}\par
\textcolor{colM}{You see a shape.} \textcolor{colC}{You press <<C>>} \textcolor{colM}{. The correct answer is revealed.}\par

\textcolor{colM}{You see a shape.} \textcolor{colC}{You press <<}\textcolor{midgray}{\underline{\phantom{MM}}}
\end{promptbox}
\end{minipage}%
\hfill
\begin{minipage}[t]{0.24\textwidth}
\begin{promptbox}[History-only]
\textcolor{colI}{You will respond with R or C.}\par
\par\smallskip
\textcolor{colC}{You press <<R>>.}\par
\textcolor{colC}{You press <<R>>.}\par
\textcolor{colC}{You press <<C>>.}\par

\textcolor{colC}{You press <<}\textcolor{midgray}{\underline{\phantom{MM}}}
\end{promptbox}
\end{minipage}%
\hfill
\begin{minipage}[t]{0.2\textwidth}
\begin{promptbox}[Choice-only]
\textcolor{colC}{<<R>>}\par
\textcolor{colC}{<<R>>}\par
\textcolor{colC}{<<C>>}\par

\textcolor{colC}{<<}\textcolor{midgray}{\underline{\phantom{MM}}}
\end{promptbox}
\end{minipage}
\vspace{2pt}
\begin{center}
\sffamily\fontsize{6.5}{8}\selectfont\color{darkgray}
\textcolor{colI}{\rule{5pt}{5pt}}\;\,Instruction / $I_\mathrm{min}$\quad
\textcolor{colS}{\rule{5pt}{5pt}}\;\,Stimuli\quad
\textcolor{colF}{\rule{5pt}{5pt}}\;\,Feedback\quad
\textcolor{colC}{\rule{5pt}{5pt}}\;\,Choice\quad
\textcolor{colM}{\rule{5pt}{5pt}}\;\,Masked\quad
\textcolor{midgray}{\underline{\phantom{MM}}}\;\,Model prediction
\end{center}
\vspace{8pt}
\includegraphics[width=\textwidth]{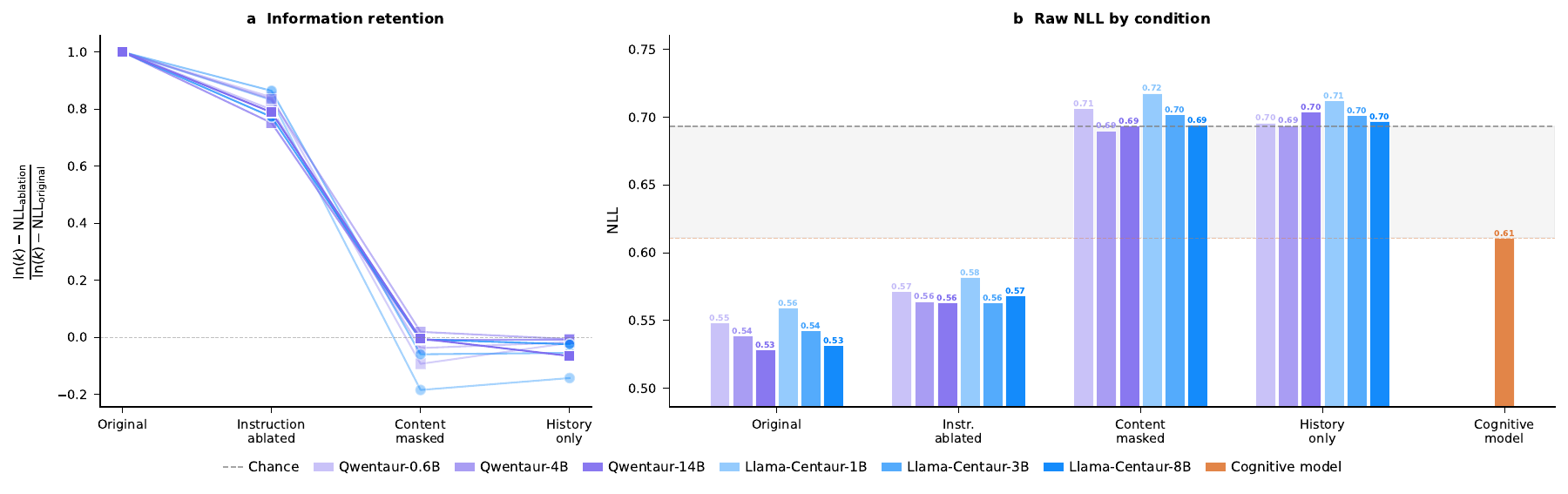}
\vspace{-4pt}
\caption{\textbf{Supervised learning: Shepard categorization} \citep{badham2017deficits}.}
\label{fig:ablation_badham2017deficits}
\end{figure}

\begin{figure}[ht!]
\centering
\begin{minipage}[t]{0.48\textwidth}
\begin{promptbox}[Original]
\textcolor{colI}{You are going to visit four different casinos (named 1, 2, 3, and 4) 24 times each. Each casino owns two slot machines that return either 0 or 0.5 points stochastically with different probabilities. You can play one of the machines in order to win points by pressing the corresponding key. Your goal is to maximize the sum of received points within all visits.}\par\smallskip
\textcolor{colS}{You go to casino 3. You can choose between machines A and V.} \textcolor{colC}{You press <<V>>} \textcolor{colF}{and receive 0.5 points.}\par
\textcolor{colS}{You go to casino 4. You can choose between machines B and U.} \textcolor{colC}{You press <<B>>} \textcolor{colF}{and receive 0.0 points.}\par
\textcolor{colS}{You go to casino 2. You can choose between machines K and C.} \textcolor{colC}{You press <<C>>} \textcolor{colF}{and receive 0.0 points.}\par

\textcolor{colS}{You go to casino 1. You can choose between machines L and F.} \textcolor{colC}{You press <<}\textcolor{midgray}{\underline{\phantom{MM}}}
\end{promptbox}
\end{minipage}%
\hfill
\begin{minipage}[t]{0.48\textwidth}
\begin{promptbox}[Instruction-ablated]
\textcolor{colS}{You go to casino 3. You can choose between machines A and V.} \textcolor{colC}{You press <<V>>} \textcolor{colF}{and receive 0.5 points.}\par
\textcolor{colS}{You go to casino 4. You can choose between machines B and U.} \textcolor{colC}{You press <<B>>} \textcolor{colF}{and receive 0.0 points.}\par
\textcolor{colS}{You go to casino 2. You can choose between machines K and C.} \textcolor{colC}{You press <<C>>} \textcolor{colF}{and receive 0.0 points.}\par

\textcolor{colS}{You go to casino 1. You can choose between machines L and F.} \textcolor{colC}{You press <<}\textcolor{midgray}{\underline{\phantom{MM}}}
\end{promptbox}
\end{minipage}
\vspace{2pt}

\begin{minipage}[t]{0.48\textwidth}
\begin{promptbox}[Content-masked]
\textcolor{colI}{You will respond with V, B, C, F, etc.}\par
\par\smallskip
\textcolor{colM}{You go to a casino. You can choose between machines.} \textcolor{colC}{You press <<V>>} \textcolor{colM}{and receive some points.}\par
\textcolor{colM}{You go to a casino. You can choose between machines.} \textcolor{colC}{You press <<B>>} \textcolor{colM}{and receive some points.}\par
\textcolor{colM}{You go to a casino. You can choose between machines.} \textcolor{colC}{You press <<C>>} \textcolor{colM}{and receive some points.}\par

\textcolor{colM}{You go to a casino. You can choose between machines.} \textcolor{colC}{You press <<}\textcolor{midgray}{\underline{\phantom{MM}}}
\end{promptbox}
\end{minipage}%
\hfill
\begin{minipage}[t]{0.24\textwidth}
\begin{promptbox}[History-only]
\textcolor{colI}{You will respond with V, B, C, F, etc.}\par
\par\smallskip
\textcolor{colC}{You press <<V>>.}\par
\textcolor{colC}{You press <<B>>.}\par
\textcolor{colC}{You press <<C>>.}\par

\textcolor{colC}{You press <<}\textcolor{midgray}{\underline{\phantom{MM}}}
\end{promptbox}
\end{minipage}%
\hfill
\begin{minipage}[t]{0.2\textwidth}
\begin{promptbox}[Choice-only]
\textcolor{colC}{<<V>>}\par
\textcolor{colC}{<<B>>}\par
\textcolor{colC}{<<C>>}\par

\textcolor{colC}{<<}\textcolor{midgray}{\underline{\phantom{MM}}}
\end{promptbox}
\end{minipage}
\vspace{2pt}
\begin{center}
\sffamily\fontsize{6.5}{8}\selectfont\color{darkgray}
\textcolor{colI}{\rule{5pt}{5pt}}\;\,Instruction / $I_\mathrm{min}$\quad
\textcolor{colS}{\rule{5pt}{5pt}}\;\,Stimuli\quad
\textcolor{colF}{\rule{5pt}{5pt}}\;\,Feedback\quad
\textcolor{colC}{\rule{5pt}{5pt}}\;\,Choice\quad
\textcolor{colM}{\rule{5pt}{5pt}}\;\,Masked\quad
\textcolor{midgray}{\underline{\phantom{MM}}}\;\,Model prediction
\end{center}
\vspace{8pt}
\includegraphics[width=\textwidth]{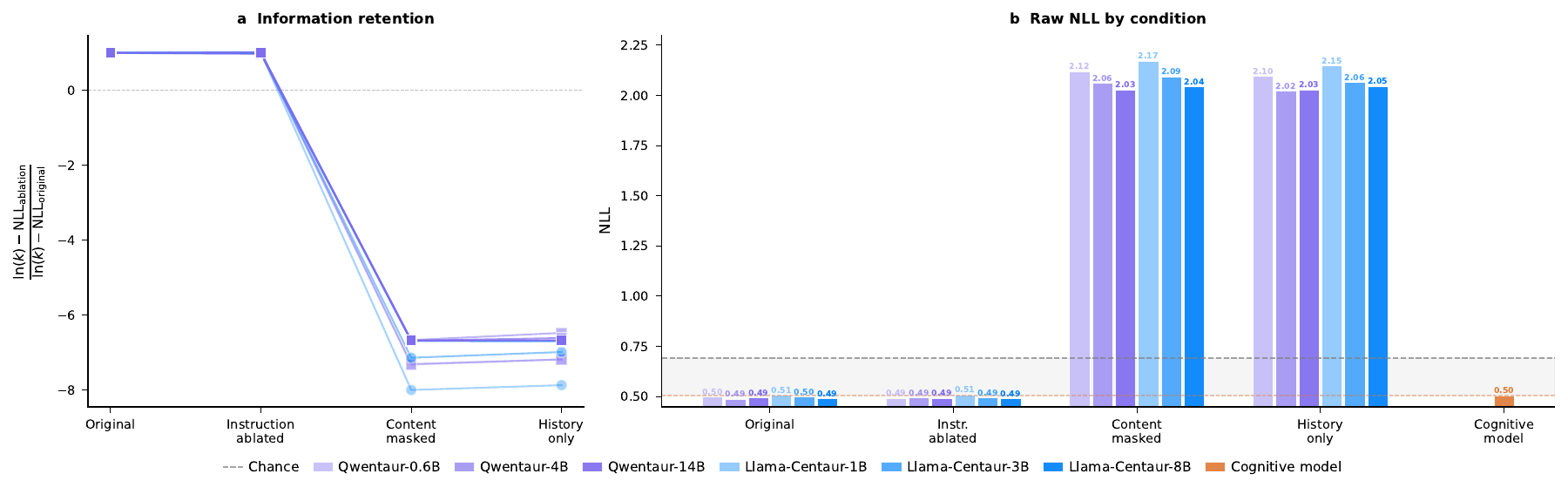}
\vspace{-4pt}
\caption{\textbf{Multi-armed bandits: Prob.\ instrumental learning} \citep{lefebvre2017behavioural}.}
\label{fig:ablation_lefebvre2017behavioural}
\end{figure}

\begin{figure}[ht!]
\centering
\begin{minipage}[t]{0.48\textwidth}
\begin{promptbox}[Original]
\textcolor{colI}{You will be playing a game in which you pretend to be a weather forecaster. In each trial, you will see between one and three tarot cards. Your task is to decide if the combination of cards presented predicts rainy weather (by pressing T) or fine weather (by pressing S). You are an amnesic patient.}\par\smallskip
\textcolor{colS}{You are seeing the following: card 1, card 2.} \textcolor{colC}{You press <<S>>} \textcolor{colF}{. You are correct, the weather is indeed fine.}\par
\textcolor{colS}{You are seeing the following: card 2.} \textcolor{colC}{You press <<S>>} \textcolor{colF}{. You are correct, the weather is indeed fine.}\par
\textcolor{colS}{You are seeing the following: card 4.} \textcolor{colC}{You press <<T>>} \textcolor{colF}{. You are correct, the weather is indeed rainy.}\par

\textcolor{colS}{You are seeing the following: card 1, card 3, card 4.} \textcolor{colC}{You press <<}\textcolor{midgray}{\underline{\phantom{MM}}}
\end{promptbox}
\end{minipage}%
\hfill
\begin{minipage}[t]{0.48\textwidth}
\begin{promptbox}[Instruction-ablated]
\textcolor{colS}{You are seeing the following: card 1, card 2.} \textcolor{colC}{You press <<S>>} \textcolor{colF}{. You are correct, the weather is indeed fine.}\par
\textcolor{colS}{You are seeing the following: card 2.} \textcolor{colC}{You press <<S>>} \textcolor{colF}{. You are correct, the weather is indeed fine.}\par
\textcolor{colS}{You are seeing the following: card 4.} \textcolor{colC}{You press <<T>>} \textcolor{colF}{. You are correct, the weather is indeed rainy.}\par

\textcolor{colS}{You are seeing the following: card 1, card 3, card 4.} \textcolor{colC}{You press <<}\textcolor{midgray}{\underline{\phantom{MM}}}
\end{promptbox}
\end{minipage}
\vspace{2pt}

\begin{minipage}[t]{0.48\textwidth}
\begin{promptbox}[Content-masked]
\textcolor{colI}{You will respond with S or T.}\par
\par\smallskip
\textcolor{colM}{You are seeing some cards.} \textcolor{colC}{You press <<S>>} \textcolor{colM}{. Feedback is given.}\par
\textcolor{colM}{You are seeing some cards.} \textcolor{colC}{You press <<S>>} \textcolor{colM}{. Feedback is given.}\par
\textcolor{colM}{You are seeing some cards.} \textcolor{colC}{You press <<T>>} \textcolor{colM}{. Feedback is given.}\par

\textcolor{colM}{You are seeing some cards.} \textcolor{colC}{You press <<}\textcolor{midgray}{\underline{\phantom{MM}}}
\end{promptbox}
\end{minipage}%
\hfill
\begin{minipage}[t]{0.24\textwidth}
\begin{promptbox}[History-only]
\textcolor{colI}{You will respond with S or T.}\par
\par\smallskip
\textcolor{colC}{You press <<S>>.}\par
\textcolor{colC}{You press <<S>>.}\par
\textcolor{colC}{You press <<T>>.}\par

\textcolor{colC}{You press <<}\textcolor{midgray}{\underline{\phantom{MM}}}
\end{promptbox}
\end{minipage}%
\hfill
\begin{minipage}[t]{0.2\textwidth}
\begin{promptbox}[Choice-only]
\textcolor{colC}{<<S>>}\par
\textcolor{colC}{<<S>>}\par
\textcolor{colC}{<<T>>}\par

\textcolor{colC}{<<}\textcolor{midgray}{\underline{\phantom{MM}}}
\end{promptbox}
\end{minipage}
\vspace{2pt}
\begin{center}
\sffamily\fontsize{6.5}{8}\selectfont\color{darkgray}
\textcolor{colI}{\rule{5pt}{5pt}}\;\,Instruction / $I_\mathrm{min}$\quad
\textcolor{colS}{\rule{5pt}{5pt}}\;\,Stimuli\quad
\textcolor{colF}{\rule{5pt}{5pt}}\;\,Feedback\quad
\textcolor{colC}{\rule{5pt}{5pt}}\;\,Choice\quad
\textcolor{colM}{\rule{5pt}{5pt}}\;\,Masked\quad
\textcolor{midgray}{\underline{\phantom{MM}}}\;\,Model prediction
\end{center}
\vspace{8pt}
\includegraphics[width=\textwidth]{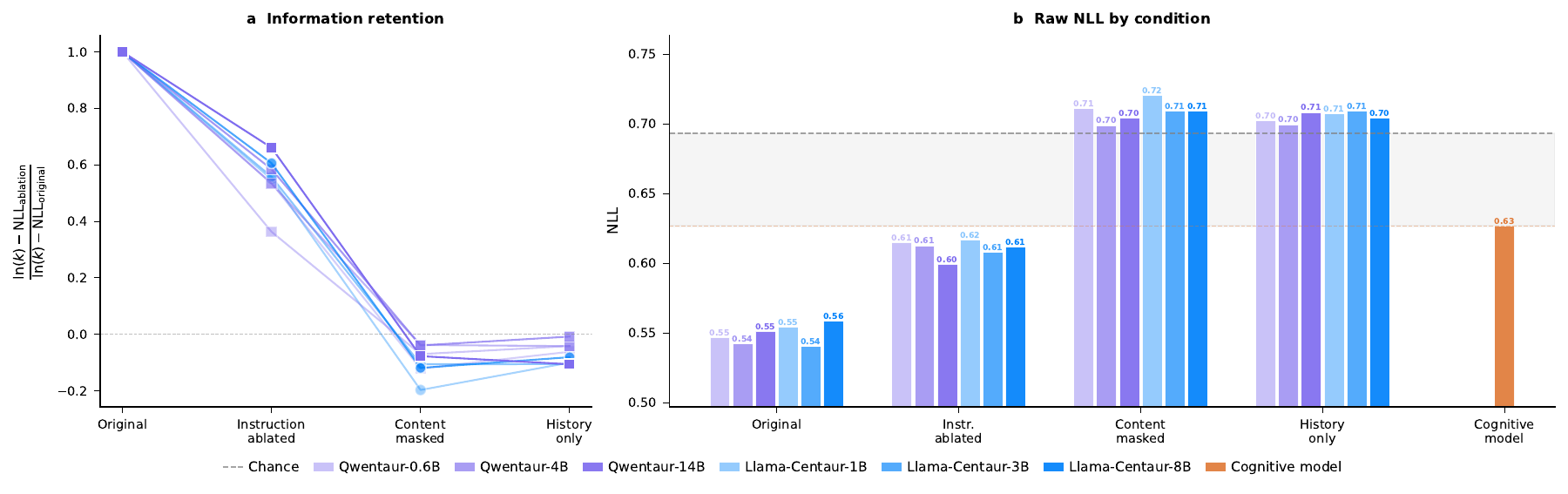}
\vspace{-4pt}
\caption{\textbf{Supervised learning: Weather prediction task} \citep{speekenbrink2008learning}.}
\label{fig:ablation_speekenbrink2008learning}
\end{figure}

\begin{figure}[ht!]
\centering
\begin{minipage}[t]{0.48\textwidth}
\begin{promptbox}[Original]
\textcolor{colI}{You are going to plant trees in two different gardens labeled North and South. The trees look different from each other regarding their leafiness and branchiness. There are 5 levels of leafiness (0, 1, 2, 3, 4) and 5 levels of branchiness (0, 1, 2, 3, 4). In each trial, you will be shown a tree and have to accept or reject it for planting. Press I to accept the tree and J to reject the tree. If you press neither, you press N. Your goal is to earn as many points as possible.}\par\smallskip
\textcolor{colS}{You get a tree with level 4 of leafiness and level 1 of branchiness in the North garden.} \textcolor{colC}{You press <<J>>} \textcolor{colF}{and get 0 points. You would have gotten 50 points, had you accepted to plant the tree.}\par
\textcolor{colS}{You get a tree with level 2 of leafiness and level 1 of branchiness in the South garden.} \textcolor{colC}{You press <<I>>} \textcolor{colF}{and get -25 points.}\par
\textcolor{colS}{You get a tree with level 0 of leafiness and level 0 of branchiness in the North garden.} \textcolor{colC}{You press <<I>>} \textcolor{colF}{and get -50 points.}\par

\textcolor{colS}{You get a tree with level 4 of leafiness and level 4 of branchiness in the South garden.} \textcolor{colC}{You press <<}\textcolor{midgray}{\underline{\phantom{MM}}}
\end{promptbox}
\end{minipage}%
\hfill
\begin{minipage}[t]{0.48\textwidth}
\begin{promptbox}[Instruction-ablated]
\textcolor{colS}{You get a tree with level 4 of leafiness and level 1 of branchiness in the North garden.} \textcolor{colC}{You press <<J>>} \textcolor{colF}{and get 0 points. You would have gotten 50 points, had you accepted to plant the tree.}\par
\textcolor{colS}{You get a tree with level 2 of leafiness and level 1 of branchiness in the South garden.} \textcolor{colC}{You press <<I>>} \textcolor{colF}{and get -25 points.}\par
\textcolor{colS}{You get a tree with level 0 of leafiness and level 0 of branchiness in the North garden.} \textcolor{colC}{You press <<I>>} \textcolor{colF}{and get -50 points.}\par

\textcolor{colS}{You get a tree with level 4 of leafiness and level 4 of branchiness in the South garden.} \textcolor{colC}{You press <<}\textcolor{midgray}{\underline{\phantom{MM}}}
\end{promptbox}
\end{minipage}
\vspace{2pt}

\begin{minipage}[t]{0.48\textwidth}
\begin{promptbox}[Content-masked]
\textcolor{colI}{You will respond with J, I, or N.}\par
\par\smallskip
\textcolor{colM}{You see a tree.} \textcolor{colC}{You press <<J>>} \textcolor{colM}{and get some points. The correct answer is revealed.}\par
\textcolor{colM}{You see a tree.} \textcolor{colC}{You press <<I>>} \textcolor{colM}{and get some points.}\par
\textcolor{colM}{You see a tree.} \textcolor{colC}{You press <<I>>} \textcolor{colM}{and get some points.}\par

\textcolor{colM}{You see a tree.} \textcolor{colC}{You press <<}\textcolor{midgray}{\underline{\phantom{MM}}}
\end{promptbox}
\end{minipage}%
\hfill
\begin{minipage}[t]{0.24\textwidth}
\begin{promptbox}[History-only]
\textcolor{colI}{You will respond with J, I, or N.}\par
\par\smallskip
\textcolor{colC}{You press <<J>>.}\par
\textcolor{colC}{You press <<I>>.}\par
\textcolor{colC}{You press <<I>>.}\par

\textcolor{colC}{You press <<}\textcolor{midgray}{\underline{\phantom{MM}}}
\end{promptbox}
\end{minipage}%
\hfill
\begin{minipage}[t]{0.2\textwidth}
\begin{promptbox}[Choice-only]
\textcolor{colC}{<<J>>}\par
\textcolor{colC}{<<I>>}\par
\textcolor{colC}{<<I>>}\par

\textcolor{colC}{<<}\textcolor{midgray}{\underline{\phantom{MM}}}
\end{promptbox}
\end{minipage}
\vspace{2pt}
\begin{center}
\sffamily\fontsize{6.5}{8}\selectfont\color{darkgray}
\textcolor{colI}{\rule{5pt}{5pt}}\;\,Instruction / $I_\mathrm{min}$\quad
\textcolor{colS}{\rule{5pt}{5pt}}\;\,Stimuli\quad
\textcolor{colF}{\rule{5pt}{5pt}}\;\,Feedback\quad
\textcolor{colC}{\rule{5pt}{5pt}}\;\,Choice\quad
\textcolor{colM}{\rule{5pt}{5pt}}\;\,Masked\quad
\textcolor{midgray}{\underline{\phantom{MM}}}\;\,Model prediction
\end{center}
\vspace{8pt}
\includegraphics[width=\textwidth]{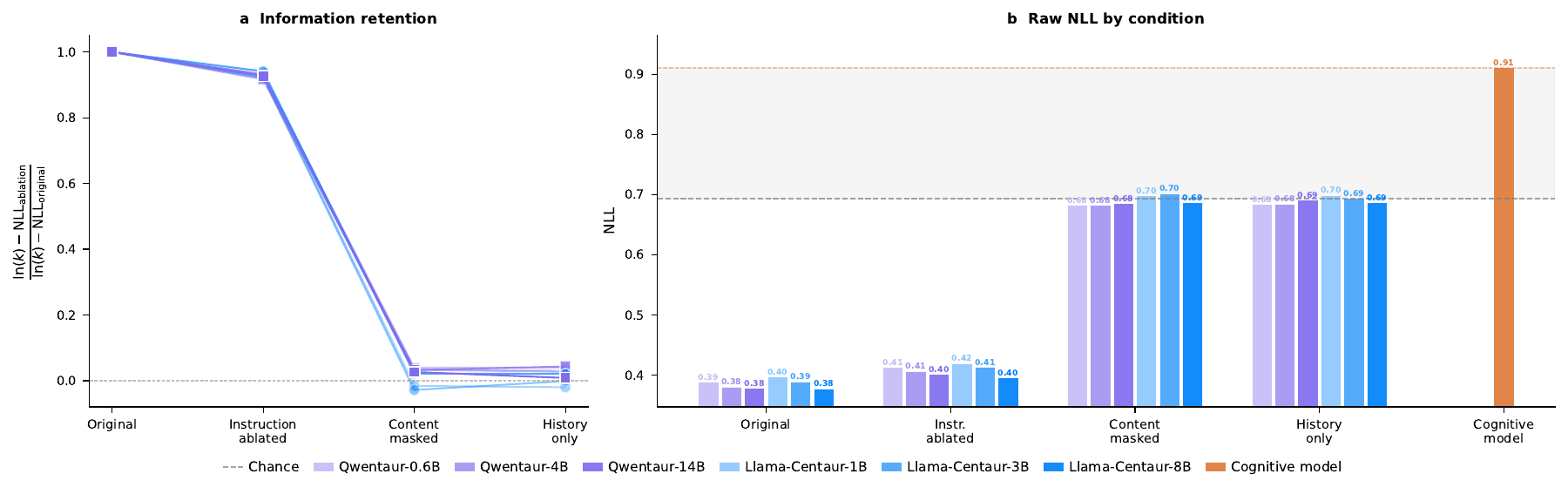}
\vspace{-4pt}
\caption{\textbf{Decision-making: Gardening task} \citep{flesch2018comparing}.}
\label{fig:ablation_flesch2018comparing}
\end{figure}

\begin{figure}[ht!]
\centering
\begin{minipage}[t]{0.48\textwidth}
\begin{promptbox}[Original]
\textcolor{colI}{You are repeatedly presented with two options, labeled P and G. Each option represents a fictitious product and you have to infer which product is superior in terms of quality. Both products are rated by four independent experts. A positive rating is indicated by 1, a negative rating is indicated by 0. The ratings of experts are given in descending order of their validity (having validities of 90\%, 80\%, 70\%, and 60\%).}\par\smallskip
\textcolor{colS}{Product P ratings: [0 0 0 1]. Product G ratings: [1 0 0 0].} \textcolor{colC}{You press <<G>>.}\par
\textcolor{colS}{Product P ratings: [0 0 0 0]. Product G ratings: [1 1 0 0].} \textcolor{colC}{You press <<G>>.}\par
\textcolor{colS}{Product P ratings: [1 1 0 0]. Product G ratings: [0 1 0 1].} \textcolor{colC}{You press <<P>>.}\par

\textcolor{colS}{Product P ratings: [1 0 1 0]. Product G ratings: [0 1 1 0].} \textcolor{colC}{You press <<}\textcolor{midgray}{\underline{\phantom{MM}}}
\end{promptbox}
\end{minipage}%
\hfill
\begin{minipage}[t]{0.48\textwidth}
\begin{promptbox}[Instruction-ablated]
\textcolor{colS}{Product P ratings: [0 0 0 1]. Product G ratings: [1 0 0 0].} \textcolor{colC}{You press <<G>>.}\par
\textcolor{colS}{Product P ratings: [0 0 0 0]. Product G ratings: [1 1 0 0].} \textcolor{colC}{You press <<G>>.}\par
\textcolor{colS}{Product P ratings: [1 1 0 0]. Product G ratings: [0 1 0 1].} \textcolor{colC}{You press <<P>>.}\par

\textcolor{colS}{Product P ratings: [1 0 1 0]. Product G ratings: [0 1 1 0].} \textcolor{colC}{You press <<}\textcolor{midgray}{\underline{\phantom{MM}}}
\end{promptbox}
\end{minipage}
\vspace{2pt}

\begin{minipage}[t]{0.48\textwidth}
\begin{promptbox}[Content-masked]
\textcolor{colI}{You will respond with G or P.}\par
\par\smallskip
\textcolor{colM}{Product P ratings: [some ratings]. Product G ratings: [some ratings].} \textcolor{colC}{You press <<G>>.}\par
\textcolor{colM}{Product P ratings: [some ratings]. Product G ratings: [some ratings].} \textcolor{colC}{You press <<G>>.}\par
\textcolor{colM}{Product P ratings: [some ratings]. Product G ratings: [some ratings].} \textcolor{colC}{You press <<P>>.}\par

\textcolor{colM}{Product P ratings: [some ratings]. Product G ratings: [some ratings].} \textcolor{colC}{You press <<}\textcolor{midgray}{\underline{\phantom{MM}}}
\end{promptbox}
\end{minipage}%
\hfill
\begin{minipage}[t]{0.24\textwidth}
\begin{promptbox}[History-only]
\textcolor{colI}{You will respond with G or P.}\par
\par\smallskip
\textcolor{colC}{You press <<G>>.}\par
\textcolor{colC}{You press <<G>>.}\par
\textcolor{colC}{You press <<P>>.}\par

\textcolor{colC}{You press <<}\textcolor{midgray}{\underline{\phantom{MM}}}
\end{promptbox}
\end{minipage}%
\hfill
\begin{minipage}[t]{0.2\textwidth}
\begin{promptbox}[Choice-only]
\textcolor{colC}{<<G>>}\par
\textcolor{colC}{<<G>>}\par
\textcolor{colC}{<<P>>}\par

\textcolor{colC}{<<}\textcolor{midgray}{\underline{\phantom{MM}}}
\end{promptbox}
\end{minipage}
\vspace{2pt}
\begin{center}
\sffamily\fontsize{6.5}{8}\selectfont\color{darkgray}
\textcolor{colI}{\rule{5pt}{5pt}}\;\,Instruction / $I_\mathrm{min}$\quad
\textcolor{colS}{\rule{5pt}{5pt}}\;\,Stimuli\quad
\textcolor{colF}{\rule{5pt}{5pt}}\;\,Feedback\quad
\textcolor{colC}{\rule{5pt}{5pt}}\;\,Choice\quad
\textcolor{colM}{\rule{5pt}{5pt}}\;\,Masked\quad
\textcolor{midgray}{\underline{\phantom{MM}}}\;\,Model prediction
\end{center}
\vspace{8pt}
\includegraphics[width=\textwidth]{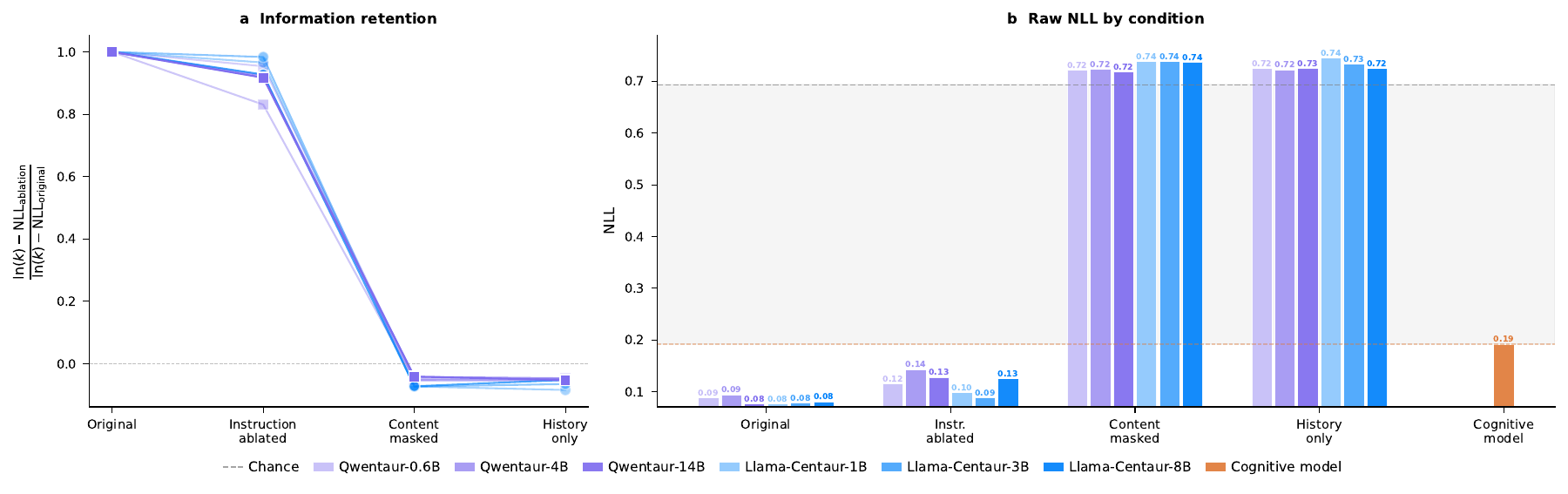}
\vspace{-4pt}
\caption{\textbf{Decision-making: Multi-attribute decision} \citep{hilbig2014generalized}.}
\label{fig:ablation_hilbig2014generalized}
\end{figure}

Five of the nine non-monotonic cases are not substantive. All five invert only at the content-masked~$\to$~history-only step, with both conditions already at or below chance, so the ordering is between two states in which the model has nothing left to lose. In four of them the inversion is smaller than $0.03$ retention units, well inside the spread across models (Figures~\ref{fig:ablation_badham2017deficits}, \ref{fig:ablation_speekenbrink2008learning}, \ref{fig:ablation_flesch2018comparing}, \ref{fig:ablation_hilbig2014generalized}). The fifth, probabilistic instrumental learning (Figure~\ref{fig:ablation_lefebvre2017behavioural}), is an artefact of the normalisation: $R$ divides by the headroom $\ln k - \mathrm{NLL}_{\mathrm{original}}$, which is only $0.197$ nats here, so a difference of a few hundredths of a nat is inflated into a large difference in $R$.

\begin{figure}[ht!]
\centering
\begin{minipage}[t]{0.48\textwidth}
\begin{promptbox}[Original]
\textcolor{colI}{You see in front of you four decks of cards labeled E, Z, S, and O. You get a loan of 2000\$ of play money. You have to select one card at a time, from any of the four decks, for 100 trials. You select a card from a deck by pressing the corresponding key. After turning a card, you win some money, the amount varies with the deck. You sometimes also have to pay a penalty, which also varies with the deck. Your goal is to maximize profit on the loan of the play money.}\par\smallskip
\textcolor{colC}{You press <<O>>} \textcolor{colF}{. You win 50.0\$ and lose 0.0\$.}\par
\textcolor{colC}{You press <<Z>>} \textcolor{colF}{. You win 100.0\$ and lose 0.0\$.}\par
\textcolor{colC}{You press <<S>>} \textcolor{colF}{. You win 50.0\$ and lose 50.0\$.}\par

\textcolor{colC}{You press <<}\textcolor{midgray}{\underline{\phantom{MM}}}
\end{promptbox}
\end{minipage}%
\hfill
\begin{minipage}[t]{0.48\textwidth}
\begin{promptbox}[Instruction-ablated]
\textcolor{colC}{You press <<O>>} \textcolor{colF}{. You win 50.0\$ and lose 0.0\$.}\par
\textcolor{colC}{You press <<Z>>} \textcolor{colF}{. You win 100.0\$ and lose 0.0\$.}\par
\textcolor{colC}{You press <<S>>} \textcolor{colF}{. You win 50.0\$ and lose 50.0\$.}\par

\textcolor{colC}{You press <<}\textcolor{midgray}{\underline{\phantom{MM}}}
\end{promptbox}
\end{minipage}
\vspace{2pt}

\begin{minipage}[t]{0.48\textwidth}
\begin{promptbox}[Content-masked]
\textcolor{colI}{You will respond with O, Z, S, or E.}\par
\par\smallskip
\textcolor{colC}{You press <<O>>} \textcolor{colM}{. You win some money and lose some money.}\par
\textcolor{colC}{You press <<Z>>} \textcolor{colM}{. You win some money and lose some money.}\par
\textcolor{colC}{You press <<S>>} \textcolor{colM}{. You win some money and lose some money.}\par

\textcolor{colC}{You press <<}\textcolor{midgray}{\underline{\phantom{MM}}}
\end{promptbox}
\end{minipage}%
\hfill
\begin{minipage}[t]{0.24\textwidth}
\begin{promptbox}[History-only]
\textcolor{colI}{You will respond with O, Z, S, or E.}\par
\par\smallskip
\textcolor{colC}{You press <<O>>.}\par
\textcolor{colC}{You press <<Z>>.}\par
\textcolor{colC}{You press <<S>>.}\par

\textcolor{colC}{You press <<}\textcolor{midgray}{\underline{\phantom{MM}}}
\end{promptbox}
\end{minipage}%
\hfill
\begin{minipage}[t]{0.2\textwidth}
\begin{promptbox}[Choice-only]
\textcolor{colC}{<<O>>}\par
\textcolor{colC}{<<Z>>}\par
\textcolor{colC}{<<S>>}\par

\textcolor{colC}{<<}\textcolor{midgray}{\underline{\phantom{MM}}}
\end{promptbox}
\end{minipage}
\vspace{2pt}
\begin{center}
\sffamily\fontsize{6.5}{8}\selectfont\color{darkgray}
\textcolor{colI}{\rule{5pt}{5pt}}\;\,Instruction / $I_\mathrm{min}$\quad
\textcolor{colS}{\rule{5pt}{5pt}}\;\,Stimuli\quad
\textcolor{colF}{\rule{5pt}{5pt}}\;\,Feedback\quad
\textcolor{colC}{\rule{5pt}{5pt}}\;\,Choice\quad
\textcolor{colM}{\rule{5pt}{5pt}}\;\,Masked\quad
\textcolor{midgray}{\underline{\phantom{MM}}}\;\,Model prediction
\end{center}
\vspace{8pt}
\includegraphics[width=\textwidth]{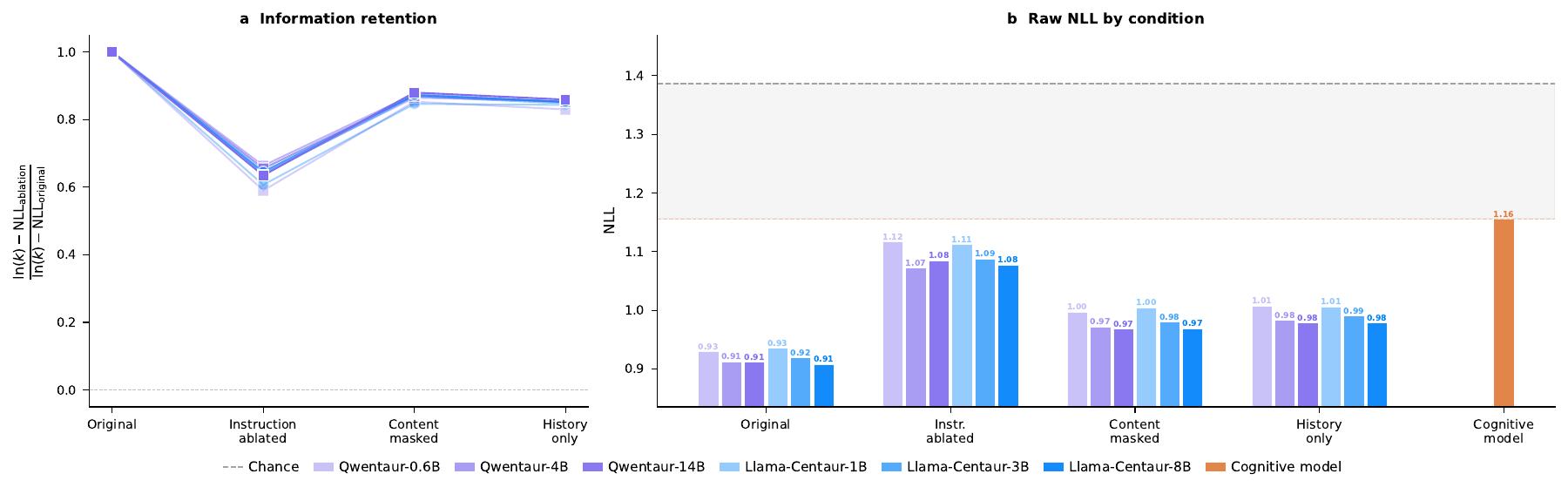}
\vspace{-4pt}
\caption{\textbf{Multi-armed bandits: Iowa gambling task} \citep{steingroever2015data}.}
\label{fig:ablation_steingroever2015data}
\end{figure}

\begin{figure}[ht!]
\centering
\begin{minipage}[t]{0.48\textwidth}
\begin{promptbox}[Original]
\textcolor{colI}{You will choose from two monetary lotteries by pressing O or I. The lotteries offer different points with different probabilities. Your choice will trigger a random draw from the chosen lottery that will be added to your bonus. Your goal is to maximize your bonus. You will be presented with multiple choice problems consisting of different lotteries varying in outcomes and probabilities.}\par\smallskip
\textcolor{colS}{Lottery O offers 55.0 points with 80.0\% probability or 0.0 points with 20.0\% probability.}\par
\textcolor{colS}{Lottery I offers 50.0 points with 100.0\% probability.}\par
\textcolor{colC}{You press <<I>>} \textcolor{colF}{.}\par
\textcolor{colC}{You press <<I>>} \textcolor{colF}{.}\par
\textcolor{colC}{You press <<I>>} \textcolor{colF}{.}\par

\textcolor{colC}{You press <<}\textcolor{midgray}{\underline{\phantom{MM}}}
\end{promptbox}
\end{minipage}%
\hfill
\begin{minipage}[t]{0.48\textwidth}
\begin{promptbox}[Instruction-ablated]
\textcolor{colS}{Lottery O offers 55.0 points with 80.0\% probability or 0.0 points with 20.0\% probability.}\par
\textcolor{colS}{Lottery I offers 50.0 points with 100.0\% probability.}\par
\textcolor{colC}{You press <<I>>} \textcolor{colF}{.}\par
\textcolor{colC}{You press <<I>>} \textcolor{colF}{.}\par
\textcolor{colC}{You press <<I>>} \textcolor{colF}{.}\par

\textcolor{colC}{You press <<}\textcolor{midgray}{\underline{\phantom{MM}}}
\end{promptbox}
\end{minipage}
\vspace{2pt}

\begin{minipage}[t]{0.48\textwidth}
\begin{promptbox}[Content-masked]
\textcolor{colI}{You will respond with I or O.}\par
\par\smallskip
\textcolor{colM}{Lottery O has some outcomes.}\par
\textcolor{colM}{Lottery I has some outcomes.}\par
\textcolor{colC}{You press <<I>>} \textcolor{colM}{.}\par
\textcolor{colC}{You press <<I>>} \textcolor{colM}{.}\par
\textcolor{colC}{You press <<I>>} \textcolor{colM}{.}\par

\textcolor{colC}{You press <<}\textcolor{midgray}{\underline{\phantom{MM}}}
\end{promptbox}
\end{minipage}%
\hfill
\begin{minipage}[t]{0.24\textwidth}
\begin{promptbox}[History-only]
\textcolor{colI}{You will respond with I or O.}\par
\par\smallskip
\textcolor{colC}{You press <<I>>.}\par
\textcolor{colC}{You press <<I>>.}\par
\textcolor{colC}{You press <<I>>.}\par

\textcolor{colC}{You press <<}\textcolor{midgray}{\underline{\phantom{MM}}}
\end{promptbox}
\end{minipage}%
\hfill
\begin{minipage}[t]{0.2\textwidth}
\begin{promptbox}[Choice-only]
\textcolor{colC}{<<I>>}\par
\textcolor{colC}{<<I>>}\par
\textcolor{colC}{<<I>>}\par

\textcolor{colC}{<<}\textcolor{midgray}{\underline{\phantom{MM}}}
\end{promptbox}
\end{minipage}
\vspace{2pt}
\begin{center}
\sffamily\fontsize{6.5}{8}\selectfont\color{darkgray}
\textcolor{colI}{\rule{5pt}{5pt}}\;\,Instruction / $I_\mathrm{min}$\quad
\textcolor{colS}{\rule{5pt}{5pt}}\;\,Stimuli\quad
\textcolor{colF}{\rule{5pt}{5pt}}\;\,Feedback\quad
\textcolor{colC}{\rule{5pt}{5pt}}\;\,Choice\quad
\textcolor{colM}{\rule{5pt}{5pt}}\;\,Masked\quad
\textcolor{midgray}{\underline{\phantom{MM}}}\;\,Model prediction
\end{center}
\vspace{8pt}
\includegraphics[width=\textwidth]{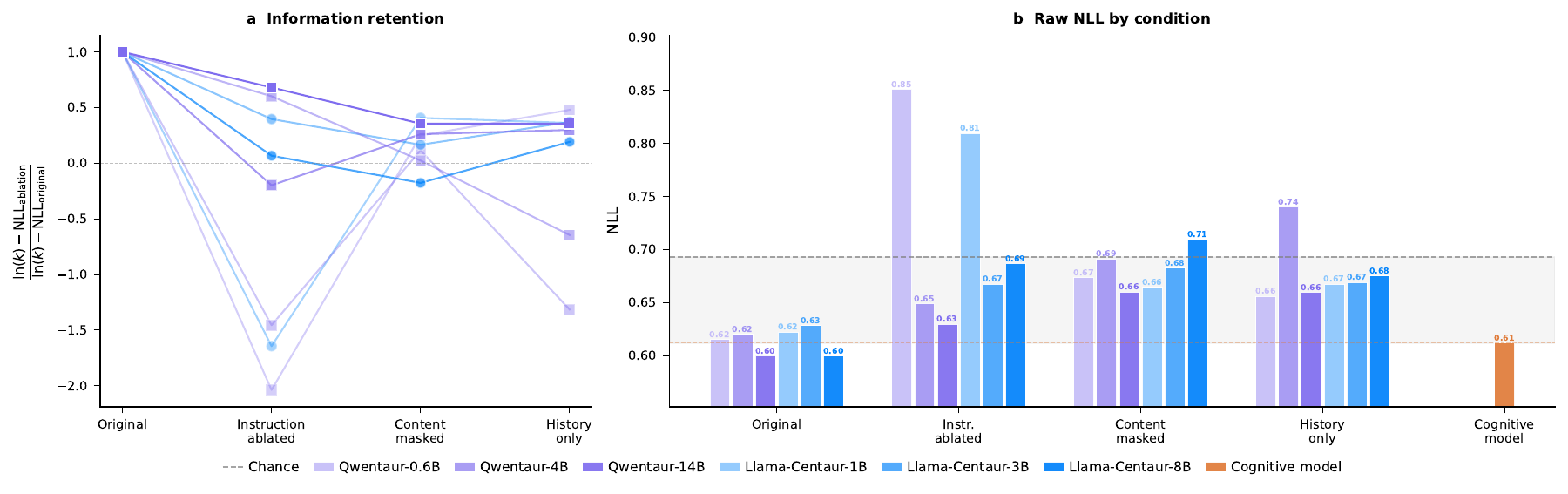}
\vspace{-4pt}
\caption{\textbf{Decision-making: Decisions from description} \citep{wulff2018sampling}.}
\label{fig:ablation_wulff2018description}
\end{figure}

\begin{figure}[ht!]
\centering
\begin{minipage}[t]{0.48\textwidth}
\begin{promptbox}[Original]
\textcolor{colI}{You are participating in multiple games involving two slot machines, labeled T and E. The two slot machines are different in different games. Each time you choose a slot machine, you get points (choosing the same slot machine will not always give you the same points). You select a slot machine by pressing the corresponding key. The expected points change randomly, abruptly, and independently with a hazard rate (which you will be told). When the points change, the new expected point value assigned to that slot machine is sampled from a uniform distribution (from 1 to 99 points). For example, if the hazard rate is 0.1, the expected points of the machines change with 10\%. Your goal is to choose the slot machine that will give you the most points.}\par\smallskip
\textcolor{colI}{Game 1.}\par
\textcolor{colS}{The hazard rate is 0.2.}\par
\textcolor{colI}{There are 100 trials in this game.}\par
\textcolor{colC}{You press <<T>>} \textcolor{colF}{and get 57 points.}\par
\textcolor{colC}{You press <<E>>} \textcolor{colF}{and get 41 points.}\par
\textcolor{colC}{You press <<T>>} \textcolor{colF}{and get 8 points.}\par

\textcolor{colC}{You press <<}\textcolor{midgray}{\underline{\phantom{MM}}}
\end{promptbox}
\end{minipage}%
\hfill
\begin{minipage}[t]{0.48\textwidth}
\begin{promptbox}[Instruction-ablated]
\textcolor{colI}{Game 1.}\par
\textcolor{colS}{The hazard rate is 0.2.}\par
\textcolor{colC}{You press <<T>>} \textcolor{colF}{and get 57 points.}\par
\textcolor{colC}{You press <<E>>} \textcolor{colF}{and get 41 points.}\par
\textcolor{colC}{You press <<T>>} \textcolor{colF}{and get 8 points.}\par

\textcolor{colC}{You press <<}\textcolor{midgray}{\underline{\phantom{MM}}}
\end{promptbox}
\end{minipage}
\vspace{2pt}

\begin{minipage}[t]{0.48\textwidth}
\begin{promptbox}[Content-masked]
\textcolor{colI}{You will respond with T or E.}\par
\textcolor{colM}{Game 1.}\par
\textcolor{colM}{The hazard rate is some probability.}\par
\textcolor{colM}{There are some trials in this game.}\par
\par\smallskip
\textcolor{colC}{You press <<T>>} \textcolor{colM}{and get some points.}\par
\textcolor{colC}{You press <<E>>} \textcolor{colM}{and get some points.}\par
\textcolor{colC}{You press <<T>>} \textcolor{colM}{and get some points.}\par

\textcolor{colC}{You press <<}\textcolor{midgray}{\underline{\phantom{MM}}}
\end{promptbox}
\end{minipage}%
\hfill
\begin{minipage}[t]{0.24\textwidth}
\begin{promptbox}[History-only]
\textcolor{colI}{You will respond with T or E.}\par
\par\smallskip
\textcolor{colC}{You press <<T>>.}\par
\textcolor{colC}{You press <<E>>.}\par
\textcolor{colC}{You press <<T>>.}\par

\textcolor{colC}{You press <<}\textcolor{midgray}{\underline{\phantom{MM}}}
\end{promptbox}
\end{minipage}%
\hfill
\begin{minipage}[t]{0.2\textwidth}
\begin{promptbox}[Choice-only]
\textcolor{colC}{<<T>>}\par
\textcolor{colC}{<<E>>}\par
\textcolor{colC}{<<T>>}\par

\textcolor{colC}{<<}\textcolor{midgray}{\underline{\phantom{MM}}}
\end{promptbox}
\end{minipage}
\vspace{2pt}
\begin{center}
\sffamily\fontsize{6.5}{8}\selectfont\color{darkgray}
\textcolor{colI}{\rule{5pt}{5pt}}\;\,Instruction / $I_\mathrm{min}$\quad
\textcolor{colS}{\rule{5pt}{5pt}}\;\,Stimuli\quad
\textcolor{colF}{\rule{5pt}{5pt}}\;\,Feedback\quad
\textcolor{colC}{\rule{5pt}{5pt}}\;\,Choice\quad
\textcolor{colM}{\rule{5pt}{5pt}}\;\,Masked\quad
\textcolor{midgray}{\underline{\phantom{MM}}}\;\,Model prediction
\end{center}
\vspace{8pt}
\includegraphics[width=\textwidth]{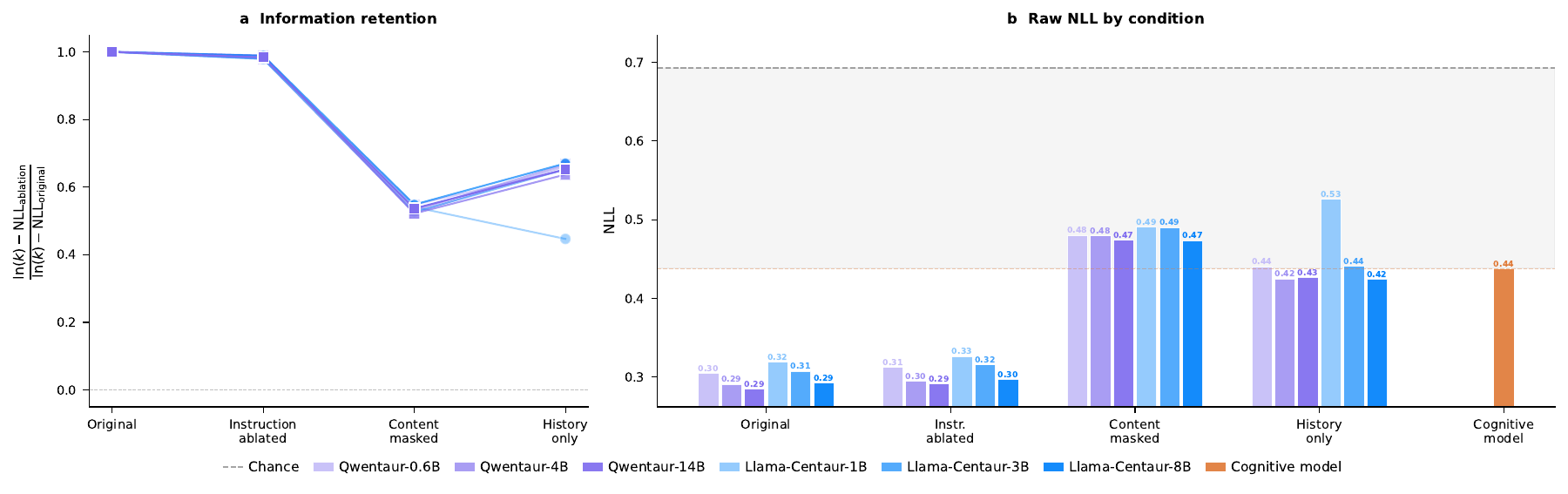}
\vspace{-4pt}
\caption{\textbf{Multi-armed bandits: Changing bandit} \citep{xiong2023neural}.}
\label{fig:ablation_xiong2023neural}
\end{figure}

\begin{figure}[ht!]
\centering
\begin{minipage}[t]{0.48\textwidth}
\begin{promptbox}[Original]
\textcolor{colI}{Imagine that you are a tourist and you have to navigate the subway network. You can go north by pressing X, west by pressing P, south by pressing N, and east by pressing E. Once you think you are at the goal station, press Y to confirm.}\par\smallskip
\textcolor{colI}{The new starting station is 4 and the goal station is 6.}\par
\textcolor{colS}{Your station: 4. Neighboring stations: 3 on the north, circle on the east, 5 on the south, and circle on the west.} \textcolor{colC}{You press <<N>>.}\par
\textcolor{colS}{Your station: 5. Neighboring stations: 4 on the north, circle on the east, circle on the south, and 6 on the west.} \textcolor{colC}{You press <<P>>.}\par
\textcolor{colF}{You are successful.}\par
\textcolor{colI}{The new starting station is 8 and the goal station is 3.}\par
\textcolor{colS}{Your station: 8. Neighboring stations: 7 on the north, 9 on the east, circle on the south, and circle on the west.} \textcolor{colC}{You press <<}\textcolor{midgray}{\underline{\phantom{MM}}}
\end{promptbox}
\end{minipage}%
\hfill
\begin{minipage}[t]{0.48\textwidth}
\begin{promptbox}[Instruction-ablated]
\textcolor{colI}{The new starting station is 4 and the goal station is 6.}\par
\textcolor{colS}{Your station: 4. Neighboring stations: 3 on the north, circle on the east, 5 on the south, and circle on the west.} \textcolor{colC}{You press <<N>>.}\par
\textcolor{colS}{Your station: 5. Neighboring stations: 4 on the north, circle on the east, circle on the south, and 6 on the west.} \textcolor{colC}{You press <<P>>.}\par
\textcolor{colF}{You are successful.}\par
\textcolor{colI}{The new starting station is 8 and the goal station is 3.}\par
\textcolor{colS}{Your station: 8. Neighboring stations: 7 on the north, 9 on the east, circle on the south, and circle on the west.} \textcolor{colC}{You press <<}\textcolor{midgray}{\underline{\phantom{MM}}}
\end{promptbox}
\end{minipage}
\vspace{2pt}

\begin{minipage}[t]{0.48\textwidth}
\begin{promptbox}[Content-masked]
\textcolor{colI}{You will respond with P, Y, E, X, or N.}\par
\par\smallskip
\textcolor{colM}{A new round begins.}\par
\textcolor{colM}{You are at a station.} \textcolor{colC}{You press <<N>>.}\par
\textcolor{colM}{You are at a station.} \textcolor{colC}{You press <<P>>.}\par
\textcolor{colM}{An outcome is reached.}\par
\textcolor{colM}{A new round begins.}\par
\textcolor{colM}{You are at a station.} \textcolor{colC}{You press <<}\textcolor{midgray}{\underline{\phantom{MM}}}
\end{promptbox}
\end{minipage}%
\hfill
\begin{minipage}[t]{0.24\textwidth}
\begin{promptbox}[History-only]
\textcolor{colI}{You will respond with P, Y, E, X, or N.}\par
\par\smallskip
\textcolor{colC}{You press <<N>>.}\par
\textcolor{colC}{You press <<P>>.}\par
\textcolor{colC}{You press <<X>>.}\par

\textcolor{colC}{You press <<}\textcolor{midgray}{\underline{\phantom{MM}}}
\end{promptbox}
\end{minipage}%
\hfill
\begin{minipage}[t]{0.2\textwidth}
\begin{promptbox}[Choice-only]
\textcolor{colC}{<<N>>}\par
\textcolor{colC}{<<P>>}\par
\textcolor{colC}{<<X>>}\par

\textcolor{colC}{<<}\textcolor{midgray}{\underline{\phantom{MM}}}
\end{promptbox}
\end{minipage}
\vspace{2pt}
\begin{center}
\sffamily\fontsize{6.5}{8}\selectfont\color{darkgray}
\textcolor{colI}{\rule{5pt}{5pt}}\;\,Instruction / $I_\mathrm{min}$\quad
\textcolor{colS}{\rule{5pt}{5pt}}\;\,Stimuli\quad
\textcolor{colF}{\rule{5pt}{5pt}}\;\,Feedback\quad
\textcolor{colC}{\rule{5pt}{5pt}}\;\,Choice\quad
\textcolor{colM}{\rule{5pt}{5pt}}\;\,Masked\quad
\textcolor{midgray}{\underline{\phantom{MM}}}\;\,Model prediction
\end{center}
\vspace{8pt}
\includegraphics[width=\textwidth]{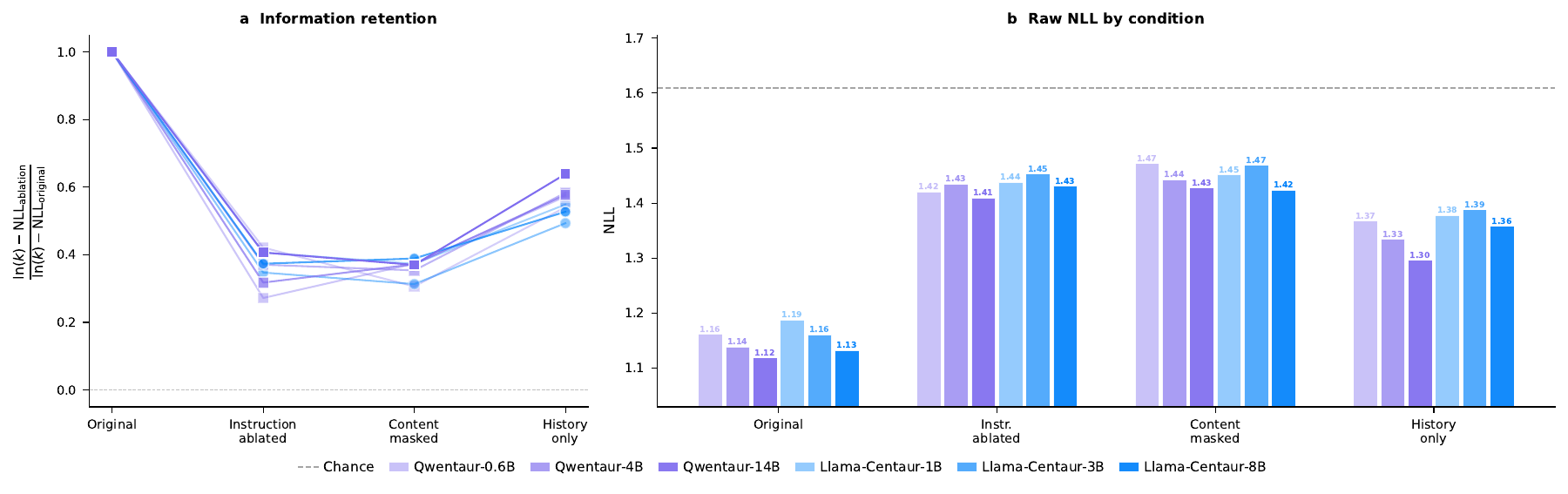}
\vspace{-4pt}
\caption{\textbf{MDP: Subway navigation} \citep{tomov2020discovery}.}
\label{fig:ablation_tomov2020discovery}
\end{figure}

The remaining four are genuine reversals, and both pairs point the same way. In two experiments (Figures~\ref{fig:ablation_xiong2023neural}, \ref{fig:ablation_tomov2020discovery}) the models predict better with the trials removed entirely than with their content masked ($R=0.42$ vs.\ $0.20$ and $0.57$ vs.\ $0.37$, in 8/9 and 9/9 models). In two more (Figures~\ref{fig:ablation_steingroever2015data}, \ref{fig:ablation_wulff2018description}) they predict better with the content masked than with only the instruction removed ($R=0.87$ vs.\ $0.63$ and $0.16$ vs.\ $-0.31$, in 9/9 models each). In both pairs the condition that leaves \textit{more} text in the prompt is the worse one, which is what one expects if a placeholder such as ``some points'' is read as a genuine stimulus value, or if full stimulus text without its instruction is interpreted under the wrong task. Partial information appears to be worse than none, so these reversals do not weaken the hierarchy in the aggregate result. 

\newpage

\begin{sidewaystable}[t!]
\centering
{\tiny
\setlength{\tabcolsep}{2pt}
\renewcommand{\arraystretch}{1.15}
\begin{tabular}{@{}l l  r r r r  r r r r  r r r r  r r r r  r r r r  r r r r  r r r r  r r r r  r@{}}
\toprule
& & \multicolumn{20}{c}{\texttt{\textbf{Qwentaur}}} & \multicolumn{12}{c}{\texttt{\textbf{Llama-Centaur}}} & \textbf{Chance} \\
\cmidrule(lr){3-22}\cmidrule(lr){23-34}
& & \multicolumn{4}{c}{\texttt{0.6B}} & \multicolumn{4}{c}{\texttt{1.7B}} & \multicolumn{4}{c}{\texttt{4B}} & \multicolumn{4}{c}{\texttt{8B}} & \multicolumn{4}{c}{\texttt{14B}} & \multicolumn{4}{c}{\texttt{1B}} & \multicolumn{4}{c}{\texttt{3B}} & \multicolumn{4}{c}{\texttt{8B}} & \\
\cmidrule(lr){3-6} \cmidrule(lr){7-10} \cmidrule(lr){11-14} \cmidrule(lr){15-18} \cmidrule(lr){19-22} \cmidrule(lr){23-26} \cmidrule(lr){27-30} \cmidrule(lr){31-34}
\textbf{Experiment} & \textbf{Type} & {\fontsize{4}{5}\selectfont orig} & {\fontsize{4}{5}\selectfont inst} & {\fontsize{4}{5}\selectfont cont} & {\fontsize{4}{5}\selectfont hist} & {\fontsize{4}{5}\selectfont orig} & {\fontsize{4}{5}\selectfont inst} & {\fontsize{4}{5}\selectfont cont} & {\fontsize{4}{5}\selectfont hist} & {\fontsize{4}{5}\selectfont orig} & {\fontsize{4}{5}\selectfont inst} & {\fontsize{4}{5}\selectfont cont} & {\fontsize{4}{5}\selectfont hist} & {\fontsize{4}{5}\selectfont orig} & {\fontsize{4}{5}\selectfont inst} & {\fontsize{4}{5}\selectfont cont} & {\fontsize{4}{5}\selectfont hist} & {\fontsize{4}{5}\selectfont orig} & {\fontsize{4}{5}\selectfont inst} & {\fontsize{4}{5}\selectfont cont} & {\fontsize{4}{5}\selectfont hist} & {\fontsize{4}{5}\selectfont orig} & {\fontsize{4}{5}\selectfont inst} & {\fontsize{4}{5}\selectfont cont} & {\fontsize{4}{5}\selectfont hist} & {\fontsize{4}{5}\selectfont orig} & {\fontsize{4}{5}\selectfont inst} & {\fontsize{4}{5}\selectfont cont} & {\fontsize{4}{5}\selectfont hist} & {\fontsize{4}{5}\selectfont orig} & {\fontsize{4}{5}\selectfont inst} & {\fontsize{4}{5}\selectfont cont} & {\fontsize{4}{5}\selectfont hist} & $\ln(k)$ \\
\midrule
Gardening task {\fontsize{4}{5}\selectfont\citep{flesch2018comparing}} & Decision & \hc{0.39} & \hc{0.41} & \hc{0.68} & \hc{0.68} & \hc{0.38} & \hc{0.41} & \hc{0.68} & \hc{0.68} & \hc{0.38} & \hc{0.41} & \hc{0.68} & \hc{0.68} & \hc{0.38} & \hc{0.40} & \hc{0.68} & \hc{0.68} & \hcu{0.38} & \hc{0.40} & \hc{0.68} & \hc{0.69} & \hc{0.40} & \hc{0.42} & \hc{0.70} & \hc{0.70} & \hc{0.39} & \hc{0.41} & \hc{0.70} & \hc{0.69} & \hcbu{0.38} & \hc{0.40} & \hc{0.69} & \hc{0.69} & \hc{0.69} \\
Columbia card task {\fontsize{4}{5}\selectfont\citep{frey2017risk}} & Decision & \hc{0.20} & \hc{0.24} & \hc{0.28} & \hc{0.30} & \hc{0.20} & \hc{0.24} & \hc{0.28} & \hc{0.30} & \hc{0.20} & \hc{0.24} & \hc{0.29} & \hc{0.30} & \hc{0.20} & \hc{0.24} & \hc{0.28} & \hc{0.29} & \hcu{0.20} & \hc{0.24} & \hc{0.29} & \hc{0.29} & \hc{0.21} & \hc{0.24} & \hc{0.28} & \hc{0.30} & \hc{0.20} & \hc{0.24} & \hc{0.28} & \hc{0.29} & \hcbu{0.20} & \hc{0.23} & \hc{0.28} & \hc{0.29} & \hc{0.69} \\
Experiential-symbolic task {\fontsize{4}{5}\selectfont\citep{garcia2023experiential}} & Decision & \hc{0.48} & \hc{0.48} & \hc{0.91} & \hc{1.54} & \hc{0.47} & \hc{0.48} & \hc{0.89} & \hc{1.51} & \hc{0.46} & \hc{0.47} & \hc{0.86} & \hc{1.49} & \hc{0.46} & \hc{0.47} & \hc{0.86} & \hc{1.47} & \hcbu{0.46} & \hc{0.47} & \hc{0.84} & \hc{1.47} & \hc{0.47} & \hc{0.47} & \hc{0.89} & \hc{1.56} & \hc{0.47} & \hc{0.47} & \hc{0.87} & \hc{1.50} & \hcu{0.46} & \hc{0.47} & \hc{0.85} & \hc{1.47} & \hcn \\
Multi-attribute DM {\fontsize{4}{5}\selectfont\citep{hilbig2014generalized}} & Decision & \hc{0.09} & \hc{0.12} & \hc{0.72} & \hc{0.72} & \hc{0.08} & \hc{0.19} & \hc{0.73} & \hc{0.73} & \hc{0.09} & \hc{0.14} & \hc{0.72} & \hc{0.72} & \hc{0.08} & \hc{0.13} & \hc{0.72} & \hc{0.72} & \hcu{0.08} & \hc{0.13} & \hc{0.72} & \hc{0.73} & \hcbu{0.08} & \hc{0.10} & \hc{0.74} & \hc{0.74} & \hc{0.08} & \hc{0.09} & \hc{0.74} & \hc{0.73} & \hc{0.08} & \hc{0.13} & \hc{0.74} & \hc{0.72} & \hc{0.69} \\
Risky choice {\fontsize{4}{5}\selectfont\citep{krueger2024identifying}} & Decision & \hc{0.48} & \hc{0.70} & \hc{0.91} & \hc{1.03} & \hc{0.47} & \hc{0.69} & \hc{0.87} & \hc{0.99} & \hc{0.44} & \hc{0.65} & \hc{0.83} & \hc{0.96} & \hc{0.43} & \hc{0.66} & \hc{0.84} & \hc{0.93} & \hcbu{0.42} & \hc{0.65} & \hc{0.81} & \hc{0.92} & \hc{0.49} & \hc{0.70} & \hc{0.89} & \hc{1.06} & \hc{0.46} & \hc{0.65} & \hc{0.84} & \hc{0.98} & \hcu{0.43} & \hc{0.64} & \hc{0.82} & \hc{0.93} & \hcn \\
choices13k {\fontsize{4}{5}\selectfont\citep{peterson2021using}} & Decision & \hc{0.44} & \hc{0.48} & \hc{0.51} & \hc{0.55} & \hc{0.44} & \hc{0.46} & \hc{0.51} & \hc{0.57} & \hc{0.43} & \hc{0.44} & \hc{0.51} & \hc{0.57} & \hc{0.43} & \hc{0.44} & \hc{0.51} & \hc{0.56} & \hcbu{0.43} & \hc{0.44} & \hc{0.51} & \hc{0.56} & \hc{0.44} & \hc{0.45} & \hc{0.51} & \hc{0.56} & \hc{0.44} & \hc{0.44} & \hc{0.51} & \hc{0.56} & \hcu{0.43} & \hc{0.44} & \hc{0.51} & \hc{0.56} & \hc{0.69} \\
CPC18 {\fontsize{4}{5}\selectfont\citep{plonsky2018when}} & Decision & \hc{0.35} & \hc{0.35} & \hc{0.38} & \hc{0.42} & \hc{0.34} & \hc{0.35} & \hc{0.38} & \hc{0.41} & \hc{0.34} & \hc{0.34} & \hc{0.38} & \hc{0.41} & \hc{0.34} & \hc{0.34} & \hc{0.38} & \hc{0.41} & \hcbu{0.33} & \hcu{0.34} & \hc{0.38} & \hc{0.41} & \hc{0.36} & \hc{0.36} & \hc{0.38} & \hc{0.41} & \hc{0.35} & \hc{0.35} & \hc{0.38} & \hc{0.41} & \hc{0.34} & \hc{0.34} & \hc{0.38} & \hc{0.41} & \hc{0.69} \\
Decisions from description {\fontsize{4}{5}\selectfont\citep{wulff2018sampling}} & Decision & \hc{0.62} & \hc{0.85} & \hc{0.67} & \hc{0.66} & \hc{0.63} & \hc{0.78} & \hc{0.69} & \hc{0.77} & \hc{0.62} & \hc{0.65} & \hc{0.69} & \hc{0.74} & \hc{0.61} & \hc{0.71} & \hc{0.67} & \hc{0.67} & \hcu{0.60} & \hc{0.63} & \hc{0.66} & \hc{0.66} & \hc{0.62} & \hc{0.81} & \hc{0.66} & \hc{0.67} & \hc{0.63} & \hc{0.67} & \hc{0.68} & \hc{0.67} & \hcbu{0.60} & \hc{0.69} & \hc{0.71} & \hc{0.68} & \hc{0.69} \\
\midrule
Two-step task {\fontsize{4}{5}\selectfont\citep{kool2016when}} & MDP & \hc{0.51} & \hc{0.53} & \hc{1.20} & \hc{1.32} & \hc{0.50} & \hc{0.52} & \hc{1.19} & \hc{1.30} & \hc{0.49} & \hc{0.52} & \hc{1.17} & \hc{1.28} & \hc{0.49} & \hc{0.52} & \hc{1.16} & \hc{1.29} & \hcbu{0.48} & \hc{0.52} & \hc{1.16} & \hc{1.23} & \hc{0.52} & \hc{0.52} & \hc{1.21} & \hc{1.34} & \hc{0.50} & \hc{0.51} & \hc{1.17} & \hc{1.36} & \hcu{0.49} & \hc{0.51} & \hc{1.18} & \hc{1.30} & \hc{0.69} \\
Two-step task {\fontsize{4}{5}\selectfont\citep{kool2017cost}} & MDP & \hc{0.54} & \hc{0.55} & \hc{1.16} & \hc{1.27} & \hc{0.53} & \hc{0.55} & \hc{1.15} & \hc{1.24} & \hcbu{0.53} & \hc{0.54} & \hc{1.13} & \hc{1.20} & \hc{0.53} & \hc{0.54} & \hc{1.12} & \hc{1.21} & \hc{0.53} & \hc{0.53} & \hc{1.11} & \hc{1.15} & \hc{0.55} & \hc{0.55} & \hc{1.15} & \hc{1.29} & \hc{0.53} & \hc{0.54} & \hc{1.13} & \hc{1.32} & \hcu{0.53} & \hc{0.53} & \hc{1.14} & \hc{1.25} & \hc{0.69} \\
Virtual subway network {\fontsize{4}{5}\selectfont\citep{tomov2020discovery}} & MDP & \hc{1.16} & \hc{1.42} & \hc{1.47} & \hc{1.37} & \hc{1.15} & \hc{1.49} & \hc{1.44} & \hc{1.35} & \hc{1.14} & \hc{1.43} & \hc{1.44} & \hc{1.33} & \hcu{1.13} & \hc{1.46} & \hc{1.43} & \hc{1.33} & \hcbu{1.12} & \hc{1.41} & \hc{1.43} & \hc{1.30} & \hc{1.19} & \hc{1.44} & \hc{1.45} & \hc{1.38} & \hc{1.16} & \hc{1.45} & \hc{1.47} & \hc{1.39} & \hc{1.13} & \hc{1.43} & \hc{1.42} & \hc{1.36} & \hc{1.61} \\
Multi-task RL {\fontsize{4}{5}\selectfont\citep{tomov2021multitask}} & MDP & \hc{0.60} & \hc{0.69} & \hc{0.81} & \hc{0.87} & \hc{0.59} & \hc{0.67} & \hc{0.80} & \hc{0.84} & \hc{0.58} & \hc{0.65} & \hc{0.78} & \hc{0.83} & \hc{0.58} & \hc{0.65} & \hc{0.77} & \hc{0.83} & \hcbu{0.57} & \hc{0.64} & \hc{0.76} & \hc{0.81} & \hc{0.60} & \hc{0.69} & \hc{0.82} & \hc{0.85} & \hc{0.59} & \hc{0.68} & \hc{0.78} & \hc{0.87} & \hcu{0.57} & \hc{0.67} & \hc{0.79} & \hc{0.84} & \hc{1.10} \\
Two-step task {\fontsize{4}{5}\selectfont\citep{zorowitz2023data}} & MDP & \hc{0.52} & \hc{0.52} & \hc{1.00} & \hc{1.23} & \hc{0.51} & \hc{0.52} & \hc{0.98} & \hc{1.16} & \hc{0.51} & \hc{0.53} & \hc{0.96} & \hc{1.10} & \hcbu{0.51} & \hc{0.52} & \hc{0.95} & \hc{1.13} & \hc{0.51} & \hc{0.52} & \hc{0.94} & \hc{1.02} & \hc{0.52} & \hc{0.53} & \hc{1.00} & \hc{1.24} & \hcu{0.51} & \hc{0.52} & \hc{0.97} & \hc{1.28} & \hc{0.51} & \hc{0.51} & \hc{0.97} & \hc{1.16} & \hc{0.69} \\
\midrule
Drifting four-armed bandit {\fontsize{4}{5}\selectfont\citep{bahrami2020four}} & Bandit & \hc{0.73} & \hc{0.88} & \hc{0.88} & \hc{0.88} & \hc{0.72} & \hc{0.85} & \hc{0.86} & \hc{0.88} & \hc{0.71} & \hc{0.83} & \hc{0.85} & \hc{0.86} & \hc{0.71} & \hc{0.84} & \hc{0.85} & \hc{0.87} & \hcbu{0.71} & \hc{0.87} & \hc{0.84} & \hc{0.87} & \hc{0.73} & \hc{0.88} & \hc{0.88} & \hc{0.88} & \hc{0.72} & \hc{0.84} & \hc{0.85} & \hc{0.87} & \hcu{0.71} & \hc{0.84} & \hc{0.84} & \hc{0.86} & \hc{1.39} \\
Horizon task {\fontsize{4}{5}\selectfont\citep{feng2021dynamics}} & Bandit & \hc{0.23} & \hc{0.25} & \hc{0.45} & \hc{0.55} & \hc{0.23} & \hc{0.25} & \hc{0.44} & \hc{0.55} & \hc{0.22} & \hc{0.24} & \hc{0.44} & \hc{0.55} & \hc{0.22} & \hc{0.24} & \hc{0.44} & \hc{0.55} & \hcu{0.22} & \hc{0.24} & \hc{0.44} & \hc{0.55} & \hc{0.26} & \hc{0.27} & \hc{0.45} & \hc{0.55} & \hc{0.24} & \hc{0.26} & \hc{0.45} & \hc{0.55} & \hcbu{0.22} & \hc{0.24} & \hc{0.44} & \hc{0.54} & \hc{0.69} \\
Two-armed bandit {\fontsize{4}{5}\selectfont\citep{gershman2018deconstructing}} & Bandit & \hc{0.30} & \hc{0.36} & \hc{0.39} & \hc{0.46} & \hc{0.30} & \hc{0.36} & \hc{0.39} & \hc{0.47} & \hc{0.29} & \hc{0.36} & \hc{0.39} & \hc{0.45} & \hcu{0.29} & \hc{0.36} & \hc{0.39} & \hc{0.45} & \hcbu{0.29} & \hc{0.35} & \hc{0.38} & \hc{0.45} & \hc{0.30} & \hc{0.34} & \hc{0.39} & \hc{0.46} & \hc{0.29} & \hc{0.34} & \hc{0.39} & \hc{0.46} & \hc{0.29} & \hc{0.35} & \hc{0.39} & \hc{0.46} & \hc{0.69} \\
Prob.\ instrumental learning {\fontsize{4}{5}\selectfont\citep{lefebvre2017behavioural}} & Bandit & \hc{0.50} & \hc{0.49} & \hc{2.12} & \hc{2.10} & \hc{0.50} & \hc{0.50} & \hc{2.09} & \hc{2.06} & \hcbu{0.49} & \hc{0.49} & \hc{2.06} & \hc{2.02} & \hcu{0.49} & \hc{0.49} & \hc{2.06} & \hc{2.04} & \hc{0.49} & \hc{0.49} & \hc{2.03} & \hc{2.03} & \hc{0.51} & \hc{0.51} & \hc{2.17} & \hc{2.15} & \hc{0.50} & \hc{0.49} & \hc{2.09} & \hc{2.06} & \hc{0.49} & \hc{0.49} & \hc{2.04} & \hc{2.05} & \hc{0.69} \\
Horizon task {\fontsize{4}{5}\selectfont\citep{sadeghiyeh2020temporal}} & Bandit & \hc{0.59} & \hc{0.59} & \hc{0.63} & \hc{0.68} & \hc{0.58} & \hc{0.59} & \hc{0.63} & \hc{0.68} & \hc{0.58} & \hc{0.59} & \hc{0.63} & \hc{0.67} & \hcu{0.58} & \hc{0.58} & \hc{0.63} & \hc{0.67} & \hcbu{0.58} & \hc{0.58} & \hc{0.63} & \hc{0.68} & \hc{0.59} & \hc{0.59} & \hc{0.63} & \hc{0.68} & \hc{0.58} & \hc{0.59} & \hc{0.63} & \hc{0.68} & \hc{0.58} & \hc{0.58} & \hc{0.63} & \hc{0.68} & \hc{0.69} \\
Structured bandit {\fontsize{4}{5}\selectfont\citep{schulz2020finding}} & Bandit & \hc{0.67} & \hc{0.71} & \hc{0.88} & \hc{1.04} & \hc{0.66} & \hc{0.70} & \hc{0.88} & \hc{1.04} & \hc{0.65} & \hc{0.69} & \hc{0.86} & \hc{1.01} & \hc{0.65} & \hc{0.68} & \hc{0.85} & \hc{0.99} & \hcbu{0.64} & \hc{0.68} & \hc{0.87} & \hc{1.00} & \hc{0.67} & \hc{0.70} & \hc{0.87} & \hc{1.05} & \hc{0.66} & \hc{0.69} & \hc{0.86} & \hc{1.03} & \hcu{0.65} & \hc{0.68} & \hc{0.85} & \hc{1.00} & \hc{2.08} \\
Horizon task {\fontsize{4}{5}\selectfont\citep{somerville2017charting}} & Bandit & \hc{0.34} & \hc{0.36} & \hc{0.53} & \hc{0.61} & \hc{0.34} & \hc{0.35} & \hc{0.53} & \hc{0.60} & \hc{0.34} & \hc{0.35} & \hc{0.52} & \hc{0.60} & \hcu{0.34} & \hc{0.34} & \hc{0.52} & \hc{0.60} & \hcbu{0.33} & \hc{0.35} & \hc{0.52} & \hc{0.60} & \hc{0.35} & \hc{0.36} & \hc{0.52} & \hc{0.60} & \hc{0.34} & \hc{0.35} & \hc{0.52} & \hc{0.60} & \hc{0.34} & \hc{0.34} & \hc{0.52} & \hc{0.60} & \hc{0.69} \\
Iowa gambling task {\fontsize{4}{5}\selectfont\citep{steingroever2015data}} & Bandit & \hc{0.93} & \hc{1.12} & \hc{1.00} & \hc{1.01} & \hc{0.93} & \hc{1.09} & \hc{0.99} & \hc{1.00} & \hc{0.91} & \hc{1.07} & \hc{0.97} & \hc{0.98} & \hc{0.91} & \hc{1.08} & \hc{0.97} & \hc{0.98} & \hcu{0.91} & \hc{1.08} & \hc{0.97} & \hc{0.98} & \hc{0.93} & \hc{1.11} & \hc{1.00} & \hc{1.01} & \hc{0.92} & \hc{1.09} & \hc{0.98} & \hc{0.99} & \hcbu{0.91} & \hc{1.08} & \hc{0.97} & \hc{0.98} & \hc{1.39} \\
Horizon task {\fontsize{4}{5}\selectfont\citep{waltz2020differential}} & Bandit & \hc{0.15} & \hc{0.16} & \hc{0.33} & \hc{0.44} & \hc{0.15} & \hc{0.16} & \hc{0.33} & \hc{0.44} & \hc{0.15} & \hc{0.17} & \hc{0.33} & \hc{0.44} & \hc{0.15} & \hc{0.16} & \hc{0.33} & \hc{0.44} & \hcu{0.15} & \hc{0.16} & \hc{0.32} & \hc{0.44} & \hc{0.16} & \hc{0.17} & \hc{0.33} & \hc{0.44} & \hc{0.16} & \hc{0.16} & \hc{0.33} & \hc{0.44} & \hcbu{0.14} & \hc{0.15} & \hc{0.33} & \hc{0.44} & \hc{0.69} \\
Horizon task {\fontsize{4}{5}\selectfont\citep{wilson2014humans}} & Bandit & \hc{0.46} & \hc{0.47} & \hc{0.56} & \hc{0.62} & \hc{0.46} & \hc{0.47} & \hc{0.55} & \hc{0.62} & \hc{0.45} & \hc{0.46} & \hc{0.55} & \hc{0.62} & \hc{0.45} & \hc{0.46} & \hc{0.54} & \hc{0.62} & \hcbu{0.45} & \hc{0.46} & \hc{0.54} & \hc{0.62} & \hc{0.47} & \hc{0.48} & \hc{0.56} & \hc{0.62} & \hc{0.46} & \hc{0.47} & \hc{0.55} & \hc{0.62} & \hcu{0.45} & \hc{0.46} & \hc{0.55} & \hc{0.62} & \hc{0.69} \\
Spatially correlated MAB {\fontsize{4}{5}\selectfont\citep{wu2018generalization}} & Bandit & \hc{2.01} & \hc{2.06} & \hc{2.41} & \hc{2.56} & \hc{1.96} & \hc{2.05} & \hc{2.36} & \hc{2.53} & \hc{1.93} & \hc{2.02} & \hc{2.37} & \hc{2.49} & \hcu{1.87} & \hc{1.99} & \hc{2.33} & \hc{2.49} & \hcbu{1.86} & \hc{1.97} & \hc{2.38} & \hc{2.50} & \hc{2.16} & \hc{2.20} & \hc{2.54} & \hc{2.73} & \hc{2.00} & \hc{2.04} & \hc{2.52} & \hc{2.68} & \hc{1.93} & \hc{2.03} & \hc{2.46} & \hc{2.61} & \hc{3.40} \\
Decisions from experience {\fontsize{4}{5}\selectfont\citep{wulff2018sampling}} & Bandit & \hc{0.49} & \hc{0.95} & \hc{0.56} & \hc{0.58} & \hc{0.49} & \hc{0.94} & \hc{0.54} & \hc{0.59} & \hc{0.49} & \hc{0.93} & \hc{0.57} & \hc{0.59} & \hc{0.49} & \hc{0.93} & \hc{0.54} & \hc{0.57} & \hcu{0.48} & \hc{0.95} & \hc{0.53} & \hc{0.56} & \hc{0.50} & \hc{0.99} & \hc{0.54} & \hc{0.60} & \hc{0.49} & \hc{0.90} & \hc{0.56} & \hc{0.59} & \hcbu{0.48} & \hc{0.97} & \hc{0.55} & \hc{0.58} & \hcn \\
Changing bandit {\fontsize{4}{5}\selectfont\citep{xiong2023neural}} & Bandit & \hc{0.30} & \hc{0.31} & \hc{0.48} & \hc{0.44} & \hc{0.30} & \hc{0.30} & \hc{0.49} & \hc{0.43} & \hcu{0.29} & \hc{0.30} & \hc{0.48} & \hc{0.42} & \hc{0.29} & \hc{0.30} & \hc{0.48} & \hc{0.44} & \hcbu{0.29} & \hc{0.29} & \hc{0.47} & \hc{0.43} & \hc{0.32} & \hc{0.33} & \hc{0.49} & \hc{0.53} & \hc{0.31} & \hc{0.32} & \hc{0.49} & \hc{0.44} & \hc{0.29} & \hc{0.30} & \hc{0.47} & \hc{0.42} & \hc{0.69} \\
\midrule
Cond.\ assoc.\ learning {\fontsize{4}{5}\selectfont\citep{collins2014working}} & Memory & \hc{0.54} & \hc{0.57} & \hc{1.03} & \hc{1.07} & \hc{0.53} & \hc{0.56} & \hc{1.03} & \hc{1.07} & \hc{0.52} & \hc{0.55} & \hc{1.02} & \hc{1.07} & \hcu{0.52} & \hc{0.56} & \hc{1.02} & \hc{1.06} & \hcbu{0.51} & \hc{0.55} & \hc{1.01} & \hc{1.06} & \hc{0.55} & \hc{0.58} & \hc{1.05} & \hc{1.08} & \hc{0.53} & \hc{0.56} & \hc{1.03} & \hc{1.07} & \hc{0.52} & \hc{0.55} & \hc{1.02} & \hc{1.06} & \hc{1.10} \\
\midrule
Shepard categorization {\fontsize{4}{5}\selectfont\citep{badham2017deficits}} & Sup.\ learn. & \hc{0.55} & \hc{0.57} & \hc{0.71} & \hc{0.70} & \hc{0.55} & \hc{0.58} & \hc{0.70} & \hc{0.70} & \hc{0.54} & \hc{0.56} & \hc{0.69} & \hc{0.69} & \hc{0.53} & \hc{0.57} & \hc{0.69} & \hc{0.69} & \hcbu{0.53} & \hc{0.56} & \hc{0.69} & \hc{0.70} & \hc{0.56} & \hc{0.58} & \hc{0.72} & \hc{0.71} & \hc{0.54} & \hc{0.56} & \hc{0.70} & \hc{0.70} & \hcu{0.53} & \hc{0.57} & \hc{0.69} & \hc{0.70} & \hc{0.69} \\
Multiple-cue judgment {\fontsize{4}{5}\selectfont\citep{collsioo2023numerical}} & Sup.\ learn. & \hc{1.17} & \hc{1.23} & \hc{1.99} & \hc{1.99} & \hc{1.16} & \hc{1.23} & \hc{1.97} & \hc{1.97} & \hc{1.14} & \hc{1.20} & \hc{1.95} & \hc{1.96} & \hc{1.14} & \hc{1.19} & \hc{1.96} & \hc{1.97} & \hcu{1.13} & \hc{1.19} & \hc{1.95} & \hc{1.96} & \hc{1.17} & \hc{1.23} & \hc{1.96} & \hc{1.97} & \hc{1.15} & \hc{1.20} & \hc{1.98} & \hc{1.98} & \hcbu{1.12} & \hc{1.19} & \hc{1.96} & \hc{1.97} & \hc{2.20} \\
Medin categorization {\fontsize{4}{5}\selectfont\citep{levering2020revisiting}} & Sup.\ learn. & \hc{0.51} & \hc{0.61} & \hc{0.79} & \hc{0.94} & \hc{0.51} & \hc{0.61} & \hc{0.79} & \hc{0.89} & \hc{0.50} & \hc{0.61} & \hc{0.78} & \hc{0.91} & \hc{0.50} & \hc{0.60} & \hc{0.78} & \hc{0.88} & \hcbu{0.50} & \hc{0.59} & \hc{0.77} & \hc{0.87} & \hc{0.52} & \hc{0.62} & \hc{0.79} & \hc{0.95} & \hc{0.50} & \hc{0.59} & \hc{0.79} & \hc{0.92} & \hcu{0.50} & \hc{0.58} & \hc{0.78} & \hc{0.91} & \hcn \\
Weather prediction task {\fontsize{4}{5}\selectfont\citep{speekenbrink2008learning}} & Sup.\ learn. & \hc{0.55} & \hc{0.61} & \hc{0.71} & \hc{0.70} & \hc{0.56} & \hc{0.65} & \hc{0.70} & \hc{0.70} & \hcu{0.54} & \hc{0.61} & \hc{0.70} & \hc{0.70} & \hc{0.55} & \hc{0.61} & \hc{0.70} & \hc{0.69} & \hc{0.55} & \hc{0.60} & \hc{0.70} & \hc{0.71} & \hc{0.55} & \hc{0.62} & \hc{0.72} & \hc{0.71} & \hcbu{0.54} & \hc{0.61} & \hc{0.71} & \hc{0.71} & \hc{0.56} & \hc{0.61} & \hc{0.71} & \hc{0.70} & \hc{0.69} \\
Aversive learning {\fontsize{4}{5}\selectfont\citep{wise2019computational}} & Sup.\ learn. & \hc{4.71} & \hc{4.72} & \hc{4.86} & \hc{4.88} & \hc{4.55} & \hc{4.60} & \hc{4.72} & \hc{4.75} & \hc{4.27} & \hc{4.30} & \hc{4.46} & \hc{4.53} & \hc{4.24} & \hc{4.25} & \hc{4.42} & \hc{4.49} & \hc{3.81} & \hc{3.86} & \hc{4.05} & \hc{4.15} & \hc{4.58} & \hc{4.59} & \hc{4.77} & \hc{4.73} & \hc{4.80} & \hc{4.80} & \hc{4.97} & \hc{4.96} & \hcbu{3.66} & \hcu{3.69} & \hc{3.88} & \hc{3.91} & \hcn \\
\midrule
\textbf{Mean (32)} & & \hc{0.69} & \hc{0.76} & \hc{1.00} & \hc{1.07} & \hc{0.68} & \hc{0.76} & \hc{0.99} & \hc{1.06} & \hc{0.66} & \hc{0.73} & \hc{0.97} & \hc{1.04} & \hc{0.66} & \hc{0.73} & \hc{0.97} & \hc{1.03} & \hcu{0.64} & \hc{0.71} & \hc{0.95} & \hc{1.01} & \hc{0.70} & \hc{0.76} & \hc{1.00} & \hc{1.08} & \hc{0.69} & \hc{0.75} & \hc{1.00} & \hc{1.08} & \hcbu{0.64} & \hc{0.71} & \hc{0.95} & \hc{1.02} & \hc{1.02} \\
\bottomrule
\end{tabular}%
}
\vspace{4pt}
\begin{center}\footnotesize
\colorbox{c1!45}{\strut\hspace{5pt}} \scriptsize$<$0.20 \quad
\colorbox{c2!40}{\strut\hspace{5pt}} \scriptsize 0.20--0.40 \quad
\colorbox{c3!40}{\strut\hspace{5pt}} \scriptsize 0.40--0.60 \quad
\colorbox{c4!45}{\strut\hspace{5pt}} \scriptsize 0.60--0.80 \quad
\colorbox{c5!55}{\strut\hspace{5pt}} \scriptsize 0.80--1.00 \quad
\colorbox{c6!50}{\strut\hspace{5pt}} \scriptsize 1.00--1.20 \quad
\colorbox{c7!45}{\strut\hspace{5pt}} \scriptsize 1.20--1.50 \quad
\colorbox{c8!40}{\strut\hspace{5pt}} \scriptsize 1.50--2.00 \quad
\colorbox{c9!35}{\strut\hspace{5pt}} \scriptsize$>$2.00
\end{center}
\vspace{-2pt}
{\tiny \underline{\textbf{Bold+underline}}\,=\,best; \underline{underline}\,=\,second-best. Lower is better. Condition labels: \textbf{orig}\,=\,original prompt, \textbf{inst}\,=\,instruction ablated, \textbf{cont}\,=\,content masked, \textbf{hist}\,=\,history only.}
\caption{\textbf{Per-experiment NLL under sequential ablation conditions on Psych-101 (in-distribution).} Each model size shows four columns corresponding to progressive prompt degradation: \emph{orig} retains the full prompt; \emph{inst} removes task instructions; \emph{cont} additionally masks stimulus values and feedback; \emph{hist} further removes trial structure, leaving only the response history. The $\ln(k)$ column shows the random-guessing baseline where $k$ is the number of per-trial response options. Experiments with mixed or continuous response formats have no well-defined $k$ and are shown as { -- }.}
\label{tab:psych101_pe_ablation}
\end{sidewaystable}

\newpage

\begin{table}[t!]
\centering
\resizebox{\columnwidth}{!}{%
\renewcommand{\arraystretch}{1.15}
\begin{tabular}{@{}l l r r r r r@{}}
\toprule
\textbf{Experiment} & \textbf{Type} & \textbf{orig} & \textbf{inst} & \textbf{cont} & \textbf{hist} & $\ln(k)$ \\
\midrule
Gardening task {\fontsize{6}{7}\selectfont\citep{flesch2018comparing}} & Decision & \hc{0.49} & \hc{0.49} & \hc{0.69} & \hc{0.69} & \hc{0.69} \\
Columbia card task {\fontsize{6}{7}\selectfont\citep{frey2017risk}} & Decision & \hc{0.21} & \hc{0.24} & \hc{0.28} & \hc{0.29} & \hc{0.69} \\
Experiential-symbolic task {\fontsize{6}{7}\selectfont\citep{garcia2023experiential}} & Decision & \hc{0.46} & \hc{0.46} & \hc{0.85} & \hc{1.43} & \hcn \\
Multi-attribute DM {\fontsize{6}{7}\selectfont\citep{hilbig2014generalized}} & Decision & \hc{0.06} & \hc{0.08} & \hc{0.74} & \hc{0.74} & \hc{0.69} \\
Risky choice {\fontsize{6}{7}\selectfont\citep{krueger2024identifying}} & Decision & \hc{0.43} & \hc{0.64} & \hc{0.82} & \hc{0.89} & \hcn \\
choices13k {\fontsize{6}{7}\selectfont\citep{peterson2021using}} & Decision & \hc{0.43} & \hc{0.44} & \hc{0.52} & \hc{0.56} & \hc{0.69} \\
CPC18 {\fontsize{6}{7}\selectfont\citep{plonsky2018when}} & Decision & \hc{0.35} & \hc{0.35} & \hc{0.38} & \hc{0.41} & \hc{0.69} \\
Decisions from description {\fontsize{6}{7}\selectfont\citep{wulff2018sampling}} & Decision & \hc{0.59} & \hc{0.61} & \hc{0.69} & \hc{0.68} & \hc{0.69} \\
\midrule
Two-step task {\fontsize{6}{7}\selectfont\citep{kool2016when}} & MDP & \hc{0.48} & \hc{0.53} & \hc{1.18} & \hc{1.23} & \hc{0.69} \\
Two-step task {\fontsize{6}{7}\selectfont\citep{kool2017cost}} & MDP & \hc{0.53} & \hc{0.54} & \hc{1.15} & \hc{1.16} & \hc{0.69} \\
Virtual subway network {\fontsize{6}{7}\selectfont\citep{tomov2020discovery}} & MDP & \hc{1.16} & \hc{1.46} & \hc{1.41} & \hc{1.30} & \hc{1.61} \\
Multi-task RL {\fontsize{6}{7}\selectfont\citep{tomov2021multitask}} & MDP & \hc{0.57} & \hc{0.66} & \hc{0.77} & \hc{0.82} & \hc{1.10} \\
Two-step task {\fontsize{6}{7}\selectfont\citep{zorowitz2023data}} & MDP & \hc{0.51} & \hc{0.52} & \hc{0.97} & \hc{1.03} & \hc{0.69} \\
\midrule
Drifting four-armed bandit {\fontsize{6}{7}\selectfont\citep{bahrami2020four}} & Bandit & \hc{0.71} & \hc{0.85} & \hc{0.84} & \hc{0.86} & \hc{1.39} \\
Horizon task {\fontsize{6}{7}\selectfont\citep{feng2021dynamics}} & Bandit & \hc{0.40} & \hc{0.39} & \hc{0.53} & \hc{0.55} & \hc{0.69} \\
Two-armed bandit {\fontsize{6}{7}\selectfont\citep{gershman2018deconstructing}} & Bandit & \hc{0.30} & \hc{0.36} & \hc{0.39} & \hc{0.46} & \hc{0.69} \\
Prob.\ instrumental learning {\fontsize{6}{7}\selectfont\citep{lefebvre2017behavioural}} & Bandit & \hc{0.50} & \hc{0.49} & \hc{2.07} & \hc{2.01} & \hc{0.69} \\
Horizon task {\fontsize{6}{7}\selectfont\citep{sadeghiyeh2020temporal}} & Bandit & \hc{0.58} & \hc{0.59} & \hc{0.63} & \hc{0.68} & \hc{0.69} \\
Structured bandit {\fontsize{6}{7}\selectfont\citep{schulz2020finding}} & Bandit & \hc{0.64} & \hc{0.68} & \hc{0.84} & \hc{1.00} & \hc{2.08} \\
Horizon task {\fontsize{6}{7}\selectfont\citep{somerville2017charting}} & Bandit & \hc{0.35} & \hc{0.36} & \hc{0.53} & \hc{0.61} & \hc{0.69} \\
Iowa gambling task {\fontsize{6}{7}\selectfont\citep{steingroever2015data}} & Bandit & \hc{0.91} & \hc{1.08} & \hc{0.97} & \hc{0.97} & \hc{1.39} \\
Horizon task {\fontsize{6}{7}\selectfont\citep{waltz2020differential}} & Bandit & \hc{0.15} & \hc{0.15} & \hc{0.34} & \hc{0.43} & \hc{0.69} \\
Horizon task {\fontsize{6}{7}\selectfont\citep{wilson2014humans}} & Bandit & \hc{0.48} & \hc{0.48} & \hc{0.56} & \hc{0.62} & \hc{0.69} \\
Spatially correlated MAB {\fontsize{6}{7}\selectfont\citep{wu2018generalization}} & Bandit & \hc{1.82} & \hc{1.95} & \hc{2.45} & \hc{2.62} & \hc{3.40} \\
Decisions from experience {\fontsize{6}{7}\selectfont\citep{wulff2018sampling}} & Bandit & \hc{0.47} & \hc{0.95} & \hc{0.54} & \hc{0.57} & \hcn \\
Changing bandit {\fontsize{6}{7}\selectfont\citep{xiong2023neural}} & Bandit & \hc{0.45} & \hc{0.45} & \hc{1.28} & \hc{0.98} & \hc{0.69} \\
\midrule
Cond.\ assoc.\ learning {\fontsize{6}{7}\selectfont\citep{collins2014working}} & Memory & \hc{0.52} & \hc{0.55} & \hc{1.03} & \hc{1.06} & \hc{1.10} \\
\midrule
Shepard categorization {\fontsize{6}{7}\selectfont\citep{badham2017deficits}} & Sup.\ learn. & \hc{0.54} & \hc{0.58} & \hc{0.70} & \hc{0.70} & \hc{0.69} \\
Multiple-cue judgment {\fontsize{6}{7}\selectfont\citep{collsioo2023numerical}} & Sup.\ learn. & \hc{1.14} & \hc{1.19} & \hc{1.94} & \hc{1.94} & \hc{2.20} \\
Medin categorization {\fontsize{6}{7}\selectfont\citep{levering2020revisiting}} & Sup.\ learn. & \hc{0.50} & \hc{0.58} & \hc{0.79} & \hc{0.90} & \hcn \\
Weather prediction task {\fontsize{6}{7}\selectfont\citep{speekenbrink2008learning}} & Sup.\ learn. & \hc{0.56} & \hc{0.60} & \hc{0.72} & \hc{0.72} & \hc{0.69} \\
Aversive learning {\fontsize{6}{7}\selectfont\citep{wise2019computational}} & Sup.\ learn. & \hc{4.13} & \hc{4.17} & \hc{4.31} & \hc{4.29} & \hcn \\
\midrule
\textbf{Mean (32)} & & \hc{0.67} & \hc{0.73} & \hc{1.00} & \hc{1.04} & \hc{1.02} \\
\bottomrule
\end{tabular}%
}
\vspace{4pt}
\begin{center}\footnotesize
\colorbox{c1!45}{\strut\hspace{5pt}} \scriptsize$<$0.20 \quad
\colorbox{c2!40}{\strut\hspace{5pt}} \scriptsize 0.20--0.40 \quad
\colorbox{c3!40}{\strut\hspace{5pt}} \scriptsize 0.40--0.60 \quad
\colorbox{c4!45}{\strut\hspace{5pt}} \scriptsize 0.60--0.80 \quad
\colorbox{c5!55}{\strut\hspace{5pt}} \scriptsize 0.80--1.00 \quad
\colorbox{c6!50}{\strut\hspace{5pt}} \scriptsize 1.00--1.20 \quad
\colorbox{c7!45}{\strut\hspace{5pt}} \scriptsize 1.20--1.50 \quad
\colorbox{c8!40}{\strut\hspace{5pt}} \scriptsize 1.50--2.00 \quad
\colorbox{c9!35}{\strut\hspace{5pt}} \scriptsize$>$2.00
\end{center}
\vspace{-2pt}
{\tiny Lower is better. Condition labels: \textbf{orig}\,=\,original prompt, \textbf{inst}\,=\,instruction ablated, \textbf{cont}\,=\,content masked, \textbf{hist}\,=\,history only.}
\caption{\textbf{Per-experiment NLL under sequential ablation conditions for Centaur-70B \citep{binz2025foundation} on Psych-101 (in-distribution).} Columns correspond to progressive prompt degradation: \emph{orig} retains the full prompt; \emph{inst} removes task instructions; \emph{cont} additionally masks stimulus values and feedback; \emph{hist} further removes trial structure, leaving only the response history. The $\ln(k)$ column shows the random-guessing baseline where $k$ is the number of per-trial response options. Experiments with mixed or continuous response formats have no well-defined $k$ and are shown as { -- }.}
\label{tab:tab:psych101_pe_ablation_centaur70b}
\end{table}

\clearpage

\clearpage

\subsection{Order Permutation (non-sequential tasks)}
\label{app:order_permutation}

Figures~\ref{fig:permutation_hebart2023things} and~\ref{fig:permutation_ruggeri2022globalizability} show original against permuted trial orderings for the two experiments that anchor opposite ends of the exchangeability spectrum.



\begin{figure}[ht!]
\centering
\begin{minipage}[t]{0.48\textwidth}
\begin{promptbox}[Original]
\textcolor{colI}{You will be presented with triplets of objects, which will be assigned to the keys T, Z, and R. In each trial, please indicate which object you think is the odd one out by pressing the corresponding key. In other words, please choose the object that is the least similar to the other two.}\par\smallskip
\textcolor{colS}{T: bottle opener, Z: ironing board, and R: goalpost.} \textcolor{colC}{You press <<R>>.}\par
\textcolor{colS}{T: calculator, Z: wrist, and R: laptop.} \textcolor{colC}{You press <<Z>>.}\par
\textcolor{colS}{T: tool, Z: chain, and R: quad.} \textcolor{colC}{You press <<R>>.}\par
\textcolor{colS}{T: birdbath, Z: bison, and R: jellyfish.} \textcolor{colC}{You press <<}\textcolor{midgray}{\underline{\phantom{MM}}}
\end{promptbox}
\end{minipage}%
\hfill
\begin{minipage}[t]{0.48\textwidth}
\begin{promptbox}[Permuted order]
\textcolor{colI}{You will be presented with triplets of objects, which will be assigned to the keys T, Z, and R. In each trial, please indicate which object you think is the odd one out by pressing the corresponding key. In other words, please choose the object that is the least similar to the other two.}\par\smallskip
\textcolor{colS}{T: tool, Z: chain, and R: quad.} \textcolor{colC}{You press <<R>>.}\par
\textcolor{colS}{T: calculator, Z: wrist, and R: laptop.} \textcolor{colC}{You press <<Z>>.}\par
\textcolor{colS}{T: bottle opener, Z: ironing board, and R: goalpost.} \textcolor{colC}{You press <<R>>.}\par
\textcolor{colS}{T: birdbath, Z: bison, and R: jellyfish.} \textcolor{colC}{You press <<}\textcolor{midgray}{\underline{\phantom{MM}}}
\end{promptbox}
\end{minipage}
\vspace{4pt}
\begin{center}
\sffamily\fontsize{6.5}{8}\selectfont\color{darkgray}
\textcolor{colI}{\rule{5pt}{5pt}}\;\,Instruction\quad
\textcolor{colS}{\rule{5pt}{5pt}}\;\,Stimuli\quad
\textcolor{colC}{\rule{5pt}{5pt}}\;\,Choice\quad
\textcolor{midgray}{\underline{\phantom{MM}}}\;\,Model prediction
\end{center}
\vspace{-2pt}
\caption{\textbf{Miscellaneous: THINGS odd-one-out} \citep{hebart2023things}.}
\label{fig:permutation_hebart2023things}
\end{figure}

\begin{figure}[ht!]
\centering
\begin{minipage}[t]{0.48\textwidth}
\begin{promptbox}[Original]
\textcolor{colI}{In the following you will be presented with multiple choices between two options Q and L. Please name which option you would prefer by pressing the corresponding key.}\par\smallskip
\textcolor{colS}{You have the choice between receiving 500\$ immediately (press Q) or receiving 550\$ in one year (press L).} \textcolor{colC}{You press <<Q>>.}\par
\textcolor{colS}{You have the choice between receiving 500\$ immediately (press Q) or receiving 600\$ in one year (press L).} \textcolor{colC}{You press <<L>>.}\par
\textcolor{colS}{You have the choice between paying 500\$ immediately (press Q) or paying 550\$ in one year (press L).} \textcolor{colC}{You press <<L>>.}\par
\textcolor{colS}{You have the choice between paying 500\$ immediately (press Q) or paying 600\$ in one year (press L).} \textcolor{colC}{You press <<}\textcolor{midgray}{\underline{\phantom{MM}}}
\end{promptbox}
\end{minipage}%
\hfill
\begin{minipage}[t]{0.48\textwidth}
\begin{promptbox}[Permuted order]
\textcolor{colI}{In the following you will be presented with multiple choices between two options Q and L. Please name which option you would prefer by pressing the corresponding key.}\par\smallskip
\textcolor{colS}{You have the choice between paying 500\$ immediately (press Q) or paying 550\$ in one year (press L).} \textcolor{colC}{You press <<L>>.}\par
\textcolor{colS}{You have the choice between receiving 500\$ immediately (press Q) or receiving 600\$ in one year (press L).} \textcolor{colC}{You press <<L>>.}\par
\textcolor{colS}{You have the choice between receiving 500\$ immediately (press Q) or receiving 550\$ in one year (press L).} \textcolor{colC}{You press <<Q>>.}\par
\textcolor{colS}{You have the choice between paying 500\$ immediately (press Q) or paying 600\$ in one year (press L).} \textcolor{colC}{You press <<}\textcolor{midgray}{\underline{\phantom{MM}}}
\end{promptbox}
\end{minipage}
\vspace{4pt}
\begin{center}
\sffamily\fontsize{6.5}{8}\selectfont\color{darkgray}
\textcolor{colI}{\rule{5pt}{5pt}}\;\,Instruction\quad
\textcolor{colS}{\rule{5pt}{5pt}}\;\,Stimuli\quad
\textcolor{colC}{\rule{5pt}{5pt}}\;\,Choice\quad
\textcolor{midgray}{\underline{\phantom{MM}}}\;\,Model prediction
\end{center}
\vspace{-2pt}
\caption{\textbf{Decision-making: Intertemporal choice} \citep{ruggeri2022globalizability}.}
\label{fig:permutation_ruggeri2022globalizability}
\end{figure}

\clearpage

\end{document}